\documentclass[10pt]{article} 
\usepackage[accepted]{tmlr}

\usepackage{amsmath,amsfonts,bm}

\def\eqref#1{equation~\ref{#1}}

\def\1{\bm{1}}

\DeclareMathAlphabet{\mathsfit}{\encodingdefault}{\sfdefault}{m}{sl}
\SetMathAlphabet{\mathsfit}{bold}{\encodingdefault}{\sfdefault}{bx}{n}

\usepackage{hyperref}
\usepackage{url}

\usepackage{algorithm}
\usepackage{algpseudocode}
\usepackage{float}
\usepackage{placeins}
\usepackage{booktabs} 
\usepackage{graphicx}
\usepackage{subcaption}
\usepackage{amssymb}
\usepackage{mathtools}
\usepackage{amsthm}
\usepackage{multirow}
\usepackage[table]{xcolor}
\usepackage{siunitx}
\newcommand{\alignbold}[2]{\makebox[0pt][l]{\phantom{#1}\makebox[0pt][r]{\textbf{#1}}\makebox[0pt][l]{\textbf{.#2}}}\phantom{#1.#2}}
\newcommand{\alignunder}[2]{\makebox[0pt][l]{\phantom{#1}\makebox[0pt][r]{\underline{#1}}\makebox[0pt][l]{\underline{.#2}}}\phantom{#1.#2}}

\usepackage{pifont}
\newcommand{\cmark}{\ding{51}}
\newcommand{\xmark}{\ding{55}}

\title{Shiva-DiT: Residual-Based Differentiable Top-$k$ Selection for Efficient Diffusion Transformers\thanks{Large language models were used solely for language editing.}}

\author{\name Jiaji Zhang \email zhangjiaji@zju.edu.cn \\
      \addr College of Computer Science and Technology, Zhejiang University
      \AND
      \name Hailiang Zhao\thanks{Corresponding authors.} \email hliangzhao@zju.edu.cn \\
      \addr School of Software Technology, Zhejiang University
      \AND
      \name Jiaju Wu \email wujiajucn@gmail.com \\
      \addr College of Computing and Data Science, Nanyang Technological University
      \AND
      \name Ruichao Sun \email csunray@zju.edu.cn \\
      \addr College of Computer Science and Technology, Zhejiang University
      \AND
      \name Xinkui Zhao \email zhaoxinkui@zju.edu.cn \\
      \addr School of Software Technology, Zhejiang University
      \AND
      \name Shuiguang Deng\footnotemark[2] \email dengsg@zju.edu.cn\\
      \addr College of Computer Science and Technology, Zhejiang University}

\def\month{11}  
\def\year{2026} 
\def\openreview{\url{https://openreview.net/forum?id=uAAaT1KajI}} 

\begin{document}

\maketitle

\begin{abstract}
Diffusion Transformers (DiTs) are costly at high resolution because self-attention scales quadratically with token sequence length.
Existing pruning methods do not jointly provide end-to-end learnability, low training overhead, and deterministic token counts for predictable token-dependent computation.
We propose \emph{Shiva-DiT}, based on \emph{Residual-Based Differentiable Top-$k$ Selection}.
Its forward pass executes hard top-$k$ selection, while a residual-aware straight-through estimator propagates gradients to both token scores and the budget $k$ without evaluating a second backbone path.
A Context-Aware Router ranks tokens based on the input, while an Adaptive Ratio Policy learns layer- and timestep-dependent retention ratios under a target average budget.
Experiments on SD3-Medium, Flux.1-dev, and PixArt-$\Sigma$ show consistent reductions in FLOPs and measured latency. On SD3-Medium, Shiva-DiT provides four fidelity-latency operating points and reaches a 1.54$\times$ wall-clock speedup with competitive fidelity.
\end{abstract}

\raggedbottom
\section{Introduction}
\label{sec:intro}

Diffusion Transformers (DiTs) are widely used in high-quality image and video generation~\citep{dit2023, pixartsigma2024, sd32024, flux2024, sora2024}. Their transformer blocks repeatedly process flattened spatial-token sequences, so self-attention cost grows quadratically with sequence length. At $1024\times1024$, a typical latent grid contains 4096 image tokens that are processed across many layers and denoising steps. This repeated computation makes high-resolution DiT inference expensive.

\begin{samepage}
Dynamic token pruning can reduce this cost, but a practical method must balance three requirements, which we call the \emph{Trilemma of Sparse Learning} (Figure~\ref{fig:trilemma}): \textbf{1) Differentiability}, so the pruning policy can be learned end to end; \textbf{2) Efficiency}, with low training and inference overhead; and \textbf{3) Strict Budget}, with a deterministic top-$k$ count and regular tensor shapes.
Once a layer and timestep budget is fixed, the token-dependent attention and FFN workload is known before those operators execute, and selection can use standard \texttt{gather}/\texttt{scatter} primitives. Actual latency still depends on hardware and implementation.
\par
\end{samepage}

Existing approaches typically compromise on at least one of these vertices. Heuristic methods~\citep{tomesd2023, ibtm2025, toma2025, sdtm2025} ensure hardware compatibility but lack learnability, relying on static rules that fail to capture semantic nuances.
Conversely, learnable masking methods inspired by Gumbel-Softmax~\citep{dydit2024} achieve differentiability but treat pruning as a thresholding problem.
This results in \textit{ragged tensors} with variable sequence lengths, so the realized budget is only met in expectation and the cost of a given input cannot be determined in advance.
DiffCR~\citep{diffcr2025} learns the budget through interpolation between discrete ratio bins, requiring two backbone forward passes during training.
Similarly, soft-ranking operators like NeuralSort~\citep{neuralsort2019} offer differentiable sorting, but constructing and multiplying the required $N \times N$ permutation matrix incurs prohibitive $\mathcal{O}(N^2D)$ computational and memory footprints.
Furthermore, they fail to propagate gradients to the token budget $k$, rendering them intractable for adaptive high-resolution vision tasks.

\begin{figure}[t]
    \centering
    \begin{subfigure}[b]{0.352\textwidth}
        \centering
        \includegraphics[width=\linewidth]{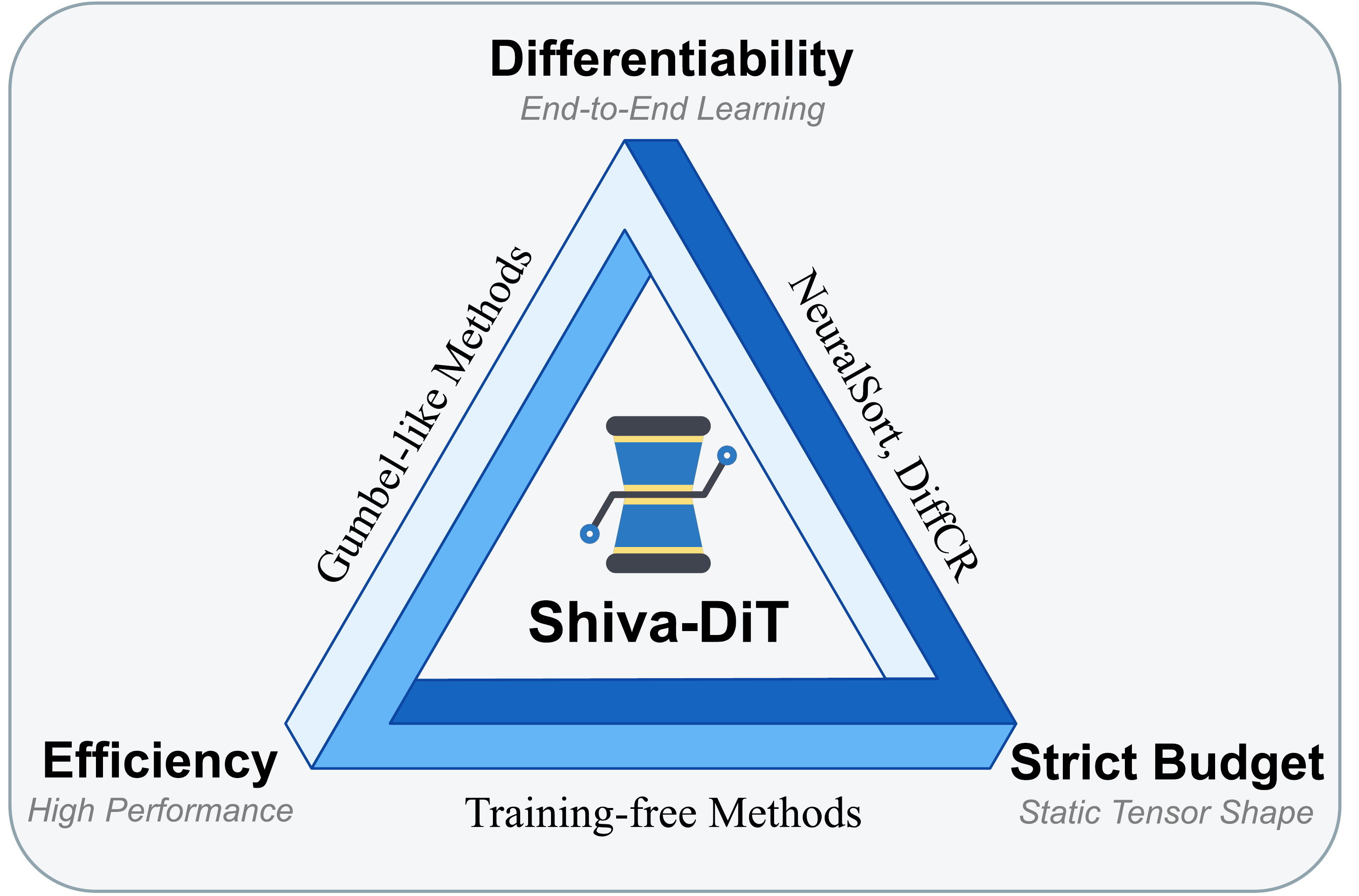} 
        \caption{The Trilemma of Sparse Learning.}
        \label{fig:trilemma}
    \end{subfigure}
    \hfill
    \begin{subfigure}[b]{0.628\textwidth}
        \centering
        \includegraphics[width=\linewidth]{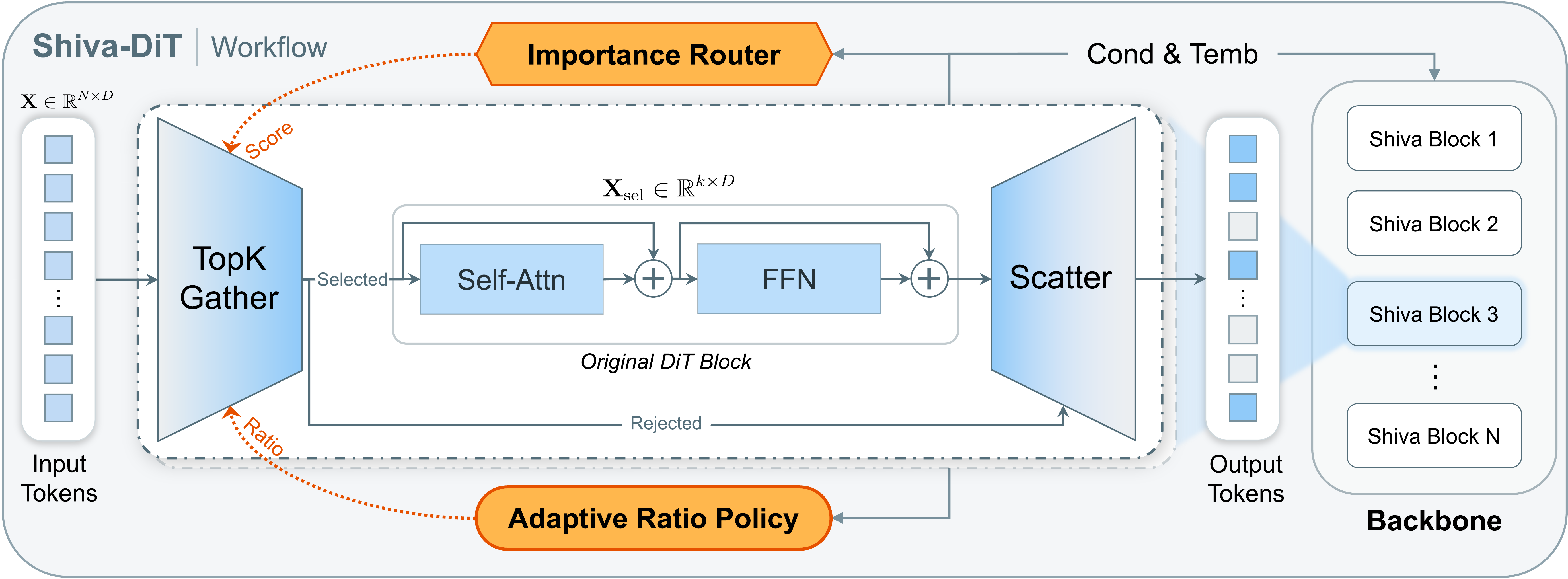}
        \caption{Shiva-DiT Workflow.}
        \label{fig:dit_workflow}
    \end{subfigure}
    \caption{\textbf{Overview of Shiva-DiT.} 
    (a) Shiva simultaneously addresses the sparse learning trilemma: Differentiability, Efficiency, and Strict Budget. 
    (b) We inject a lightweight \emph{Context-Aware Router} and \emph{Adaptive Ratio Policy} into the frozen backbone. Using residual-based differentiable Top-$k$ selection, Shiva ensures learnability while maintaining static tensor shapes for hardware efficiency.}
    \label{fig:shiva_architecture}
\end{figure}

In this paper, we introduce \emph{Shiva-DiT}, a framework that addresses this trilemma through \textit{Residual-Based Differentiable Top-$k$ Selection}. By leveraging the backbone's residual nature, we formulate a gradient estimator that models pruning as a learnable selection between the active block and the identity skip connection. This design enables the end-to-end optimization of a deterministic token budget $k$, enforcing hardware-friendly static tensor shapes. To fully exploit this, we incorporate a \emph{Context-Aware Router} using a pairwise layer-sharing strategy and an \emph{Adaptive Ratio Policy} that learns a decoupled spatiotemporal schedule, thereby automatically allocating computation to information-dense generative stages without manual heuristics.

\textbf{Our contributions are summarized as follows:}
(1) We propose \emph{Residual-Based Differentiable Top-$k$ Selection}, which jointly learns token scores and the budget $k$ using a single backbone pass while retaining deterministic hard selection in the forward pass.
(2) We introduce a \emph{Context-Aware Router} that scores tokens using prompt, timestep, and layer information, and an \emph{Adaptive Ratio Policy} that learns retention budgets across layers and denoising timesteps.
(3) Experiments on SD3-Medium, Flux.1-dev, and PixArt-$\Sigma$ evaluate quality, FLOPs, and end-to-end latency. On SD3-Medium, Shiva-DiT reaches a 1.54$\times$ wall-clock speedup while maintaining competitive generative fidelity.

\section{Related Work}
\label{sec:related_work}

\subsection{Token Reduction in Vision Transformers}
Token reduction methods for Vision Transformers (ViTs)~\citep{vit2020} use dynamic pruning, token pooling, and bipartite matching to reduce the number of tokens processed by subsequent layers~\citep{dyvit2021, tokenpooling2023, tome2022, zhang2024star, zeroprune2024, tran2024pitome}. Token Fusion~\citep{tofu2023} combines pruning and merging, selecting operations based on functional linearity and using MLERP to preserve feature norms. These methods were primarily developed for classification. Applying them to DiTs requires preserving the spatial information needed for dense image generation~\citep{dit2023}.

\subsection{Heuristic and Training-Free Acceleration}
To adapt reduction techniques to diffusion, ToMeSD~\citep{tomesd2023} extended token merging to Stable Diffusion, introducing unmerge operations to preserve spatial consistency. To better preserve semantic details, subsequent works incorporated advanced saliency metrics. AT-EDM~\citep{atedm2024} evaluates token importance using attention maps, while IBTM~\citep{ibtm2025} leverages classifier-free guidance (CFG) scores to protect high-information tokens. Other approaches explore spectral or optimization perspectives: FreqTS~\citep{freqts2025} prioritizes high-frequency components encoding structural details, and ToMA~\citep{toma2025} reformulates token selection as a submodular optimization problem. Leveraging the temporal dynamics, SDTM~\citep{sdtm2025} utilizes the ``structure-then-detail" denoising prior, applying a hand-crafted schedule to merge tokens aggressively in early structural stages. While effective as post-training optimizations, such heuristic policies are inherently static and may not optimally adapt to complex redundancy patterns. CoReDiT~\citep{Li_2026_CVPR} uses spatial coherence to prune tokens and reconstruct skipped attention outputs.

\paragraph{Kernel Incompatibility.}
A practical bottleneck for methods relying on explicit attention maps (e.g., ranking via $QK^T$~\citep{atedm2024, zeroprune2024, sdtm2025}) is their conflict with hardware-efficient attention algorithms. State-of-the-art implementations like FlashAttention~\citep{dao2022flashattention} employ tiling-based parallel algorithms that compute attention scores block-by-block in fast on-chip SRAM. Crucially, these algorithms \emph{never materialize} the complete $N \times N$ attention map in High-Bandwidth Memory (HBM). Pruning algorithms that mandate access to the global attention map inevitably force a fallback to quadratic-memory implementations, thereby negating the scalability and speed benefits of modern attention kernels.

\subsection{Temporal Feature Caching}
Accelerating inference by exploiting feature correlation between adjacent timesteps has also gained traction. DiTFastAttn~\citep{ditfastattn2024} identifies redundancies across spatial and temporal dimensions, proposing cross-timestep attention sharing. ToCa~\citep{toca2024} and DaTo~\citep{dato2024} introduce token-wise caching based on dynamics, updating only critical tokens while reusing cached features for others. Similarly, CA-ToMe~\citep{catome2025} and AsymRnR~\citep{asymrnr2024} extend merging results across steps. While effective for standard sampling regimes, these strategies fundamentally rely on the assumption of \textit{temporal continuity}. This dependency may limit their applicability in accelerated inference scenarios, such as few-step distillation, where the feature distribution evolves rapidly between discrete timesteps. By avoiding reliance on historical feature alignment, our approach maintains a generalized design suitable for diverse sampling schedules. DiffSparse~\citep{zhu2026diffsparse} learns layer-wise token sparsity for feature caching, while SODA~\citep{Shao_2026_CVPR} jointly schedules caching and pruning using offline sensitivity estimates.

\subsection{Learning-Based Dynamic Architectures}
Recent work makes dynamic pruning learnable, with different budget and execution trade-offs. Mixture-of-Depths (MoD)~\citep{raposo2024mixtureofdepths} uses fixed-capacity top-$k$ routing: selected tokens execute a transformer block, while the others bypass that block through the residual path. DyDiT~\citep{dydit2024} uses threshold-based masking, so the realized token count varies with the input. SparseDiT~\citep{sparsedit2025} fine-tunes manually specified sparse-dense blocks with a prescribed timestep schedule, which reduces computation during both training and inference but does not learn the token budget end to end. DiffRate~\citep{diffrate2023} interpolates between discrete ratio bins, and DiffCR~\citep{diffcr2025} adapts this idea to diffusion using two backbone paths. General-purpose differentiable sorting and ranking methods include NeuralSort~\citep{neuralsort2019}, SoftSort~\citep{prillo2020softsort}, and the $O(N\log N)$ operator of Blondel et al.~\citep{blondel2020fastsort}. NeuralSort constructs an $N\times N$ relaxation, while SOFT top-$k$~\citep{softtopk2020} uses iterative solves. Shiva-DiT instead executes one hard top-$k$ backbone path and uses differentiable surrogates for token scores and the budget. In our unified benchmark, its training time is $1.07\times$ that of Vanilla, compared with $1.65\times$ for DiffCR; Appendix~\ref{app:training_efficiency} reports the full protocol.

\section{Method}
\label{sec:method}

\subsection{Preliminaries}
\label{subsec:preliminaries}

\paragraph{Unified Perspective on Generative Models.}
We unify Denoising Diffusion Probabilistic Models (DDPMs)~\citep{ddpm2020} and Flow Matching frameworks~\citep{flowmatching2023} under a common objective: learning a time-dependent backbone $F_\theta(\mathbf{x}_t, t, c)$ to minimize the regression error
\begin{equation}
    \mathcal{L} = \mathbb{E}_{t, \mathbf{x}_0, \mathbf{x}_1} \left[ \| F_\theta(\mathbf{x}_t, t, c) - \mathbf{y}_t \|^2 \right].
\end{equation}
Here, the regression target $\mathbf{y}_t$ varies by architecture, representing either the noise $\epsilon$ in legacy models (e.g., SD1.5~\citep{sd1.5_2022}, SDXL~\citep{sdxl2023}) or the velocity field $v_t$ in modern flow matching approaches (e.g., SD3-Medium~\citep{sd32024}, Flux~\citep{flux2024}). Since \emph{Shiva-DiT} optimizes the internal token processing efficiency of $F_\theta$, it is fundamentally agnostic to the specific prediction target (noise vs. velocity) and compatible with off-the-shelf samplers (e.g., DDIM~\citep{ddim2022}, Euler~\citep{edm2022} and DPM~\citep{dpm2022}).

\paragraph{Backbone Architecture.}
We focus on models where the backbone $F_\theta$ is instantiated as a Diffusion Transformer (DiT)~\citep{dit2023}. The input latent is flattened into a sequence of $N$ spatial tokens $\mathbf{X} \in \mathbb{R}^{N \times D}$, where $D$ is the hidden dimension. These tokens are processed by $L$ transformer layers. A generic block at layer $\ell$ updates the sequence via:
\begin{equation}
    \begin{aligned}
        \mathbf{Y}_\ell &= \mathbf{X}_\ell + \text{Attention}(\text{Norm}(\mathbf{X}_\ell, c, t)), \\
        \mathbf{X}_{\ell+1} &= \mathbf{Y}_\ell + \text{FFN}(\text{Norm}(\mathbf{Y}_\ell, c, t)).
    \end{aligned}
\end{equation}
\paragraph{Problem Formulation.}
Regardless of the prediction target, the computational bottleneck lies in the sequence length $N$, as Attention~\citep{attn2017} scales with $\mathcal{O}(N^2)$ and FFN with $\mathcal{O}(N)$. Our goal is to introduce a differentiable selection mechanism within these blocks to dynamically identify a sparse subset $\mathbf{X}_{\text{sel}} \in \mathbb{R}^{k \times D}$ ($k \ll N$) for computation, thereby accelerating the entire iterative generation process.

\subsection{Residual-Based Differentiable Top-\texorpdfstring{$k$}{k} Selection}
\label{subsec:differentiable_topk}

We propose \textit{Residual-Based Differentiable Top-$k$ Selection}, which executes hard selection in the forward pass and propagates surrogate gradients to token scores and the budget $k$ using a single backbone pass.

\paragraph{Forward Pass: Hard Selection.}
The forward pass uses hard top-$k$ selection to enforce the token budget. Let $\mathbf{X} \in \mathbb{R}^{N \times D}$ denote the input tokens and $\mathbf{s} \in \mathbb{R}^N$ be the corresponding importance scores predicted by the Router. Given a target budget $k$, we identify the index set of the top-$k$ scores:
\begin{equation}
    \mathcal{I}_{\text{topk}} = \operatorname{argtopk}(\mathbf{s}, k).
\end{equation}
The token set is physically partitioned into selected ($\mathbf{X}_{\text{sel}}$) and rejected ($\mathbf{X}_{\text{rej}}$) subsets via a gather operation:
\begin{equation}
    \mathbf{X}_{\text{sel}} = \{ \mathbf{x}_i \mid i \in \mathcal{I}_{\text{topk}} \}, \quad \mathbf{X}_{\text{rej}} = \mathbf{X} \setminus \mathbf{X}_{\text{sel}}.
\end{equation}
Within each Transformer block, only $\mathbf{X}_{\text{sel}}$ is processed by Attention and the FFN. The outputs are then scattered back to the full token layout before the next block, reducing computation without permanently shortening the sequence.

\paragraph{Backward Pass: Gradient Estimation.}
Since the discrete indicator function $\mathbb{I}(i \in \mathcal{I}_{\text{topk}})$ has zero gradients, we construct a differentiable surrogate graph. We inject Gaussian noise $\epsilon \sim \mathcal{N}(0, \sigma_{\mathrm{noise}}^2)$ into the scores ($\tilde{\mathbf{s}} = \mathbf{s} + \epsilon$) and compute a \textit{descending soft rank} $\tilde{r}_i$ via pairwise sigmoid comparisons inspired by RankNet~\citep{softrank2005} (detailed in Algorithm~\ref{alg:softrank}):
\begin{equation}
    \tilde{r}_i(\tilde{\mathbf{s}}) = 1 + \sum_{j \neq i} \sigma\left(\frac{\tilde{s}_j - \tilde{s}_i}{\tau_{\text{rank}}}\right),
    \label{eq:rank}
\end{equation}
where $\tau_{\text{rank}}$ is the temperature. During training, Eq.~\ref{eq:rank} entails $O(N^2)$ comparisons, but the operation is restricted to the scalar scores $\tilde{\mathbf{s}} \in \mathbb{R}^{N \times 1}$. This avoids the prohibitive $O(N^2 D)$ cost of feature level soft sorting methods such as NeuralSort~\citep{neuralsort2019}.

To enable gradient flow to the learnable budget $k$, we formulate the inclusion score $\pi_i$ as a continuous relaxation~\citep{gumbel2017} of the binary selection indicator:
\begin{equation}
    \pi_i(\tilde{\mathbf{s}}, k) = \sigma\left( \frac{k - \tilde{r}_i(\tilde{\mathbf{s}})}{\tau_{\text{sel}}} \right),
    \label{eq:prob_select}
\end{equation}
where $\tau_{\text{sel}}$ is a temperature parameter.
This creates a fully differentiable path: $\frac{\partial \pi_i}{\partial \mathbf{s}}$ and $\frac{\partial \pi_i}{\partial k}$ are non-zero. Note that in this relaxed view, $k$ acts as a continuous threshold against the soft ranks.

\paragraph{Residual-Based Gradient Estimator.}
To propagate gradients through the non-differentiable bottleneck, we employ a \textit{Residual-Based Straight-Through Estimator} (STE). Since the deterministic hard selection restricts observation to a single path per token, we define the gradient of the selection probability $\pi_i$ based on the feature sensitivity of the actually executed path. A detailed discussion on this estimator and its gradient formulation is provided in Appendix~\ref{app:grad_derivation}.
\begin{equation}
    \frac{\partial \mathcal{L}}{\partial \pi_i} \approx 
    \begin{cases}
        \left\langle \nabla_{\mathbf{x}_i}^{\text{sel}} \mathcal{L}, \; \mathbf{x}_i \right\rangle & \text{if } i \in \mathcal{I}_{\text{topk}} \\
        -\left\langle \nabla_{\mathbf{x}_i}^{\text{rej}} \mathcal{L}, \; \mathbf{x}_i \right\rangle & \text{if } i \notin \mathcal{I}_{\text{topk}}
    \end{cases},
    \label{eq:grad_injection_main}
\end{equation}
where $\nabla^{\text{sel}}$ and $\nabla^{\text{rej}}$ denote the gradients of the loss $\mathcal{L}$ with respect to the effective input features under the selected and rejected paths, respectively.
In practice, this piecewise formulation is implemented with vectorized tensor operations. The negative sign expresses rejected-path sensitivity in the same inclusion coordinate as selected-path sensitivity.

\begin{algorithm}[htbp]
   \caption{Differentiable Soft Rank Calculation}
   \label{alg:softrank}
   \begin{algorithmic}[1]
      \State \textbf{Input:} Perturbed scores $\tilde{\mathbf{s}} \in \mathbb{R}^{N}$, temperature $\tau_{\text{rank}}$
      \State \textbf{Output:} Soft ranks $\tilde{\mathbf{r}} \in \mathbb{R}^{N}$
      \State $D_{ij} \gets (\tilde{s}_j - \tilde{s}_i) / \tau_{\text{rank}}$, for $i,j \in \{1,\ldots,N\}$
      \State $P_{ij} \gets \sigma(D_{ij}) \cdot (1 - \mathbb{I}[i=j])$
      \Comment{Exclude self-comparison}
      \State $\tilde{r}_i \gets 1 + \sum_{j=1}^{N} P_{ij}$
      \State \Return $\tilde{\mathbf{r}}$
   \end{algorithmic}
\end{algorithm}

\paragraph{Comparison with Existing Paradigms.}
Our mechanism addresses two critical limitations of prior work.
\emph{(1) Budget Controllability:} Gumbel-based methods like DyDiT~\citep{dydit2024} rely on prediction-then-threshold masking, suffering from budget instability, since the realized token count is only constrained in expectation. Our explicit top-$k$ selection guarantees exact budget adherence and regularly shaped tensors.
\emph{(2) Training Cost with Learnable Budgets:} Interpolation-based methods like DiffCR~\citep{diffcr2025} require two backbone passes to obtain gradients for the budget. Our residual-based estimator uses the executed path, while the soft inclusion function in Eq.~\ref{eq:prob_select} propagates gradients to both token scores and $k$, requiring only one backbone pass.

Table~\ref{tab:topk_comparison} makes the resulting distinction explicit. Among the compared mechanisms, Shiva-DiT is the only one that combines a deterministic token count, a continuously learnable budget, and a single backbone pass. At $N=4096$, its selector requires 0.264~ms forward, 3.384~ms backward, and 80.6~MB of saved activations, compared with 7.622~ms, 15.135~ms, and 580.8~MB for NeuralSort-STE; Appendix~\ref{sec:topk_comparison} gives the full scaling study and method-by-method discussion.

\begin{table}[t]
\centering
\small
\setlength{\tabcolsep}{4pt}
\caption{Comparison of differentiable selection mechanisms. ``Static count'' means that each forward pass executes a deterministic token count, and ``passes'' counts backbone evaluations per training step.}
\begin{tabular}{lccccc}
\toprule
 & NeuralSort & SOFT Top-$k$ & DyDiT & DiffCR & \textbf{Shiva-DiT} \\
\midrule
Forward selection  & Soft permutation & Soft OT plan & Binary mask & Hard top-$k$ & Hard top-$k$ \\
Learnable $k$      & \xmark & \xmark & \xmark & \cmark~(interpolation) & \cmark~(continuous) \\
Router gradient    & Permutation & Sinkhorn & Gumbel-STE & Post-multiply & Residual STE \\
Backbone passes    & 1 & 1 & 1 & 2 & 1 \\
Forward memory     & $O(N^2)$ & $O(N)$ & $O(N)$ & $O(N)$ & $O(N)$ \\
Static count       & \cmark & \cmark & \xmark & \cmark & \cmark \\
\bottomrule
\end{tabular}
\label{tab:topk_comparison}
\end{table}

\subsection{Context-Aware Token Importance Router}
\label{subsec:router}

Token importance in diffusion is inherently contextual, depending on the denoising stage, semantic alignment, and network abstraction level. Unlike prior heuristics based on local metrics (e.g., L2 norms, attention scores), we propose \emph{ShivaRouter}, a lightweight context-conditioned scoring network that captures these multi-dimensional dependencies to predict token informativeness.

\paragraph{Architecture.}
To efficiently capture multi-dimensional dependencies, the router predicts a scalar importance score $s_i$ by conditioning the local token feature $\mathbf{x}_i \in \mathbb{R}^{D}$ on three global signals: global semantics $\mathbf{c}$ and temporal embedding $t$ (both reused from the pre-trained backbone), alongside a learnable layer identity embedding $l$. We project these inputs into a low-dimensional bottleneck space ($d'=64$) and apply an additive fusion to minimize overhead:
\begin{equation}
    \mathbf{h}_{\text{ctx}} = \text{Proj}_t(t) + \text{Proj}_p(\mathbf{c}) + \text{Proj}_l(l),
\end{equation}
\begin{equation}
    s_i = \mathbf{w}^\text{T} \cdot \text{LayerNorm}\left( \operatorname{SiLU}( \text{Proj}_x(\mathbf{x}_i) + \mathbf{h}_{\text{ctx}} ) \right),
\end{equation}
where $\text{Proj}_*$ denote lightweight linear projections. This conditional shift modulates token representations based on context, with the layer embedding $l$ enabling the router to distinguish nuances between layers within a shared group.

\paragraph{Locally Shared Parameters.}
We evaluate router sharing strategies across network depths. Empirically, \emph{Pairwise Sharing} (adjacent layer pairs share one router) achieves the optimal trade-off, capturing local feature consistency between consecutive blocks while reducing parameter overhead, outperforming both layer-specific and globally-shared alternatives.

\subsection{Adaptive Ratio Policy}
\label{subsec:policy}

Determining optimal pruning ratios across layers and timesteps is challenging. Prior methods often rely on static heuristics~\citep{sparsedit2025} or discrete timestep buckets~\citep{diffcr2025}, failing to capture the fine-grained dynamics of the diffusion process. To address this, we introduce a \emph{Ratio Policy Network}, a lightweight controller that predicts a continuous retention ratio $r_{t,l} \in (0, 1]$ from the timestep and layer index. This ratio determines the discrete budget $k = \lfloor N \cdot r_{t,l} \rfloor$ (as discussed in Sec.~\ref{subsec:differentiable_topk}), enabling precise and continuous control over hardware-constrained token counts throughout the generation.

\paragraph{Continuous Decoupled Modulation.}
We hypothesize that the optimal sparsity profile can be decomposed into a superposition of a static spatial prior and a global temporal trend. This separability is not a new assumption: SparseDiT~\citep{sparsedit2025} arrives at its hand-crafted schedule by designing a depth profile and a temporal ramp as independent axes, and we make the same structure learnable rather than fixed.
Unlike multiplicative mechanisms such as FiLM~\citep{film2018}, which inherently assume that temporal dynamics exert complex, layer-specific scaling effects (i.e., different timesteps modulate different layers non-uniformly), we argue that the redundancy trend across timesteps is largely consistent across network depths.
Therefore, we adopt a \textit{Decoupled Additive} architecture to explicitly treat spatial and temporal redundancies as independent, additive factors. Appendix~\ref{app:more_ablation} reports a second, practical reason for this choice: the multiplicative form admits a continuous scale symmetry that leaves its factors unidentified, and a trained FiLM policy is found sitting on that flat direction.

Let $\mathbf{e}_\ell \in \mathbb{R}^{d}$ be a learnable embedding for layer $\ell$, and $\mathbf{e}_t = \text{MLP}(t)$ be the projected timestep embedding. The policy predicts the logit of the retention ratio via:
\begin{equation}
    \text{logit}(r_{t,\ell}) = \Phi_{\text{time}}(\mathbf{e}_t) + \Phi_{\text{layer}}(\mathbf{e}_\ell) + b_{\text{anchor}},
\end{equation}
\begin{equation}
    r_{t,\ell} = \sigma(\text{logit}(r_{t,\ell})),
\end{equation}
where $\Phi$ denotes lightweight MLPs and $b_{\text{anchor}}$ is a learnable scalar initialized to $\sigma^{-1}(R_{\text{target}})$.
Crucially, since the policy network $\Phi$ depends solely on the timestep $t$ and layer index $\ell$, and is independent of dynamic input tokens or text prompts, the predicted ratios are deterministic for a given checkpoint.
Consequently, during inference, the entire policy network is compiled into a lightweight \emph{Look-Up Table} (LUT), introducing negligible computational overhead to the generation process.

\paragraph{Stabilized Budget Constraint.}
Optimizing a dynamic policy under a global budget $R_{\text{target}}$ requires balancing stability and flexibility. DyDiT~\citep{dydit2024} produces unpredictable costs via thresholding, while DiffCR~\citep{diffcr2025} penalizes mini-batch deviations ($\|\bar{r}_{\text{batch}} - R_{\text{target}}\|^2$). This rigid constraint suits static policies but is ill-suited for time-dependent schedules, as it prevents allocating varying budgets across diffusion timesteps.

To support time-dependent pruning, we introduce an \emph{Exponential Moving Average (EMA)} $\mu_{\text{global}}$ to decouple the budget constraint from the sparse mini-batch sampling. At iteration $u$, it tracks the long-term retention ratio via:
\begin{equation}
    \mu_{\text{global}}^{(u)} = \beta \cdot \bar{r}_{\text{batch}}^{(u)} + (1 - \beta) \cdot \mu_{\text{global}}^{(u-1)},
\label{eq:ema_avg}
\end{equation}
where $\bar{r}_{\text{batch}}$ is the globally synchronized batch mean. We then formulate an EMA-guided proxy budget loss:
\begin{equation}
    \mathcal{L}_{\text{budget}} = \bar{r}_{\text{batch}} \cdot \text{sg}\left[ 2 \cdot (\mu_{\text{global}} - R_{\text{target}}) \right],
\label{eq:budget_loss}
\end{equation}
where $\text{sg}[\cdot]$ denotes the stop-gradient operator and $\mu_{\text{global}}$ is the updated EMA estimate. The resulting gradient steers the long-run average retention toward $R_{\text{target}}$ without forcing every mini-batch to match it, while hard top-$k$ selection still enforces the predicted token count exactly at each layer and timestep. The stability of this mechanism is further ensured by \emph{Stratified Timestep Sampling} (Sec.~\ref{subsec:training}), which minimizes the variance of $\bar{r}_{\text{batch}}$ and provides a stable signal for the EMA.

\subsection{Training Objectives and Strategy}
\label{subsec:training}

To effectively optimize the router's discrete decisions and the policy's continuous constraints alongside the backbone, we employ a composite objective function and a multi-stage training process.

\paragraph{Composite Loss with Distillation.}
The total objective integrates diffusion reconstruction $\mathcal{L}_{\text{diff}}$, budget constraint $\mathcal{L}_{\text{budget}}$, and distillation terms:
\begin{equation}
    \mathcal{L}_{\text{total}} = \mathcal{L}_{\text{diff}} + \lambda_b \mathcal{L}_{\text{budget}} + \lambda_d \mathcal{L}_{\text{distill}}.
\end{equation}
To ensure semantic consistency, we incorporate multi-scale distillation from the frozen teacher model. Specifically, we apply feature alignment losses every 4 transformer blocks~\citep{dydit2024} alongside a final output distillation loss, defined as Normalized Feature Distillation (see Eq.~\ref{eq:distill_cal} in Appendix~\ref{app:training_details}), providing dense supervision to guide the router's optimization.

\paragraph{Staged Training Protocol.}
Directly optimizing all components from a cold start can lead to instability. We therefore adopt a progressive three-stage strategy, consisting of \textit{Router Warmup}, \textit{Policy Learning}, and \textit{Joint Tuning}, to sequentially establish ranking capability, optimize the budget schedule, and recover generation fidelity. Detailed configurations for each stage are provided in Appendix~\ref{app:training_details}.

\paragraph{Stratified Sampling Strategy.}
To stabilize the joint optimization of the modules, we introduce a \emph{Stratified Sampling} mechanism to rectify the high variance of standard uniform sampling.

\textit{1) Stratified Timestep Sampling.}
Standard sampling $t \sim \mathcal{U}(0, T)$ often leads to temporal clustering within a mini-batch, destabilizing the budget controller. We instead enforce uniform coverage by partitioning the domain $[0, T]$ into $B$ equi-width intervals for a global batch size $B$, drawing one sample $t_i$ from each:
\begin{equation}
    t_i \sim \mathcal{U}\left( \frac{i \cdot T}{B}, \frac{(i+1) \cdot T}{B} \right), \quad i \in \{0, \dots, B-1\}.
\end{equation}
As detailed in Appendix~\ref{subsec:variance_analysis}, this ensures $\bar{r}_{\text{batch}}$ acts as a low-variance estimator of the global expectation, providing a stable gradient signal for optimization.

\textit{2) Stratified Ratio Sampling.}
During router warmup, we partition each batch into $K$ groups with target ratios $r_k$ sampled from disjoint intervals over $[r_{\min}, r_{\max}]$. This prevents overfitting to fixed-ratio decision boundaries, ensuring compatibility with the subsequent dynamic schedule.

\raggedbottom
\section{Experiments}
\subsection{Main Results}
\label{sec:main_exp}

\paragraph{Experimental Settings.}
We evaluate Shiva-DiT on SD3-Medium~\citep{sd32024}, Flux.1-dev~\citep{flux2024}, and PixArt-$\Sigma$~\citep{pixartsigma2024} using PEFT~\citep{peft} on the MJHQ-30K~\citep{mjhq30k2024} dataset. We additionally evaluate transfer without further training on DrawBench~\citep{drawbench} and PartiPrompts~\citep{pali}, and at $512\times512$ and $1024\times768$ resolutions. Training uses 8$\times$ NVIDIA H200 GPUs for 30 to 50 hours. Inference is measured on an NVIDIA RTX 4090. The main latency benchmark times end-to-end 50-step generation in eager mode and includes routing, sorting, gather, and scatter operations. We compare training-free methods, including ToMeSD~\citep{tomesd2023}, ToFu~\citep{tofu2023}, SDTM~\citep{sdtm2025}, ToMA~\citep{toma2025}, and IBTM~\citep{ibtm2025}, with training-based methods, including NeuralSort~\citep{neuralsort2019}, SparseDiT~\citep{sparsedit2025}, DiffCR~\citep{diffcr2025}, and DyDiT~\citep{dydit2024}. All methods are evaluated in our unified implementation using their official objectives and hyperparameters. Appendix~\ref{app:baseline_configs} provides configurations, and Appendix~\ref{app:overhead_anatomy} reports the pruning overhead separately.

\begin{table}[htbp]
    \centering
    
    \small
    \setlength{\tabcolsep}{4pt}
    
    \setlength{\aboverulesep}{0pt}
    \setlength{\belowrulesep}{0pt}
    \renewcommand{\arraystretch}{1.2}
    \caption{Quantitative results on SD3-Medium ($1024 \times 1024$, 50 steps, RTX 4090). IR: ImageReward; IQA: CLIP-IQA. \textit{Finetuned}: dense LoRA-finetuned baseline with the same inference cost as \textit{Vanilla}; training details are provided in Appendix~\ref{app:training_details}. $^\dagger$Zero-shot budget transfer from Shiva-80\% and Shiva-60\% via LUT bias $b_\text{anchor}$, without additional training (Appendix~\ref{app:generalization}).}
    
    \begin{tabular}{l cccc S[table-format=2.2] cc S[table-format=2.1]} 
        \toprule
        \multirow{2}{*}{\textbf{Methods}} & \multicolumn{4}{c}{\textbf{Image Quality Metrics}} & \multicolumn{4}{c}{\textbf{Efficiency Metrics}} \\
        \cmidrule(lr){2-5} \cmidrule(lr){6-9}
        & FID $\downarrow$ & IR $\uparrow$ & CLIP $\uparrow$ & IQA $\uparrow$ & {FLOPs (T)} & Latency (ms) & Speedup & {$\Delta$ FLOPs (\%)} \\
        \midrule
        
        Vanilla & 16.86 & 0.9385 & 30.89 & 0.4567 & 
        \multicolumn{1}{c}{\multirow{2}{*}{11.80}} & 
        \multirow{2}{*}{7659.9} & 
        \multirow{2}{*}{1.00$\times$} & 
        \multicolumn{1}{c}{\multirow{2}{*}{0.0}} \\ 
        
        \textit{Finetuned} & \textit{13.22} & \textit{1.0179} & \textit{31.18} & \textit{0.4743} & 
        & & & \\ 
        \midrule
        
        ToMeSD~\citeyearpar{tomesd2023} & 40.22 & 0.6727 & 30.88 & 0.4632 & 8.98 & 7352.9 & 1.04$\times$ & 23.9 \\
        ToFu~\citeyearpar{tofu2023}     & 47.43 & 0.4157 & 30.36 & 0.4655 & 8.98 & 7296.7 & 1.05$\times$ & 23.9 \\
        SDTM~\citeyearpar{sdtm2025}     & 47.08 & 0.3257 & 29.96 & 0.4705 & 8.83 & 7180.6 & 1.07$\times$ & 25.2 \\
        ToMA~\citeyearpar{toma2025}     & 30.06 & 0.7033 & 30.83 & 0.4633 & 8.23 & 7406.5 & 1.03$\times$ & 30.3 \\
        IBTM~\citeyearpar{ibtm2025}     & 19.96 & 0.8610 & 30.88 & 0.4695 & 10.18 & 7526.6 & 1.02$\times$ & 13.7 \\
        NeuralSort~\citeyearpar{neuralsort2019} & 22.78 & 0.8379 & 30.78 & 0.4764 & 9.47 & 6377.3 & 1.20$\times$ & 19.7 \\
        DiffCR~\citeyearpar{diffcr2025} & 17.51 & 0.7884 & 30.81 & 0.4941 & 9.93 & 6709.8 & 1.14$\times$ & 15.8 \\
        
        DyDiT~\citeyearpar{dydit2024}   & 15.29 & 0.7844 & 30.65 & 0.4824 & 9.42 & 6553.9 & 1.17$\times$ & 20.2 \\
 
        SparseDiT~\citeyearpar{sparsedit2025}   & 25.71 & 0.7751 & 30.56 & 0.4795 & 8.95 & 6211.3 & 1.24$\times$ & 24.2 \\
        
        \midrule
        
        \rowcolor{gray!20} \textbf{Shiva-90\%}$^\dagger$ & \alignbold{12}{15} & 0.9228 & \alignunder{31}{26} & \alignunder{0}{4959} & 10.19 & 6996.2 & 1.09$\times$ & 13.6 \\
 
        \rowcolor{gray!20} \textbf{Shiva-80\%} & \alignunder{13}{83} & 0.8915 & 31.10 & \alignbold{0}{5051} & 8.92 & 6112.6 & 1.25$\times$ & 24.4 \\
 
        \rowcolor{gray!20} \textbf{Shiva-70\%}$^\dagger$ & 15.74 & \alignunder{0}{9466} & 31.16 & 0.4871 & {\phantom{0}\underline{\tablenum[table-format=1.2]{7.38}}} & \alignunder{5603}{5} & \underline{1.37$\times$} & {\underline{\tablenum[table-format=2.1]{37.5}}} \\
        
        \rowcolor{gray!20} \textbf{Shiva-60\%} & 16.42 & \alignbold{0}{9974} & \alignbold{31}{35} & 0.4952 & 
        \bfseries 6.96 & 
        \alignbold{4989}{0} & 
        \textbf{1.54$\times$} & 
        \bfseries 41.0 \\ 
        
        \bottomrule
    \end{tabular}
    \label{tab:main_results}
\end{table}

\paragraph{Main Results on SD3-Medium.}
Table~\ref{tab:main_results} and Figures~\ref{fig:effi_scatter} and~\ref{fig:qualitative_main} summarize the fidelity-latency trade-off. \textit{Finetuned} denotes the dense LoRA-finetuned baseline. Shiva-DiT provides four operating points on the quality--latency Pareto frontier among the evaluated configurations, from $1.09\times$ to $1.54\times$ wall-clock speedup. Figure~\ref{fig:effi_scatter} sweeps DiffCR, DyDiT, and IBTM; Appendix~\ref{app:baseline_configs} gives their settings and realized compute. Only Shiva-80\% and Shiva-60\% are trained directly. Shiva-90\% and Shiva-70\% are obtained without additional training by shifting the compiled LUT bias $b_\text{anchor}$ (Appendix~\ref{app:generalization}). Shiva-90\% has the lowest FID in the table, Shiva-80\% has the highest IQA, and Shiva-60\% reaches $1.54\times$ speedup with a CLIP score of 31.35. Figure~\ref{fig:qualitative_main} shows that Shiva-DiT retains facial structure, beard strands, and garment texture close to Vanilla and the dense finetuned reference, while DiffCR introduces visible distortion in this example. Both Shiva-60\% and Shiva-80\% also outperform DiffCR and DyDiT on all five targeted Qwen-Image-Bench~\citep{li2026qwenimagebench} criteria (Appendix~\ref{app:qwen_bench}).

\begin{figure}[t]
    \centering
    \begin{minipage}[t]{0.47\linewidth}
        \centering
        \vspace{0pt}
        \includegraphics[width=\linewidth]{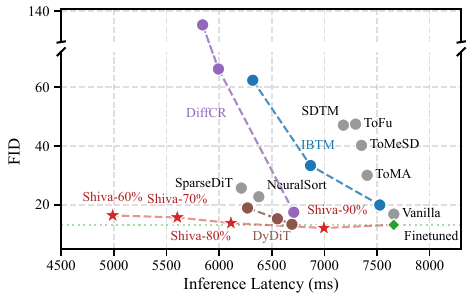}
    \end{minipage}
    \hfill
    \begin{minipage}[t]{0.51\linewidth}
        \centering
        \vspace{0pt}
        \begin{subfigure}[b]{0.235\linewidth}
            \includegraphics[width=\linewidth]{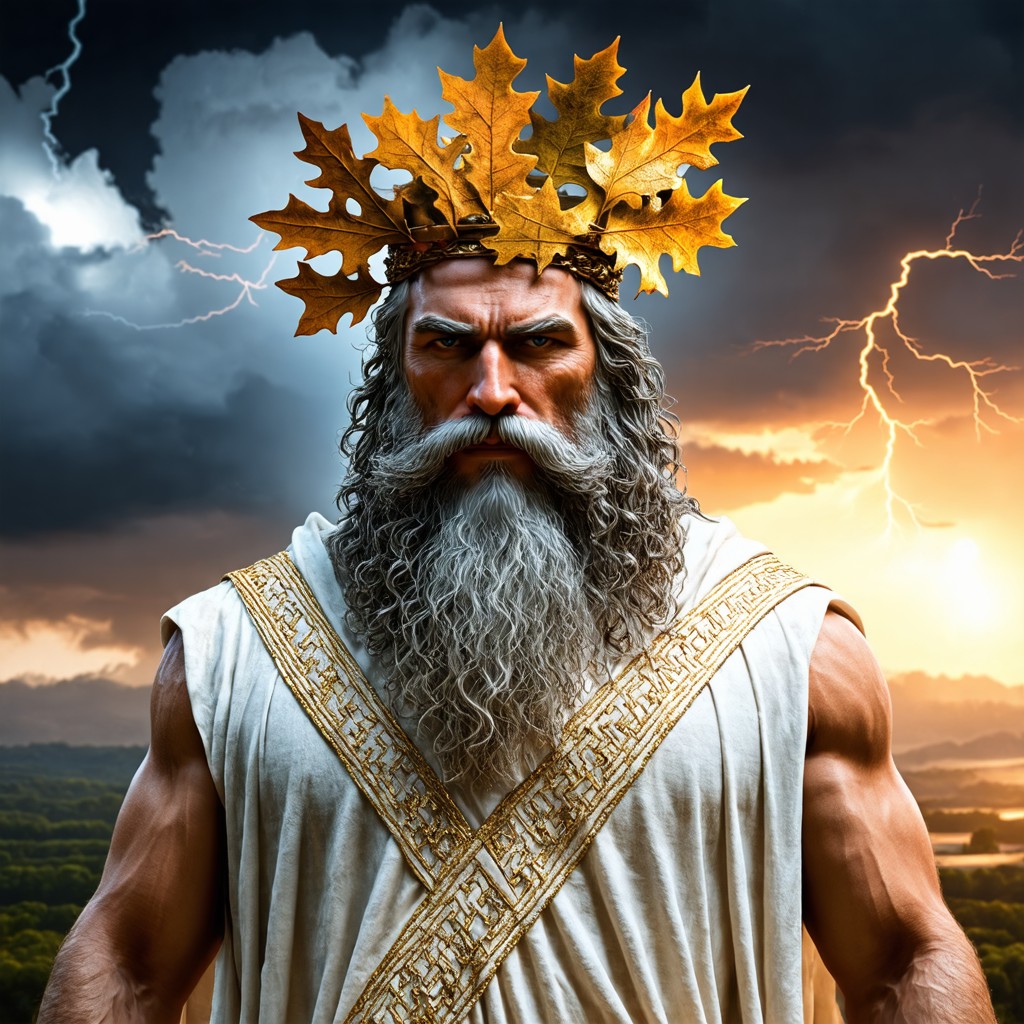}
            \caption*{Vanilla}
        \end{subfigure}
        \hfill
        \begin{subfigure}[b]{0.235\linewidth}
            \includegraphics[width=\linewidth]{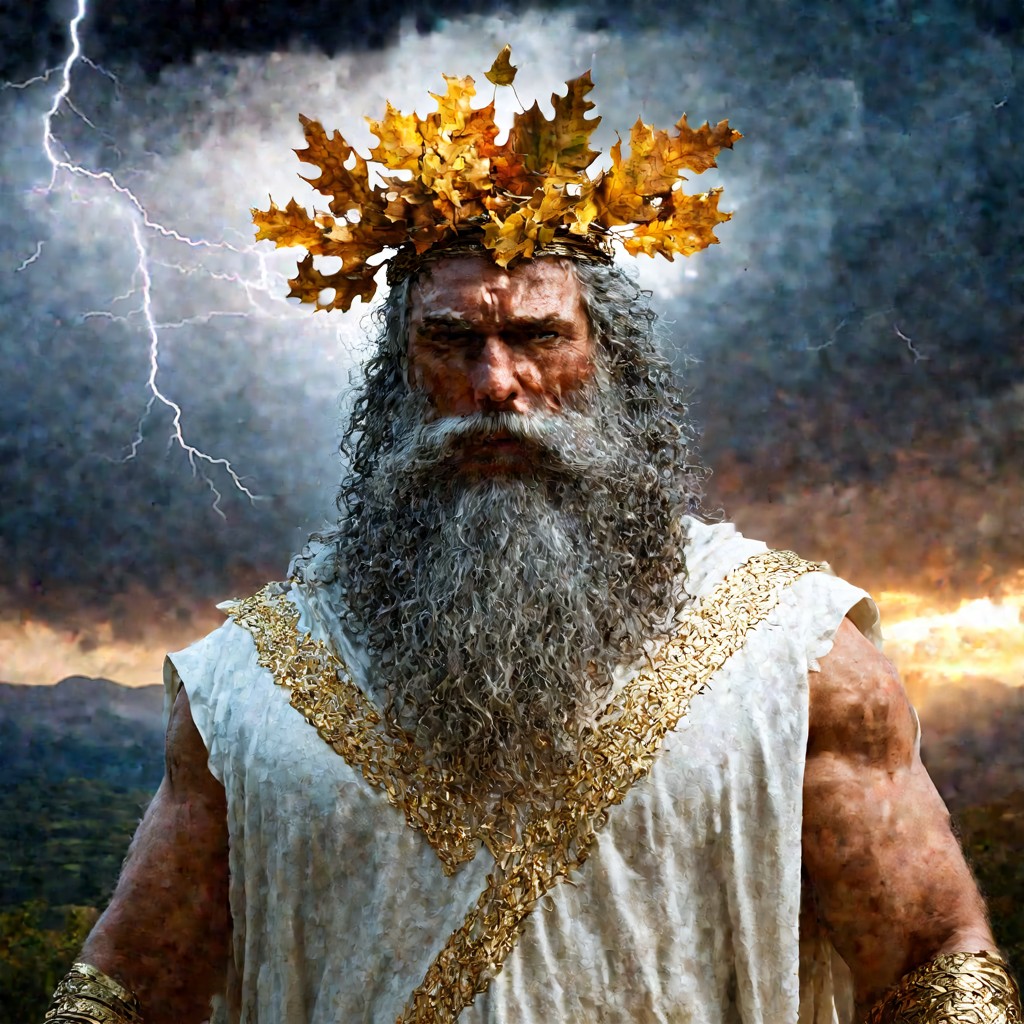}
            \caption*{ToMeSD}
        \end{subfigure}
        \hfill
        \begin{subfigure}[b]{0.235\linewidth}
            \includegraphics[width=\linewidth]{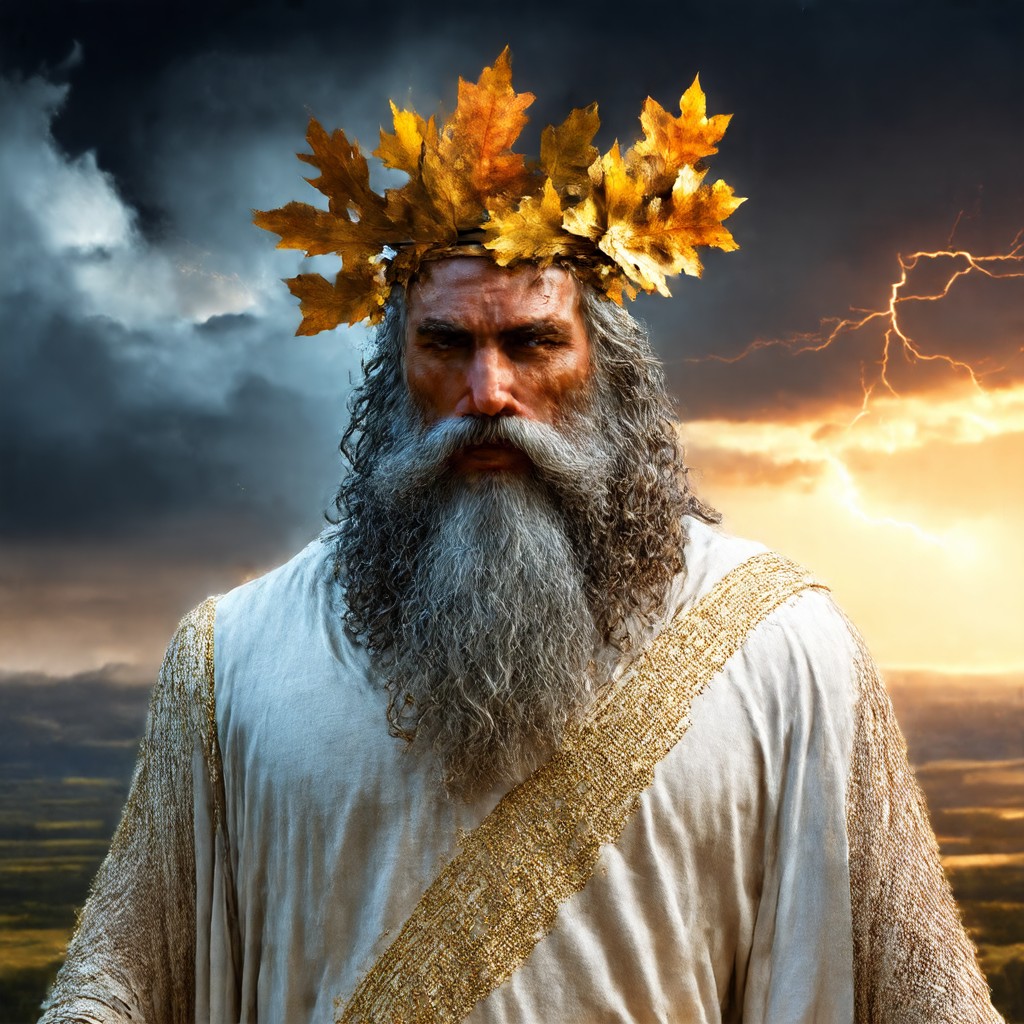}
            \caption*{IBTM}
        \end{subfigure}
        \hfill
        \begin{subfigure}[b]{0.235\linewidth}
            \includegraphics[width=\linewidth]{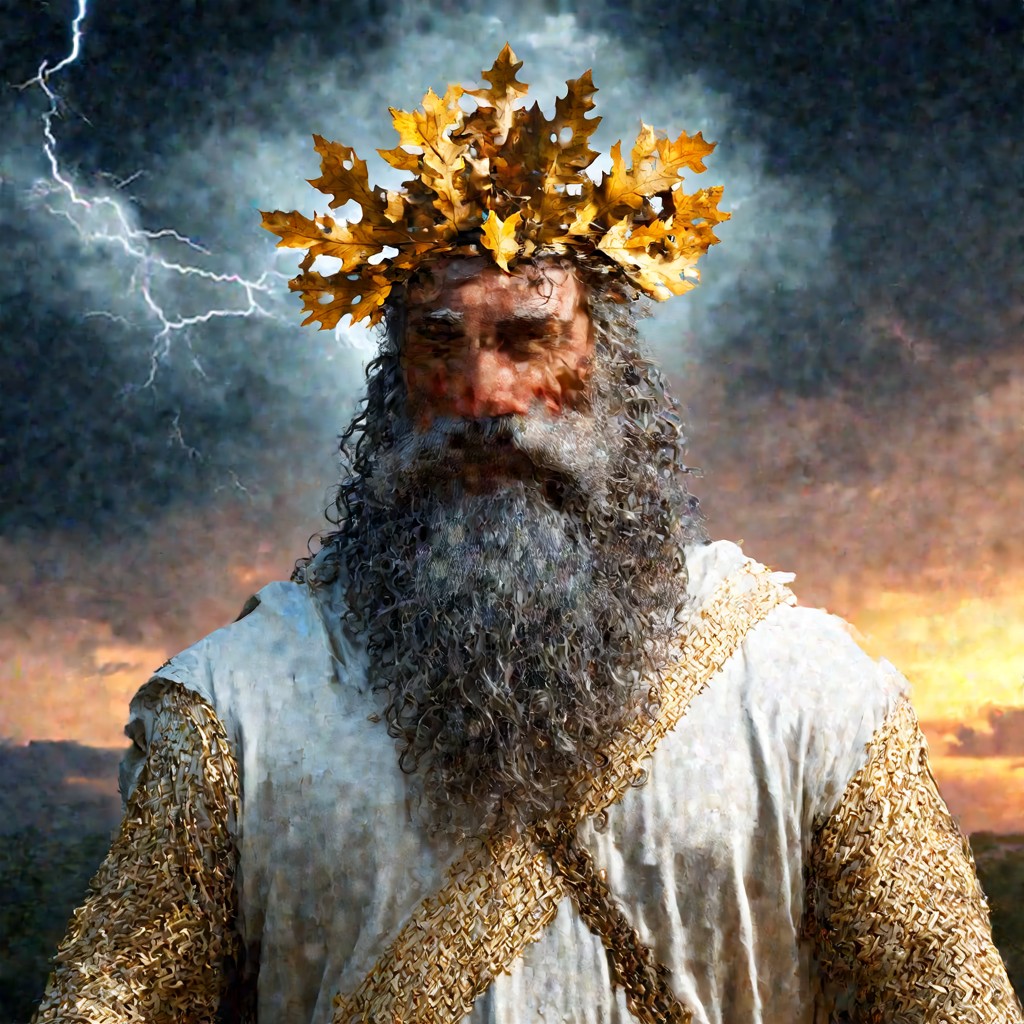}
            \caption*{ToFu}
        \end{subfigure}
        \par\vspace{0.1em}
        \begin{subfigure}[b]{0.235\linewidth}
            \includegraphics[width=\linewidth]{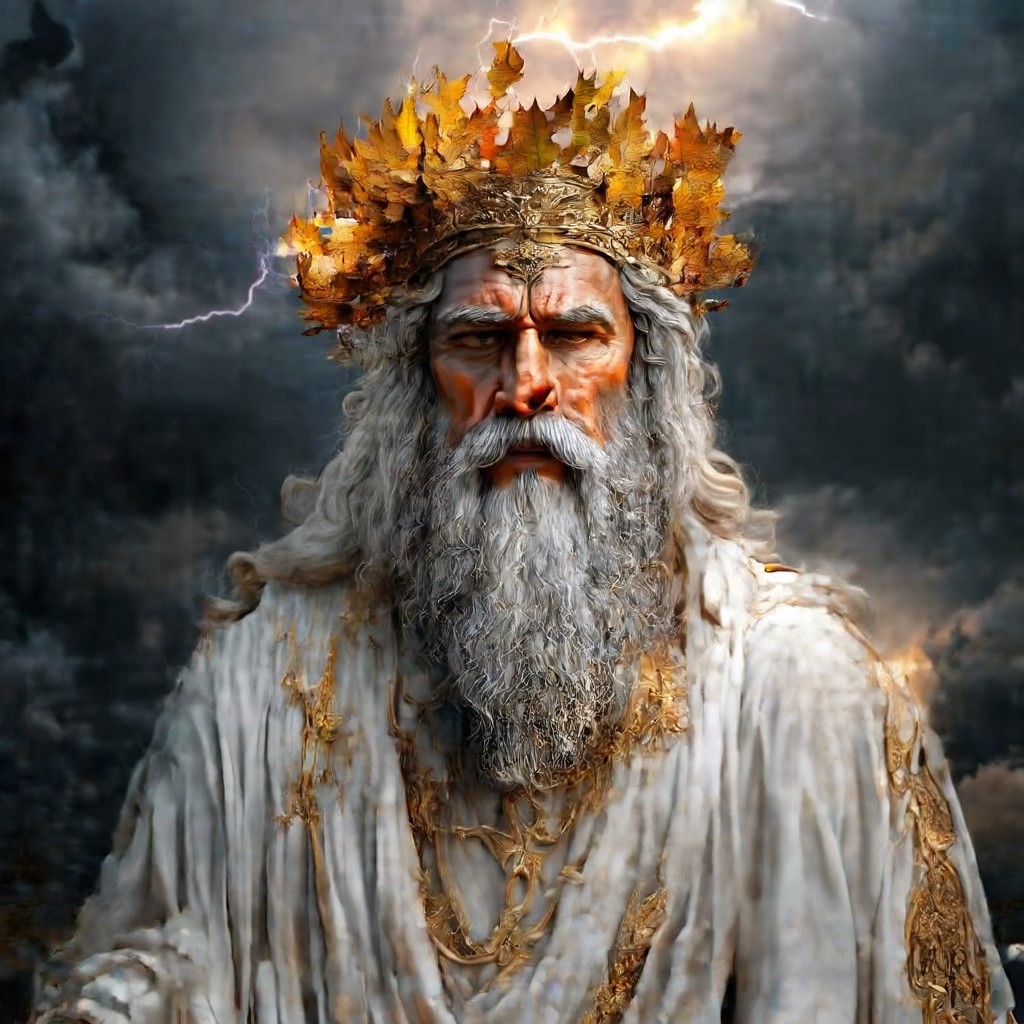}
            \caption*{DiffCR}
        \end{subfigure}
        \hfill
        \begin{subfigure}[b]{0.235\linewidth}
            \includegraphics[width=\linewidth]{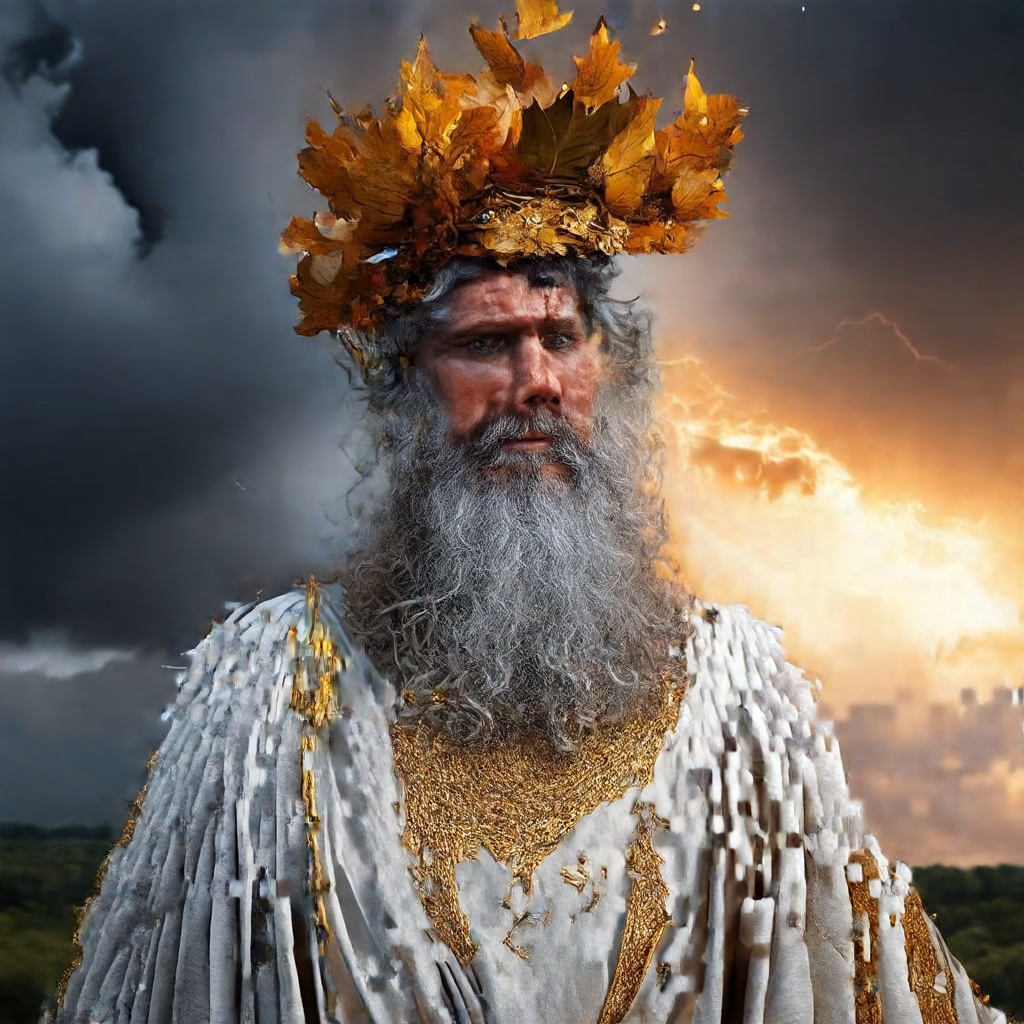}
            \caption*{NeuralSort}
        \end{subfigure}
        \hfill
        \begin{subfigure}[b]{0.235\linewidth}
            \includegraphics[width=\linewidth]{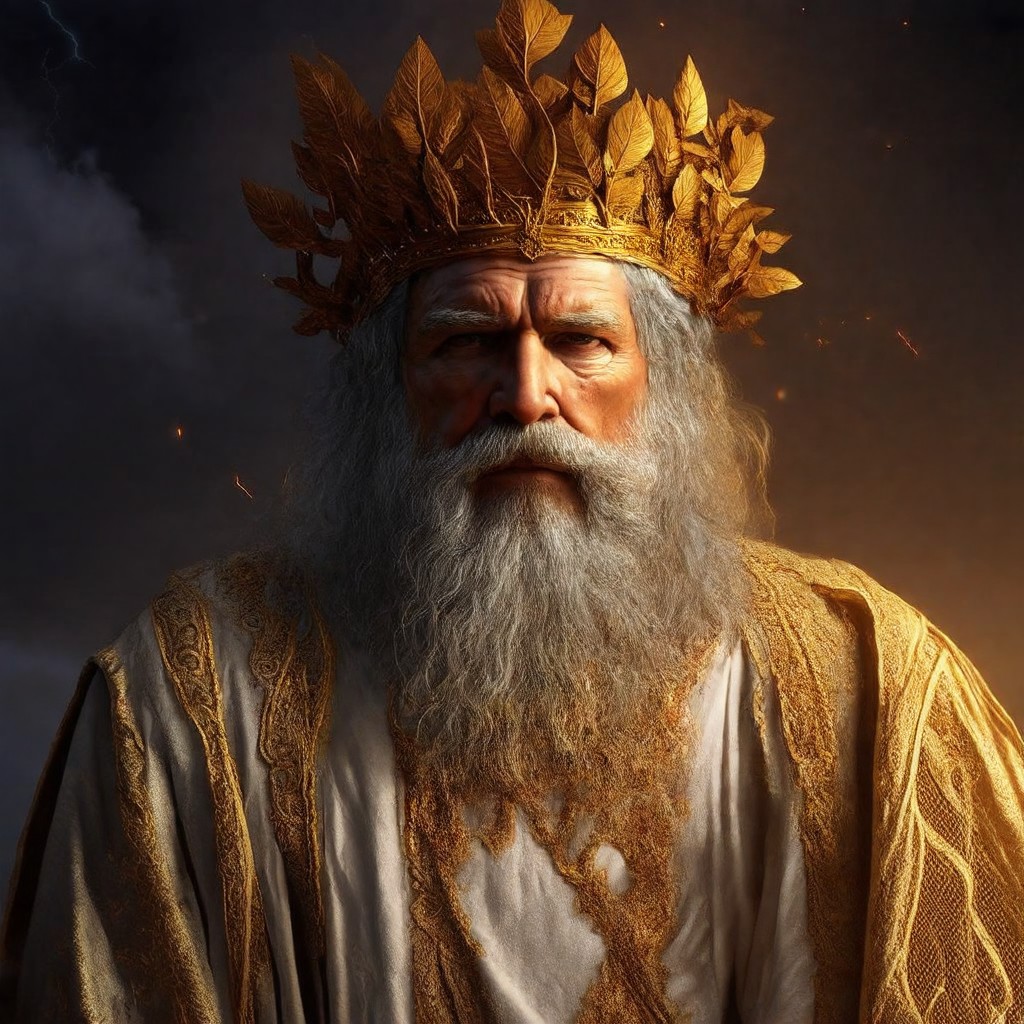}
            \caption*{\textbf{Shiva-DiT}}
        \end{subfigure}
        \hfill
        \begin{subfigure}[b]{0.235\linewidth}
            \includegraphics[width=\linewidth]{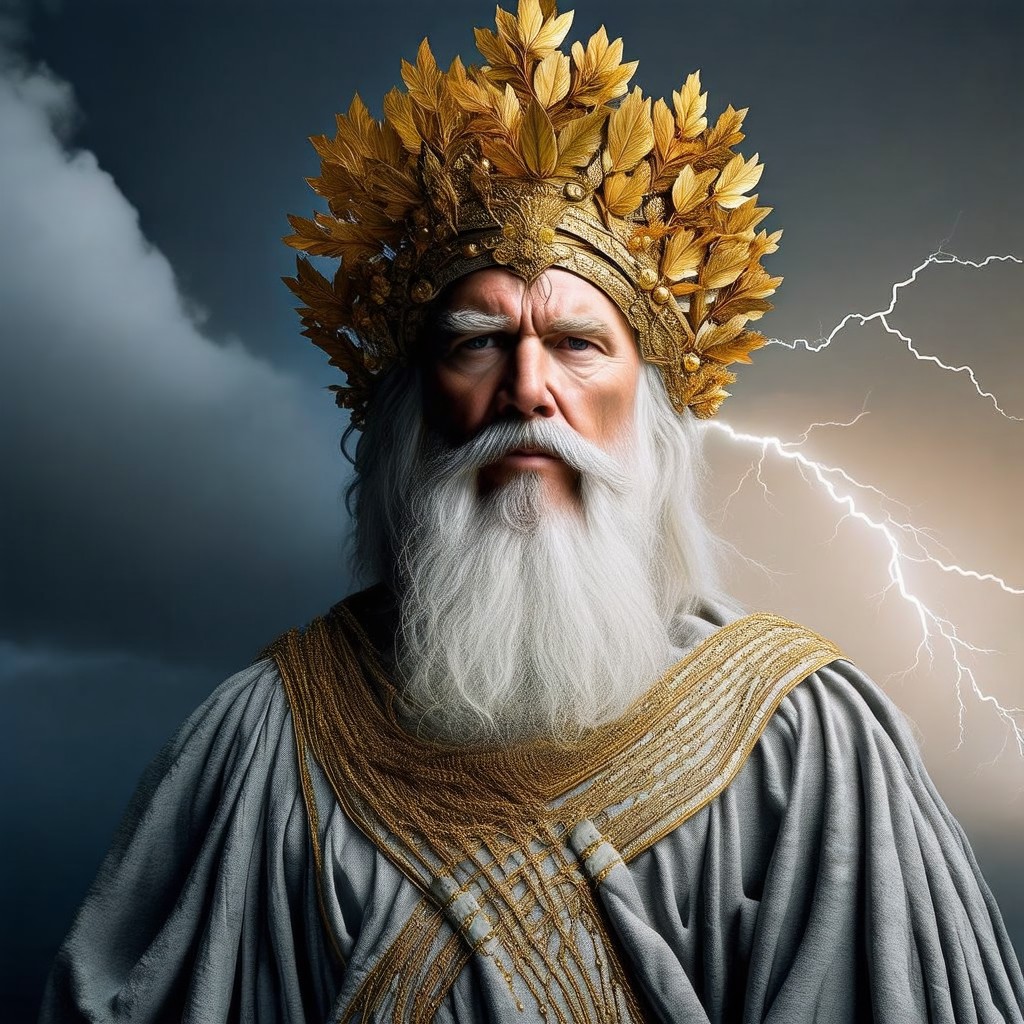}
            \caption*{\textit{Finetuned}}
        \end{subfigure}
    \end{minipage}
    \par\vspace{0.2em}
    \addtocounter{figure}{-1}
    \begin{minipage}[t]{0.47\linewidth}
        \vspace{0pt}
        \captionsetup{justification=raggedright,singlelinecheck=false}
        \captionof{figure}{Quality--latency trade-off on SD3-Medium. Shiva-DiT lies on the Pareto frontier of evaluated configurations (Appendix~\ref{app:baseline_configs}).}
        \label{fig:effi_scatter}
    \end{minipage}
    \hfill
    \begin{minipage}[t]{0.51\linewidth}
        \vspace{0pt}
        \captionsetup{justification=raggedright,singlelinecheck=false}
        \captionof{figure}{Qualitative comparison on SD3-Medium for the prompt ``God Zeus wearing a golden oak leaves crown on his head, grey beard.''}
        \label{fig:qualitative_main}
    \end{minipage}
\end{figure}

\paragraph{Run-to-Run Stability.}To verify that the reported gains exceed sampling noise, we regenerate each configuration with five matched random seeds. Across configurations, the standard deviation is $0.07$--$0.27$ for FID and $0.03$--$0.07$ for CLIP Score. Shiva-90\% outperforms the dense finetuned reference on both metrics for all five seeds, and its smallest paired FID margin exceeds three times the seed-to-seed deviation. Shiva-60\% likewise improves CLIP Score in all five runs. Appendix~\ref{app:multiseed} reports the paired differences and the corresponding comparison for Shiva-80\%.

\paragraph{Hardware-Aware Efficiency.} 
For Shiva-60\%, attention time drops by $38.5\%$ and FFN time by $30.5\%$; attention accounts for $45.4\%$ of the Vanilla wall-clock. The pruning path costs approximately 190~ms per image, so lower budgets amortize it more effectively. With \texttt{torch.compile}, Shiva-60\% remains $1.29\times$ faster than compiled Vanilla, confirming that the gain persists under graph optimization. Appendices~\ref{app:overhead_anatomy} and~\ref{app:compilation} give the stage-level and compiled results.

Table~\ref{tab:train_cost} shows that Shiva reaches 1.73 iterations per second with 7\% time overhead and lower peak memory than Vanilla. DiffCR takes $1.65\times$ the Vanilla iteration time because it evaluates two backbone paths. SparseDiT takes $0.94\times$ because its prescribed sparse-dense structure removes computation during fine-tuning as well as inference. This faster training point has FID 25.71, compared with 16.86 for Vanilla in Table~\ref{tab:main_results}. Appendix~\ref{app:training_efficiency} gives the protocol and additional discussion.

\begin{table}[!htbp]
    \centering
    \small
    \setlength{\tabcolsep}{2pt}
    \renewcommand{\arraystretch}{1.23}
    \caption{Training efficiency on SD3-Medium with 8$\times$ H200 GPUs (global batch size 16).}
    \begin{tabular*}{\linewidth}{@{\extracolsep{\fill}}lccccc@{}}
        \toprule
        Method & Learnable $k$ & Peak VRAM (MB) $\downarrow$ & Throughput (it/s) $\uparrow$ & Time/Iter (ms) $\downarrow$ & Relative \\
        \midrule
        Vanilla & \xmark & 51564 & 1.86 & 538 & $1.00\times$ \\
        NeuralSort~\cite{neuralsort2019} & \xmark & 58088 & 1.05 & 952 & $1.77\times$ \\
        DyDiT~\cite{dydit2024} & \xmark & 59610 & 1.43 & 699 & $1.30\times$ \\
        SparseDiT~\cite{sparsedit2025} & \xmark & 54826 & 1.98 & 505 & $0.94\times$ \\
        DiffCR~\cite{diffcr2025} & \cmark & 66374 & 1.13 & 885 & $1.65\times$ \\
        \textbf{Shiva (ours)} & \cmark & \textbf{47650} & \alignbold{1}{73} & \textbf{578} & $\mathbf{1.07\times}$ \\
        \bottomrule
    \end{tabular*}
    \label{tab:train_cost}
\end{table}

\paragraph{Generalization Across Architectures and Settings.}
\label{sec:generalization_main}

Table~\ref{tab:generalization_arch_main} extends Shiva-DiT beyond SD3-Medium. The achieved speedup depends on both the chosen retention ratio and the backbone size. Flux.1-dev reaches a $1.25\times$ wall-clock speedup with 24.0\% fewer FLOPs. PixArt-$\Sigma$ has fewer parameters, and more aggressive pruning substantially degrades image quality in our experiments. We therefore use a more conservative retention ratio, yielding a 17.2\% FLOPs reduction and a $1.12\times$ speedup. Shiva improves FID and latency over both added baselines. ImageReward and CLIP favor Shiva on Flux and IBTM on PixArt; ToMA has slightly higher IQA on Flux. Appendix~\ref{app:baseline_configs} gives configurations, and Appendix~\ref{app:add_models} provides qualitative comparisons.

\begin{table}[!htbp]
    \centering
    \caption{Generalization to Flux.1-dev and PixArt-$\Sigma$ at $1024\times1024$ (50 steps, RTX 4090).}
    \small
    \setlength{\tabcolsep}{2.8pt}
    \renewcommand{\arraystretch}{1.23}
    \begin{tabular*}{\linewidth}{@{\extracolsep{\fill}}llcccccccc@{}}
        \toprule
        \multirow{2}{*}{Architecture} & \multirow{2}{*}{Method} & \multicolumn{4}{c}{Quality} & \multicolumn{4}{c}{Efficiency} \\
        \cmidrule(lr){3-6} \cmidrule(lr){7-10}
        & & FID$\downarrow$ & IR$\uparrow$ & CLIP$\uparrow$ & IQA$\uparrow$ & FLOPs (T) & Latency (ms) & Speedup & $\Delta$ FLOPs (\%) \\
        \midrule
        \multirow{4}{*}{Flux.1-dev}
        & Vanilla & 14.25 & 0.93 & 30.56 & 0.50 & \multirow{2}{*}{44.63} & \multirow{2}{*}{33691} & \multirow{2}{*}{1.00$\times$} & \multirow{2}{*}{0.0} \\
        & \textit{Finetuned} & \textit{9.29} & \textit{0.33} & \textit{29.19} & \textit{0.56} & & & & \\
        & ToMA & 26.72 & 0.78 & 30.24 & 0.54 & 36.04 & 28091 & 1.20$\times$ & 19.3 \\
        & \textbf{Shiva} & 15.03 & 0.92 & 30.29 & 0.53 & 33.93 & 26891 & 1.25$\times$ & 24.0 \\
        \midrule
        \multirow{4}{*}{PixArt-$\Sigma$}
        & Vanilla & 11.08 & 0.93 & 31.37 & 0.46 & \multirow{2}{*}{12.97} & \multirow{2}{*}{5910} & \multirow{2}{*}{1.00$\times$} & \multirow{2}{*}{0.0} \\
        & \textit{Finetuned} & \textit{9.63} & \textit{0.97} & \textit{31.32} & \textit{0.48} & & & & \\
        & IBTM & 13.17 & 0.83 & 31.08 & 0.47 & 11.27 & 5806 & 1.02$\times$ & 13.1 \\
        & \textbf{Shiva} & 11.84 & 0.72 & 30.63 & 0.47 & 10.74 & 5281 & 1.12$\times$ & 17.2 \\
        \bottomrule
    \end{tabular*}
    \label{tab:generalization_arch_main}
     
\end{table}
 
We further evaluate transfer across held-out prompt sets and resolutions without additional training (Table~\ref{tab:generalization}). Shiva-80\% improves IQA over Vanilla from 0.459 to 0.496 on DrawBench and from 0.468 to 0.508 on PartiPrompts. It also lowers FID from 19.02 to 17.05 at $512\times512$ and from 16.21 to 13.75 at $1024\times768$, showing that the learned router and timestep-layer policy remain effective beyond the training prompt distribution and token-grid resolution.

\begin{table}[!t]
\centering
\caption{Dataset and resolution transfer on SD3-Medium without retraining.}
\small
\begin{minipage}[t]{0.46\linewidth}
    \centering
    \setlength{\tabcolsep}{4pt}
    \begin{tabular}[t]{llccc}
        \toprule
        \textbf{Dataset} & \textbf{Method} & IR $\uparrow$ & CLIP $\uparrow$ & IQA $\uparrow$ \\
        \midrule
        \multirow{6}{*}{DrawBench}
          & Vanilla    & 0.980 & 34.126 & 0.459 \\
          \cmidrule(l){2-5}
          & ToMeSD    & 0.830 & \alignbold{33}{778} & 0.471 \\
          & DiffCR    & 0.832 & 32.598 & 0.477 \\
          & DyDiT    & 0.615 & 32.210 & 0.484 \\
          & Shiva-80\% & \alignbold{0}{875} & \alignunder{33}{569} & \alignbold{0}{496} \\
          & Shiva-60\% & \alignunder{0}{853} & 33.309 & \alignunder{0}{487} \\
        \midrule
        \multirow{6}{*}{PartiPrompts}
          & Vanilla    & 1.182 & 33.103 & 0.468 \\
          \cmidrule(l){2-5}
          & ToMeSD    & 0.961 & \alignbold{33}{622} & 0.466 \\
          & DiffCR    & 0.971 & 32.900 & \alignunder{0}{500} \\
          & DyDiT    & 0.904 & 32.907 & 0.495 \\
          & Shiva-80\% & \alignbold{1}{170} & 33.232 & \alignbold{0}{508} \\
          & Shiva-60\% & \alignunder{1}{144} & \alignunder{33}{246} & 0.498 \\
        \bottomrule
    \end{tabular}
\end{minipage}
\hfill
\begin{minipage}[t]{0.50\linewidth}
    \centering
    \setlength{\tabcolsep}{4pt}
    \begin{tabular}[t]{llcccc}
        \toprule
        \textbf{Res.} & \textbf{Method} & FID $\downarrow$ & IR $\uparrow$ & CLIP $\uparrow$ & IQA $\uparrow$ \\
        \midrule
        \multirow{6}{*}{$512{\times}512$}
          & Vanilla    & 19.02 & 0.840 & 30.654 & 0.423 \\
          \cmidrule(l){2-6}
          & ToMeSD & 36.42 & 0.468 & 30.314 & 0.400 \\
          & DiffCR & 26.83 & 0.627 & 30.245 & 0.413 \\
          & DyDiT & 26.53 & 0.662 & 30.059 & 0.444 \\
          & Shiva-80\% & \alignbold{17}{05} & \alignbold{0}{733} & \alignunder{30}{812} & \alignunder{0}{452} \\
          & Shiva-60\% & \alignunder{19}{32} & \alignunder{0}{716} & \alignbold{30}{860} & \alignbold{0}{454} \\
        \midrule
        \multirow{6}{*}{\small$1024{\times}768$}
          & Vanilla    & 16.21 & 0.933 & 30.796 & 0.456 \\
          \cmidrule(l){2-6}
          & ToMeSD & 39.73 & 0.646 & 30.717 & 0.461 \\
          & DiffCR & 32.18 & 0.747 & 30.585 & 0.482 \\
          & DyDiT & \alignunder{16}{13} & 0.665 & 30.453 & 0.478 \\
          & Shiva-80\% & \alignbold{13}{75} & \alignunder{0}{877} & \alignunder{31}{008} & \alignbold{0}{495} \\
          & Shiva-60\% & 16.55 & \alignbold{0}{954} & \alignbold{31}{220} & \alignunder{0}{483} \\
        \bottomrule
    \end{tabular}
\end{minipage}
\label{tab:generalization}
\end{table}

\begin{table}[!t]
\centering
\small
\setlength{\tabcolsep}{6pt}
\caption{Directional agreement on a trained SD3-Medium block over 50 draws. The exact marginal is measured on 512 tokens; chance cosine has standard deviation $0.044$.}
\begin{tabular}{llccc}
\toprule
Estimator & Reference & Mean & Std & $>0$ \\
\midrule
Single-path (ours) & Two-path & $\phantom{-}0.536$ & 0.065 & 100\% \\
\midrule
Two-path           & Exact marginal & $\phantom{-}0.144$ & 0.423 & 64\% \\
Single-path (ours) & Exact marginal & $\phantom{-}0.120$ & 0.295 & 68\% \\
Gumbel-Sigmoid STE & Exact marginal & $\phantom{-}0.114$ & 0.313 & 64\% \\
Random direction   & Exact marginal & $-0.008$ & 0.040 & 36\% \\
Feature norm       & Exact marginal & $\phantom{-}0.019$ & 0.471 & 48\% \\
\bottomrule
\end{tabular}
\label{tab:estimator_real}
\end{table}
\FloatBarrier

\subsection{Ablation Study}
\label{sec:ablation}

\paragraph{Gradient Estimator Validation.}
We examine the two approximations in our backward pass: using the executed path to estimate the two-path surrogate, and propagating this estimate through the soft-rank relaxation. Appendix~\ref{app:synthetic_experiments} isolates both approximations on controlled residual tasks across budgets. Table~\ref{tab:estimator_real} then evaluates the estimator on a trained SD3-Medium block, with Appendix~\ref{app:real_block_validation} providing the evaluation protocol. The single-path estimate reaches $0.536$ cosine similarity with the two-path surrogate and remains positively aligned in all 50 draws. Against the exact fixed-budget marginal on 512 tokens, it is positive in 68\% of draws ($p\!\approx\!0.008$), while random-direction and feature-norm controls remain at chance. The Gumbel-Sigmoid baseline shows comparable directional agreement; its formulation here does not learn an explicit budget $k$. Together, these results support our biased single-path surrogate for deterministic hard top-$k$ selection. Further diagnostics in Appendix~\ref{app:real_block_validation} show that directional agreement varies across depths, denoising timesteps, and retention ratios.

\raggedbottom \paragraph{Router Architecture Analysis.}
Table~\ref{tab:ablation_router} examines parameter sharing and router context. Global sharing ($G=1$) gives higher reconstruction error, while the fully independent Max configuration also underperforms despite having more parameters. Pairwise sharing achieves the lowest MSE among the sharing strategies ($0.894$), consistent with useful local similarity between adjacent layers. For panel (b), every context variant is trained separately with the same recipe, and token features are always present. Full conditioning gives the highest similarity. Removing prompt or layer conditioning raises MSE by 0.049 and 0.008, respectively. Removing timestep conditioning gives nearly unchanged MSE but lower similarity, while removing all context raises MSE and lowers similarity. Figure~\ref{fig:router_viz} in Appendix~\ref{app:more_ablation} visualizes the resulting token selections.

\begin{table}[htbp]
    \centering
    \small
    \caption{Router design ablations at a $0.6$ retention ratio on 1,980 validation images. Panel (b) uses matched training recipes and always provides token features.}
    \begin{minipage}[t]{0.61\linewidth}
        \centering
        \caption*{(a) Parameter sharing}
        \setlength{\tabcolsep}{3.5pt}
        \begin{tabular}{lcccccc}
            \toprule
            Config & Grouping & $G$ & $d^{'}$ & Params & MSE $\downarrow$ & Sim $\uparrow$ \\
            \midrule
            Tiny & Global & 1 & 64 & 0.3M & 2.102 & 0.829 \\
            Shared & Global & 1 & 512 & 2.6M & 1.254 & 0.862 \\
            Max & Independent & 24 & 512 & 64M & 1.831 & 0.805 \\
            Slim & Independent & 24 & 64 & 8.0M & 1.072 & 0.862 \\
            Group-4 & 6-Layers & 4 & 128 & 2.7M & 1.074 & 0.858 \\
            \midrule
            Group-8 & 3-Layers & 8 & 64 & 2.7M & \alignunder{0}{906} & \alignunder{0}{879} \\
            \textbf{Group-12} & Pairwise & 12 & 64 & 4.1M & \alignbold{0}{894} & \alignbold{0}{881} \\
            \bottomrule
        \end{tabular}
    \end{minipage}
    \hfill
    \begin{minipage}[t]{0.36\linewidth}
        \centering
        \caption*{(b) Context inputs}
        \setlength{\tabcolsep}{4pt}
        \begin{tabular}{ccccc}
            \toprule
            Time & Prompt & Layer & MSE $\downarrow$ & Sim $\uparrow$ \\
            \midrule
            \cmark & \cmark & \cmark & \alignunder{0}{894} & \alignbold{0}{881} \\
            \xmark & \cmark & \cmark & \alignbold{0}{893} & \alignunder{0}{879} \\
            \cmark & \cmark & \xmark & 0.902 & 0.878 \\
            \cmark & \xmark & \cmark & 0.943 & 0.875 \\
            \xmark & \xmark & \cmark & 0.914 & 0.874 \\
            \xmark & \xmark & \xmark & 0.934 & 0.871 \\
            \bottomrule
        \end{tabular}
    \end{minipage}
    \label{tab:ablation_router}
    \label{tab:hctx_ablation}
\end{table}

\paragraph{Ratio Generalization.}The adaptive policy requires one router to rank tokens under layer- and timestep-dependent budgets. Training only at a fixed ratio can overfit a single sorting threshold, whereas stratified sampling over $[r_{\min},r_{\max}]$ exposes the router to the full operating range. It produces a broader score distribution and more discriminative rankings across retention levels, as visualized by the matched heatmaps and histograms in Appendix~\ref{app:diff_strategy}. This comparison connects the stratified training protocol to the adaptive budgets evaluated below.

\begin{table}[!t]
    \centering
    \caption{Fixed, hand-designed, and adaptive ratio policies on MJHQ-30K at $\bar{r}\approx0.6$. The hand-designed policy uses the schedule formula of \citet{sparsedit2025}; the variants differ only in the ratio policy. This ablation is a separate experiment from the Shiva-60\% result in Table~\ref{tab:main_results} and has a different realized retention cost.}
    \newcommand{\metriccell}[2]{\makebox[1.4em][r]{#1}\makebox[2.25em][l]{#2}}
    \begin{tabular}{lcccc}
        \toprule
        Policy & FID $\downarrow$ & ImageReward $\uparrow$ & CLIP Score $\uparrow$ & CLIP IQA $\uparrow$ \\
        \midrule
        Fixed Ratio & \metriccell{21}{.76} & \metriccell{0}{.835} & \metriccell{30}{.75} & \metriccell{0}{.497} \\
        Hand-designed & \metriccell{\underline{17}}{\underline{.83}} & \metriccell{\textbf{0}}{\textbf{.882}} & \metriccell{\underline{30}}{\underline{.88}} & \metriccell{\underline{0}}{\underline{.498}} \\
        \textbf{Adaptive} & \metriccell{\textbf{17}}{\textbf{.36}} & \metriccell{\underline{0}}{\underline{.880}} & \metriccell{\textbf{30}}{\textbf{.94}} & \metriccell{\textbf{0}}{\textbf{.500}} \\
        \bottomrule 
    \end{tabular}
    \label{tab:ratio_ablation}
\end{table}

\begin{figure}[!ht]
    \centering
    \begin{subfigure}[b]{0.35\textwidth}
        \centering
        \includegraphics[width=\linewidth]{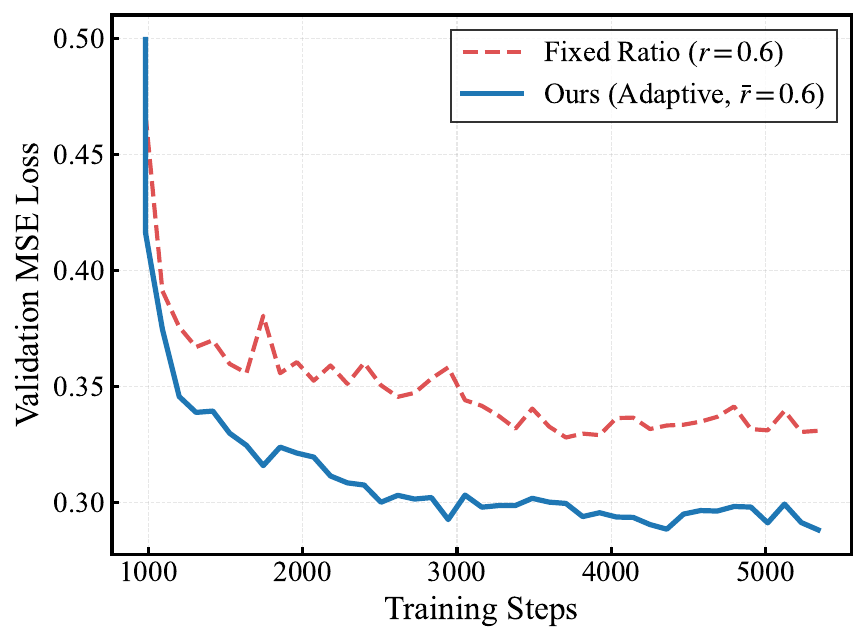}
        \caption{Validation loss during training.}
        \label{fig:loss_curve}
    \end{subfigure}
    \hfill
    \begin{subfigure}[b]{0.63\textwidth}
        \centering
        \includegraphics[width=\linewidth]{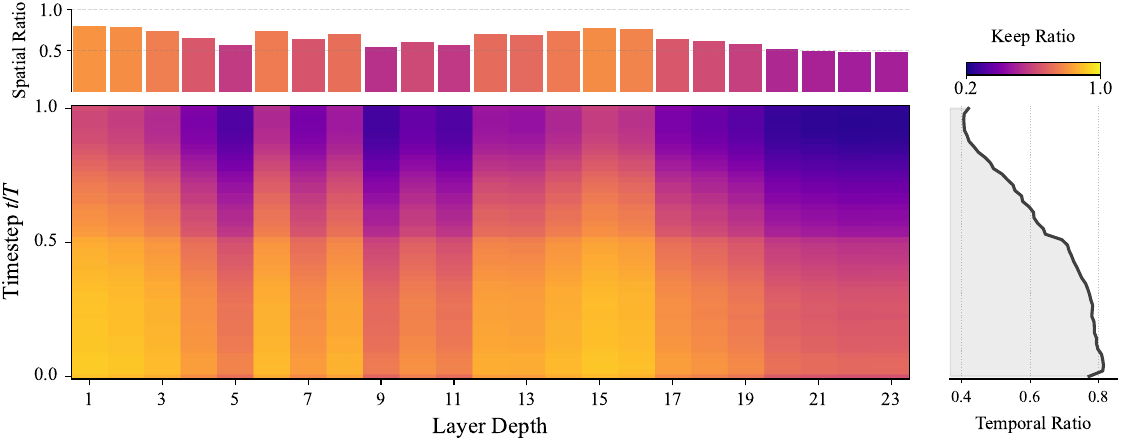}
        \caption{Learned pruning-ratio schedule.}
        \label{fig:ratio_heatmap}
    \end{subfigure}
    \caption{Adaptive policy analysis. (a) The adaptive policy converges to lower validation MSE than the matched fixed-ratio baseline. (b) The learned schedule varies across layers ($x$-axis) and diffusion timesteps ($y$-axis), revealing a ``structure-first, detail-later'' allocation.}
    \label{fig:ratio_ablation}
\end{figure}

\paragraph{Impact of Adaptive Ratio Policy.}
We compare adaptive, fixed, and hand-designed ratio policies under a strict budget ($\bar{r} \approx 0.6$). The variants use the same router architecture, residual estimator, distillation objectives, backbone adaptation, training stages, optimization settings, and matched average retention ratio; only the ratio policy differs. As shown in Figure~\ref{fig:ratio_ablation}(a) and Table~\ref{tab:ratio_ablation}, the adaptive policy improves validation MSE and perceptual metrics over the fixed-ratio baseline by allocating retention budgets across layers and denoising timesteps. Visualizing the learned schedule (Figure~\ref{fig:ratio_ablation}) reveals a convergence to a \emph{structure-first} spatial strategy (preserving shallow layers) and a progressive temporal pattern: distinct from the uniform baseline, it allocates maximum budget ($\approx 0.8$) to high-frequency refinement stages for detail recovery while aggressively pruning ($\approx 0.4$) during early chaotic structural phases.
\FloatBarrier

\raggedbottom
\begin{samepage}
\section{Limitations and Broader Impact}
Shiva-DiT relies on spatial redundancy, which weakens in high-density settings such as dense text rendering. At the 60\% budget, performance is weakest on the \texttt{logo} category, where most prompts involve text rendering (Appendix~\ref{app:percat}). Rejected tokens remain on the identity path, so detail can be lost when redundancy is limited; token-fusion strategies may reduce this failure mode. Training-time soft ranking incurs quadratic computational costs at larger token counts (Table~\ref{tab:topk_measured}). Reported speedups also depend on the backbone, hardware, batch size, and execution stack. Our cross-architecture and compiled measurements show that the gains persist in the tested settings, but they do not imply identical acceleration on every deployment. Although Shiva-DiT reduces computational and energy costs, it inherits the misuse risks of text-conditioned image models, including deceptive synthetic imagery, so provenance and content-safety safeguards should remain in place.
\par
\end{samepage}

\section{Conclusion}

Shiva-DiT uses Residual-Based Differentiable Top-$k$ Selection to learn token scores and budgets while executing a deterministic number of tokens. Across the tested DiT backbones, reductions in token-dependent FLOPs translate into measured end-to-end latency gains. On SD3-Medium, Shiva-DiT reaches up to a $1.54\times$ wall-clock speedup while maintaining competitive generative quality.

\section*{Acknowledgments}
This research was supported in part by the National Science Foundation of China (62502441, 62125206), the Major Program of the National Natural Science Foundation of Zhejiang (LD25F020002), and the Zhejiang Key Laboratory Project (2024E10001). Hailiang Zhao’s work was supported in part by the Zhejiang University Education Foundation Qizhen Scholar Foundation.

\bibliography{main}

@misc{li2026qwenimagebench,
  title={{Qwen-Image-Bench}: From Generation to Creation in Text-to-Image Evaluation},
  author={Niantong Li and Guangzheng Hu and Weixu Qiao and Ying Ba and Qichen Hong and Shijun Shen and Jinlin Wang and Fan Zhou and Jianye Kang and Xin Shang and Ziyi He and Wei Wang and Dalin Li and Jiahao Li and Jie Zhang and Kaiyuan Gao and Kun Yan and Lihan Jiang and Ningyuan Tang and Shengming Yin and Tianhe Wu and Xiao Xu and Xiaoyue Chen and Yuxiang Chen and Yan Shu and Yanran Zhang and Yilei Chen and Yixian Xu and Zekai Zhang and Zhendong Wang and Zihao Liu and Zikai Zhou and Hongzhu Shi and Yi Wang and Bing Zhao and Hu Wei and Lin Qu and Chenfei Wu},
  year={2026},
  eprint={2605.28091},
  archivePrefix={arXiv},
  primaryClass={cs.CV},
  url={https://arxiv.org/abs/2605.28091}
}

@misc{civitai_qwen_image_bench_fp8,
  author={{Civitai}},
  title={{Qwen-Image-Bench-FP8}},
  howpublished={Hugging Face model repository},
  year={2026},
  url={https://huggingface.co/Civitai/Qwen-Image-Bench-FP8},
  note={Revision 210ede152c4fc7da163dd2ff7d4f321898d0cad0; accessed 2026-09-29}
}

@misc{raposo2024mixtureofdepths,
  title={Mixture-of-Depths: Dynamically Allocating Compute in Transformer-Based Language Models},
  author={David Raposo and Sam Ritter and Blake Richards and Timothy Lillicrap and Peter Conway Humphreys and Adam Santoro},
  year={2024},
  eprint={2404.02258},
  archivePrefix={arXiv},
  url={https://arxiv.org/abs/2404.02258}
}

@inproceedings{prillo2020softsort,
  title={{SoftSort}: A Continuous Relaxation for the Argsort Operator},
  author={Sebastian Prillo and Julian Eisenschlos},
  booktitle={Proceedings of the 37th International Conference on Machine Learning},
  series={Proceedings of Machine Learning Research},
  volume={119},
  pages={7793--7802},
  year={2020},
  publisher={PMLR},
  url={https://proceedings.mlr.press/v119/prillo20a.html}
}

@inproceedings{blondel2020fastsort,
  title={Fast Differentiable Sorting and Ranking},
  author={Mathieu Blondel and Olivier Teboul and Quentin Berthet and Josip Djolonga},
  booktitle={Proceedings of the 37th International Conference on Machine Learning},
  series={Proceedings of Machine Learning Research},
  volume={119},
  pages={950--959},
  year={2020},
  publisher={PMLR},
  url={https://proceedings.mlr.press/v119/blondel20a.html}
}

@inproceedings{softrank2005,
author = {Burges, Chris and Shaked, Tal and Renshaw, Erin and Lazier, Ari and Deeds, Matt and Hamilton, Nicole and Hullender, Greg},
title = {Learning to rank using gradient descent},
year = {2005},
isbn = {1595931805},
publisher = {Association for Computing Machinery},
address = {New York, NY, USA},
url = {https://doi.org/10.1145/1102351.1102363},
doi = {10.1145/1102351.1102363},
booktitle = {Proceedings of the 22nd International Conference on Machine Learning},
pages = {89–96},
numpages = {8},
location = {Bonn, Germany},
series = {ICML '05}
}

@inproceedings{
dydit2024,
title={Dynamic Diffusion Transformer},
author={Wangbo Zhao and Yizeng Han and Jiasheng Tang and Kai Wang and Yibing Song and Gao Huang and Fan Wang and Yang You},
booktitle={The Thirteenth International Conference on Learning Representations},
year={2025},
url={https://openreview.net/forum?id=taHwqSrbrb}
}

@inproceedings{
gumbel2017,
title={Categorical Reparameterization with Gumbel-Softmax},
author={Eric Jang and Shixiang Gu and Ben Poole},
booktitle={International Conference on Learning Representations},
year={2017},
url={https://openreview.net/forum?id=rkE3y85ee}
}

@inproceedings{diffcr2025,
  author       = {Haoran You and
                  Connelly Barnes and
                  Yuqian Zhou and
                  Yan Kang and
                  Zhenbang Du and
                  Wei Zhou and
                  Lingzhi Zhang and
                  Yotam Nitzan and
                  Xiaoyang Liu and
                  Zhe Lin and
                  Eli Shechtman and
                  Sohrab Amirghodsi and
                  Yingyan Celine Lin},
  title        = {Layer- and Timestep-Adaptive Differentiable Token Compression Ratios
                  for Efficient Diffusion Transformers},
  booktitle    = {{IEEE/CVF} Conference on Computer Vision and Pattern Recognition,
                  {CVPR} 2025, Nashville, TN, USA, June 11-15, 2025},
  pages        = {18072--18082},
  publisher    = {Computer Vision Foundation / {IEEE}},
  year         = {2025},
  doi          = {10.1109/CVPR52734.2025.01684},
  bibsource    = {dblp computer science bibliography, https://dblp.org}
}

@misc{ste2013,
      title={Estimating or Propagating Gradients Through Stochastic Neurons for Conditional Computation}, 
      author={Yoshua Bengio and Nicholas Léonard and Aaron Courville},
      year={2013},
      eprint={1308.3432},
      archivePrefix={arXiv},
      primaryClass={cs.LG},
      url={https://arxiv.org/abs/1308.3432}, 
}

@inproceedings{tome2022,
  title={Token Merging: Your {ViT} but Faster},
  author={Bolya, Daniel and Fu, Cheng-Yang and Dai, Xiaoliang and Zhang, Peizhao and Feichtenhofer, Christoph and Hoffman, Judy},
  booktitle={International Conference on Learning Representations},
  year={2023}
}

@article{tomesd2023,
  title={Token Merging for Fast Stable Diffusion},
  author={Bolya, Daniel and Hoffman, Judy},
  journal={CVPR Workshop on Efficient Deep Learning for Computer Vision},
  year={2023}
}

@inproceedings{dao2022flashattention,
  title={Flash{A}ttention: Fast and Memory-Efficient Exact Attention with {IO}-Awareness},
  author={Dao, Tri and Fu, Daniel Y. and Ermon, Stefano and Rudra, Atri and R{\'e}, Christopher},
  booktitle={Advances in Neural Information Processing Systems (NeurIPS)},
  year={2022}
}

@misc{ibtm2025,
      title={Importance-Based Token Merging for Efficient Image and Video Generation}, 
      author={Haoyu Wu and Jingyi Xu and Hieu Le and Dimitris Samaras},
      year={2025},
      eprint={2411.16720},
      archivePrefix={arXiv},
      primaryClass={cs.CV},
      url={https://arxiv.org/abs/2411.16720}, 
}

@inproceedings{sdtm2025,
  author       = {Haipeng Fang and
                  Sheng Tang and
                  Juan Cao and
                  Enshuo Zhang and
                  Fan Tang and
                  Tong{-}Yee Lee},
  title        = {Attend to Not Attended: Structure-then-Detail Token Merging for Post-training
                  DiT Acceleration},
  booktitle    = {{IEEE/CVF} Conference on Computer Vision and Pattern Recognition,
                  {CVPR} 2025, Nashville, TN, USA, June 11-15, 2025},
  pages        = {18083--18092},
  publisher    = {Computer Vision Foundation / {IEEE}},
  year         = {2025},
  doi          = {10.1109/CVPR52734.2025.01685},
  bibsource    = {dblp computer science bibliography, https://dblp.org}
}

@inproceedings{
asymrnr2024,
title={AsymRnR: Video Diffusion Transformers Acceleration with Asymmetric Reduction and Restoration},
author={Wenhao Sun and Rong-Cheng Tu and Jingyi Liao and Zhao Jin and Dacheng Tao},
booktitle={Forty-second International Conference on Machine Learning},
year={2025},
url={https://openreview.net/forum?id=5PiZevq9fY}
}

@INPROCEEDINGS{atedm2024,
  author={Wang, Hongjie and Liu, Difan and Kang, Yan and Li, Yijun and Lin, Zhe and Jha, Niraj K. and Liu, Yuchen},
  booktitle={2024 IEEE/CVF Conference on Computer Vision and Pattern Recognition (CVPR)}, 
  title={Attention-Driven Training-Free Efficiency Enhancement of Diffusion Models}, 
  year={2024},
  volume={},
  number={},
  pages={16080-16089},
  doi={10.1109/CVPR52733.2024.01522}}

@INPROCEEDINGS{zeroprune2024,
  author={Wang, Hongjie and Dedhia, Bhishma and Jha, Niraj K.},
  booktitle={2024 IEEE/CVF Conference on Computer Vision and Pattern Recognition (CVPR)}, 
  title={Zero-TPrune: Zero-Shot Token Pruning Through Leveraging of the Attention Graph in Pre-Trained Transformers}, 
  year={2024},
  volume={},
  number={},
  pages={16070-16079},
  doi={10.1109/CVPR52733.2024.01521}}

@INPROCEEDINGS{dit2023,
  author={Peebles, William and Xie, Saining},
  booktitle={2023 IEEE/CVF International Conference on Computer Vision (ICCV)}, 
  title={Scalable Diffusion Models with Transformers}, 
  year={2023},
  volume={},
  number={},
  pages={4172-4182},
  doi={10.1109/ICCV51070.2023.00387}}

@article{vit2020,
  title={An Image is Worth 16x16 Words: Transformers for Image Recognition at Scale},
  author={Dosovitskiy, Alexey and Beyer, Lucas and Kolesnikov, Alexander and Weissenborn, Dirk and Zhai, Xiaohua and Unterthiner, Thomas and  Dehghani, Mostafa and Minderer, Matthias and Heigold, Georg and Gelly, Sylvain and Uszkoreit, Jakob and Houlsby, Neil},
  journal={ICLR},
  year={2021}
}

@inproceedings{dyvit2021,
  title={DynamicViT: Efficient Vision Transformers with Dynamic Token Sparsification},
  author={Rao, Yongming and Zhao, Wenliang and Liu, Benlin and Lu, Jiwen and Zhou, Jie and Hsieh, Cho-Jui},
  booktitle = {Advances in Neural Information Processing Systems (NeurIPS)},
  year = {2021}
}

@INPROCEEDINGS{tokenpooling2023,
  author={Marin, Dmitrii and Chang, Jen-Hao Rick and Ranjan, Anurag and Prabhu, Anish and Rastegari, Mohammad and Tuzel, Oncel},
  booktitle={2023 IEEE/CVF Winter Conference on Applications of Computer Vision (WACV)}, 
  title={Token Pooling in Vision Transformers for Image Classification}, 
  year={2023},
  volume={},
  number={},
  pages={12-21},
  doi={10.1109/WACV56688.2023.00010}}

@INPROCEEDINGS{diffrate2023,
  author={Chen, Mengzhao and Shao, Wenqi and Xu, Peng and Lin, Mingbao and Zhang, Kaipeng and Chao, Fei and Ji, Rongrong and Qiao, Yu and Luo, Ping},
  booktitle={2023 IEEE/CVF International Conference on Computer Vision (ICCV)}, 
  title={DiffRate : Differentiable Compression Rate for Efficient Vision Transformers}, 
  year={2023},
  volume={},
  number={},
  pages={17118-17128},
  doi={10.1109/ICCV51070.2023.01574}}

@INPROCEEDINGS{tofu2023,
  author={Kim, Minchul and Gao, Shangqian and Hsu, Yen-Chang and Shen, Yilin and Jin, Hongxia},
  booktitle={2024 IEEE/CVF Winter Conference on Applications of Computer Vision (WACV)}, 
  title={Token Fusion: Bridging the Gap between Token Pruning and Token Merging}, 
  year={2024},
  volume={},
  number={},
  pages={1372-1381},
  doi={10.1109/WACV57701.2024.00141}}

@inproceedings{
zhang2024star,
title={Synergistic Patch Pruning for Vision Transformer: Unifying Intra- \& Inter-Layer Patch Importance},
author={Yuyao Zhang and Lan Wei and Nikolaos Freris},
booktitle={The Twelfth International Conference on Learning Representations},
year={2024},
url={https://openreview.net/forum?id=COO51g41Q4}
}

@inproceedings{tran2024pitome,
author = {Lee, Seon-Ho and Wang, Jue and Zhang, Zhikang and Fan, David and Li, Xinyu},
title = {Video token merging for long-form video understanding},
year = {2024},
isbn = {9798331314385},
publisher = {Curran Associates Inc.},
address = {Red Hook, NY, USA},
booktitle = {Proceedings of the 38th International Conference on Neural Information Processing Systems},
articleno = {444},
numpages = {21},
location = {Vancouver, BC, Canada},
series = {NIPS '24}
}

@inproceedings{freqts2025,
author = {Yang, Xinye and Yang, Yuxin and Pang, Haoran and Tian, Aaron Xuxiang and Li, Luking},
title = {FreqTS: frequency-aware token selection for accelerating diffusion models},
year = {2025},
isbn = {978-1-57735-897-8},
publisher = {AAAI Press},
url = {https://doi.org/10.1609/aaai.v39i9.33008},
doi = {10.1609/aaai.v39i9.33008},
booktitle = {Proceedings of the Thirty-Ninth AAAI Conference on Artificial Intelligence and Thirty-Seventh Conference on Innovative Applications of Artificial Intelligence and Fifteenth Symposium on Educational Advances in Artificial Intelligence},
articleno = {1035},
numpages = {8},
series = {AAAI'25/IAAI'25/EAAI'25}
}

@inproceedings{
toma2025,
title={To{MA}: Token Merge with Attention for Diffusion Models},
author={Wenbo Lu and Shaoyi Zheng and Yuxuan Xia and Shengjie Wang},
booktitle={Forty-second International Conference on Machine Learning},
year={2025},
url={https://openreview.net/forum?id=51l8tvuIxo}
}

@inproceedings{
      ditfastattn2024,
      title={Di{TF}astAttn: Attention Compression for Diffusion Transformer Models},
      author={Zhihang Yuan and Hanling Zhang and Lu Pu and Xuefei Ning and Linfeng Zhang and Tianchen Zhao and Shengen Yan and Guohao Dai and Yu Wang},
      booktitle={The Thirty-eighth Annual Conference on Neural Information Processing Systems},
      year={2024},
      url={https://openreview.net/forum?id=51HQpkQy3t}
}

@article{catome2025,
      title={Cached Adaptive Token Merging: Dynamic Token Reduction and Redundant Computation Elimination in Diffusion Model}, 
      author={Omid Saghatchian and Atiyeh Gh. Moghadam and Ahmad Nickabadi},
      year={2025},
      eprint={2501.00946},
      archivePrefix={arXiv},
      primaryClass={cs.CV},
}

@inproceedings{
toca2024,
title={Accelerating Diffusion Transformers with Token-wise Feature Caching},
author={Chang Zou and Xuyang Liu and Ting Liu and Siteng Huang and Linfeng Zhang},
booktitle={The Thirteenth International Conference on Learning Representations},
year={2025},
url={https://openreview.net/forum?id=yYZbZGo4ei}
}

@inproceedings{
sparsedit2025,
title={SparseDiT: Token Sparsification for Efficient Diffusion Transformer},
author={Shuning Chang and Pichao WANG and Jiasheng Tang and Fan Wang and Yi Yang},
booktitle={The Thirty-ninth Annual Conference on Neural Information Processing Systems},
year={2025},
url={https://openreview.net/forum?id=jTBxyQempF}
}

@misc{dato2024,
      title={Token Pruning for Caching Better: 9 Times Acceleration on Stable Diffusion for Free}, 
      author={Evelyn Zhang and Bang Xiao and Jiayi Tang and Qianli Ma and Chang Zou and Xuefei Ning and Xuming Hu and Linfeng Zhang},
      year={2024},
      eprint={2501.00375},
      archivePrefix={arXiv},
      primaryClass={cs.CV},
      url={https://arxiv.org/abs/2501.00375}, 
}

@inproceedings{attn2017,
author = {Vaswani, Ashish and Shazeer, Noam and Parmar, Niki and Uszkoreit, Jakob and Jones, Llion and Gomez, Aidan N. and Kaiser, \L{}ukasz and Polosukhin, Illia},
title = {Attention is all you need},
year = {2017},
isbn = {9781510860964},
publisher = {Curran Associates Inc.},
address = {Red Hook, NY, USA},
booktitle = {Proceedings of the 31st International Conference on Neural Information Processing Systems},
pages = {6000–6010},
numpages = {11},
location = {Long Beach, California, USA},
series = {NIPS'17}
}

@inproceedings{ddpm2020,
author = {Ho, Jonathan and Jain, Ajay and Abbeel, Pieter},
title = {Denoising diffusion probabilistic models},
year = {2020},
isbn = {9781713829546},
publisher = {Curran Associates Inc.},
address = {Red Hook, NY, USA},
booktitle = {Proceedings of the 34th International Conference on Neural Information Processing Systems},
articleno = {574},
numpages = {12},
location = {Vancouver, BC, Canada},
series = {NIPS '20}
}

@inproceedings{
flowmatching2023,
title={Flow Matching for Generative Modeling},
author={Yaron Lipman and Ricky T. Q. Chen and Heli Ben-Hamu and Maximilian Nickel and Matthew Le},
booktitle={The Eleventh International Conference on Learning Representations },
year={2023},
url={https://openreview.net/forum?id=PqvMRDCJT9t}
}

@inproceedings{
ddim2022,
title={Denoising Diffusion Implicit Models},
author={Jiaming Song and Chenlin Meng and Stefano Ermon},
booktitle={International Conference on Learning Representations},
year={2021},
url={https://openreview.net/forum?id=St1giarCHLP}
}

@inproceedings{dpm2022,
author = {Lu, Cheng and Zhou, Yuhao and Bao, Fan and Chen, Jianfei and Li, Chongxuan and Zhu, Jun},
title = {DPM-solver: a fast ODE solver for diffusion probabilistic model sampling in around 10 steps},
year = {2022},
isbn = {9781713871088},
publisher = {Curran Associates Inc.},
address = {Red Hook, NY, USA},
booktitle = {Proceedings of the 36th International Conference on Neural Information Processing Systems},
articleno = {418},
numpages = {13},
location = {New Orleans, LA, USA},
series = {NIPS '22}
}

@misc{flux2024,
  author       = {{Black Forest Labs}},
  title        = {{Flux.1}},
  year         = {2024},
  url          = {https://blackforestlabs.ai/},
  note         = {Accessed: 2025-05-05}
}

@inproceedings{
sdxl2023,
title={{SDXL}: Improving Latent Diffusion Models for High-Resolution Image Synthesis},
author={Dustin Podell and Zion English and Kyle Lacey and Andreas Blattmann and Tim Dockhorn and Jonas M{\"u}ller and Joe Penna and Robin Rombach},
booktitle={The Twelfth International Conference on Learning Representations},
year={2024},
url={https://openreview.net/forum?id=di52zR8xgf}
}

@inproceedings{sd32024,
author = {Esser, Patrick and Kulal, Sumith and Blattmann, Andreas and Entezari, Rahim and M\"{u}ller, Jonas and Saini, Harry and Levi, Yam and Lorenz, Dominik and Sauer, Axel and Boesel, Frederic and Podell, Dustin and Dockhorn, Tim and English, Zion and Rombach, Robin},
title = {Scaling rectified flow transformers for high-resolution image synthesis},
year = {2024},
publisher = {JMLR.org},
booktitle = {Proceedings of the 41st International Conference on Machine Learning},
articleno = {503},
numpages = {28},
location = {Vienna, Austria},
series = {ICML'24}
}

@InProceedings{sd1.5_2022,
    author    = {Rombach, Robin and Blattmann, Andreas and Lorenz, Dominik and Esser, Patrick and Ommer, Bj\"orn},
    title     = {High-Resolution Image Synthesis With Latent Diffusion Models},
    booktitle = {Proceedings of the IEEE/CVF Conference on Computer Vision and Pattern Recognition (CVPR)},
    month     = {June},
    year      = {2022},
    pages     = {10684-10695}
}

@inproceedings{film2018,
author = {Perez, Ethan and Strub, Florian and de Vries, Harm and Dumoulin, Vincent and Courville, Aaron},
title = {FiLM: visual reasoning with a general conditioning layer},
year = {2018},
isbn = {978-1-57735-800-8},
publisher = {AAAI Press},
booktitle = {Proceedings of the Thirty-Second AAAI Conference on Artificial Intelligence and Thirtieth Innovative Applications of Artificial Intelligence Conference and Eighth AAAI Symposium on Educational Advances in Artificial Intelligence},
articleno = {483},
numpages = {10},
location = {New Orleans, Louisiana, USA},
series = {AAAI'18/IAAI'18/EAAI'18}
}

@inproceedings{
adamw2019,
title={Decoupled Weight Decay Regularization},
author={Ilya Loshchilov and Frank Hutter},
booktitle={International Conference on Learning Representations},
year={2019},
url={https://openreview.net/forum?id=Bkg6RiCqY7},
}

@inproceedings{
neuralsort2019,
title={Stochastic Optimization of Sorting Networks via Continuous Relaxations},
author={Aditya Grover and Eric Wang and Aaron Zweig and Stefano Ermon},
booktitle={International Conference on Learning Representations},
year={2019},
url={https://openreview.net/forum?id=H1eSS3CcKX},
}

@InProceedings{pixartsigma2024,
author="Chen, Junsong
and Ge, Chongjian
and Xie, Enze
and Wu, Yue
and Yao, Lewei
and Ren, Xiaozhe
and Wang, Zhongdao
and Luo, Ping
and Lu, Huchuan
and Li, Zhenguo",
editor="Leonardis, Ale{\v{s}}
and Ricci, Elisa
and Roth, Stefan
and Russakovsky, Olga
and Sattler, Torsten
and Varol, G{\"u}l",
title="PIXART-$\Sigma$: Weak-to-Strong Training of Diffusion Transformer for 4K Text-to-Image Generation",
booktitle="Computer Vision -- ECCV 2024",
year="2025",
publisher="Springer Nature Switzerland",
address="Cham",
pages="74--91",
isbn="978-3-031-73411-3"
}

@article{sora2024,
  title={Video generation models as world simulators},
  author={Tim Brooks and Bill Peebles and Connor Holmes and Will DePue and Yufei Guo and Li Jing and David Schnurr and Joe Taylor and Troy Luhman and Eric Luhman and Clarence Ng and Ricky Wang and Aditya Ramesh},
  year={2024},
  url={https://openai.com/research/video-generation-models-as-world-simulators},
}

@inproceedings{edm2022,
author = {Karras, Tero and Aittala, Miika and Laine, Samuli and Aila, Timo},
title = {Elucidating the design space of diffusion-based generative models},
year = {2022},
isbn = {9781713871088},
publisher = {Curran Associates Inc.},
address = {Red Hook, NY, USA},
booktitle = {Proceedings of the 36th International Conference on Neural Information Processing Systems},
articleno = {1926},
numpages = {13},
location = {New Orleans, LA, USA},
series = {NIPS '22}
}

@misc{diffusers2022,
  author = {Patrick von Platen and Suraj Patil and Anton Lozhkov and Pedro Cuenca and Nathan Lambert and Kashif Rasul and Mishig Davaadorj and Dhruv Nair and Sayak Paul and William Berman and Yiyi Xu and Steven Liu and Thomas Wolf},
  title = {Diffusers: State-of-the-art diffusion models},
  year = {2022},
  publisher = {GitHub},
  journal = {GitHub repository},
  howpublished = {\url{https://github.com/huggingface/diffusers}}
}

@inproceedings{dora2024,
author = {Liu, Shih-Yang and Wang, Chien-Yi and Yin, Hongxu and Molchanov, Pavlo and Wang, Yu-Chiang Frank and Cheng, Kwang-Ting and Chen, Min-Hung},
title = {DoRA: weight-decomposed low-rank adaptation},
year = {2024},
publisher = {JMLR.org},
booktitle = {Proceedings of the 41st International Conference on Machine Learning},
articleno = {1299},
numpages = {22},
location = {Vienna, Austria},
series = {ICML'24}
}

@misc{cfg2022,
      title={Classifier-Free Diffusion Guidance}, 
      author={Jonathan Ho and Tim Salimans},
      year={2022},
      eprint={2207.12598},
      archivePrefix={arXiv},
      primaryClass={cs.LG},
      url={https://arxiv.org/abs/2207.12598}, 
}

@article{sgd1951,
 ISSN = {00034851},
 author = {Herbert Robbins and Sutton Monro},
 journal = {The Annals of Mathematical Statistics},
 number = {3},
 pages = {400--407},
 publisher = {Institute of Mathematical Statistics},
 title = {A Stochastic Approximation Method},
 urldate = {2026-01-04},
 volume = {22},
 year = {1951}
}

@inproceedings{adam2015,
  author       = {Diederik P. Kingma and
                  Jimmy Ba},
  editor       = {Yoshua Bengio and
                  Yann LeCun},
  title        = {Adam: {A} Method for Stochastic Optimization},
  booktitle    = {3rd International Conference on Learning Representations, {ICLR} 2015,
                  San Diego, CA, USA, May 7-9, 2015, Conference Track Proceedings},
  year         = {2015},
  bibsource    = {dblp computer science bibliography, https://dblp.org}
}

@Misc{peft,
  title =        {{PEFT}: State-of-the-art Parameter-Efficient Fine-Tuning methods},
  author =       {Sourab Mangrulkar and Sylvain Gugger and Lysandre Debut and Younes Belkada and Sayak Paul and Benjamin Bossan and Marian Tietz},
  howpublished = {\url{https://github.com/huggingface/peft}},
  year =         {2022}
}

@misc{mjhq30k2024,
      title={Playground v2.5: Three Insights towards Enhancing Aesthetic Quality in Text-to-Image Generation}, 
      author={Daiqing Li and Aleks Kamko and Ehsan Akhgari and Ali Sabet and Linmiao Xu and Suhail Doshi},
      year={2024},
      eprint={2402.17245},
      archivePrefix={arXiv},
      primaryClass={cs.CV},
      url={https://arxiv.org/abs/2402.17245}, 
}

@inproceedings{drawbench,
author = {Saharia, Chitwan and Chan, William and Saxena, Saurabh and Lit, Lala and Whang, Jay and Denton, Emily and Ghasemipour, Seyed Kamyar Seyed and Ayan, Burcu Karagol and Mahdavi, S. Sara and Gontijo-Lopes, Raphael and Salimans, Tim and Ho, Jonathan and Fleet, David J and Norouzi, Mohammad},
title = {Photorealistic text-to-image diffusion models with deep language understanding},
year = {2022},
isbn = {9781713871088},
publisher = {Curran Associates Inc.},
address = {Red Hook, NY, USA},
booktitle = {Proceedings of the 36th International Conference on Neural Information Processing Systems},
articleno = {2643},
numpages = {16},
location = {New Orleans, LA, USA},
series = {NIPS '22}
}

@article{
pali,
title={Scaling Autoregressive Models for Content-Rich Text-to-Image Generation},
author={Jiahui Yu and Yuanzhong Xu and Jing Yu Koh and Thang Luong and Gunjan Baid and Zirui Wang and Vijay Vasudevan and Alexander Ku and Yinfei Yang and Burcu Karagol Ayan and Ben Hutchinson and Wei Han and Zarana Parekh and Xin Li and Han Zhang and Jason Baldridge and Yonghui Wu},
journal={Transactions on Machine Learning Research},
issn={2835-8856},
year={2022},
url={https://openreview.net/forum?id=AFDcYJKhND},
note={Featured Certification}
}

@inproceedings{softtopk2020,
 author = {Xie, Yujia and Dai, Hanjun and Chen, Minshuo and Dai, Bo and Zhao, Tuo and Zha, Hongyuan and Wei, Wei and Pfister, Tomas},
 booktitle = {Advances in Neural Information Processing Systems},
 editor = {H. Larochelle and M. Ranzato and R. Hadsell and M.F. Balcan and H. Lin},
 pages = {20520--20531},
 publisher = {Curran Associates, Inc.},
 title = {Differentiable Top-k with Optimal Transport},
 url = {https://proceedings.neurips.cc/paper_files/paper/2020/file/ec24a54d62ce57ba93a531b460fa8d18-Paper.pdf},
 volume = {33},
 year = {2020}
}

@InProceedings{Li_2026_CVPR,
    author    = {Li, Zhuojin and Cheng, Hsin-Pai and Cai, Hong and Han, Shizhong and Porikli, Fatih},
    title     = {CoReDiT: Spatial Coherence-Guided Token Pruning and Reconstruction for Efficient Diffusion Transformers},
    booktitle = {Proceedings of the IEEE/CVF Conference on Computer Vision and Pattern Recognition (CVPR) Workshops},
    month     = {June},
    year      = {2026},
    pages     = {3438-3447}
}

@inproceedings{
zhu2026diffsparse,
title={DiffSparse: Accelerating Diffusion Transformers with Learned Token Sparsity},
author={Haowei Zhu and Ji Liu and Ziqiong Liu and Dong Li and Jun-Hai Yong and Bin Wang and Emad Barsoum},
booktitle={The Fourteenth International Conference on Learning Representations},
year={2026},
url={https://openreview.net/forum?id=V3eUas3VCL}
}

@InProceedings{Shao_2026_CVPR,
    author    = {Shao, Tong and Fu, Yusen and Sun, Guoying and Kong, Jingde and Tian, Zhuotao and Su, Jingyong},
    title     = {SODA: Sensitivity-Oriented Dynamic Acceleration for Diffusion Transformer},
    booktitle = {Proceedings of the IEEE/CVF Conference on Computer Vision and Pattern Recognition (CVPR)},
    month     = {June},
    year      = {2026},
    pages     = {33012-33021}
}

@inproceedings{hypersd2024,
  title={{Hyper-SD}: Trajectory Segmented Consistency Model for Efficient Image Synthesis},
  author={Yuxi Ren and Xin Xia and Yanzuo Lu and Jiacheng Zhang and Jie Wu and Pan Xie and Xing Wang and Xuefeng Xiao},
  booktitle={Advances in Neural Information Processing Systems},
  volume={37},
  year={2024},
  doi={10.52202/079017-3725},
  url={https://proceedings.neurips.cc/paper_files/paper/2024/hash/d4e1c24ac41ff0b82ca1b171731f0b23-Abstract-Conference.html}
}

@inproceedings{gu2023mixofshow,
  title={{Mix-of-Show}: Decentralized Low-Rank Adaptation for Multi-Concept Customization of Diffusion Models},
  author={Yuchao Gu and Xintao Wang and Jay Zhangjie Wu and Yujun Shi and Yunpeng Chen and Zihan Fan and Wuyou Xiao and Rui Zhao and Shuning Chang and Weijia Wu and Yixiao Ge and Ying Shan and Mike Zheng Shou},
  booktitle={Advances in Neural Information Processing Systems},
  volume={36},
  year={2023},
  doi={10.52202/075280-0699},
  url={https://papers.nips.cc/paper_files/paper/2023/hash/3340ee1e4a8bad8d32c35721712b4d0a-Abstract-Conference.html}
}

@inproceedings{simsar2025loraclr,
  title={{LoRACLR}: Contrastive Adaptation for Customization of Diffusion Models},
  author={Enis Simsar and Thomas Hofmann and Federico Tombari and Pinar Yanardag},
  booktitle={Proceedings of the IEEE/CVF Conference on Computer Vision and Pattern Recognition},
  pages={13189--13198},
  year={2025},
  url={https://openaccess.thecvf.com/content/CVPR2025/html/Simsar_LoRACLR_Contrastive_Adaptation_for_Customization_of_Diffusion_Models_CVPR_2025_paper.html}
}
\bibliographystyle{tmlr}

\newpage
\appendix
\section{Method Analysis and Validation}
\label{app:derivations}

In this appendix, we derive the gradient estimators and validate their core approximations. Sections~\ref{app:soft_rank_jacobian} and~\ref{app:grad_derivation} analyze the soft-ranking Jacobian and residual-based estimator. Section~\ref{subsec:variance_analysis} analyzes stratified sampling, Section~\ref{app:synthetic_experiments} reports controlled synthetic validation, Section~\ref{app:real_block_validation} specifies the trained-block protocol, and Section~\ref{sec:topk_comparison} compares differentiable top-$k$ alternatives.

\subsection{Jacobian and Consistency of the Soft Ranking Matrix}
\label{app:soft_rank_jacobian}

The differentiability of our method hinges on the continuous mapping from scores to ranks. Here, we derive the Jacobian of the soft ranking operation to elucidate how gradients propagate through the sorting mechanism. 

Recall that the \textit{descending soft rank} $\tilde{r}_i$ of token $i$ is defined based on pairwise comparisons with all other tokens $j$:
\begin{equation}
    \tilde{r}_i(\mathbf{s}) = 1 + \sum_{j \neq i} \sigma(D_{ji}), \quad \text{where } D_{ji} = \frac{s_j - s_i}{\tau_{\text{rank}}}.
    \label{eq:appendix_soft_rank}
\end{equation}
In this descending formulation, $D_{ji} > 0$ implies $s_j > s_i$, contributing to a larger (worse) rank index for token $i$. Let $\pi_i = \sigma((k - \tilde{r}_i)/\tau_{\text{sel}})$ be the selection probability. To update the router scores $\mathbf{s}$, we must compute the gradient of the rank vector $\tilde{\mathbf{r}}$ with respect to the score vector $\mathbf{s}$.

Let $\sigma'(z) = \sigma(z)(1-\sigma(z))$ denote the derivative of the sigmoid function. We define the pairwise gradient scalar as $\delta_{ji} \triangleq \sigma'(D_{ji})$. The partial derivative of the rank $\tilde{r}_i$ with respect to an arbitrary score $s_m$ is:
\begin{equation}
    \frac{\partial \tilde{r}_i}{\partial s_m} = \sum_{j \neq i} \frac{\partial \sigma(D_{ji})}{\partial s_m} = \frac{1}{\tau_{\text{rank}}} \sum_{j \neq i} \delta_{ji} \cdot \frac{\partial (s_j - s_i)}{\partial s_m}.
\end{equation}
We analyze the two distinct cases for the index $m$:

\paragraph{Case 1: Self-Sensitivity ($m = i$).}
We consider the gradient of a token's rank with respect to its own score. In the summation over $j$, $s_i$ appears in every term as $-s_i$:
\begin{equation}
    \frac{\partial \tilde{r}_i}{\partial s_i} = \frac{1}{\tau_{\text{rank}}} \sum_{j \neq i} \delta_{ji} \cdot (-1) = -\frac{1}{\tau_{\text{rank}}} \sum_{j \neq i} \sigma'\left(\frac{s_j - s_i}{\tau_{\text{rank}}}\right).
\label{eq:self_sensitivity}
\end{equation}
Since $\sigma' > 0$, this diagonal term of the Jacobian is strictly negative. This mathematically confirms that increasing a token's score $s_i$ consistently decreases (improves) its rank index $\tilde{r}_i$.

\paragraph{Case 2: Cross-Sensitivity ($m \neq i$).}
We consider the gradient of token $i$'s rank with respect to another token $m$'s score. The summation over $j$ vanishes for all terms except where $j=m$:
\begin{equation}
    \frac{\partial \tilde{r}_i}{\partial s_m} = \frac{1}{\tau_{\text{rank}}} \cdot \delta_{mi} \cdot (1) = \frac{1}{\tau_{\text{rank}}} \sigma'\left(\frac{s_m - s_i}{\tau_{\text{rank}}}\right).
\end{equation}
This off-diagonal term is strictly positive, indicating that increasing the score of a competitor token $m$ increases (worsens) the rank of token $i$.

\paragraph{Global Gradient Flow.}
By combining these terms via the chain rule, the gradient of the loss $\mathcal{L}$ with respect to a specific score $s_m$ aggregates feedback from the entire sequence:
\begin{equation}
    \frac{\partial \mathcal{L}}{\partial s_m} = \sum_{i=1}^N \frac{\partial \mathcal{L}}{\partial \pi_i} \cdot \underbrace{\frac{\partial \pi_i}{\partial \tilde{r}_i}}_{\text{Selection}} \cdot \underbrace{\frac{\partial \tilde{r}_i}{\partial s_m}}_{\text{Ranking}}.
\end{equation}
This formulation highlights the \emph{context-aware} nature of our gradient estimation: updating $s_m$ not only affects the probability of token $m$ being selected (via $\frac{\partial \tilde{r}_m}{\partial s_m}$) but also impacts the selection probabilities of all other tokens $i$ (via $\frac{\partial \tilde{r}_i}{\partial s_m}$), thereby encouraging the router to learn a global sorting policy rather than independent scoring.
 
\subsection{Derivation of the Residual-Based Gradient Estimator}
\label{app:grad_derivation}

To enable end-to-end optimization of the router and ratio policy parameters despite the discrete hard selection, we employ a Straight-Through Estimator (STE) based on a differentiable surrogate graph.

\paragraph{Surrogate Input-Gating Model.}
We define the surrogate effective inputs for the selected path ($\mathbf{x}_i^{\text{sel}}$) and the rejected path ($\mathbf{x}_i^{\text{rej}}$) as functions of the inclusion score $\pi_i$. The selected path executes the residual block, whereas the rejected path follows the identity skip. The inclusion score $\pi_i$ serves as a continuous gate between these paths:
\begin{equation}
    \begin{aligned}
        \mathbf{x}_i^{\text{sel}} &= \pi_i \cdot \mathbf{x}_i, \\
        \mathbf{x}_i^{\text{rej}} &= (1 - \pi_i) \cdot \mathbf{x}_i.
    \end{aligned}
    \label{eq:surrogate_def}
\end{equation}
This complementary gating is a modelling choice rather than a derived identity. As $\pi_i$ moves from 0 to 1, it transfers the token continuously from the rejected path to the selected path, while the hard top-$k$ forward pass remains unchanged.
The resulting gradient directions are properties of this surrogate and are assessed empirically in App.~\ref{app:synthetic_experiments}.

\paragraph{Gradient Derivation.}
We derive the gradient of the loss $\mathcal{L}$ with respect to $\pi_i$ via the chain rule. By differentiating Eq.~\ref{eq:surrogate_def}, we observe that $\frac{\partial \mathbf{x}_i^{\text{sel}}}{\partial \pi_i} = \mathbf{x}_i$ and $\frac{\partial \mathbf{x}_i^{\text{rej}}}{\partial \pi_i} = -\mathbf{x}_i$. 
Let $\nabla^{\text{sel}}$ and $\nabla^{\text{rej}}$ denote the backpropagated gradients $\frac{\partial \mathcal{L}}{\partial \mathbf{x}^{\text{sel}}}$ and $\frac{\partial \mathcal{L}}{\partial \mathbf{x}^{\text{rej}}}$, respectively. We denote this two-path gradient of the surrogate by $g^*_i$, derived via the chain rule as:
\begin{equation}
    \begin{aligned}
        \frac{\partial \mathcal{L}}{\partial \pi_i} 
        &= \left\langle \frac{\partial \mathcal{L}}{\partial \mathbf{x}_i^{\text{sel}}}, \frac{\partial \mathbf{x}_i^{\text{sel}}}{\partial \pi_i} \right\rangle + \left\langle \frac{\partial \mathcal{L}}{\partial \mathbf{x}_i^{\text{rej}}}, \frac{\partial \mathbf{x}_i^{\text{rej}}}{\partial \pi_i} \right\rangle \\
        &= \left\langle \nabla_{\mathbf{x}_i}^{\text{sel}} \mathcal{L}, \; \mathbf{x}_i \right\rangle + \left\langle \nabla_{\mathbf{x}_i}^{\text{rej}} \mathcal{L}, \; -\mathbf{x}_i \right\rangle \\
        &= \left\langle \nabla_{\mathbf{x}_i}^{\text{sel}} \mathcal{L} - \nabla_{\mathbf{x}_i}^{\text{rej}} \mathcal{L}, \; \mathbf{x}_i \right\rangle.
    \end{aligned}
    \label{eq:ideal_grad}
\end{equation}

\paragraph{Stochastic Single-Path Estimator.}
In the actual training computation, deterministic top-$k$ selection activates exactly one path per token. Thus, only the corresponding term in Eq.~\ref{eq:ideal_grad} is observed at position $i$, while obtaining the counterfactual term would require evaluating the other path. We therefore use the observed term as the stochastic single-path estimate:
\begin{equation}
    \hat{g}_i = \frac{\partial \mathcal{L}}{\partial \pi_i} \approx
    \begin{cases}
        \left\langle \nabla_{\mathbf{x}_i}^{\mathrm{sel}} \mathcal{L}, \; \mathbf{x}_i \right\rangle & \text{if } i \in \mathcal{I}_{\mathrm{topk}}, \\
        -\left\langle \nabla_{\mathbf{x}_i}^{\mathrm{rej}} \mathcal{L}, \; \mathbf{x}_i \right\rangle & \text{if } i \notin \mathcal{I}_{\mathrm{topk}}.
    \end{cases}
    \label{eq:estimator}
\end{equation}
The implementation scatters the selected and rejected feature gradients back to their original token positions before taking a row-wise inner product with $\mathbf{X}$. This produces Eq.~\ref{eq:estimator} without evaluating the counterfactual path. The estimator is biased and is not intended to recover $g_i^*$ in expectation. Stratified ratio sampling and score perturbation vary the executed path across training. We assess the resulting directional agreement empirically in App.~\ref{app:synthetic_experiments}.

\paragraph{Gradient Flow to the Budget $k$.}
A key advantage of our differentiable sorting formulation is the explicit gradient flow through the continuous budget $k$ to the ratio policy. Applying the chain rule, the gradient for $k$ aggregates the sensitivities of all tokens:
\begin{equation}
    \frac{\partial \mathcal{L}}{\partial k} = \sum_{i=1}^{N} \frac{\partial \mathcal{L}}{\partial \pi_i} \cdot \frac{\partial \pi_i}{\partial k}.
\end{equation}
Since $\pi_i$ is a sigmoid function of $k$ (Eq.~\ref{eq:prob_select}), the term $\frac{\partial \pi_i}{\partial k}$ is strictly positive.
Consequently, the update direction for $k$ is determined by the positively weighted aggregate above: under gradient descent, a negative value increases $k$, whereas a positive value decreases it. Tokens for which greater inclusion reduces the task loss therefore contribute to increasing the budget. This task signal is combined at the ratio policy with the EMA budget loss in Eq.~\ref{eq:budget_loss}, which steers the long-run average retention toward $R_{\text{target}}$.

\paragraph{Differentiable Cost Control via $k$-Relaxation.}
Our formulation explicitly differentiates the inclusion scores with respect to the policy-predicted budget $k$. This propagates the task gradient through $k$ to the Ratio Policy Network, while hard top-$k$ enforces the predicted integer token count in the forward pass. Together with the EMA budget loss in Eq.~\ref{eq:budget_loss}, the policy learns how to distribute the target average retention ratio across layers and timesteps.

\paragraph{Analysis of Bias and Variance.}
We analyse $\hat{g}_i$ relative to $g_i^*$, the two-path gradient of the surrogate model in Eq.~\ref{eq:ideal_grad}. Since $g_i^*$ is a property of the gating model rather than ground truth, our experiments also compare the resulting directions with discrete marginals defined independently of the surrogate.

\begin{itemize}
    \item \textbf{Bias and Empirical Directional Alignment:}
    As a straight-through gradient estimator~\citep{ste2013}, $\hat{g}_i$ is biased by construction.
    The sign inversion maps rejected-path feedback to the same inclusion coordinate as selected-path feedback.
    On the synthetic residual task in Appendix~\ref{app:synthetic_experiments}, the single-path estimate has positive cosine similarity with the two-path surrogate across the evaluated budgets.

    \item \textbf{Variance and Control:}
    The single-path estimate varies with the sampled mask.
    Score perturbation and ratio sampling improve path coverage, while accumulation over multiple draws improves directional agreement (Table~\ref{tab:mc_accumulation}).
    Stratified timestep sampling (Sec.~\ref{subsec:training}; App.~\ref{subsec:variance_analysis}) further reduces the sampling variance of the budget signal.
\end{itemize}

\subsection{Variance Reduction Analysis of Stratified Sampling}
\label{subsec:variance_analysis}

In the main text (Section 3.5), we employ Stratified Timestep Sampling to stabilize the budget estimation. Here, we show that its variance is no greater than that of standard Uniform Sampling and is strictly lower whenever the stratum means differ.

\textbf{Problem Formulation.} 
Let $r(t): [0, T] \to [0, 1]$ denote the optimal pruning ratio at timestep $t$. In diffusion models, $r(t)$ exhibits significant temporal variation across the generation process. Our objective is to estimate the global expected pruning ratio $\mu$ over the entire diffusion process: $\mu = \mathbb{E}_{t \sim \mathcal{U}[0, T]}[r(t)]$.

\textbf{Uniform Sampling Estimator.} 
A standard mini-batch $\mathcal{B}_{uni}$ consists of $B$ independent samples drawn uniformly, i.e., $t_i \sim \mathcal{U}[0, T]$. The Monte Carlo estimator is $\hat{\mu}_{uni} = \frac{1}{B} \sum_{i=1}^B r(t_i)$. The variance of this estimator is:
\begin{equation}
    \text{Var}(\hat{\mu}_{uni}) = \frac{\sigma^2}{B},
\end{equation}
where $\sigma^2 = \text{Var}(r(t))$ represents the total variance of the pruning ratio across the entire timestep domain $[0, T]$.

\textbf{Stratified Sampling Estimator.} 
We partition the domain $[0, T]$ into $B$ disjoint strata intervals $S_1, \dots, S_B$, each of length $T/B$. We draw exactly one sample $t_j$ uniformly from each stratum $j$. The stratified estimator is $\hat{\mu}_{strat} = \frac{1}{B} \sum_{j=1}^B r(t_j)$.
According to the \textit{Law of Total Variance}, the total variance $\sigma^2$ can be decomposed into the average \textit{within-stratum} variance and the \textit{between-stratum} variance:
\begin{equation}
    \sigma^2 = \underbrace{\frac{1}{B}\sum_{j=1}^B \sigma_j^2}_{\text{Average Within-Stratum Var}} + \underbrace{\frac{1}{B}\sum_{j=1}^B (\mu_j - \mu)^2}_{\text{Between-Stratum Var}},
\end{equation}
where $\sigma_j^2$ is the variance within stratum $j$, and $\mu_j$ is the expected value within stratum $j$. Crucially, the variance of the Stratified Estimator is determined only by the within-stratum components:
\begin{equation}
    \text{Var}(\hat{\mu}_{strat}) = \frac{1}{B^2} \sum_{j=1}^B \sigma_j^2 = \frac{1}{B} \left( \frac{1}{B}\sum_{j=1}^B \sigma_j^2 \right).
\end{equation}

The reduction in variance achieved by Stratified Sampling is explicitly given by the difference between the two estimators:
\begin{equation}
    \Delta \text{Var} = \text{Var}(\hat{\mu}_{uni}) - \text{Var}(\hat{\mu}_{strat}) = \frac{1}{B} \underbrace{\left[ \frac{1}{B}\sum_{j=1}^B (\mu_j - \mu)^2 \right]}_{\text{Between-Stratum Variance}}.
\end{equation}
Since the term inside the square brackets is a sum of squares, it is non-negative. Whenever at least two stratum means differ, the term is positive and Stratified Sampling strictly reduces the estimator variance. Learned pruning schedules in DiTs vary across timesteps, so this condition is expected to hold in practice. Removing the between-stratum component reduces sampling noise and stabilizes the learnable budget.

\subsection{Validation on Synthetic Data}
\label{app:synthetic_experiments}

We use synthetic tasks to evaluate joint learning of token scores and budgets and to examine the two gradient approximations.

\begin{figure}[htbp]
    \centering
    \begin{subfigure}[t]{0.32\textwidth}
        \centering
        \includegraphics[width=\textwidth]{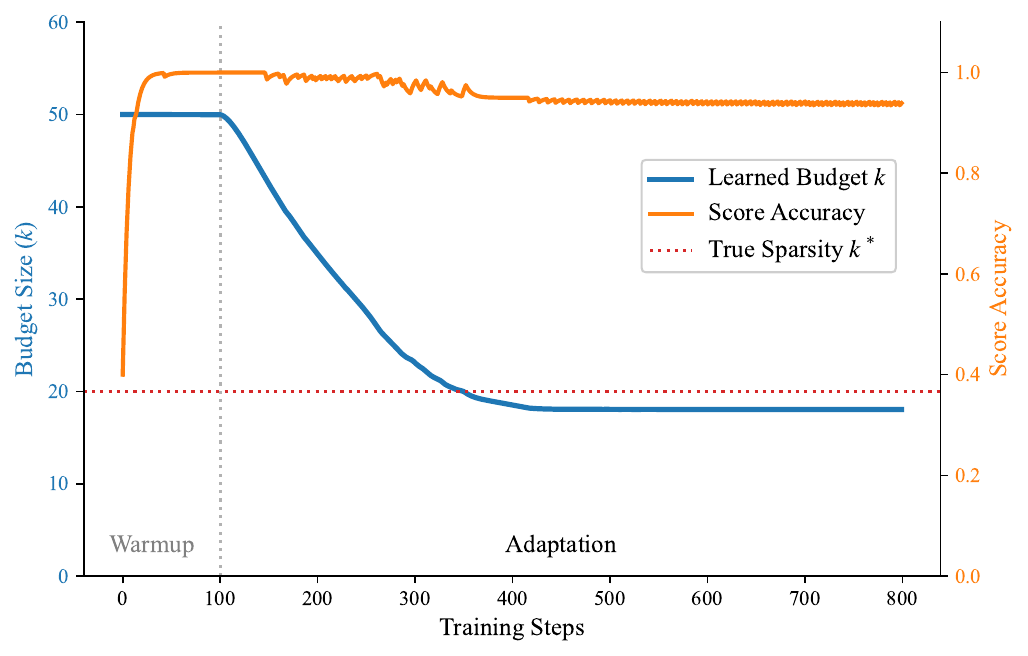}
        \caption{Budget and accuracy dynamics.}
        \label{fig:unified_dynamics}
    \end{subfigure}
    \hfill
    \begin{subfigure}[t]{0.32\textwidth}
        \centering
        \includegraphics[width=\textwidth]{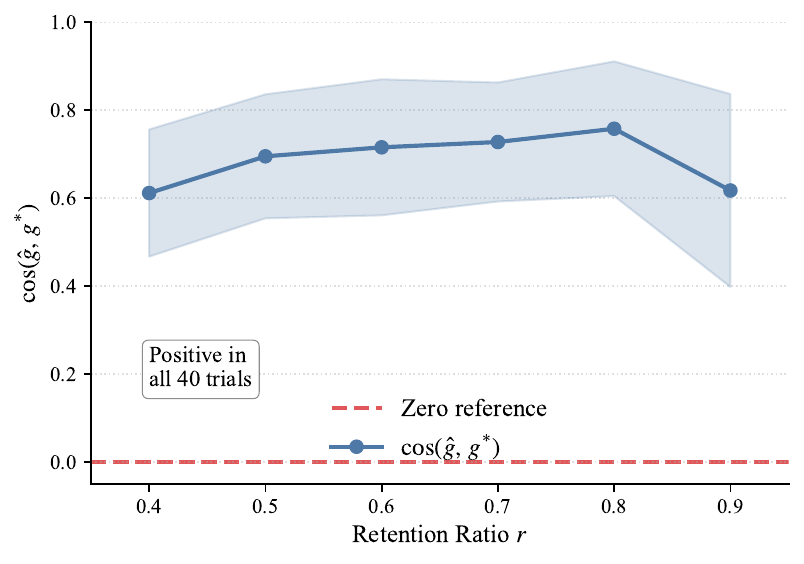}
        \caption{Single-path directional agreement across budgets.}
        \label{fig:single_path_fidelity}
    \end{subfigure}
    \hfill
    \begin{subfigure}[t]{0.32\textwidth}
        \centering
        \includegraphics[width=\textwidth]{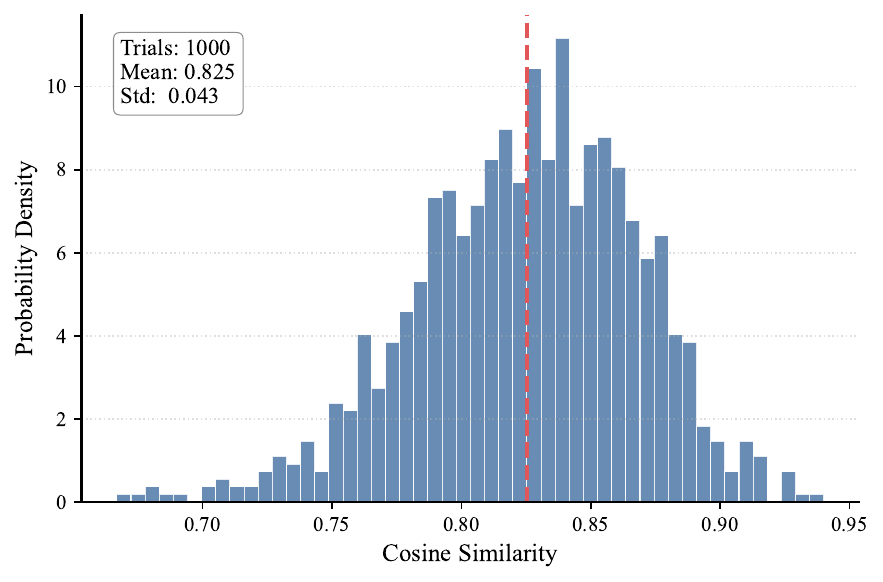}
        \caption{Soft-rank relaxation vs.\ inclusion contrast.}
        \label{fig:grad_consistency}
    \end{subfigure}
    \caption{Validation on synthetic data.
    (a) The learnable budget $k$ (blue) decays from 50 and stabilizes near the ground-truth sparsity $k^*=20$ (red dotted), while score accuracy (orange) stays $\approx 95\%$.
    (b) The single-path estimator has positive cosine similarity with the two-path surrogate at every evaluated retention ratio, with positive values in all trials.
    (c) The soft-rank relaxation retains a strongly positive alignment ($0.825 \pm 0.043$) with the coordinate-wise inclusion contrast.}
    \label{fig:synthetic_val}
\end{figure}

\paragraph{Joint Learning of Scoring and Budgeting.}
To verify the capability of the Shiva router to learn feature-aware ranking and adjust the token budget jointly, we constructed a synthetic dataset ($N=100, D=16$) comprising 20 high-magnitude signal tokens (mean 10.0) and 80 noise tokens ($k^*=20$). We initialized the budget at $k=50$ and employed a dual-optimizer strategy, using Adam~\citep{adam2015} ($\texttt{lr}=0.1$) for the router and SGD~\citep{sgd1951} ($\texttt{lr}=2.0$) for the budget $k$. The training process consisted of a 100-step warmup followed by a 700-step adaptation phase. As illustrated in Figure~\ref{fig:synthetic_val}(a), the router rapidly learns to identify signal tokens during the warmup. Subsequently, the budget $k$ automatically decays and stabilizes around $k \approx 18$. This equilibrium reflects the trade-off induced by the sparsity penalty ($\lambda=0.1$), while score accuracy remains approximately 95\%.

\paragraph{Decomposing the Estimator.}
The gradient that reaches the router traverses two distinct approximations, which we validate in isolation rather than through a single aggregate figure:
(i) the \emph{single-path drop}, whereby Eq.~\ref{eq:estimator} retains only the executed path of the two-path gradient (Eq.~\ref{eq:ideal_grad}); and
(ii) the \emph{soft-rank relaxation}, whereby the estimator of Eq.~\ref{eq:estimator}, defined with respect to the gating value as $\hat{\mathbf{g}} = \partial \mathcal{L} / \partial \boldsymbol{\pi}$, is propagated to the router logits by the chain rule $\nabla_{\mathbf{s}} \mathcal{L} = (\partial \boldsymbol{\pi} / \partial \mathbf{s})^\top \hat{\mathbf{g}}$, where the Jacobian $\partial \boldsymbol{\pi} / \partial \mathbf{s}$ follows from the soft-rank definition in Eq.~\ref{eq:appendix_soft_rank} and the selection probability in Eq.~\ref{eq:prob_select}.
We assess the single-path drop against the corresponding two-path surrogate and compare the propagated directions with discrete targets that do not depend on the gating model.

\paragraph{(i)~Directional Agreement with the Two-Path Surrogate.}
To evaluate the single-path approximation, we construct a residual task in which rejected tokens continue to influence the loss through the identity skip:
\begin{equation}
    \mathbf{y}_i(\mathcal{S}) =
    \begin{cases}
        \mathbf{x}_i + \mathrm{Block}(\mathbf{X}_{\mathcal{S}})_i & \text{if } i \in \mathcal{S},\\
        \mathbf{x}_i & \text{otherwise},
    \end{cases}
    \qquad
    \mathcal{L}(\mathcal{S}) = \left\| \mathbf{y}(\mathcal{S}) - \mathbf{y}_{\text{full}} \right\|^2,
    \label{eq:residual_toy}
\end{equation}
where $\mathrm{Block}$ is a frozen attention-plus-FFN block and $\mathbf{y}_{\text{full}}$ its unpruned output, mirroring the Stage-1 distillation objective ($N=100$, $D=16$). Since hard selection executes only one path per token, the counterfactual half of Eq.~\ref{eq:ideal_grad} is recovered from an additional forward pass with that token's membership flipped. As shown in Figure~\ref{fig:synthetic_val}(b), averaged over 40 trials the single-path estimator aligns with the two-path gradient at every operating point, with $\cos(\hat{g}, g^*)$ ranging from $0.61$ to $0.76$ and remaining strictly positive in \emph{all} trials. Thus the executed path provides a useful direction without evaluating both paths of Eq.~\ref{eq:ideal_grad} within a training step.

This directional agreement strengthens under the stochastic training protocol. Accumulating $\hat{g}$ and $g^*$ across draws with randomised budgets and perturbed scores (Sec.~\ref{subsec:training}) improves their alignment over the single-draw value (Table~\ref{tab:mc_accumulation}).

\begin{table}[t]
    \centering
    \small
    \caption{Monte Carlo accumulation under the stochastic training protocol. Averaging over draws improves alignment between the single-path estimate and the two-path direction.}
    \begin{tabular}{lccccc}
        \toprule
        Accumulated draws & 1 & 2 & 5 & 10 & 25 \\
        \midrule
        $\cos(\overline{\hat{g}},\, \overline{g^*})$
            & 0.695 & 0.781 & 0.799 & 0.799 & \alignbold{0}{811} \\
        \bottomrule
    \end{tabular}
    \label{tab:mc_accumulation}
\end{table}

\paragraph{(ii)~Directional Agreement with the Inclusion Contrast.}
We assess the second approximation via a Monte Carlo simulation over $1{,}000$ independent trials. The reference is the exact coordinate-wise inclusion contrast of the task objective,
\begin{equation}
    \Delta_i^{\mathrm{coord}} = \mathcal{L}(\mathbf{m}_{-i},m_i=1) - \mathcal{L}(\mathbf{m}_{-i},m_i=0),
    \label{eq:coordinate_marginal}
\end{equation}
where $\mathbf{m}_{-i}$ is held fixed while token $i$ is included or excluded. For $\mathcal{L} = \sum_{i \in \mathcal{I}_{\text{topk}}} \mathbf{1}^\top \mathbf{x}_i - \sum_{i \notin \mathcal{I}_{\text{topk}}} \mathbf{1}^\top \mathbf{x}_i$, this contrast is $\Delta_i^{\mathrm{coord}} = 2\,\mathbf{1}^\top \mathbf{x}_i$. The implementation uses $\mathbf{1}^\top \mathbf{x}_i$ because positive scaling does not affect cosine similarity. Figure~\ref{fig:synthetic_val}(c) shows a strictly positive distribution with a mean of $0.825 \pm 0.043$. The corresponding measurement on the residual task of Eq.~\ref{eq:residual_toy} yields $0.724 \pm 0.118$.

\subsection{Trained-Block Validation Protocol}
\label{app:real_block_validation}

The trained-block experiment reported in Table~\ref{tab:estimator_real} uses layer 1 of a trained SD3-Medium model at denoising step 2, with $N{=}4096$ image tokens and $k{=}1900$. We use the real latent and Stage-1 distillation objective. The exact fixed-budget marginal is evaluated on 512 of the 4096 tokens, and the 50 draws perturb scores and randomise budgets as during training. Appendix~\ref{app:synthetic_experiments} separately reports the controlled synthetic experiments used to isolate the two estimator approximations across budgets.

\paragraph{Extension Across Layers, Timesteps, and Budgets.}
We extend this diagnostic around the central setting (layer 11, denoising step 14, retention $0.6$), varying depth over layers 1, 11, and 23, denoising step over 2, 14, and 26, and retention over $0.4$, $0.6$, and $0.8$ one axis at a time. This gives seven distinct settings. Each uses three real inputs (seeds 42, 43, and 44) and ten score draws per input, for 30 draws per setting and 210 in total. Real block inputs are captured with the original 28-step SD3 sampling configuration (CFG 7, $1024\times1024$, FP16). The frozen-block diagnostic runs in FP32 with TF32 disabled, using $N=4096$ and 512 fixed-budget swap probes per draw. Cosine similarity compares each surrogate direction with the local fixed-budget swap-contrast direction; the positive fraction counts draws with cosine greater than zero.

\begin{table}[htbp]
\centering
\small
\setlength{\tabcolsep}{5pt}
\caption{Extended trained-block directional diagnostics. Cosines are mean $\pm$ standard deviation over 30 draws per setting; $>0$ is the positive-alignment percentage. Indices are zero-based.}
\label{tab:estimator_extended}
\begin{tabular}{rrrcccc}
\toprule
 & & & \multicolumn{2}{c}{Residual STE} & \multicolumn{2}{c}{Gumbel-Sigmoid STE} \\
\cmidrule(lr){4-5}\cmidrule(lr){6-7}
Layer & Step & Retention & Cosine & $>0$ (\%) & Cosine & $>0$ (\%) \\
\midrule
1  & 14 & 0.60 & $\phantom{-}0.158\pm0.315$ & 63.33 & $0.174\pm0.355$ & 63.33 \\
11 & 2  & 0.60 & $-0.053\pm0.408$ & 40.00 & $0.042\pm0.462$ & 63.33 \\
11 & 14 & 0.40 & $\phantom{-}0.291\pm0.151$ & 100.00 & $0.070\pm0.527$ & 56.67 \\
11 & 14 & 0.60 & $\phantom{-}0.298\pm0.184$ & 93.33 & $0.169\pm0.405$ & 70.00 \\
11 & 14 & 0.80 & $\phantom{-}0.230\pm0.141$ & 96.67 & $0.081\pm0.333$ & 76.67 \\
11 & 26 & 0.60 & $\phantom{-}0.010\pm0.317$ & 43.33 & $0.142\pm0.372$ & 70.00 \\
23 & 14 & 0.60 & $\phantom{-}0.142\pm0.270$ & 66.67 & $0.162\pm0.392$ & 56.67 \\
\bottomrule
\end{tabular}
\end{table}

Directional agreement varies with layer, denoising step, and retention ratio. At layer 11 and step 14, the residual surrogate has mean cosine similarities between 0.230 and 0.298 across the three retention ratios. At the same layer, the mean is $-0.053$ at step 2 and 0.010 at step 26.

\subsection{Comparison of Differentiable \texorpdfstring{Top-$k$}{Top-k} Methods}
\label{sec:topk_comparison}

Token pruning fundamentally requires selecting a discrete subset of tokens, an operation that is non-differentiable and thus incompatible with end-to-end learning.
Various methods have been proposed to relax this discrete selection into a differentiable surrogate.
Table~\ref{tab:topk_comparison} and Algorithm~\ref{alg:softrank} in the main paper summarize the compared mechanisms and the soft-rank computation used by our backward surrogate. The forward pass still executes ordinary hard top-$k$ indexing.

\paragraph{NeuralSort.}
NeuralSort~\citep{neuralsort2019} constructs a continuous relaxation of the permutation matrix $\hat{P} \in \mathbb{R}^{B \times N \times N}$ via pairwise score comparisons.
In its original formulation, the sorted feature sequence is obtained as $\hat{P} \cdot \mathbf{x}$, which incurs $O(N^2 D)$ computational cost and is intractable for high-resolution vision tasks.
We adapt it to token selection via STE~\citep{ste2013}: since the soft permutation produces weighted mixtures rather than clean token selections, we apply hard top-$k$ indexing in the forward pass and use the soft gates from $\hat{P}$ only to route gradients back to the score network.
However, the full $N \times N$ permutation matrix must still be materialized in GPU memory during training, resulting in $O(N^2)$ memory consumption that becomes a bottleneck at high resolutions.
Our method confines the $O(N^2)$ pairwise computation to the backward pass and operates on scalar scores. The forward pass uses hard top-$k$ indexing without constructing a soft permutation matrix.
Furthermore, $k$ remains a fixed hyperparameter with no mechanism for gradient-based adaptation.

\paragraph{SOFT Top-$k$.}
SOFT top-$k$~\citep{softtopk2020} casts selection as an optimal transport problem between the $N$ scores and two destination points representing the selected and rejected sets, with the marginal fixed at $k$.
The solution is obtained by entropic regularisation and Sinkhorn iterations, and gradients are taken analytically through the resulting fixed point rather than by unrolling the solver.
This gives a smooth relaxation with $O(N)$ memory and a deterministic count, so unlike DyDiT it does not produce ragged tensors, and unlike NeuralSort it never materialises an $N \times N$ matrix.
Its cost, however, is governed by the number of Sinkhorn iterations rather than by the sequence length: the operator takes essentially the same time at $N{=}1024$ as at $N{=}4096$ (Table~\ref{tab:topk_measured}), and at $4.0$~ms per call it dominates a diffusion backbone that invokes selection 672 times per image.
Our formulation replaces the iterative solve with a single pass of pairwise comparisons over scalar scores, which is quadratic in arithmetic but executes as one kernel and is confined to the backward pass.

\paragraph{DyDiT.}
DyDiT~\citep{dydit2024} models token selection as a set of independent binary decisions using a Gumbel-Sigmoid estimator~\citep{gumbel2017}.
During training, each token's logit is perturbed by Gumbel noise and passed through a hard sigmoid via STE~\citep{ste2013}.
While this achieves differentiability with only $O(N)$ memory, each token is scored independently without a global ranking constraint.
As a result, the number of active tokens varies unpredictably across samples and timesteps, so the realized budget is only met in expectation and the cost of a given input is not known before execution.
The absence of a strict budget also makes it impossible to provide theoretical guarantees on FLOPs reduction.
In our extension of DyDiT to SD3-Medium with classifier-free guidance (CFG), we further observe that the original implementation assumes batch size 1: the active mask is computed as a union across the batch (\texttt{any(dim=0)}), which can increase the actual number of retained tokens beyond that selected for any individual sample.

\paragraph{DiffCR.}
DiffCR~\citep{diffcr2025} introduces a learnable compression ratio by interpolating between two discrete ratio bins at training time, enabling the budget $k$ to vary across layers via gradient descent.
The ratio is discretized into a finite set of bins (e.g., with a step size of 0.1), and the gradient for the ratio parameter is the slope of a linear interpolation between two neighboring bins.
The router itself receives gradients by multiplying token scores onto the FFN output post-hoc, decoupling the gradient path from the selection operation, in contrast to NeuralSort which routes gradients via STE directly through selection.
The bin interpolation requires two backbone forward passes per training step, one at each neighboring bin.
In our benchmark, the measured iteration time is $1.65\times$ that of Vanilla (Appendix~\ref{app:training_efficiency}).

\paragraph{Shiva-DiT.}
Our method combines hard top-$k$ execution with single-pass learning of token scores and the budget $k$.
In the forward pass, tokens are selected via a standard \texttt{argsort} and \texttt{gather}, which require $O(N\log N)$ time and $O(N)$ auxiliary memory in standard implementations and produce static tensor shapes amenable to CUDA Graph compilation.
In the backward pass, we reconstruct a differentiable surrogate graph over the scalar importance scores $\mathbf{s} \in \mathbb{R}^{B \times N}$ using the soft rank in Eq.~\ref{eq:rank}. Here, $\tilde{s} = s + \epsilon$, $\epsilon \sim \mathcal{N}(0, \sigma_{\mathrm{noise}}^2)$.
This operation is $O(N^2)$ but acts only on the scalar score vector, without the $D$-dimensional feature factor present in NeuralSort.
The soft rank is then used to define a continuous inclusion probability $\pi_i = \sigma\!\left((k - \tilde{r}_i) / \tau_\text{sel}\right)$, through which gradients flow to both the router logits and the budget $k$ itself, since $\partial \pi_i / \partial k \neq 0$.
The residual-based STE (Eq.~\ref{eq:grad_injection_main} in the main paper) uses the loss sensitivity of each token's executed path to construct its surrogate gradient, without an additional backbone pass.

\begin{table}[htbp]
\centering
\footnotesize
\setlength{\tabcolsep}{5pt}
\caption{Measured cost of one selection operator on an RTX 4090 (dim $1536$, batch $2$, $r=0.6$, FP32). NeuralSort (soft) multiplies $\hat{P}\mathbf{x}$; NeuralSort (STE) uses hard indexing and routes gradients through $\hat{P}$. Retained activation storage includes input/score leaf clones and selected outputs. MB denotes $10^6$ bytes.}
\label{tab:topk_measured}
\begin{tabular}{lrrrr}
\toprule
Method & $N$ & Forward (ms) & Backward (ms) & Activation (MB) \\
\midrule
NeuralSort (soft) & 1024 & 0.442 & 0.856 & 54.6 \\
                 & 2048 & 1.624 & 3.117 & 167.9 \\
                 & 4096 & 9.283 & 17.066 & 570.6 \\
\midrule
NeuralSort (STE) & 1024 & 0.476 & 1.137 & 58.8 \\
                & 2048 & 1.232 & 2.265 & 172.9 \\
                & 4096 & 7.622 & 15.135 & 580.8 \\
\midrule
SOFT top-$k$ & 1024 & 4.119 & 11.277 & 28.6 \\
             & 2048 & 4.006 & 10.951 & 57.2 \\
             & 4096 & 4.006 & 11.427 & 114.4 \\
\midrule
Shiva & 1024 & 0.133 & 0.825 & 20.2 \\
      & 2048 & 0.133 & 0.817 & 40.3 \\
      & 4096 & 0.264 & 3.384 & 80.6 \\
      & 8192 & 0.542 & 12.667 & 161.7 \\
      & 16384 & 0.981 & 46.652 & 323.4 \\
\bottomrule
\end{tabular}
\end{table}

In the $N=1024$--4096 comparison in Table~\ref{tab:topk_measured}, Shiva has the lowest measured forward and backward times. Its retained activation storage grows approximately linearly, whereas the NeuralSort variants retain quadratic-size tensors. This retained storage does not represent peak backward memory: Shiva reconstructs quadratic pairwise comparisons during the backward pass.

The extended measurements use the same selector timer with five warmup iterations and ten timed repetitions, $\tau_{\mathrm{rank}}=\tau_{\mathrm{sel}}=10$, and FP32 features and scores. Forward time is the mean wall-clock duration, and backward time is the complete forward--backward duration minus forward time. All three sizes complete on an RTX 4090 with 24 GB of memory. As $N$ grows from 4096 to 16384, retained activation storage rises from 80.6 to 323.4 MB, while backward time rises from 3.384 to 46.652 ms. Thus the training-time pairwise operation remains a quadratic scaling constraint even though inference uses only hard selection. These are selector microbenchmarks, not end-to-end high-resolution training measurements.

\FloatBarrier
\section{Experimental Setup}
\label{app:implementation_details}

\subsection{Data and Evaluation Protocol}
\label{app:evaluation_protocol}
We randomly partition MJHQ-30K into 28{,}000 training images and 2{,}000 validation images, using the corresponding validation prompts for evaluation. These 2{,}000 prompts are disjoint from the training prompts. For evaluation, we generate one image per validation prompt under each of five fixed seeds, with prompts and initial latents matched across methods within each architecture. FID is computed once on the pooled 10{,}000 generated images using \texttt{pytorch-fid} with 2{,}048-dimensional Inception features and reference statistics from all 30{,}000 original MJHQ-30K images. ImageReward, CLIP Score, and CLIP-IQA are each averaged over the same 10{,}000 generated images. The per-seed analysis in Appendix~\ref{app:multiseed} is a separate variability check.

\subsection{Training Details}
\label{app:training_details}

To ensure the stability of the coupled optimization problem involving the discrete router ranking, the continuous ratio policy, and the generative backbone, we employ a staged training process. This curriculum learning strategy decouples the optimization targets to allow each component to initialize properly before joint optimization.

\paragraph{Stage 1: Router Warmup.}
In the initial phase, the \emph{Ratio Policy} network is bypassed. To prevent the router from overfitting to a single fixed sparsity level and to encourage it to learn a global ranking capability, we sample the target retention ratio $r$ from a stratified uniform distribution $r \sim \mathcal{U}(r_{\min}, r_{\max})$ as described in Sec.~\ref{subsec:training}. The router is forced to adapt to these varying random budgets, thereby learning to discriminate token importance across the entire sparsity spectrum.

\paragraph{Stage 2: Policy Learning.}
Once the router has converged, we activate the \emph{Ratio Policy} network. 
In this stage, the task objective guides budget allocation across layers and timesteps, while the \emph{EMA-guided proxy budget loss} (Eq.~\ref{eq:budget_loss}) aligns the average token retention ratio with the target.
We employ a global EMA estimator for the budget constraint with a momentum coefficient of $\beta = 0.2$ to ensure stability against batch-wise variance.

Crucially, to prevent the policy from collapsing into a trivial solution (e.g., a flat uniform ratio) due to the strict budget penalty, we introduce a \emph{Logit Noise Injection} mechanism. We add zero-mean Gaussian noise $\epsilon \sim \mathcal{N}(0, 1)$ to the output logits of the policy network before the sigmoid activation:
\begin{equation}
    r_{t,l} = \sigma(\text{logit}(r_{t,l}) + \epsilon).
\end{equation}
This perturbation promotes stochastic exploration of the ratio space and smooths the Stage 2 optimization objective. Because the noise passes through nonlinear transformations, its gradient is not generally an unbiased estimator of the deterministic objective. We therefore use it only as an exploration mechanism during Stage 2. The noise is removed in Stage 3, where the deterministic policy is jointly tuned with the backbone.

\paragraph{Stage 3: Joint Tuning.}
In the final stage, the exploration noise is removed. We unfreeze the backbone adaptation parameters, such as LoRA, and jointly fine-tune the Router, Policy, and Backbone. The objective is to minimize the diffusion reconstruction loss while maintaining the established budget constraints. This stage allows the backbone to adapt to the sparse token flow and recover any fine-grained details lost due to pruning.

\paragraph{Hyperparameters and Optimization.}
All models are trained using the AdamW~\citep{adamw2019} optimizer with mixed-precision (\texttt{bfloat16}) to ensure memory efficiency. We utilize a global effective batch size of 16 (micro-batch size 1 per device).
The learning rates are set to $3 \times 10^{-4}$ for the Router and $1 \times 10^{-4}$ for the Ratio Policy.
For $\mu_{\text{global}}^{(u)}$ in Eq.~\ref{eq:ema_avg}, we set $\beta$ to $0.2$.
For efficient backbone adaptation, we employ DoRA~\citep{dora2024} with a rank of $r=32$, applied to all linear projections within the Attention and FFN blocks.
The \textit{Finetuned} baseline is a dense LoRA-finetuned model trained with the same data, backbone adaptation configuration, training duration, and reconstruction and distillation losses and weights as Shiva-DiT's backbone fine-tuning.
A linear warmup of 200 steps is applied at the beginning of each training stage.
To support Classifier-Free Guidance~\citep{cfg2022} (CFG), we randomly drop text captions with a probability of $p=0.1$. During inference, we use 50 sampling steps with CFG scales of 7.0 for SD3-Medium and 4.5 for PixArt. Flux.1-dev uses a guidance scale of 3.5 through its guidance-distilled conditioning, without a separate unconditional CFG branch. We additionally bypass token reduction in the first DiT block, as empirically this stabilizes importance scoring in subsequent layers with negligible overhead.

\paragraph{Temperature Annealing with Normalized Sorting.} In our implementation, we normalize the sigmoid logit by sequence length $N$ to improve training stability across resolutions; this is equivalent to an effective selection temperature of $N\tau_{\text{sel}}$. The selection probability is defined in Eq.~\ref{eq:prob_select}.
To sharpen the soft inclusion probabilities used in the backward surrogate, we employ a deterministic annealing schedule: the ranking temperature $\tau_{\text{rank}}$ decays from 0.2, and the selection temperature $\tau_{\text{sel}}$ decays from 0.1, gradually hardening the probability distribution over the course of training.

\paragraph{Normalized Feature Distillation.}
For intermediate layer distillation, we observed that standard MSE loss is sensitive to the varying feature magnitudes across depths, often necessitating heuristically small loss weights in prior works~\citep{dydit2024}.
To stabilize optimization, we propose \textit{Normalized Feature Distillation}, which applies Layer Normalization before the distance calculation:
\begin{equation}
    \mathcal{L}_{\text{distill}} = \sum_{l} \left\| \text{LayerNorm}(\mathbf{h}_l^{\text{Student}}) - \text{LayerNorm}(\mathbf{h}_l^{\text{Teacher}}) \right\|^2_2.
    \label{eq:distill_cal}
\end{equation}
This standardization ensures that the distillation gradient is driven by semantic misalignment rather than numerical scale differences.
For comprehensive configuration files, please refer to the attached source code.

\subsection{Implementation Details}
\label{app:implementation_framework}

To ensure reproducibility and facilitate future research, we developed a unified and modular codebase based on the \texttt{diffusers} library~\citep{diffusers2022}. Below we detail the policy network, engineering framework, and model-specific adaptations.

\paragraph{Adaptive Ratio Policy Network.}
The policy network employs a decoupled additive architecture comprising two parallel branches. 
The \textit{Layer Branch} maps the layer index to a hidden embedding ($\mathbb{R}^{256}$) via a learnable lookup table, followed by a 2-layer MLP (Linear$\to$SiLU$\to$Linear) to predict the spatial logit. 
Simultaneously, the \textit{Time Branch} processes the sinusoidal timestep embedding through a mirrored MLP structure with a hidden dimension of 256. 
The final retention ratio $r_{t,l}$ is obtained by summing the logits from both branches with a learnable global bias parameter $b_{\text{anchor}}$. 
For training stability, $b_{\text{anchor}}$ is initialized to the inverse sigmoid of the target initial ratio (e.g., $\sigma^{-1}(0.6)$) to ensure the training starts from a valid operating point.

\paragraph{Unified Hook-Based Framework.}
Our \emph{plug-and-play} framework separates token reduction from the backbone without modifying model source code. A centralized \textit{Hook Manager} injects operations before and after Attention and FFN and at DiT block boundaries. The shared interface applies our method and baselines to Stable Diffusion 3~\citep{sd32024} (SD3-Medium), PixArt-$\Sigma$~\citep{pixartsigma2024}, and Flux.1-dev~\citep{flux2024}, with lightweight architecture-specific handling. Separating the pruning policy from execution also supports rapid prototyping of efficient diffusion algorithms.

\paragraph{Model-Specific Adaptations.}
We address two critical implementation details regarding positional embeddings and guidance strategies:
\begin{itemize}
    \item \textbf{Rotary Positional Embeddings (RoPE):} Handling spatial information varies by architecture. For models like SD3-Medium and PixArt, where absolute positional embeddings are added to the tokens prior to the transformer blocks, no additional handling is required as the spatial information is carried within the token features. 
    However, for models like Flux which apply RoPE at every attention layer, simply discarding tokens breaks the geometric correspondence. In our implementation, we gather the corresponding cosine and sine components of the RoPE cache based on the selected token indices ($\mathcal{I}_{\text{topk}}$). We note that exploring more sophisticated, topology-aware RoPE selection strategies remains a promising direction for future work.
    
    \item \textbf{Guidance Consistency Strategy:} Following the CFG token-alignment strategy used in ToMeSD~\citep{tomesd2023} and DaTo~\citep{dato2024}, we select the same spatial token positions for the conditional and unconditional branches in each block. Our shared selection applies hard top-$k$ to the element-wise maximum of the two branches' scores:
    \begin{equation}
        \mathbf{s}_{\text{unified}} = \max(\mathbf{s}_{\text{cond}}, \mathbf{s}_{\text{uncond}}).
    \end{equation}
\end{itemize}

\subsection{Baseline Configurations and Realized Budgets}
\label{app:baseline_configs}

\paragraph{Baseline Reproduction and Configuration.}
To ensure a fair comparison, we integrated all baseline methods into our unified framework, strictly following their official implementations and hyperparameters. Each training-based baseline is trained for 30 epochs on the same training data, following its original training configuration, including LoRA. Training-based baselines use their respective official objectives (e.g., distillation for DyDiT~\citep{dydit2024}, standard MSE for DiffCR~\citep{diffcr2025}), while Shiva-DiT follows the staged protocol described in Appendix~\ref{app:training_details}.

\paragraph{Baseline Configurations and Realized Compute.}
Table~\ref{tab:baseline_settings} lists the settings and realized FLOPs for the baselines in Table~\ref{tab:main_results}. Table~\ref{tab:budget_sweep_settings} gives the operating points in Figure~\ref{fig:effi_scatter}. A merging ratio specifies the fraction reduced in the enabled operations; DiffCR's scale and DyDiT's threshold are control parameters, not retention ratios.

\begin{table}[htbp]
\centering
\footnotesize
\setlength{\tabcolsep}{4pt}
\caption{SD3-Medium baseline configurations for Table~\ref{tab:main_results}. FLOPs are in T per denoising step. SA, CA, and FFN denote self-attention, cross-attention, and feed-forward computation; layer and step indices are zero-based.}
\label{tab:baseline_settings}
\begin{tabular}{lp{0.70\linewidth}r}
\toprule
Method & Configuration & FLOPs \\
\midrule
ToMeSD & Merge ratio: $0.2$; stride: $(2,2)$; merging: SA, FFN; CA: off. & 8.98 \\
ToFu & Merge ratio: $0.2$; stride: $(2,2)$; reduction: SA, FFN; CA: off; pruning: first 12 layers; MLERP: remaining layers. & 8.98 \\
SDTM & Stride: $(2,2)$; switch-step fraction: $0.4$; initial ratio: $0.5$; deviation: $0.2$; $\alpha_s=0.05$; MCW: $0.2$; protected steps: every third; schedule: cosine. & 8.83 \\
ToMA & Merge ratio: $0.3$; tiles: $8\times8$; skipped layers: 0--4; merging: SA, FFN; CA: off. & 8.23 \\
IBTM & Merge ratio: $0.2$; layers: 6--18; dense step: 0; reduction: SA, FFN; CA: off; pool: $0.4$; destination ratio: $0.25$; pruning-step fraction: $0.12$. & 10.18 \\
NeuralSort & Temperature: $1.0$; target retention: $0.85$. & 9.47 \\
DiffCR & Retention bins: $1.0,0.9,\ldots,0.1$; bin selection: nearest; scale: $1.0$; dense layers: native. & 9.93 \\
DyDiT & Dynamic width: off; dynamic tokens: on; threshold: $0.5$; temperature: $5.0$; token selection: FFN; attention: dense. & 9.42 \\
SparseDiT & PoolingFormer: layer 0; SDTM: layers 1--22; $r_{\min}=0.1$, $r_{\max}=0.3$. & 8.95 \\
\bottomrule
\end{tabular}
\end{table}

\begin{table}[htbp]
\centering
\small
\setlength{\tabcolsep}{4pt}
\caption{Baseline operating points in Figure~\ref{fig:effi_scatter} on SD3-Medium ($N=4096$, 50 steps). Retention is the realized mean image-token retention across layers and steps. DyDiT has input-dependent FFN counts; its realized compute is reported as FLOPs. $^\ast$Original Table~\ref{tab:main_results} configuration.}
\label{tab:budget_sweep_settings}
\begin{tabular*}{\linewidth}{@{\extracolsep{\fill}}llrrr@{}}
\toprule
Method & Control parameter & Retention (\%) & FLOPs (T) & FID \\
\midrule
DiffCR & Scale $1.00^{\ast}$ & 87.509 & 9.93 & 17.51 \\
       & Scale $0.85$ & 82.088 & 9.20 & 66.15 \\
       & Scale $0.70$ & 75.002 & 8.31 & 132.84 \\
\midrule
DyDiT & Threshold $0.25$ & --- & 9.67 & 13.43 \\
      & Threshold $0.50^{\ast}$ & --- & 9.42 & 15.29 \\
      & Threshold $0.75$ & --- & 9.09 & 18.96 \\
\midrule
IBTM & Merging ratio $0.20^{\ast}$ & 89.386 & 10.18 & 19.96 \\
     & Merging ratio $0.40$ & 78.772 & 8.77 & 33.38 \\
     & Merging ratio $0.60$ & 68.158 & 7.56 & 62.33 \\
\bottomrule
\end{tabular*}
\end{table}

\paragraph{Realized Shiva Budgets.}
Table~\ref{tab:shiva_realized_budgets} reports the deployed budgets at $N=4096$ over 50 denoising steps. Configuration names such as Shiva-60\% identify operating points; their realized retention is reported explicitly. The all-layer mean includes the dense first block, while the active-layer mean includes only prunable blocks (SD3: 1--23; PixArt: 1--27; Flux: 1--56). Targets are expressed as all-layer means. An active-layer target $R_{\mathrm{active}}$ is converted to $[100+(L-1)R_{\mathrm{active}}]/L$, where $R_{\mathrm{active}}$ is in percent and $L$ is the number of blocks. SD3 counts are computed from its deployed integer LUT; Flux and PixArt counts are reconstructed from their BF16 ratio policies and sampling timesteps. Hard top-$k$ executes the resulting integer count at each layer and timestep; the global average need not equal the requested target exactly.

\begin{table}[htbp]
\centering
\small
\setlength{\tabcolsep}{4pt}
\caption{Realized Shiva budgets. Retention and targets are percentages. $\Delta$ is all-layer actual retention minus the equivalent all-layer target, in percentage points. Mean $k$ includes all layers.}
\label{tab:shiva_realized_budgets}
\begin{tabular*}{\linewidth}{@{\extracolsep{\fill}}llrrrrr@{}}
\toprule
Model & Configuration & Target & Actual & Active layers & Mean $k$ & $\Delta$ \\
\midrule
SD3 & Shiva-60\% & 66.458 & 65.196 & 63.683 & 2670.45 & $-1.262$ \\
SD3 & Shiva-70\% & 70.000 & 69.705 & 68.388 & 2855.11 & $-0.295$ \\
SD3 & Shiva-80\% & 80.833 & 80.756 & 79.919 & 3307.75 & $-0.078$ \\
SD3 & Shiva-90\% & 90.000 & 89.328 & 88.864 & 3658.86 & $-0.672$ \\
PixArt-$\Sigma$ & Shiva & 90.357 & 87.858 & 87.408 & 3598.66 & $-2.499$ \\
Flux.1-dev & Shiva & 80.351 & 78.384 & 77.998 & 3210.61 & $-1.967$ \\
\bottomrule
\end{tabular*}
\end{table}

\paragraph{Additional Backbone Configurations.}
The quantitative results for Flux.1-dev and PixArt-$\Sigma$ are reported in Table~\ref{tab:generalization_arch_main} of the main paper.
For Flux.1-dev, ToMA uses a merging ratio of $0.5$, $16\times16$ tiles, and a merge-step interval of 2. It skips the first five double-stream blocks and the first five single-stream blocks. The configuration enables self-attention merging with parallel mode, with cross-attention and separate FFN merging disabled; the maximum text sequence length is 512. For PixArt-$\Sigma$, IBTM uses a merging ratio of $0.2$ in layers 7--20, enabling self-attention and cross-attention merging while leaving the FFN unmerged. Its pool parameter is $0$, destination ratio is $0.25$, and pruning-timestep fraction is $0.2$. Layer indices are zero-based. Both configurations use 50 steps, $1024\times1024$ images, and BF16, with the evaluation protocol in Appendix~\ref{app:evaluation_protocol}.

IBTM uses conditional/unconditional CFG information for token scoring. The guidance-distilled Flux setup used here does not execute those two branches, so we use ToMA as its compatible baseline; PixArt retains IBTM.

\clearpage
\section{Additional Experimental Analysis}
\label{app:extra_exp}

\subsection{Router and Policy Analysis}
\label{app:more_ablation}

\subsubsection{Impact of Training Strategy on Ratio Generalization}
\label{app:diff_strategy}

\begin{figure}[H] \centering \begin{subfigure}[b]{0.49\textwidth}
        \centering
        \includegraphics[width=\linewidth]{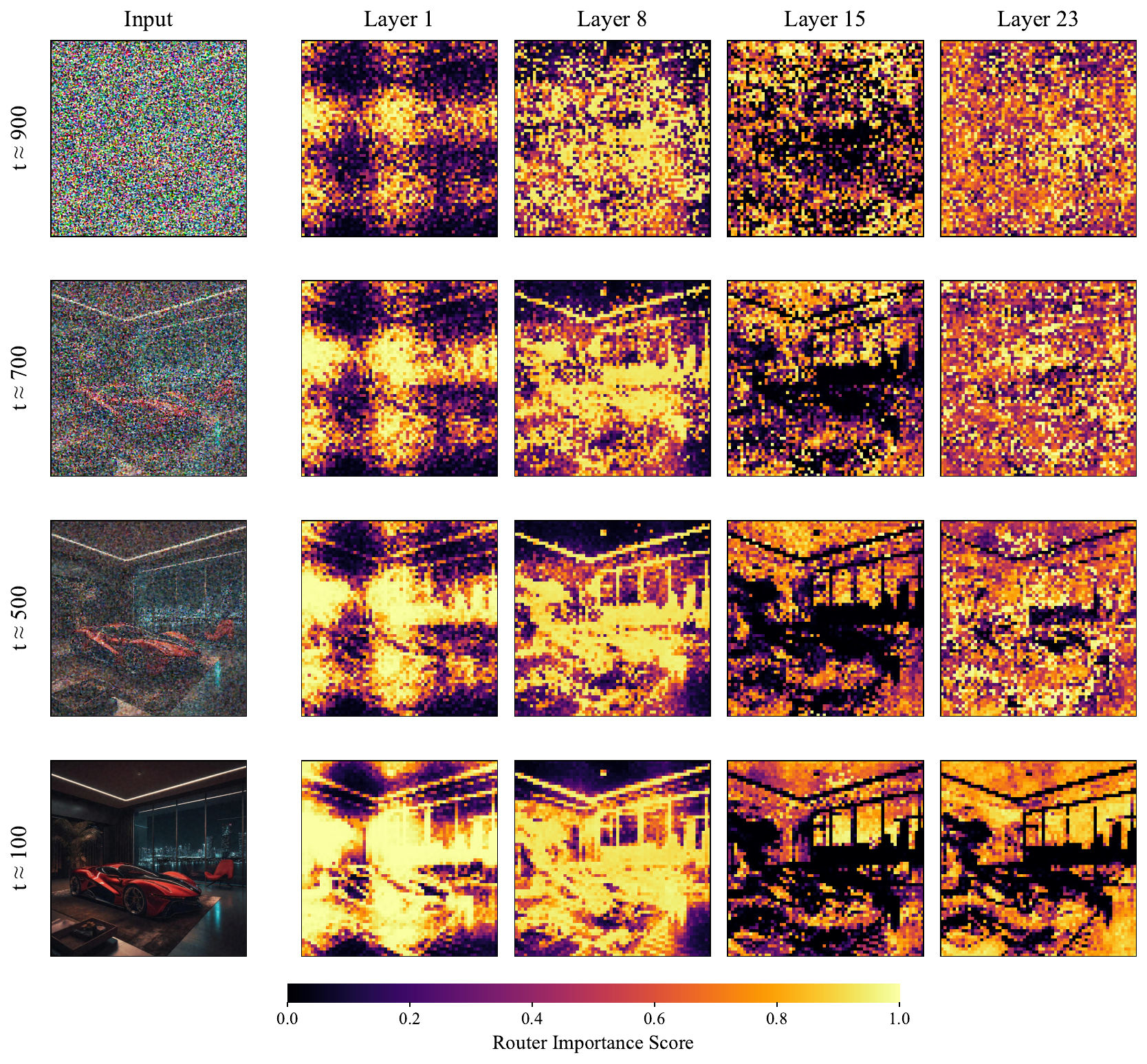}
        \caption{Heatmap (Ours / Stratified)}
        \label{fig:heatmap_ours}
    \end{subfigure}
    \hfill
    \begin{subfigure}[b]{0.49\textwidth}
        \centering
        \includegraphics[width=\linewidth]{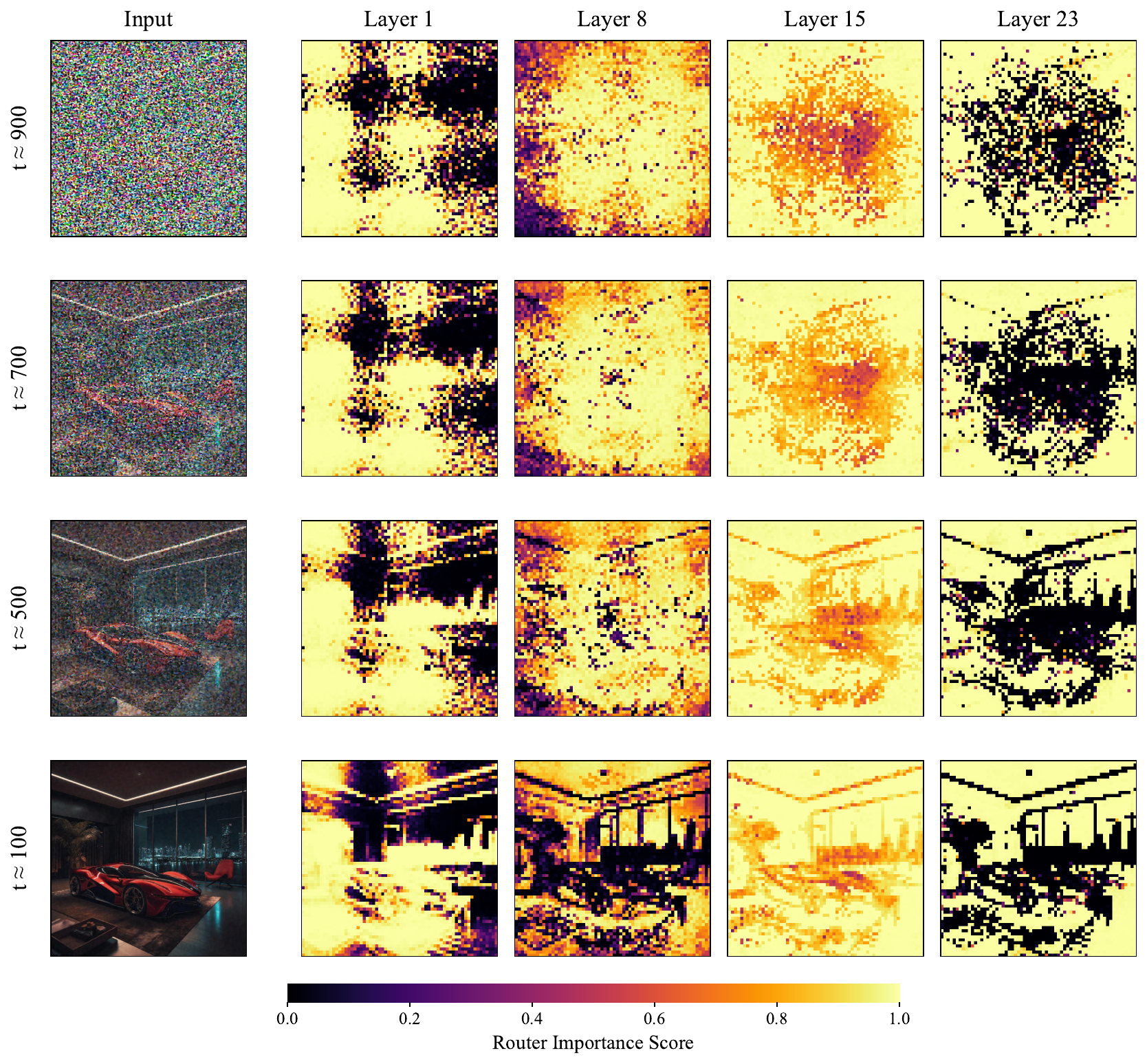}
        \caption{Heatmap (Fixed)}
        \label{fig:heatmap_fixed}
    \end{subfigure}

    \vspace{0.3cm}

    \begin{subfigure}[b]{0.49\textwidth}
        \centering
        \includegraphics[width=\linewidth]{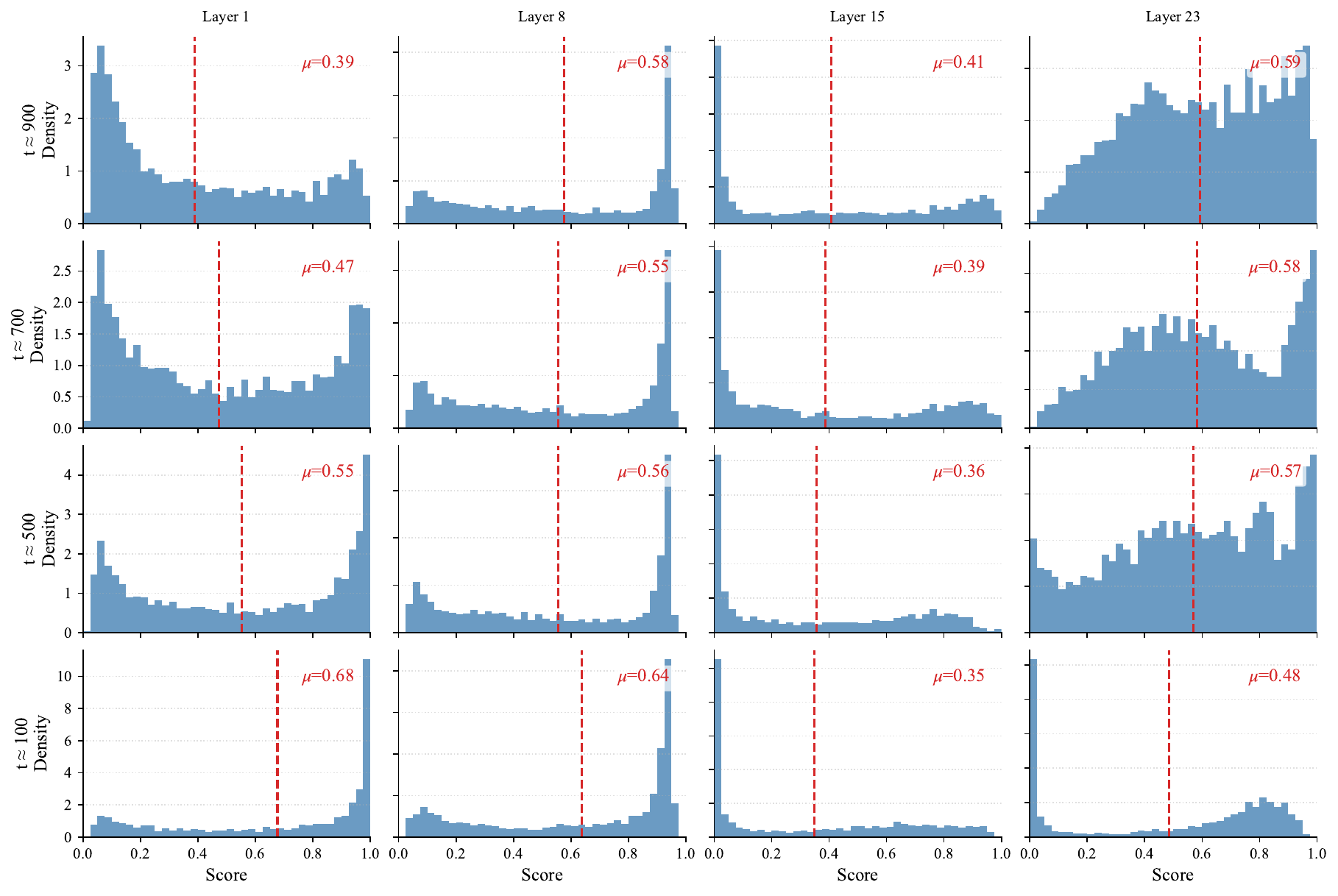}
        \caption{Histogram (Ours / Stratified)}
        \label{fig:hist_ours}
    \end{subfigure}
    \hfill
    \begin{subfigure}[b]{0.49\textwidth}
        \centering
        \includegraphics[width=\linewidth]{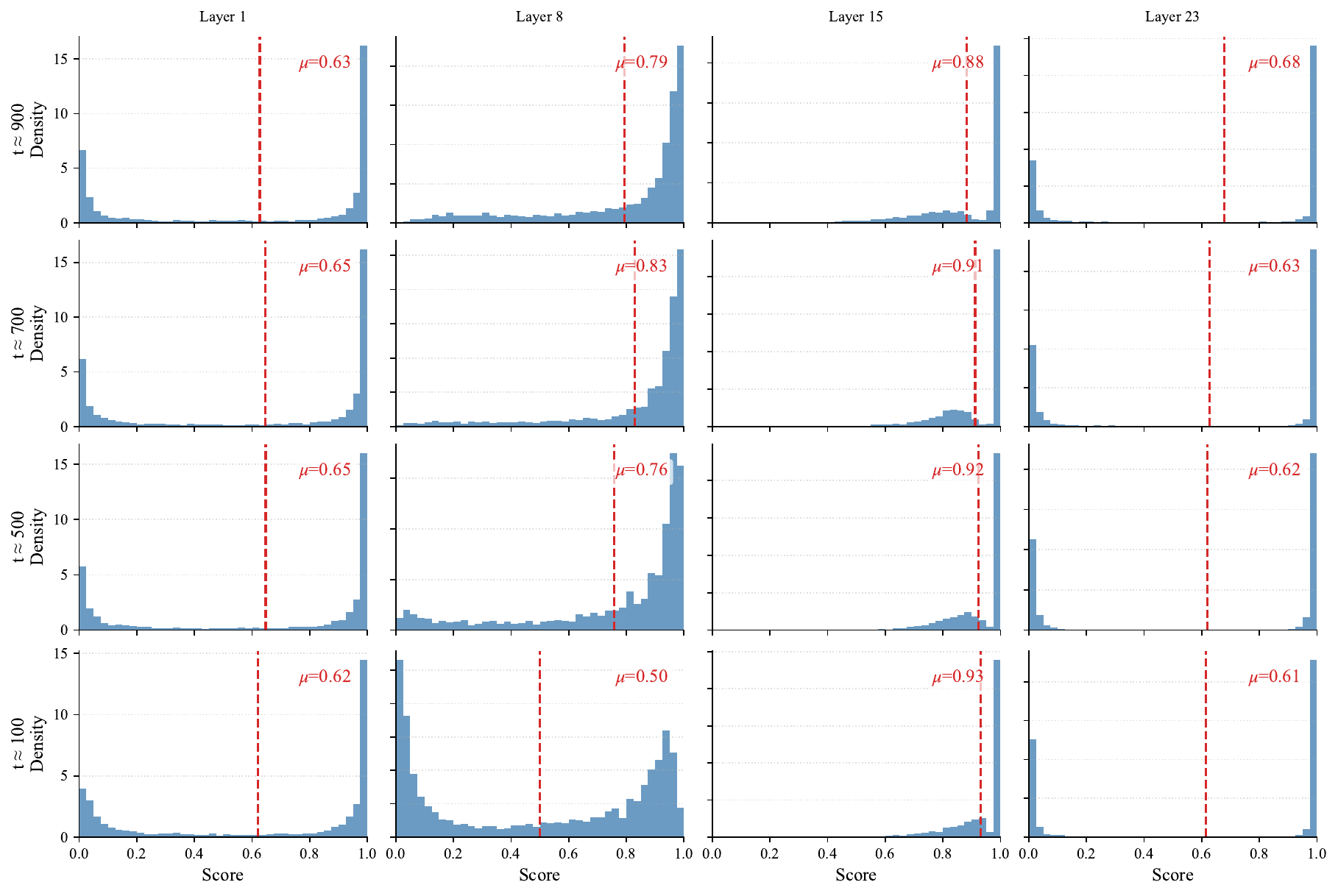}
        \caption{Histogram (Fixed)}
        \label{fig:hist_fixed}
    \end{subfigure}
    
    \caption{Comparison of Router Training Strategies: Top row (a, b) displays spatial importance heatmaps; Bottom row (c, d) shows the corresponding score distributions. Training with our proposed Stratified Sampling (Left column) exposes the router to varying sparsity levels, resulting in a more dispersed score distribution with a broader dynamic range. This indicates stronger discriminative power between informative and non-informative tokens. In contrast, the Fixed Ratio baseline (Right column) optimizes for a single sorting threshold, leading to a narrow, clustered distribution with limited differentiation.}
    \label{fig:router_comparison_grid}
\end{figure}

Figure~\ref{fig:router_comparison_grid} provides detailed spatial and score-distribution evidence for the ratio-generalization discussion in Section~\ref{sec:ablation}. The stratified and fixed-ratio routers use the same architecture and differ only in training-budget sampling. The heatmaps compare spatial selections, while the histograms expose the broader score range learned under stratification.

\subsubsection{Router Context and Token Selection}
\begin{figure}[H]
    \centering
    \includegraphics[width=0.6\linewidth]{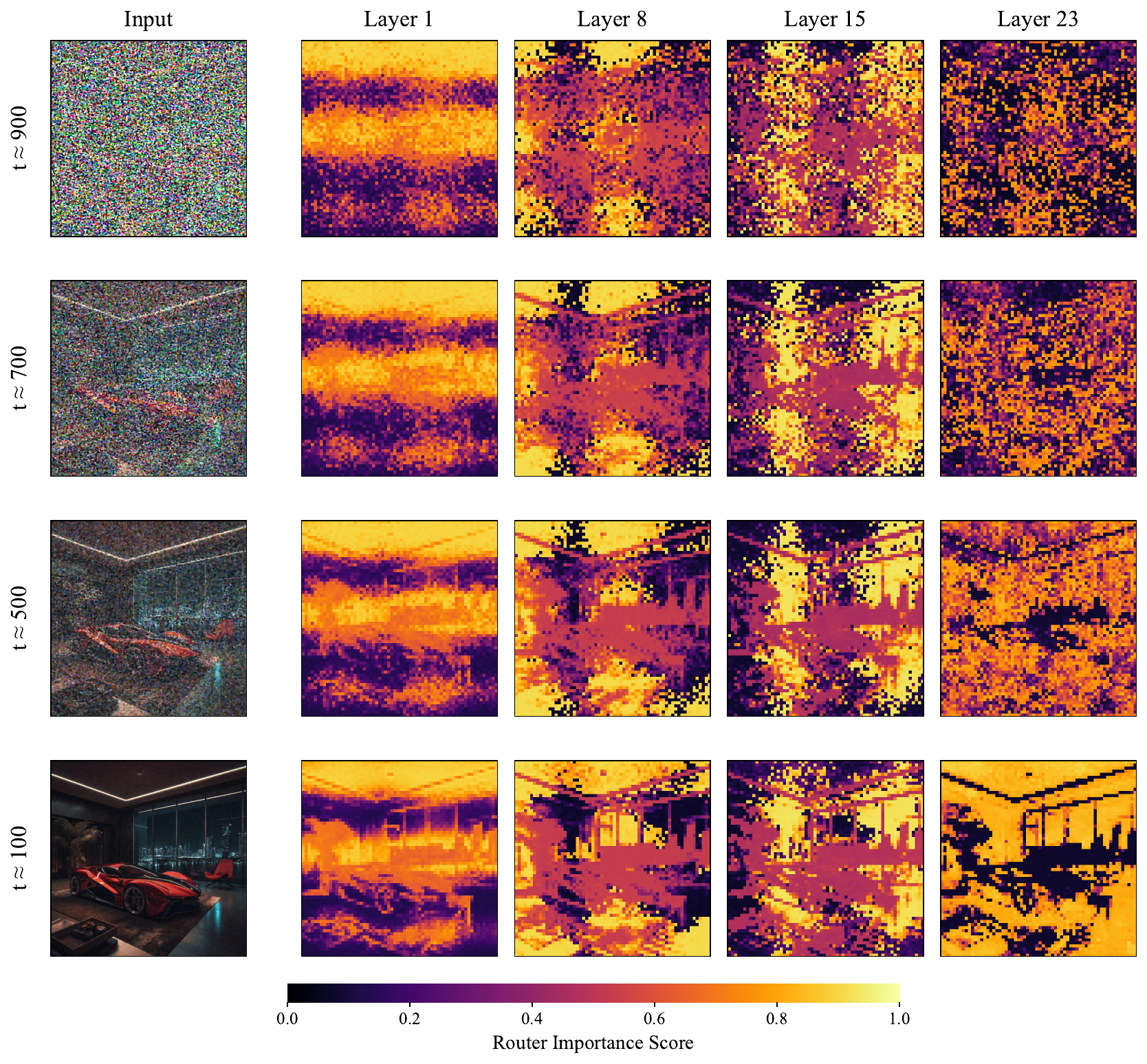}
    \caption{Learned masks from the pairwise (Group-12) router.}
    \label{fig:router_viz}
\end{figure}

The context-input ablation is reported alongside the sharing analysis in Table~\ref{tab:ablation_router}. Figure~\ref{fig:router_viz} complements it by showing that the pairwise router concentrates active tokens on semantic regions and texture-rich areas across layers and timesteps.

\subsubsection{Additive versus Multiplicative Coupling}
SparseDiT~\citep{sparsedit2025} constructs its schedule from a depth profile and a temporal ramp. The depth profile applies pooling in the bottom layers, mixes sparse and dense tokens in the middle layers, and retains dense computation in the top layers. The temporal ramp increases the token count during denoising. Our additive policy in Eq.~(10) learns the depth and temporal components, and its learned temporal branch increases monotonically over denoising steps. We also examine multiplicative coupling of these components.

The multiplicative term $(1+\gamma)\Phi_{\text{layer}}$ has a scale symmetry: multiplying $\Phi_{\text{layer}}$ by $c$ and dividing $1+\gamma$ by $c$ leaves the logit unchanged. The factors are therefore not individually identifiable from the logit. In the trained FiLM policy, the mean of $\gamma$ is $17.68$ and $\|\Phi_{\text{layer}}\|=0.063$, giving a product magnitude of approximately $1.18$. The corresponding gradient factors are $\partial\,\text{logit}/\partial\Phi_{\text{layer}}=1+\gamma\approx18.7$ and $\partial\,\text{logit}/\partial\gamma=\Phi_{\text{layer}}\approx0.063$, a ratio of $296$. For the additive form, both derivatives are one.

Separate learning rates rescale the coordinate directions but do not remove this scale symmetry. The level set is a hyperbola, so its tangent direction changes with the factor values. Changes in $\gamma$ also change the gradient scale for $\Phi_{\text{layer}}$. Weight decay favors balanced factor scales, but its estimated cumulative effect over the evaluated training schedule is three orders of magnitude too small to correct the imbalance.

These observations motivate the additive form, which trains reliably in our experiments. Normalizing $\Phi_{\text{layer}}$ is a possible way to constrain the scale of the multiplicative variant.

\subsection{Budget Transfer}
\label{app:budget_transfer}
\label{app:generalization}
 
The learned ratio policy is compiled into a Look-Up Table (LUT) indexed by timestep $t$ and layer $\ell$. To transfer to a different budget, we adjust the global bias $b_\text{anchor}$ in Eq.~(10) before the sigmoid and compile the resulting retention ratios into an updated LUT. This requires no additional training or policy-network evaluation during inference.
 
During warmup, Stratified Ratio Sampling (Section~\ref{subsec:training}) exposes the Context-Aware Router to retention ratios across $[r_\text{min},r_\text{max}]$. The router thus learns token rankings under multiple budgets. Appendix~\ref{app:diff_strategy} compares the resulting score distributions and token selections with those from fixed-ratio training.
 
Concretely, in the main experiments we train only two checkpoints 
(Shiva-80\% and Shiva-60\%) and derive two additional operating 
points by adjusting $b_\text{anchor}$:
\begin{itemize}
    \item \textbf{Shiva-90\%}: transferred from Shiva-80\% by 
    increasing $b_\text{anchor}$. Achieves FID 12.15.
    \item \textbf{Shiva-70\%}: transferred from Shiva-60\% by 
    increasing $b_\text{anchor}$. Achieves FID 15.74 at 
    $1.37\times$ speedup.
\end{itemize}
All four operating points lie on the quality--latency Pareto frontier among the evaluated configurations; 
full quantitative results are reported in Table~\ref{tab:main_results}.

\subsection{Efficiency Analysis}

\subsubsection{Training Efficiency}
\label{app:training_efficiency}

To substantiate our claims regarding training efficiency, we compare memory footprint, throughput, and latency across representative pruning methods with different budget mechanisms. All models are evaluated on the MJHQ-30K dataset using 8$\times$ NVIDIA H200 GPUs with a global batch size of 16.

Table~\ref{tab:train_cost} in the main paper reports the measured memory and time-per-iteration comparison. The corresponding throughputs are 1.86, 1.05, 1.43, 1.98, 1.13, and 1.73 iterations/s for Vanilla, NeuralSort, DyDiT, SparseDiT, DiffCR, and Shiva, respectively.

Table~\ref{tab:train_cost} compares training costs and support for a learnable token budget $k$. NeuralSort, DyDiT, and SparseDiT all forgo learnable $k$: NeuralSort's hard integer slicing severs gradient flow to the budget; DyDiT relies on threshold-based masking that produces unpredictable token counts; and SparseDiT fine-tunes a manually specified sparse architecture with a prescribed timestep-wise pruning-rate schedule. Among these methods, NeuralSort and DyDiT incur $1.77\times$ and $1.30\times$ training cost, respectively. SparseDiT instead reduces the cost to $0.94\times$ because its sparse-dense modules execute fewer tokens during fine-tuning as well as inference. The layer allocation and token-count range are fixed design choices rather than an end-to-end optimized budget; in our unified SD3-Medium setting, this operating point increases FID from 16.86 for Vanilla to 25.71 in Table~\ref{tab:main_results}. Among the compared baselines, DiffCR supports learnable $k$ through two backbone passes. Its measured iteration time is $1.65\times$ that of Vanilla, and its peak VRAM is 66.3 GB, the highest among the compared methods.

Shiva-DiT integrates $k$ into a continuous soft inclusion function and learns the budget using one backbone pass. Its iteration time is $1.07\times$ that of Vanilla, with peak VRAM of 47.6 GB compared with 51.5 GB for Vanilla. By training directly at the compressed token capacity $k$, the primary activation footprint is bounded to $\mathcal{O}(k)$. The auxiliary overhead of our differentiable top-$k$ is further limited to an $\mathcal{O}(N^2)$ scalar soft-rank matrix computed over scores rather than features, which comfortably fits within the recovered memory.

\subsubsection{Inference Overhead}
\label{app:overhead_anatomy}

The main paper reports the net speedup. Table~\ref{tab:phase_breakdown} accounts separately for the cost of the pruning machinery itself, timed inside the hook with CUDA events over the 23 pruned layers of a 50-step generation at $1024\times1024$.

\begin{table}[h]
\centering
\caption{Per-image cost of each stage of the pruning path for SD3-Medium, in milliseconds, measured on an RTX 4090.}
\begin{tabular}{lcccc}
\toprule
Stage & Shiva-90\% & Shiva-80\% & Shiva-70\% & Shiva-60\% \\
\midrule
Ratio lookup (LUT) & 1.5 & 1.4 & 1.5 & 1.5 \\
Router             & 79.4 & 79.0 & 79.1 & 78.4 \\
Sorting            & 23.8 & 23.9 & 24.0 & 23.7 \\
Gather             & 42.5 & 42.7 & 42.6 & 42.3 \\
Scatter            & 42.8 & 45.6 & 49.4 & 50.7 \\
\midrule
Total overhead     & 190.0 & 192.7 & 196.5 & 196.6 \\
Attention and FFN  & 6293.1 & 5707.9 & 4969.2 & 4602.7 \\
\bottomrule
\end{tabular}
\label{tab:phase_breakdown}
\end{table}

The device-side pruning stages contribute approximately $190$ to $197$~ms per image and vary only mildly across retention ratios. The router and sorting operate on the full token set, while the gather and scatter operations are largely dominated by kernel launch and memory access costs at this scale. Scatter becomes modestly more expensive at lower retention ratios because more skipped positions must be restored. In contrast, the attention and FFN cost decreases substantially from $6293.1$~ms to $4602.7$~ms as the token budget is reduced. Therefore, the reduction in backbone computation considerably outweighs the additional pruning cost, with larger savings at more aggressive operating points.

The speedup is not confined to single-image generation. At $1024\times1024$, processing two images together reduces the per-image latency of Shiva-60\% by a further $9.5\%$, from $5667$ to $5127$~ms, while the per-image latency of the baseline is unchanged. These batch-size measurements were collected on a separate workstation with a different CPU from the main latency benchmark, so absolute values should be compared only within this experiment. Batching therefore widens the margin rather than eroding it.

\subsubsection{Compilation and Graph Capture}
\label{app:compilation}

We compare eager and compiled execution to evaluate the effect of graph optimization on the latency gains.
Table~\ref{tab:compilation} repeats the comparison with \texttt{torch.compile} enabled (Inductor backend, \texttt{mode=`default'}), on SD3-Medium at $1024\times1024$ with 50 sampling steps in fp16 on a single RTX 4090. These measurements were collected on a different workstation from the main latency results in Table~\ref{tab:main_results}; absolute latencies should therefore be compared only within each table, not across the two tables. Shiva-60\% remains the fastest configuration in both eager and compiled execution. Relative to eager Vanilla on this workstation, compiled Shiva-60\% achieves a combined $1.49\times$ speedup; under a compiled-to-compiled comparison, it is $1.29\times$ faster than compiled Vanilla.

\begin{table}[h]
\centering
\caption{End-to-end latency with and without \texttt{torch.compile}, measured on a different workstation from Table~\ref{tab:main_results}. Speedups are relative to eager Vanilla on this workstation; absolute latencies should be compared only within this table.}
\begin{tabular}{lcccc}
\toprule
\multirow{2}{*}{Method} & \multicolumn{2}{c}{Eager} & \multicolumn{2}{c}{Compiled} \\
\cmidrule(lr){2-3} \cmidrule(lr){4-5}
 & ms & Speedup & ms & Speedup \\
\midrule
Vanilla                & 7912.7 & $1.00\times$ & 6823.1 & $1.16\times$ \\
DyDiT~\cite{dydit2024} & 6959.4 & $1.14\times$ & 5818.0 & $1.36\times$ \\
\midrule
Shiva-90\%             & 7315.3 & $1.08\times$ & 6903.6 & $1.15\times$ \\
Shiva-80\%             & 6740.8 & $1.17\times$ & 6330.9 & $1.25\times$ \\
Shiva-70\%             & 5974.1 & $1.33\times$ & 5626.1 & $1.41\times$ \\
Shiva-60\%             & 5625.5 & $1.41\times$ & 5304.7 & $\mathbf{1.49\times}$ \\
\bottomrule
\end{tabular}
\label{tab:compilation}
\end{table}

A stricter form of static execution is CUDA Graph capture, which requires every tensor shape to be fixed before the graph is recorded. A deterministic budget satisfies this along the entire pruning path: the retention count depends only on the layer and the timestep, so scoring, selection, gather, the block itself and the scatter all have known shapes, and a whole denoising step can be recorded as a single graph. Threshold-based selection implemented with data-dependent \texttt{nonzero()} cannot be captured directly in this form because its output length varies with the input.

Table~\ref{tab:cudagraph} evaluates graph replay under host contention by varying the share of a CPU core available to the process.

\begin{table}[h]
\centering
\caption{One denoising step of SD3-Medium at $1024\times1024$ captured as a single CUDA graph, covering the router, the selection, the gather, the block and the scatter. The schedule is 24 layers over 28 steps with per-layer budgets taken from the Shiva-60\% table. This separate host-contention stress test is not used for the 50-step end-to-end latency comparison above. Measured on an RTX 4090 with an Intel Xeon Gold 6430; host share is varied by co-scheduling competing processes on the same core.}
\begin{tabular}{lccc}
\toprule
Host share & Eager (ms) & Captured (ms) & Speedup \\
\midrule
$1$   & 1385.7 & 1374.0 & $1.01\times$ \\
$1/2$ & 1587.1 & 1383.8 & $1.15\times$ \\
$1/4$ & 3384.8 & 1378.4 & $2.46\times$ \\
$1/8$ & 6706.3 & 1372.7 & $\mathbf{4.89\times}$ \\
\bottomrule
\end{tabular}
\label{tab:cudagraph}
\end{table}

Captured latency varies by less than one percent across the tested CPU shares, while eager latency increases as the available CPU time decreases. Eager execution issues approximately 27,000 kernels per image, whereas the captured schedule uses 28 graph launches. Graph replay therefore reduces host dispatch overhead. On the unloaded host, the measured latencies are similar; the benefit increases under contention. The predetermined layer and timestep budgets provide the fixed tensor shapes required for capture. Recording the 28 graphs takes $6.0$~s once, after which they are replayed for each image.

\subsection{Run-to-Run Variability}
\label{app:multiseed}

To establish that the margins in Table~\ref{tab:main_results} exceed sampling noise, each configuration is regenerated under five random seeds with the prompt set held fixed, and every seed is scored independently. Within a configuration the seed-to-seed standard deviation of FID ranges from $0.07$ to $0.27$, and that of CLIP Score from $0.03$ to $0.07$.

Because the same prompts and seeds are used for every configuration, the comparison can be made pairwise, which reduces variability due to differences in prompts and sampling seeds. Table~\ref{tab:multiseed} reports the per-seed difference between Shiva-90\% and the dense finetuned reference.

\begin{table}[h]
\centering
\caption{Per-seed difference between Shiva-90\% and \textit{Finetuned}. Negative $\Delta$FID and positive $\Delta$CLIP favour Shiva-90\%.}
\begin{tabular}{lcc}
\toprule
Seed & $\Delta$FID & $\Delta$CLIP \\
\midrule
0 & $-0.741$ & $+0.081$ \\
1 & $-0.813$ & $+0.052$ \\
2 & $-0.743$ & $+0.033$ \\
3 & $-1.360$ & $+0.045$ \\
4 & $-0.919$ & $+0.077$ \\
\bottomrule
\end{tabular}
\label{tab:multiseed}
\end{table}

Shiva-90\% is ahead on both metrics under all five seeds, and the FID margin is at least three times the seed-to-seed deviation of either configuration. The same test applied to CLIP Score places Shiva-60\% ahead of \textit{Finetuned} under all five seeds as well. Shiva-80\% has higher FID than \textit{Finetuned} under all five seeds, consistent with Table~\ref{tab:main_results}.

\subsection{Additional Quality Evaluation}

\subsubsection{Per-Category Quality}
\label{app:percat}

We use the MJHQ-30K category labels to evaluate quality differences between Shiva-60 and Vanilla across prompt types. The \texttt{logo} category contains many prompts requiring rendered text.

\begin{table}[!htbp]
    \centering
    \small
    \setlength{\tabcolsep}{5pt}
    \caption{Per-category quality at the 60\% budget on SD3-Medium, as paired per-image differences (Shiva-60 minus Vanilla) over $2000$ MJHQ-30K prompts at $1024 \times 1024$, same prompt and same seed in both arms. Positive is better for all three metrics. Categories are ordered by $\Delta$CLIP.}
    \begin{tabular}{l r rrr}
        \toprule
        \textbf{Category} & $n$ & $\Delta$CLIP & $\Delta$IQA & $\Delta$IR \\
        \midrule
        \texttt{logo}      & 207 & $-0.708$ & $+0.025$ & $-0.073$ \\
        \texttt{art}       & 195 & $+0.358$ & $+0.035$ & $+0.029$ \\
        \texttt{plants}    & 195 & $+0.399$ & $+0.048$ & $+0.026$ \\
        \texttt{fashion}   & 210 & $+0.430$ & $+0.046$ & $+0.077$ \\
        \texttt{food}      & 196 & $+0.578$ & $+0.034$ & $+0.126$ \\
        \texttt{vehicles}  & 205 & $+0.686$ & $+0.050$ & $+0.083$ \\
        \texttt{people}    & 217 & $+0.789$ & $+0.047$ & $+0.062$ \\
        \texttt{animals}   & 187 & $+0.860$ & $+0.055$ & $+0.116$ \\
        \texttt{landscape} & 185 & $+0.890$ & $+0.035$ & $+0.088$ \\
        \texttt{indoor}    & 203 & $+0.979$ & $+0.028$ & $+0.151$ \\
        \midrule
        \textbf{All}       & 2000 & $+0.520$ & $+0.040$ & $+0.068$ \\
        \bottomrule
    \end{tabular}
    \label{tab:percat}
\end{table}

We report paired per-image differences in CLIP, CLIP-IQA, and ImageReward using identical prompts and seeds. Each category contains approximately 200 images.

Table~\ref{tab:percat} reports results for 2,000 paired images. The aggregate difference is positive for all three metrics. The \texttt{logo} category has the lowest difference on each metric: CLIP decreases by $0.708$ ($2.8\sigma$), ImageReward decreases by $0.073$, and CLIP-IQA increases by $0.025$, the smallest gain among the ten categories. The other nine categories have positive differences on all three metrics.

\subsubsection{Targeted Evaluation on Qwen-Image-Bench}
\label{app:qwen_bench}

To complement the aggregate image-quality metrics, we evaluate prompt following, composition, text rendering, quality, and real-world fidelity using Qwen-Image-Bench~\citep{li2026qwenimagebench}. We generate one $1024\times1024$ image for each of its 1,000 official English prompts with SD3-Medium, using 50 sampling steps, a guidance scale of 7.0, and BF16 precision. Each prompt uses the same saved initial latent across all methods. Vanilla is the base model; the pruning methods use their fixed checkpoints, with Shiva evaluated at the 60\% and 80\% budgets used in the main experiments.

We uniformly sample, without replacement, 300 applicable prompts for Alignment and 150 each for Composition, Text Rendering, Quality, and Real-world Fidelity. The prompt lists are fixed before scoring and shared across methods. Images are scored with Civitai's FP8 release of Q-Judger~\citep{civitai_qwen_image_bench_fp8}, using greedy decoding and the official checklists and score parser.

\begin{table}[htbp]
\centering
\small
\setlength{\tabcolsep}{4pt}
\setlength{\aboverulesep}{0pt}
\setlength{\belowrulesep}{0pt}
\renewcommand{\arraystretch}{1.2}
\caption{Targeted evaluation on Qwen-Image-Bench. All scores are on a 0--100 scale and higher is better. Fidelity denotes Real-world Fidelity. Vanilla is an unpruned reference.}
\label{tab:qwen_bench}
\begin{tabular}{l ccccc}
\toprule
Method & Alignment & Composition & Text Rendering & Quality & Fidelity \\
\midrule
Vanilla & 44.36 & 40.00 & 14.54 & 48.76 & 42.46 \\
\midrule
DiffCR & 37.21 & 24.53 & 3.61 & 30.95 & 41.55 \\
DyDiT & 34.42 & 26.00 & 4.43 & 42.30 & 38.46 \\
\rowcolor{gray!20} \textbf{Shiva-80\%} & \alignbold{41}{81} & \alignbold{32}{53} & \alignunder{5}{32} & \alignunder{44}{67} & \alignunder{42}{34} \\
\rowcolor{gray!20} \textbf{Shiva-60\%} & \alignunder{40}{48} & \alignunder{31}{73} & \alignbold{5}{92} & \alignbold{46}{86} & \alignbold{42}{78} \\
\bottomrule
\end{tabular}
\end{table}

Table~\ref{tab:qwen_bench} shows that both Shiva variants outperform DiffCR and DyDiT on all five evaluated criteria. Shiva-60\% achieves a Quality score of 46.86, compared with 30.95 for DiffCR and 42.30 for DyDiT, while Shiva-80\% reaches 41.81 in Alignment and 32.53 in Composition. These results extend the advantage over the tested pruning baselines to capabilities beyond the aggregate metrics in the main evaluation. Text rendering remains substantially below Vanilla for all pruning methods, identifying a shared limitation on this more demanding criterion.

\subsection{Combination with Other Acceleration Techniques}
\label{app:composition}

\subsubsection{Quantization}

We evaluate the combination of token pruning and quantization on Shiva-80\% by quantizing the feed-forward layers to INT8 weights and activations.

Table~\ref{tab:quant_quality} reports changes in FID and CLIP when 11 or 22 blocks are quantized.

\begin{table}[h]
\centering
\caption{Quality of Shiva-80\% under INT8, relative to the same model in fp16. Paired over 1{,}000 MJHQ prompts with identical seeds.}
\begin{tabular}{lcc}
\toprule
Quantized scope & $\Delta$FID & $\Delta$CLIP \\
\midrule
11 blocks & $+0.616$ & $+0.118$ \\
22 blocks & $+2.824$ & $-0.082$ \\
\bottomrule
\end{tabular}
\label{tab:quant_quality}
\end{table}

Table~\ref{tab:quant_speed} reports transformer forward latency and memory usage. Adding INT8 quantization to Shiva-80\% reduces latency from 104.5 to 96.6~ms, a further reduction of $7.5\%$. Weight memory decreases from 3.92 to 3.50 GB, and peak memory decreases from 4.21 to 3.79 GB.

\begin{table}[H]\centering\caption{Transformer forward pass at $1024\times1024$. Weights are measured after loading, activations as peak minus weights.}
\begin{tabular}{lcccc}
\toprule
Configuration & Weights (GB) & Peak (GB) & Latency (ms) & Speedup \\
\midrule
Vanilla            & 3.90 & 4.19 & 152.1 & $1.00\times$ \\
Shiva-80\%         & 3.92 & 4.21 & 104.5 & $1.46\times$ \\
Shiva-80\% + INT8  & 3.50 & 3.79 & \phantom{0}96.6 & $1.57\times$ \\
\bottomrule
\end{tabular}
\label{tab:quant_speed}
\end{table}

\pagebreak[0]
\begin{samepage}
\subsubsection{Accelerated Sampling}
\label{app:accelerated_sampling}

Step distillation reduces the number of denoising steps, while token pruning reduces the computation within each step. We evaluate their composition using the released SD3-Medium 8-step LoRA from Hyper-SD~\citep{hypersd2024}, comparing Shiva, DyDiT, and DiffCR. We use the Shiva-60\% checkpoint. Combining the distillation and pruning LoRAs required exploring several configurations; we found it more stable to keep the first four sampling steps unpruned and apply each method's pruning only during the final four steps. All three pruning methods follow this schedule.
\par
\end{samepage}

\begin{table}[htbp]
\centering
\small
\setlength{\tabcolsep}{4pt}
\setlength{\aboverulesep}{0pt}
\setlength{\belowrulesep}{0pt}
\renewcommand{\arraystretch}{1.2}
\caption{Token pruning with Hyper-SD 8-step sampling on SD3-Medium. Latency measures DiT inference, and speedup is relative to the unpruned distilled model.}
\label{tab:distill}
\begin{tabular}{l cccc S[table-format=2.2] cc}
\toprule
Method & FID $\downarrow$ & IR $\uparrow$ & CLIP $\uparrow$ & IQA $\uparrow$ & {FLOPs (T/step) $\downarrow$} & Latency (ms) $\downarrow$ & Speedup $\uparrow$ \\\midrule
Hyper-SD & 19.77 & 0.765 & 30.42 & 0.485 & 11.80 & 1233.1 & 1.00$\times$ \\
\midrule
DyDiT & 21.03 & 0.660 & 30.42 & 0.490 & 10.59 & 1150.3 & 1.07$\times$ \\
DiffCR & 26.48 & 0.653 & 30.38 & 0.489 & 10.39 & 1105.8 & 1.12$\times$ \\
\rowcolor{gray!20} \textbf{Shiva} & \alignbold{20}{90} & \alignbold{0}{728} & \alignbold{30}{62} & \alignbold{0}{494} & \bfseries 10.00 & \alignbold{1091}{4} & \textbf{1.13$\times$} \\
\bottomrule
\end{tabular}
\end{table}

Table~\ref{tab:distill} shows that Shiva achieves the best values on all four quality metrics among the pruning methods, while also requiring the fewest FLOPs and the lowest DiT latency. Relative to the unpruned 8-step model, Shiva reduces FLOPs by 15.24\% and latency from 1233.1 to 1091.4~ms, providing a further 1.13$\times$ speedup. FID and ImageReward degrade relative to this reference, whereas CLIP and IQA improve. These results demonstrate compatibility with accelerated sampling under the tested configuration and a favorable quality--efficiency trade-off relative to DyDiT and DiffCR.

Combining LoRAs trained for distillation and pruning introduces an additional adapter-compatibility problem. Related work on multi-concept diffusion customization, including Mix-of-Show~\citep{gu2023mixofshow} and LoRACLR~\citep{simsar2025loraclr}, studies how to fuse separately trained concept adapters while preserving their capabilities. Extending such fusion and adaptation strategies to adapters trained for different acceleration objectives is a promising direction for future work.

\subsubsection{Caching}

Feature caching (Sec.~\ref{sec:related_work}) reduces repeated computation by reusing features across adjacent timesteps. Combining it with our layer- and timestep-dependent token selection requires coordinating which features are refreshed and reused as the selected tokens change. We leave the design and evaluation of this combination to future work.

\FloatBarrier
\section{Qualitative Results}
\label{app:qualitative}

\subsection{SD3-Medium}

\label{app:vis_results}

\begin{figure}[H]
    \centering
    \begin{subfigure}[b]{0.13\textwidth}
        \includegraphics[width=\linewidth]{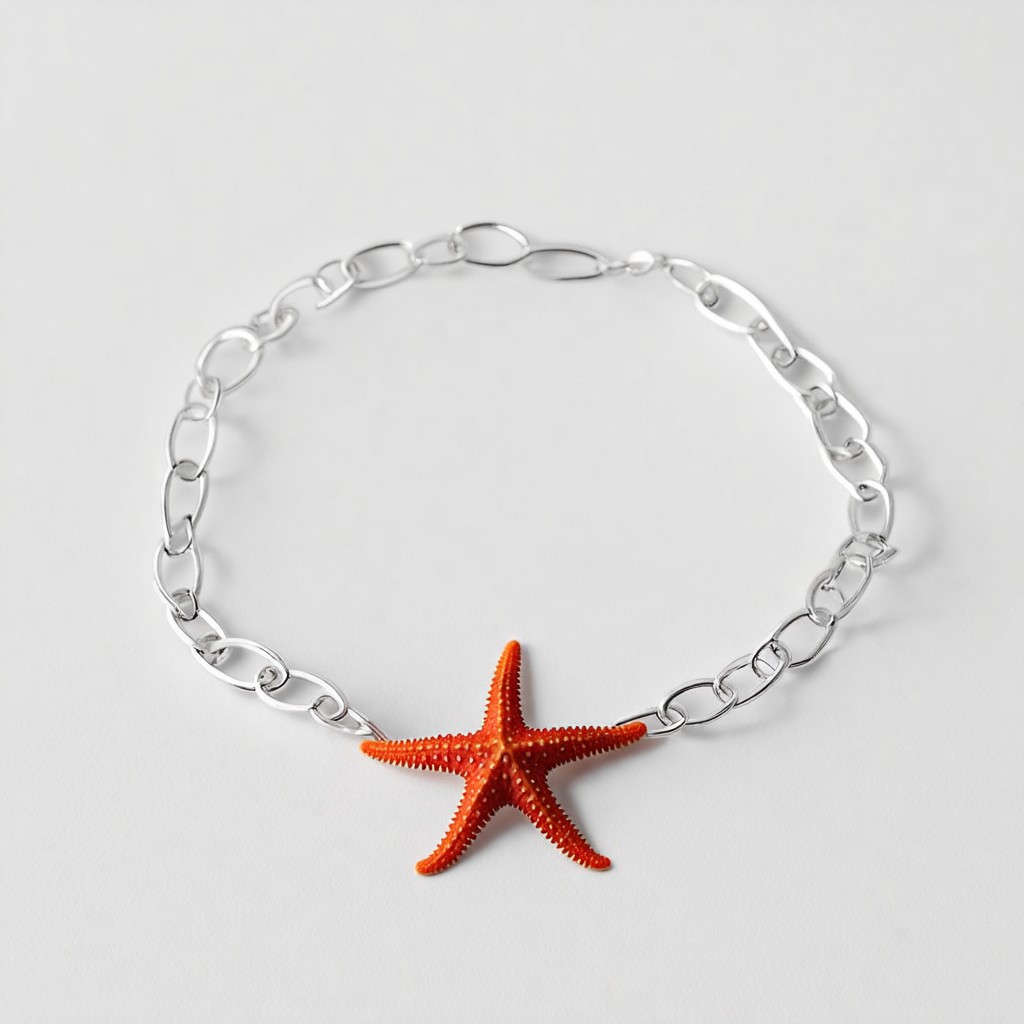}
        \caption*{Vanilla}
    \end{subfigure}\hspace{0.01\textwidth}
    \begin{subfigure}[b]{0.13\textwidth}
        \includegraphics[width=\linewidth]{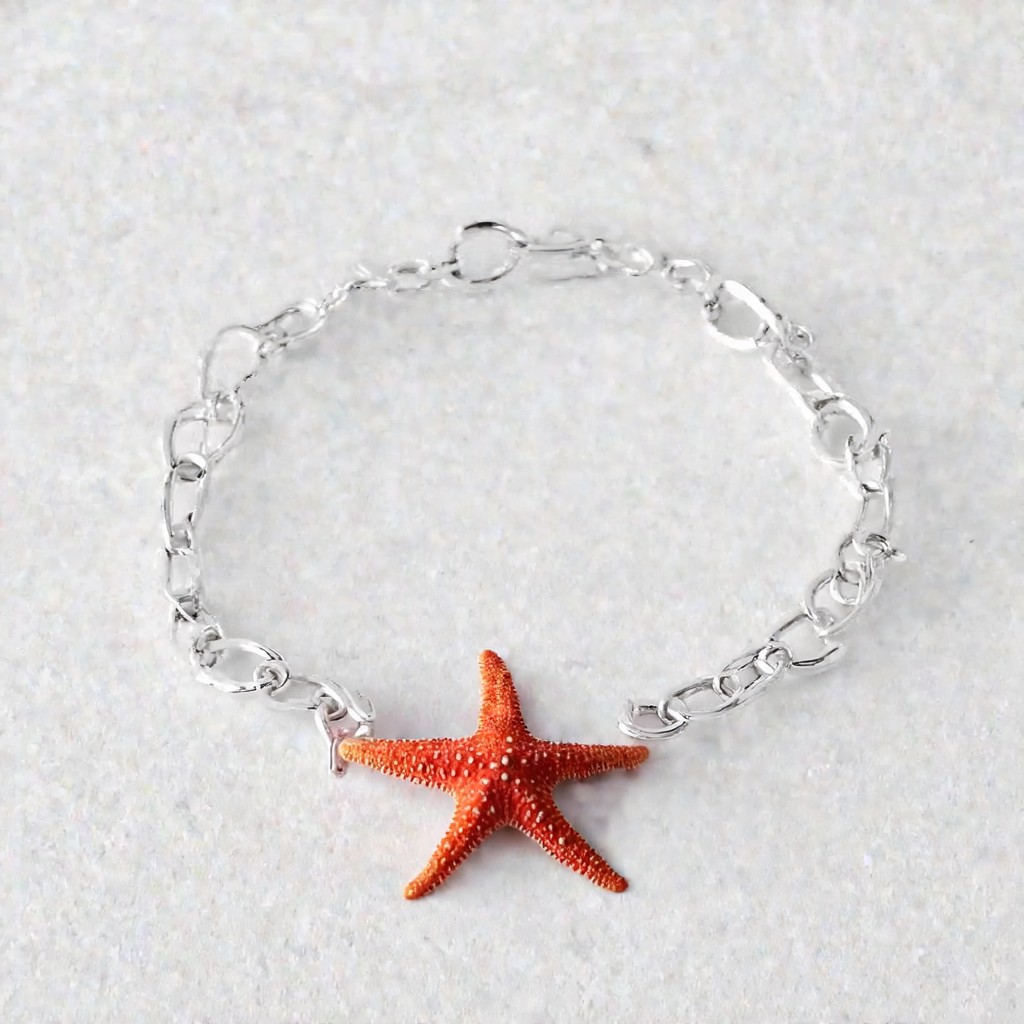}
        \caption*{ToMeSD}
    \end{subfigure}\hspace{0.01\textwidth}
    \begin{subfigure}[b]{0.13\textwidth}
        \includegraphics[width=\linewidth]{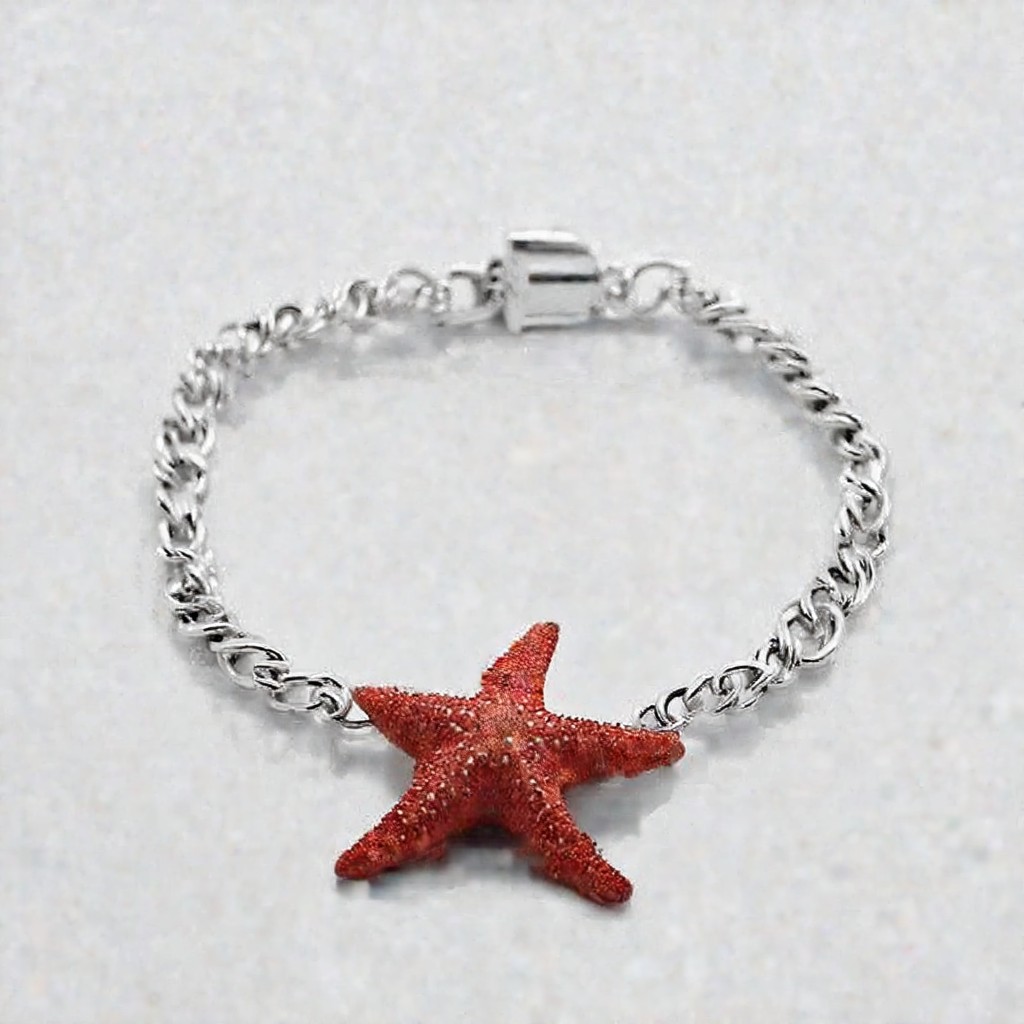}
        \caption*{ToFu}
    \end{subfigure}\hspace{0.01\textwidth}
    \begin{subfigure}[b]{0.13\textwidth}
        \includegraphics[width=\linewidth]{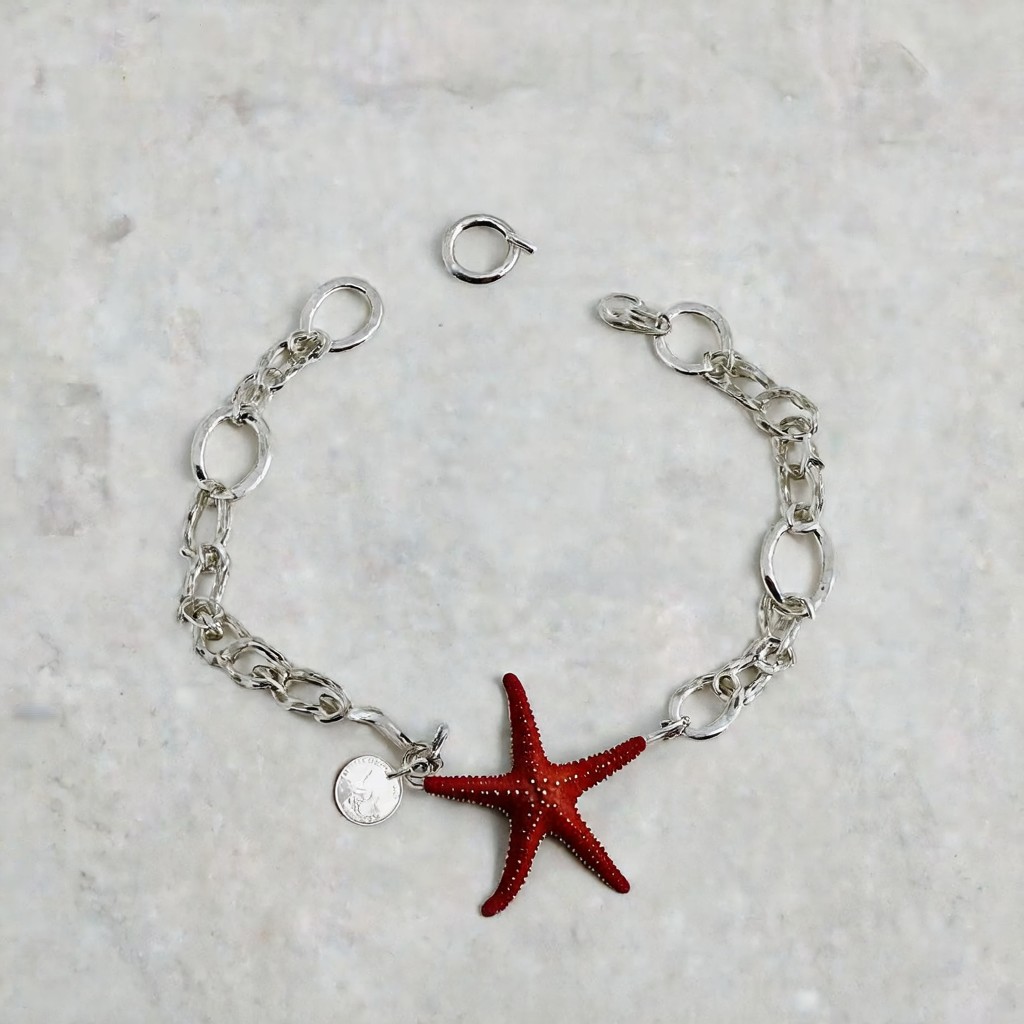}
        \caption*{SDTM}
    \end{subfigure}\hspace{0.01\textwidth}
    \begin{subfigure}[b]{0.13\textwidth}
        \includegraphics[width=\linewidth]{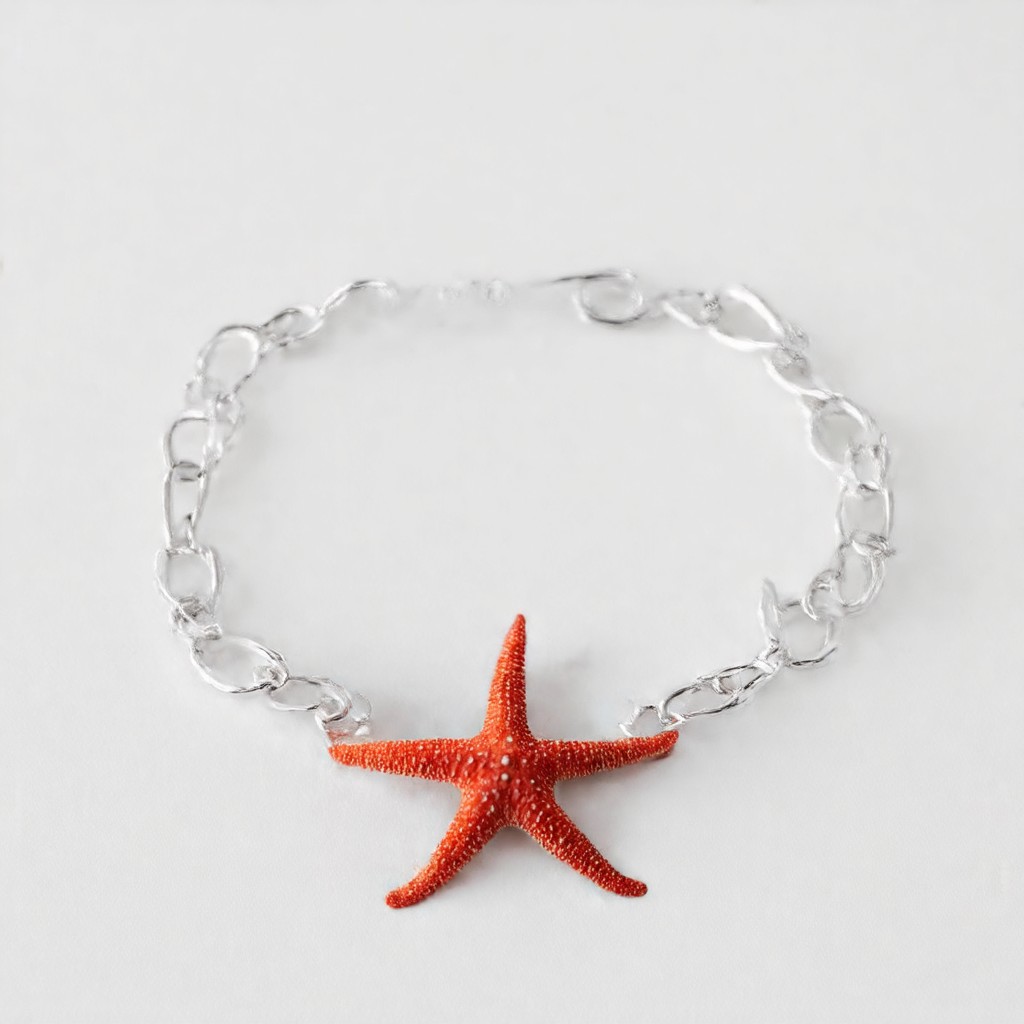}
        \caption*{IBTM}
    \end{subfigure}

    \vspace{0.2em}

    \begin{subfigure}[b]{0.13\textwidth}
        \includegraphics[width=\linewidth]{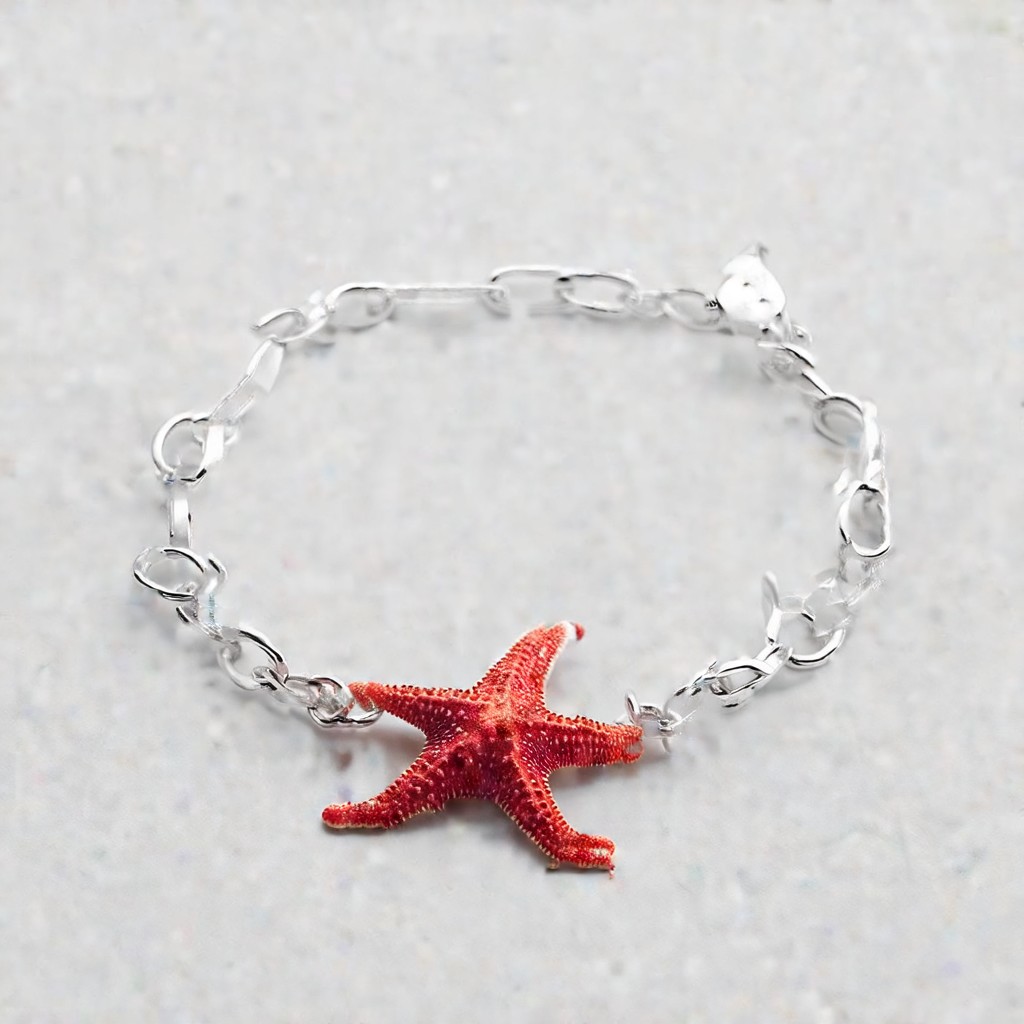}
        \caption*{ToMA}
    \end{subfigure}\hspace{0.01\textwidth}
    \begin{subfigure}[b]{0.13\textwidth}
        \includegraphics[width=\linewidth]{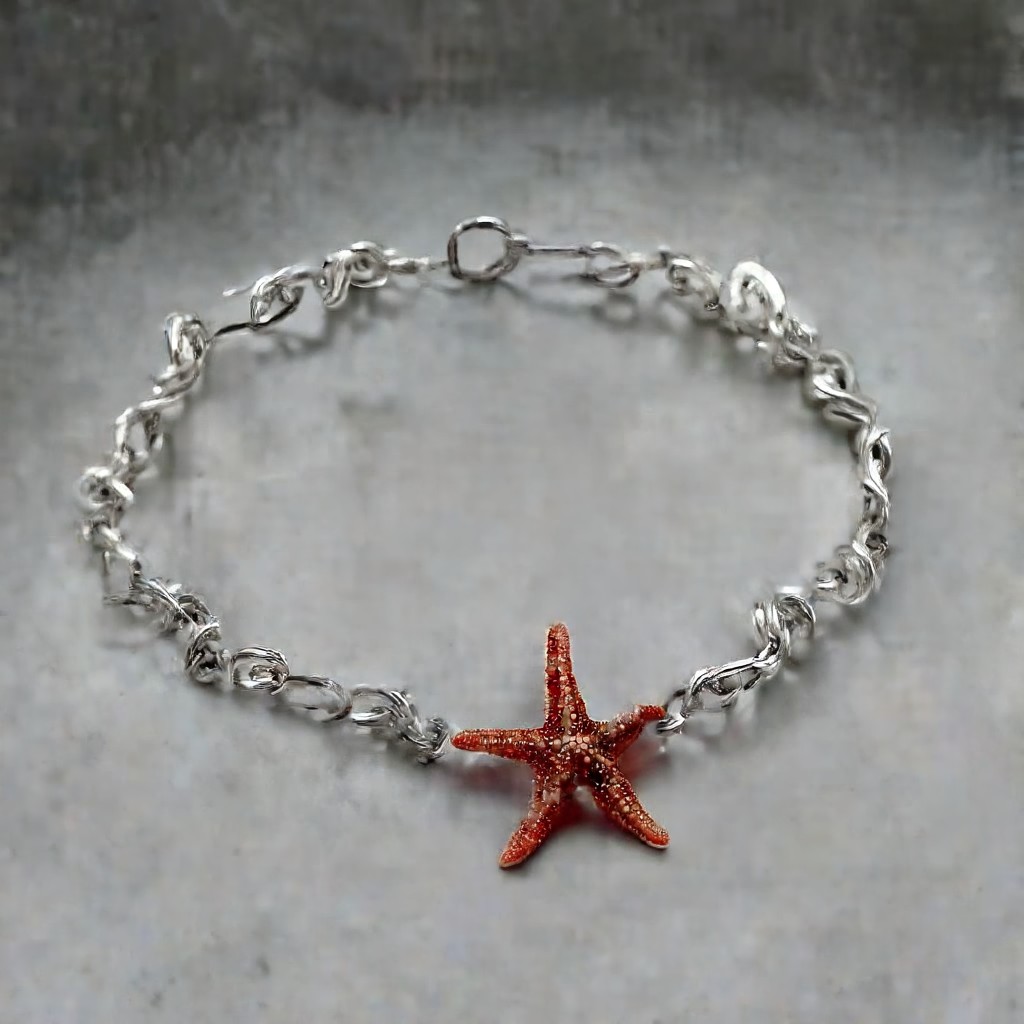}
        \caption*{DiffCR}
    \end{subfigure}\hspace{0.01\textwidth}
    \begin{subfigure}[b]{0.13\textwidth}
        \includegraphics[width=\linewidth]{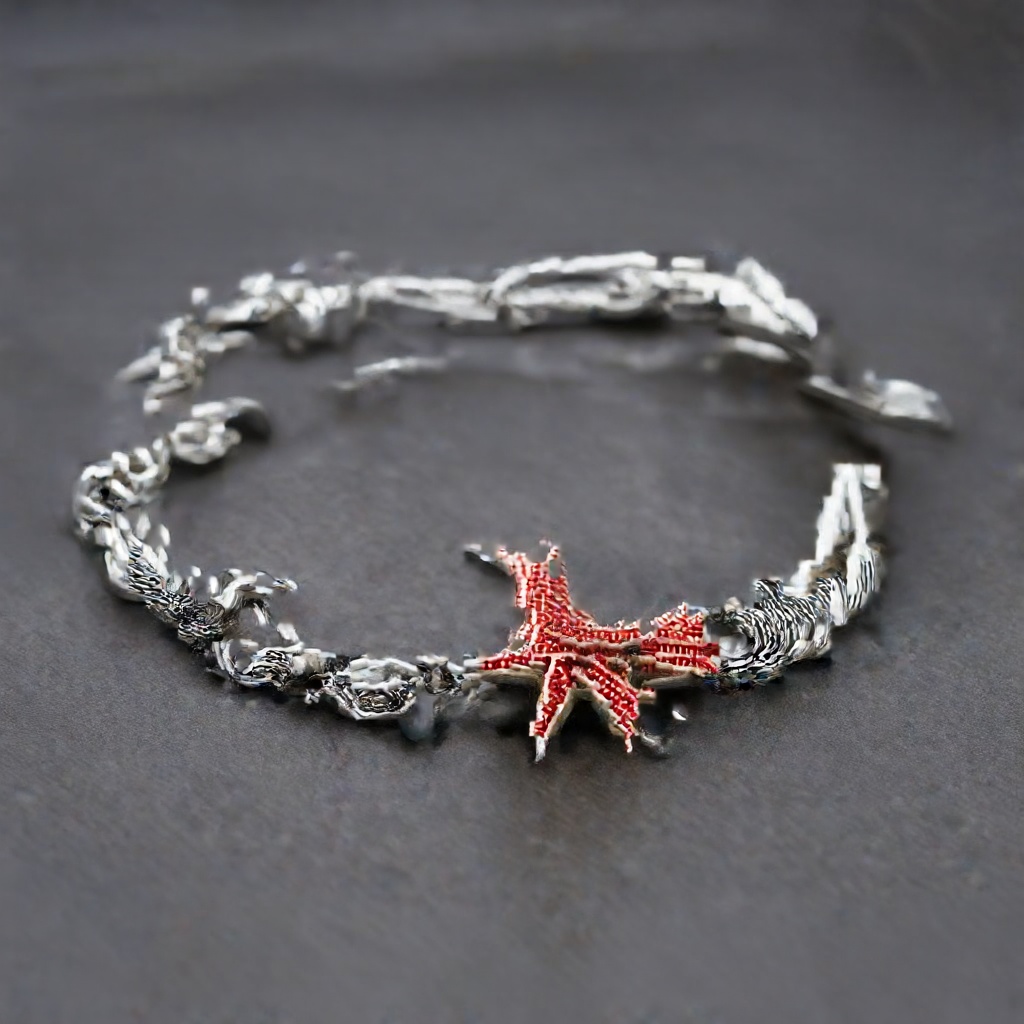}
        \caption*{NeuralSort}
    \end{subfigure}\hspace{0.01\textwidth}
    \begin{subfigure}[b]{0.13\textwidth}
        \includegraphics[width=\linewidth]{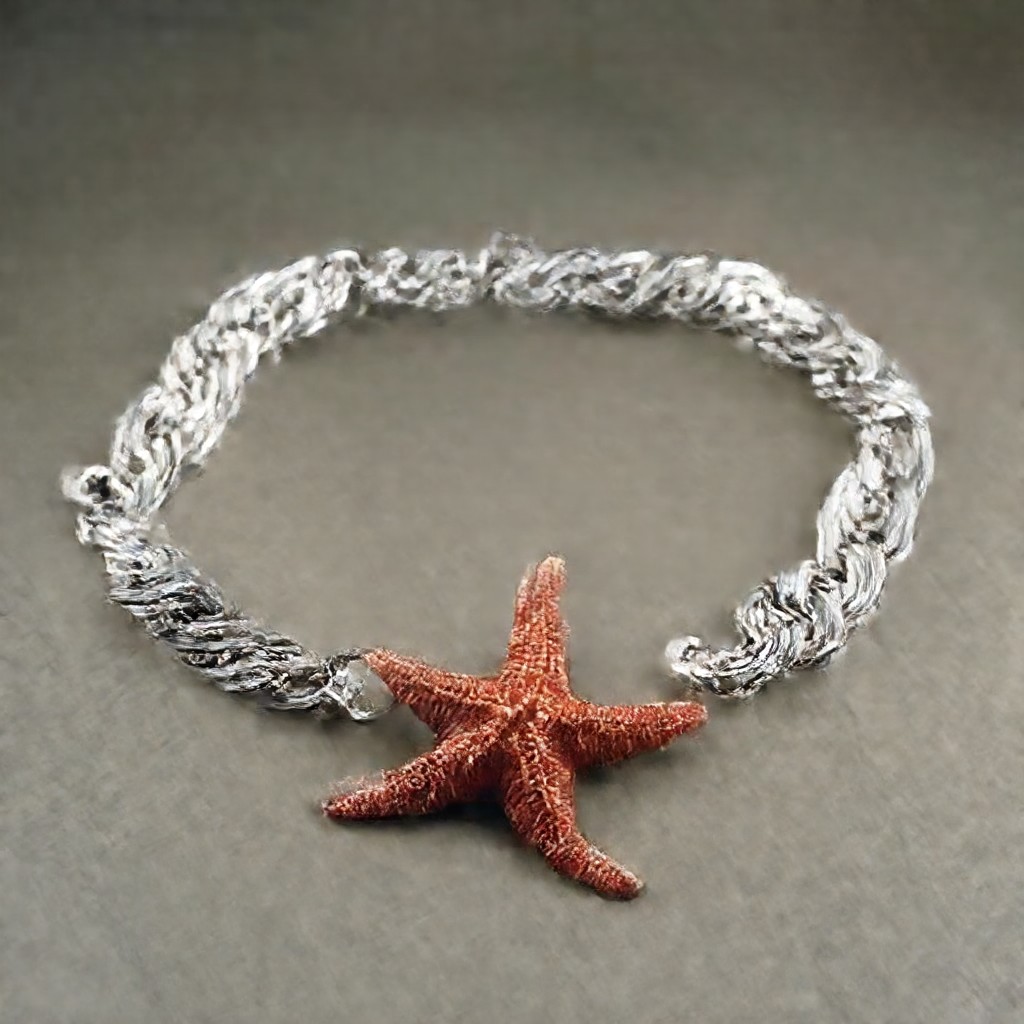}
        \caption*{SparseDiT}
    \end{subfigure}\hspace{0.01\textwidth}
    \begin{subfigure}[b]{0.13\textwidth}
        \includegraphics[width=\linewidth]{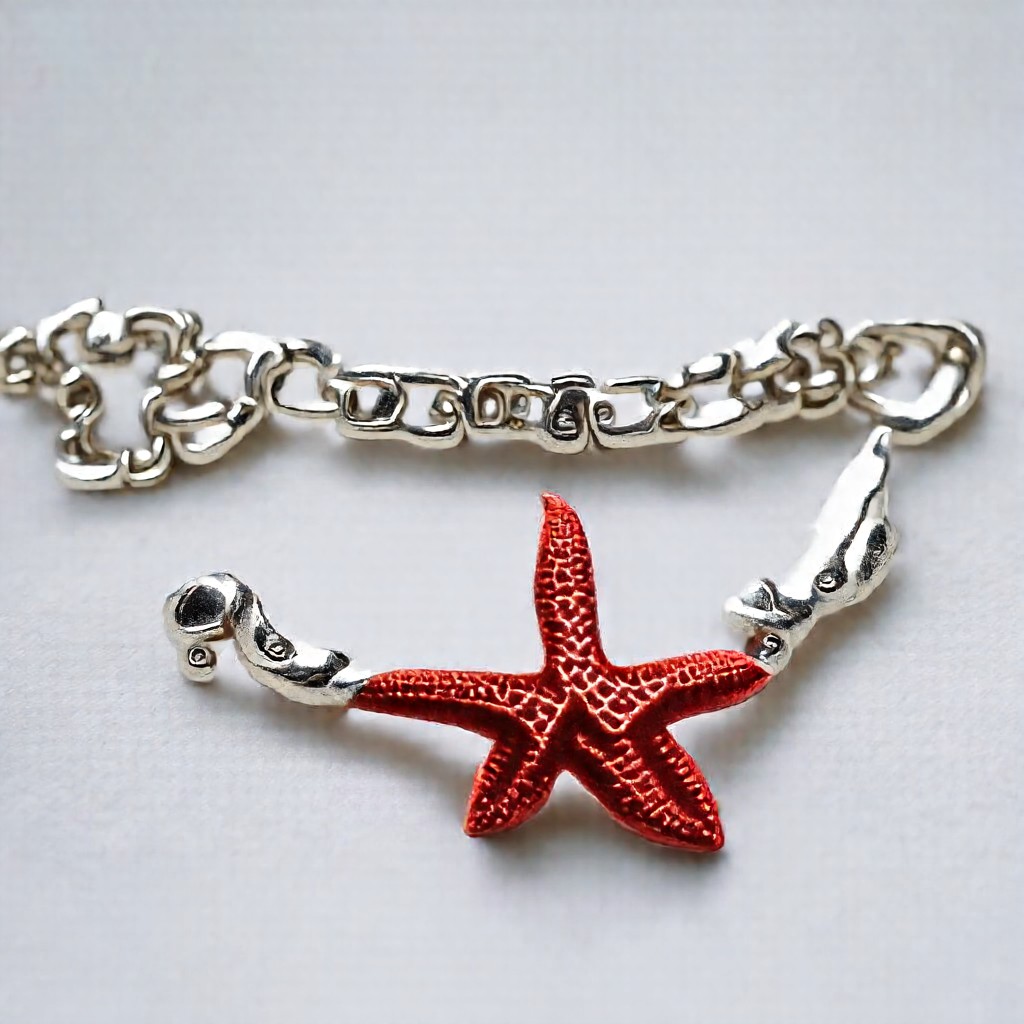}
        \caption*{DyDiT}
    \end{subfigure}

    \vspace{0.2em}

    \begin{subfigure}[b]{0.13\textwidth}
        \includegraphics[width=\linewidth]{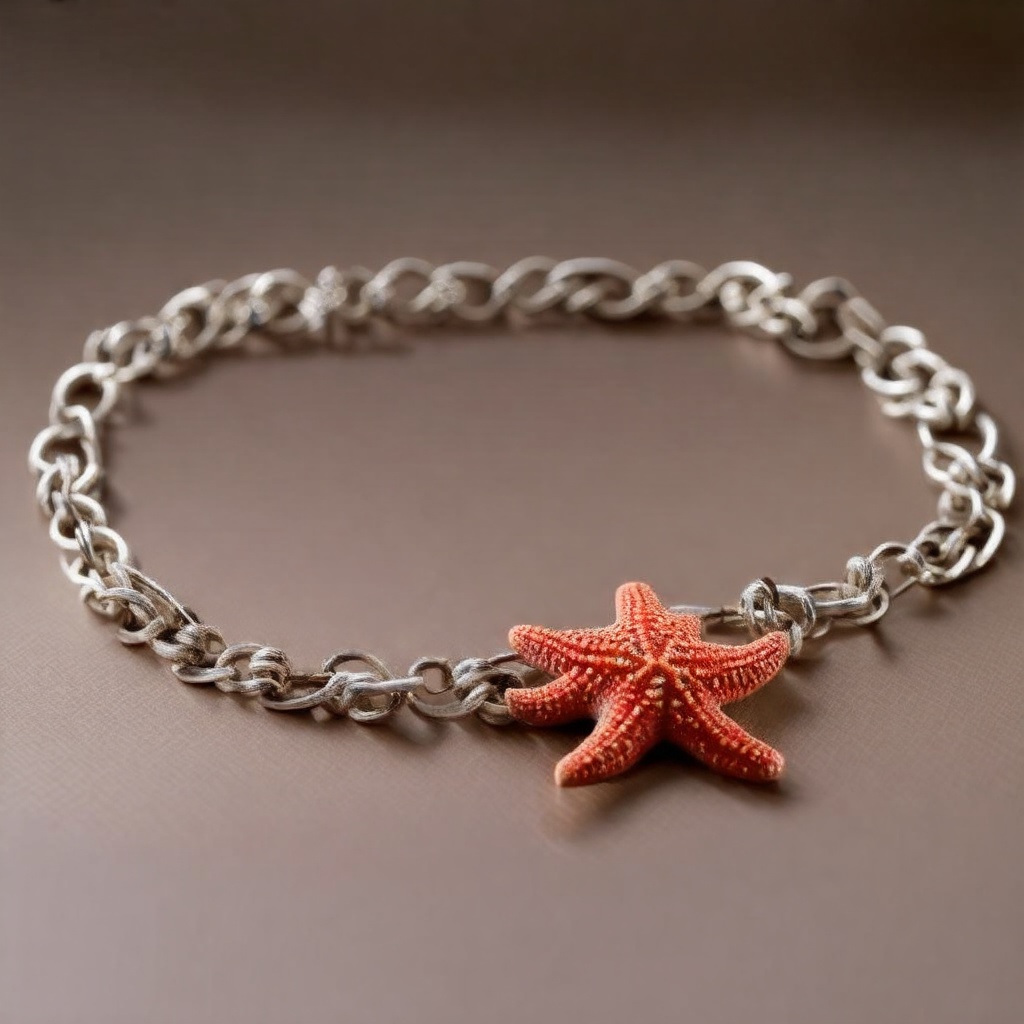}
        \caption*{\textbf{Shiva-90}}
    \end{subfigure}\hspace{0.01\textwidth}
    \begin{subfigure}[b]{0.13\textwidth}
        \includegraphics[width=\linewidth]{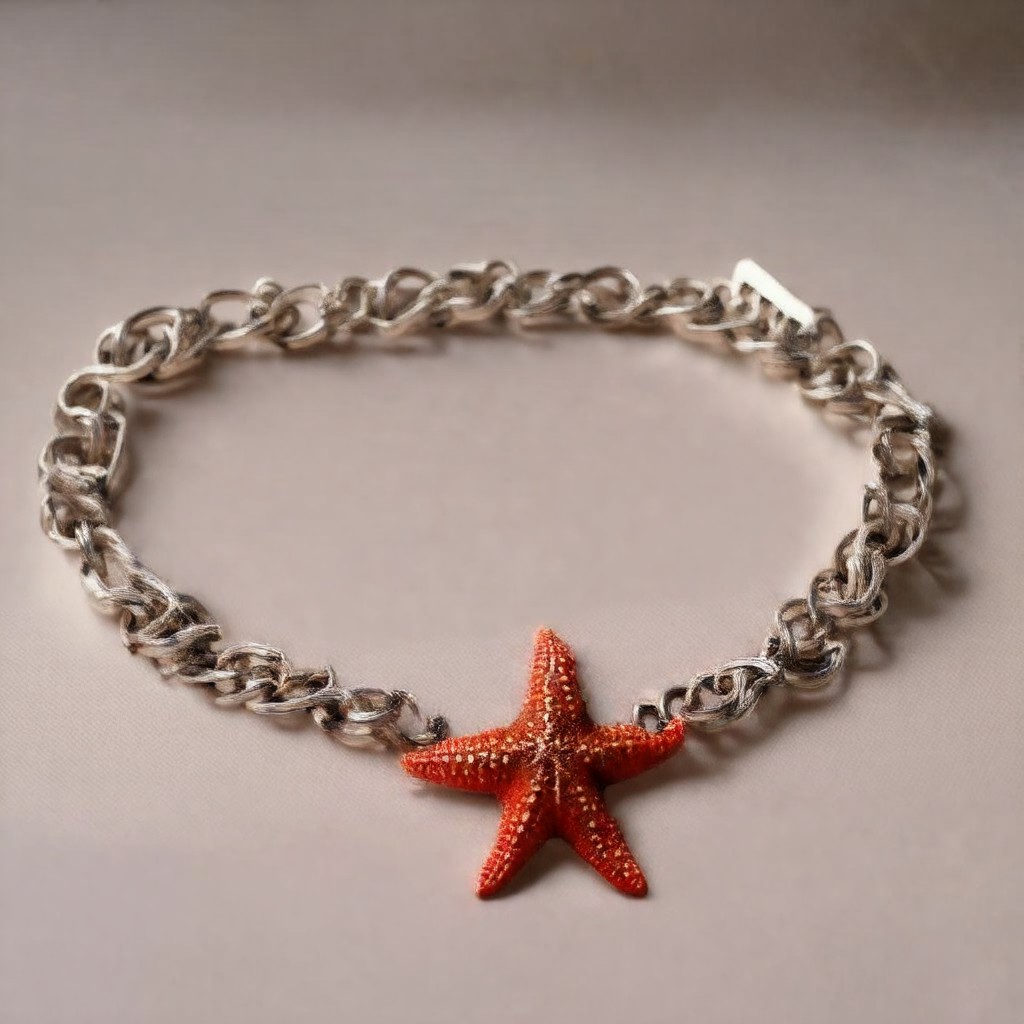}
        \caption*{\textbf{Shiva-80}}
    \end{subfigure}\hspace{0.01\textwidth}
    \begin{subfigure}[b]{0.13\textwidth}
        \includegraphics[width=\linewidth]{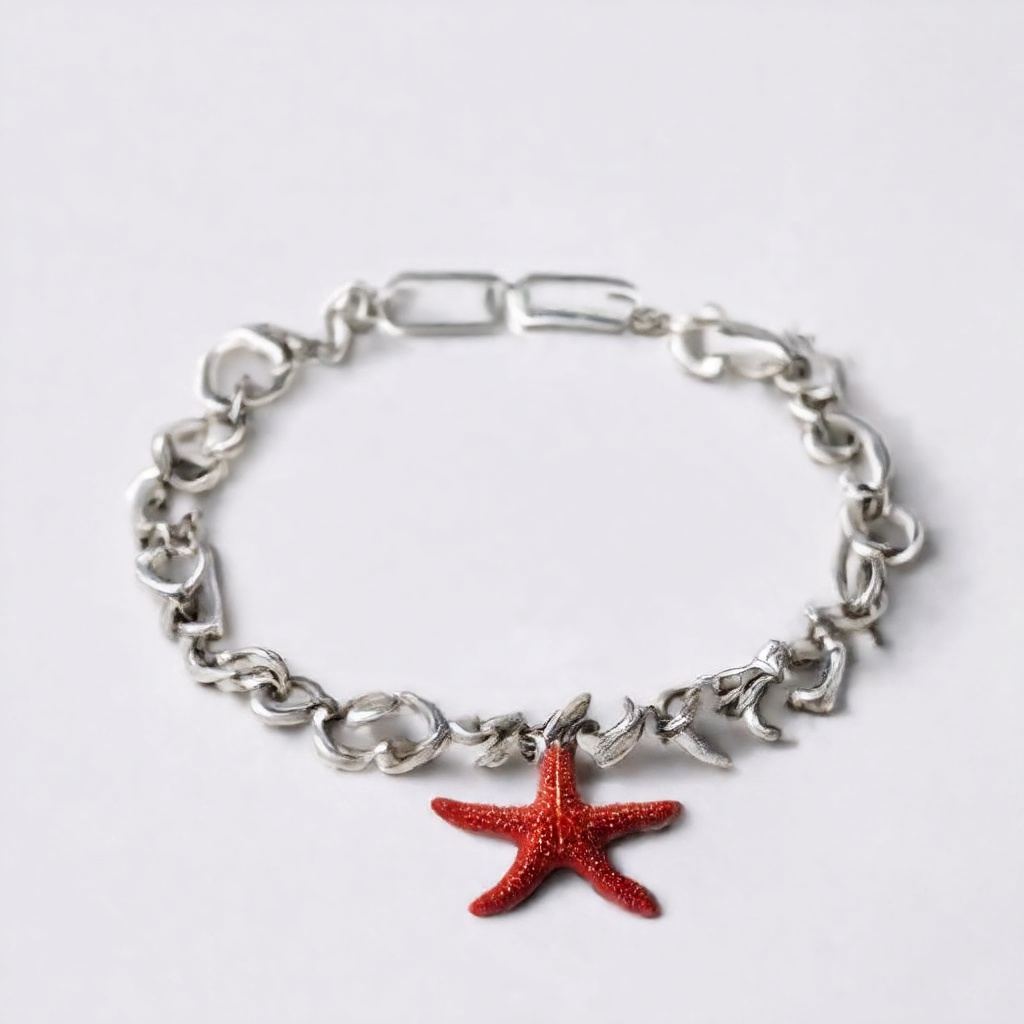}
        \caption*{\textbf{Shiva-70}}
    \end{subfigure}\hspace{0.01\textwidth}
    \begin{subfigure}[b]{0.13\textwidth}
        \includegraphics[width=\linewidth]{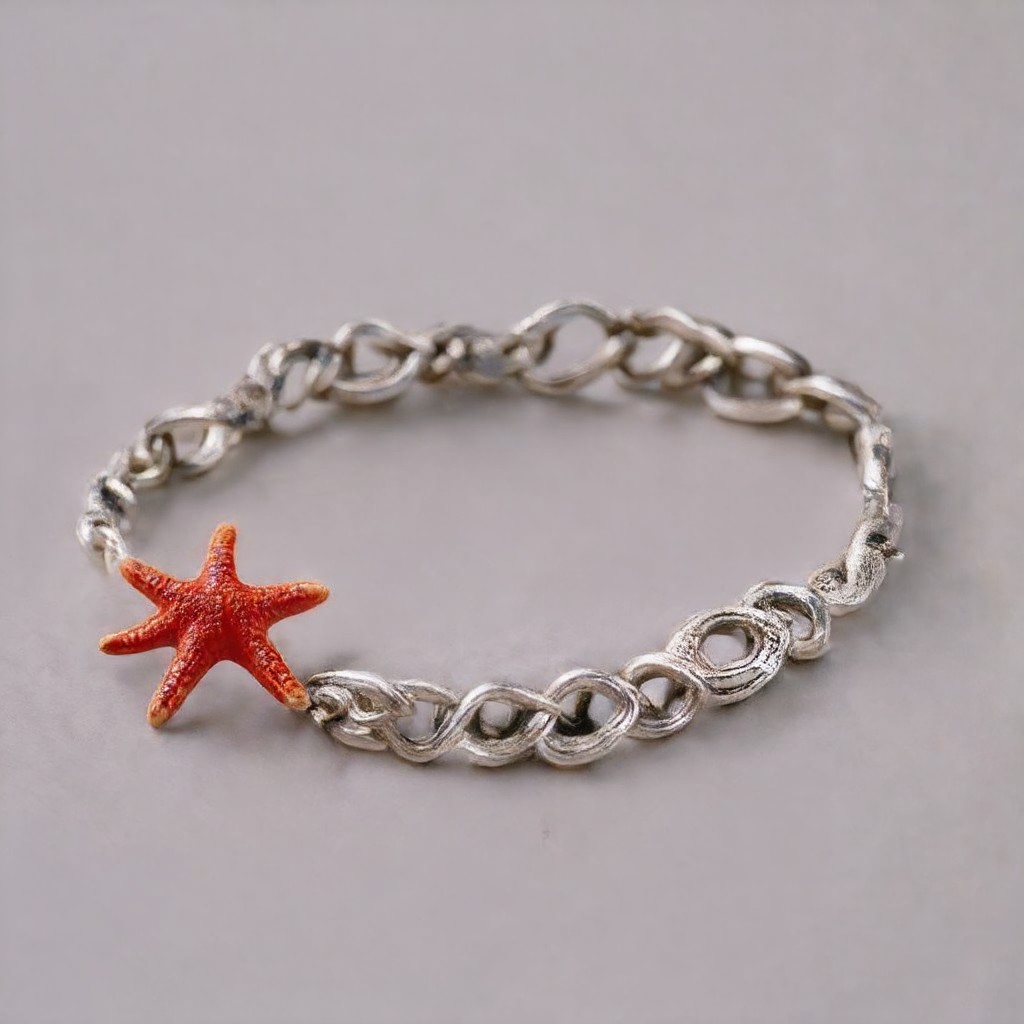}
        \caption*{\textbf{Shiva-60}}
    \end{subfigure}\hspace{0.01\textwidth}
    \begin{subfigure}[b]{0.13\textwidth}
        \includegraphics[width=\linewidth]{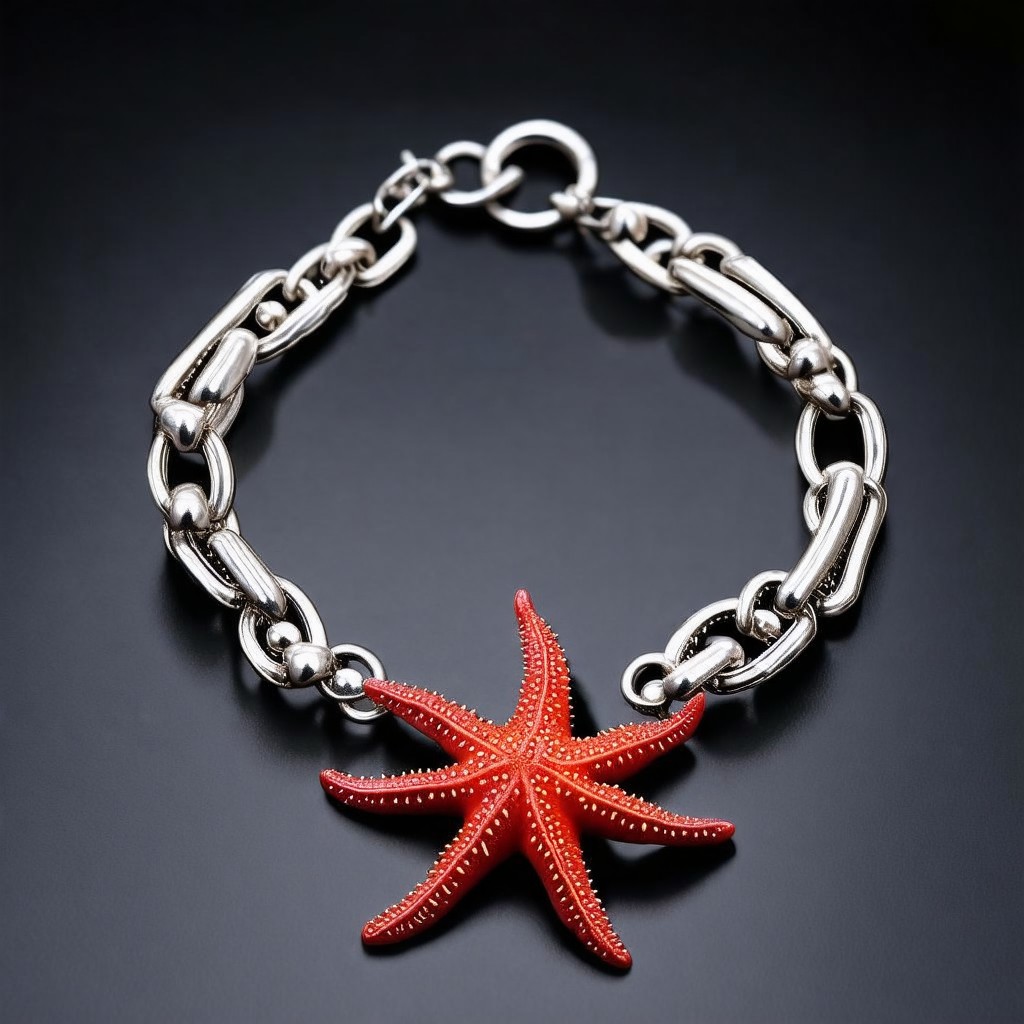}
        \caption*{Finetuned}
    \end{subfigure}

    \caption{Prompt: \textit{a delicate silver link bracelet with a small red seastar attached to it.}}
    \label{fig:appendix_fig_1}
\end{figure}

\begin{figure}[H]
    \centering
    \begin{subfigure}[b]{0.13\textwidth}
        \includegraphics[width=\linewidth]{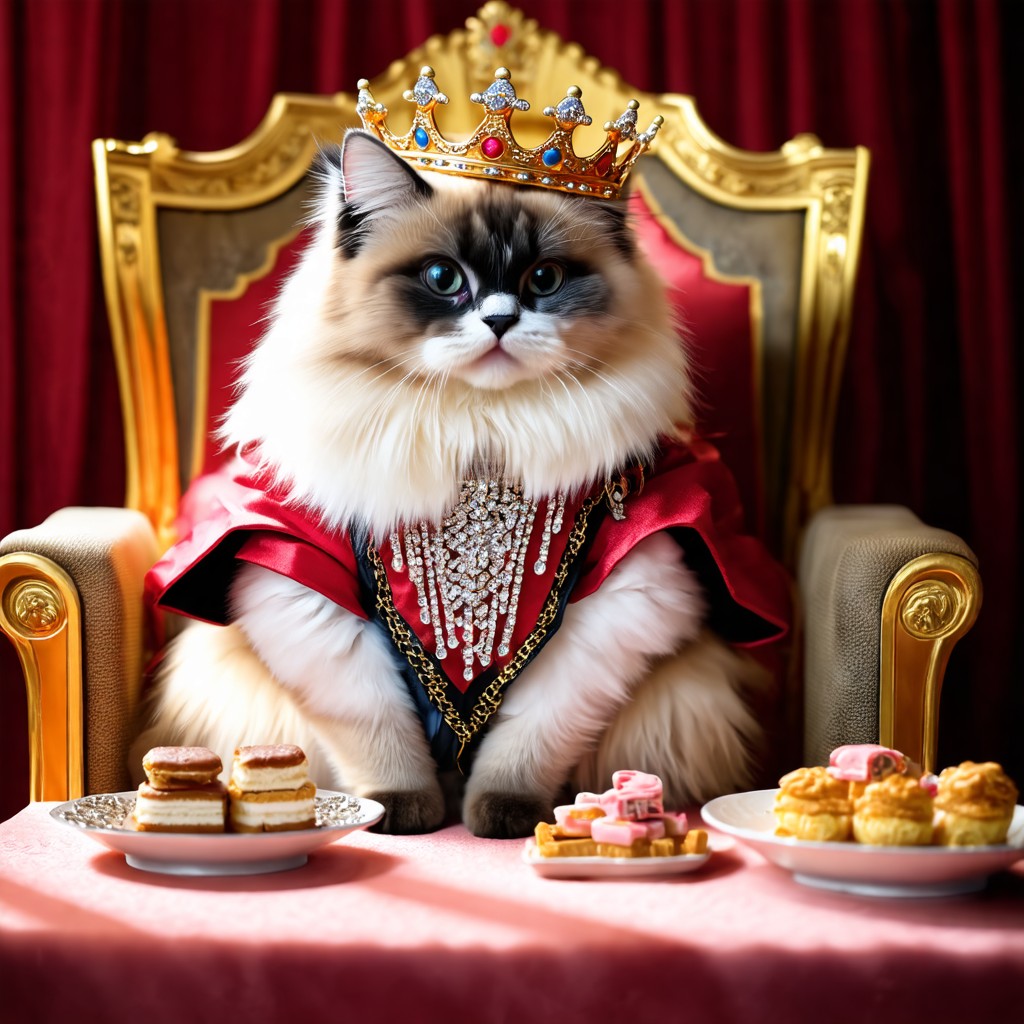}
        \caption*{Vanilla}
    \end{subfigure}\hspace{0.01\textwidth}
    \begin{subfigure}[b]{0.13\textwidth}
        \includegraphics[width=\linewidth]{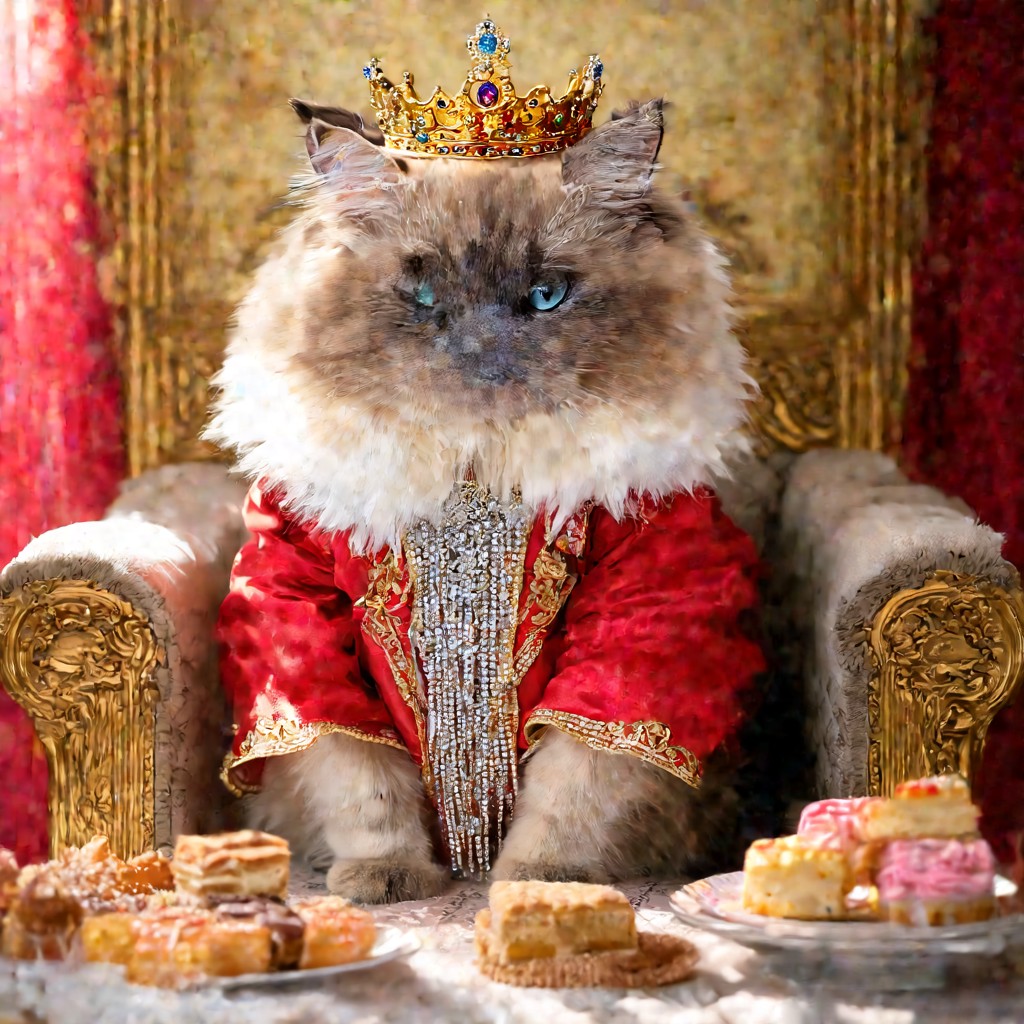}
        \caption*{ToMeSD}
    \end{subfigure}\hspace{0.01\textwidth}
    \begin{subfigure}[b]{0.13\textwidth}
        \includegraphics[width=\linewidth]{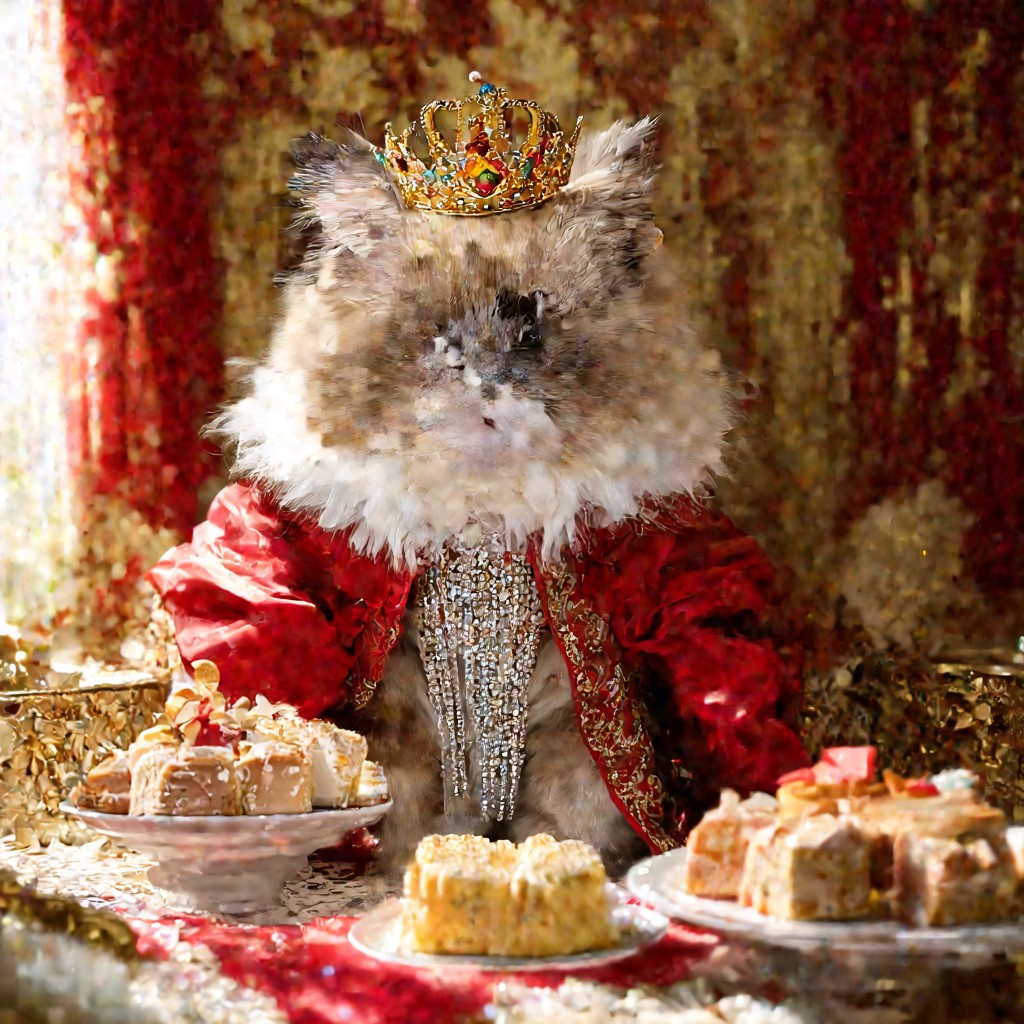}
        \caption*{ToFu}
    \end{subfigure}\hspace{0.01\textwidth}
    \begin{subfigure}[b]{0.13\textwidth}
        \includegraphics[width=\linewidth]{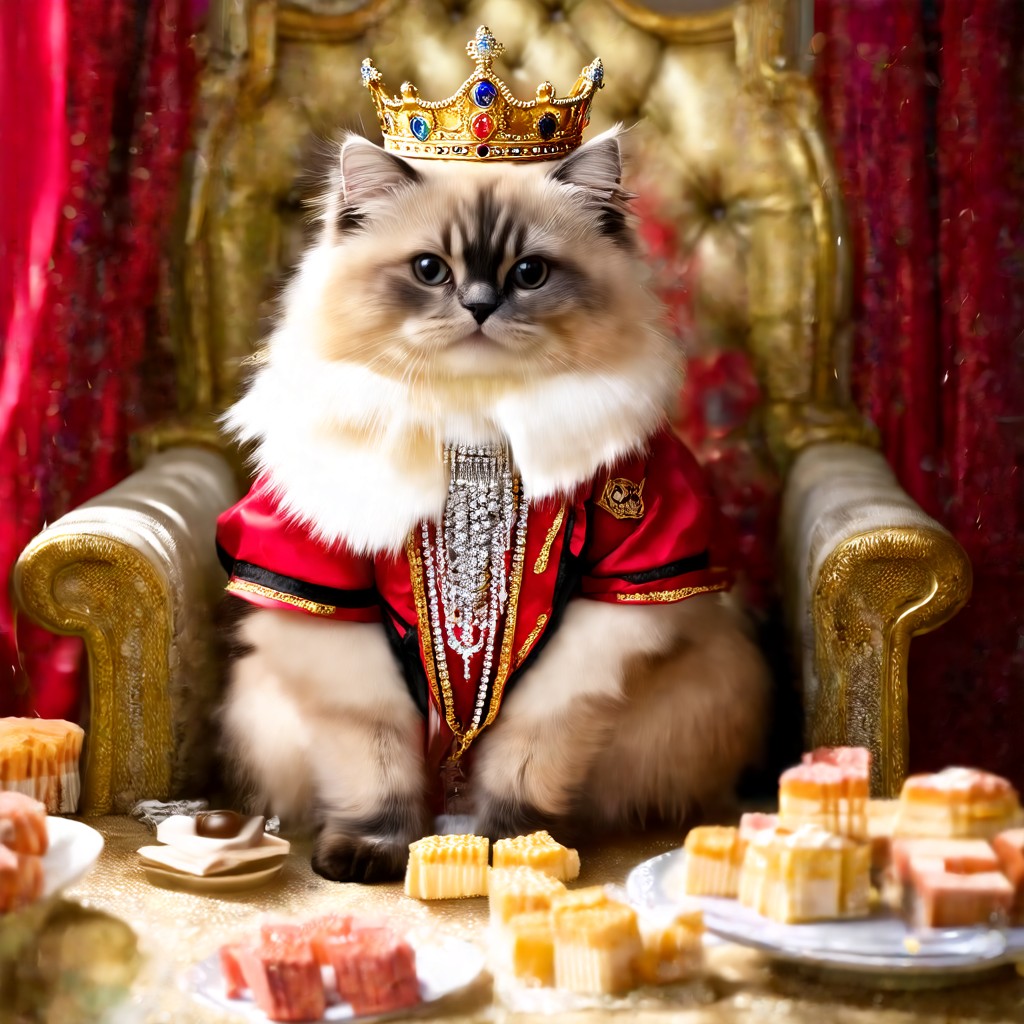}
        \caption*{SDTM}
    \end{subfigure}\hspace{0.01\textwidth}
    \begin{subfigure}[b]{0.13\textwidth}
        \includegraphics[width=\linewidth]{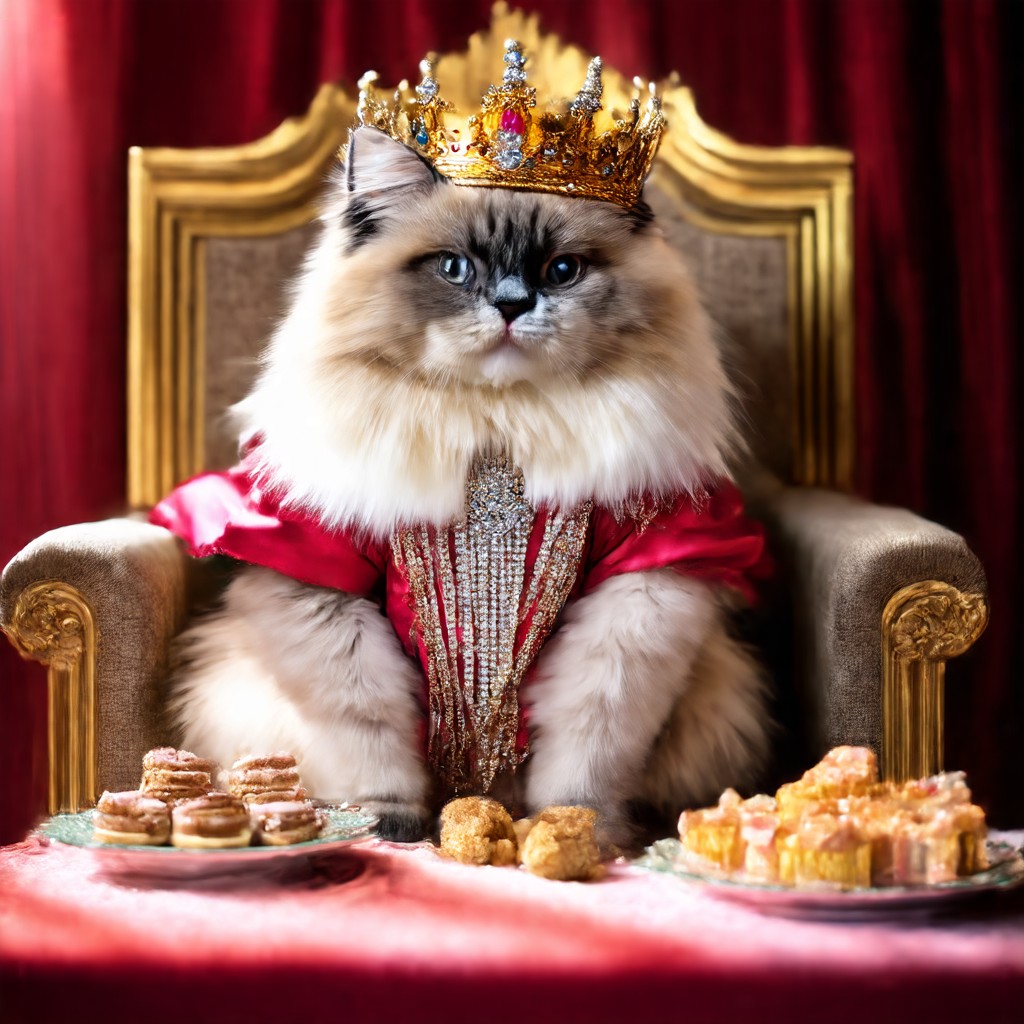}
        \caption*{IBTM}
    \end{subfigure}

    \vspace{0.2em}

    \begin{subfigure}[b]{0.13\textwidth}
        \includegraphics[width=\linewidth]{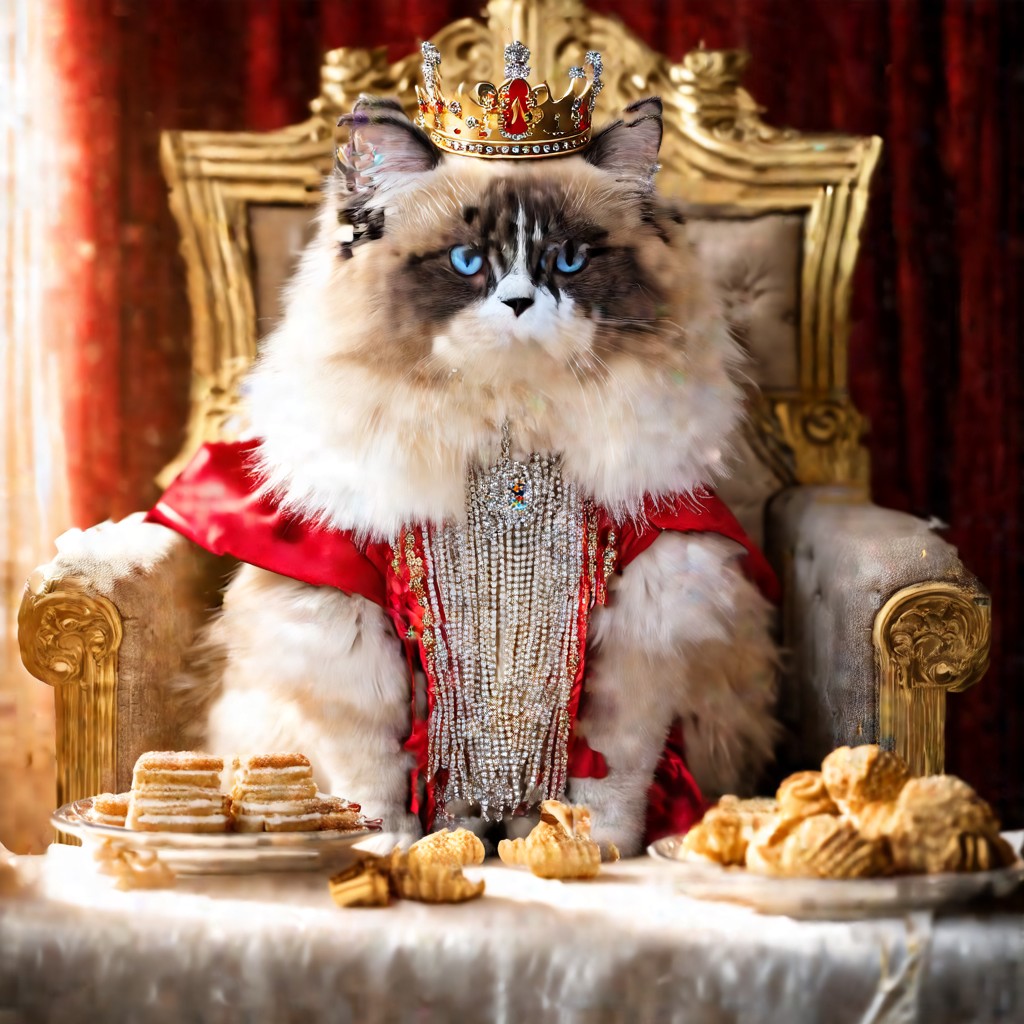}
        \caption*{ToMA}
    \end{subfigure}\hspace{0.01\textwidth}
    \begin{subfigure}[b]{0.13\textwidth}
        \includegraphics[width=\linewidth]{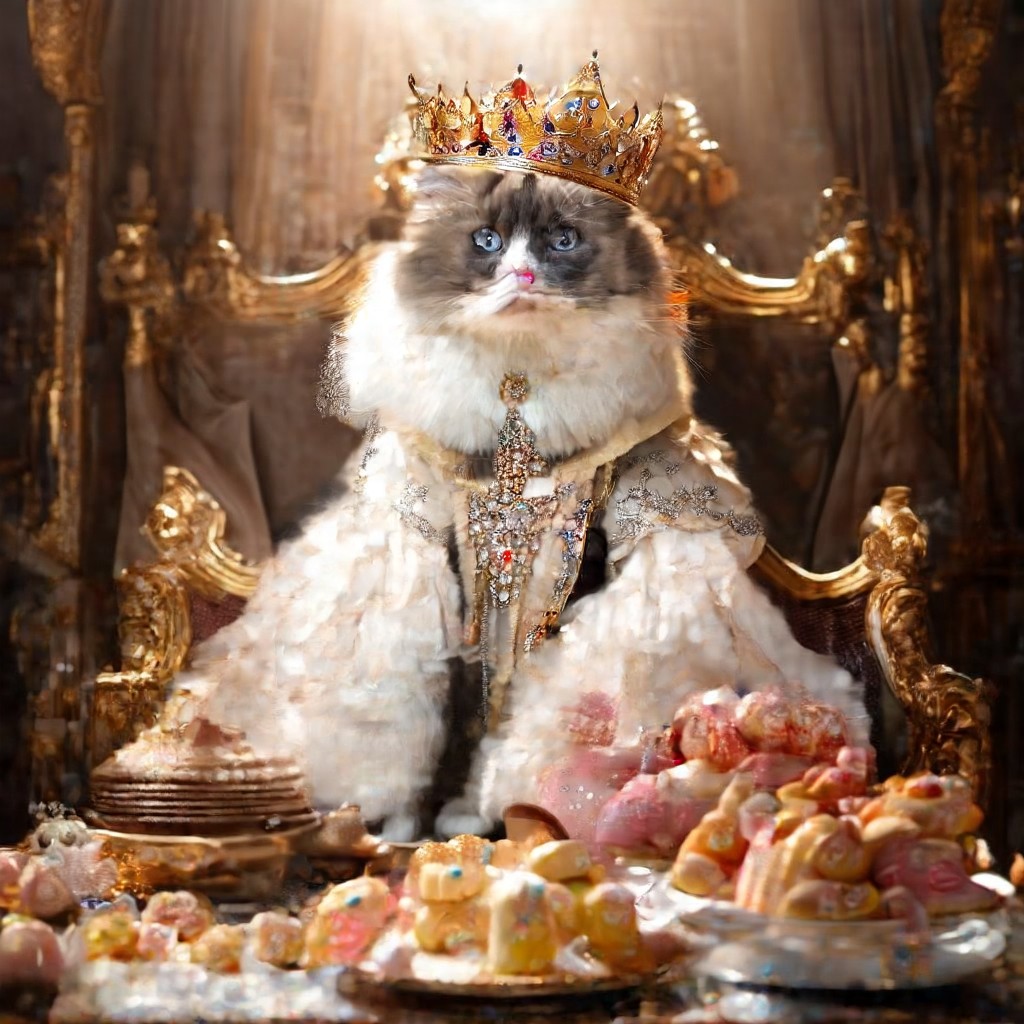}
        \caption*{DiffCR}
    \end{subfigure}\hspace{0.01\textwidth}
    \begin{subfigure}[b]{0.13\textwidth}
        \includegraphics[width=\linewidth]{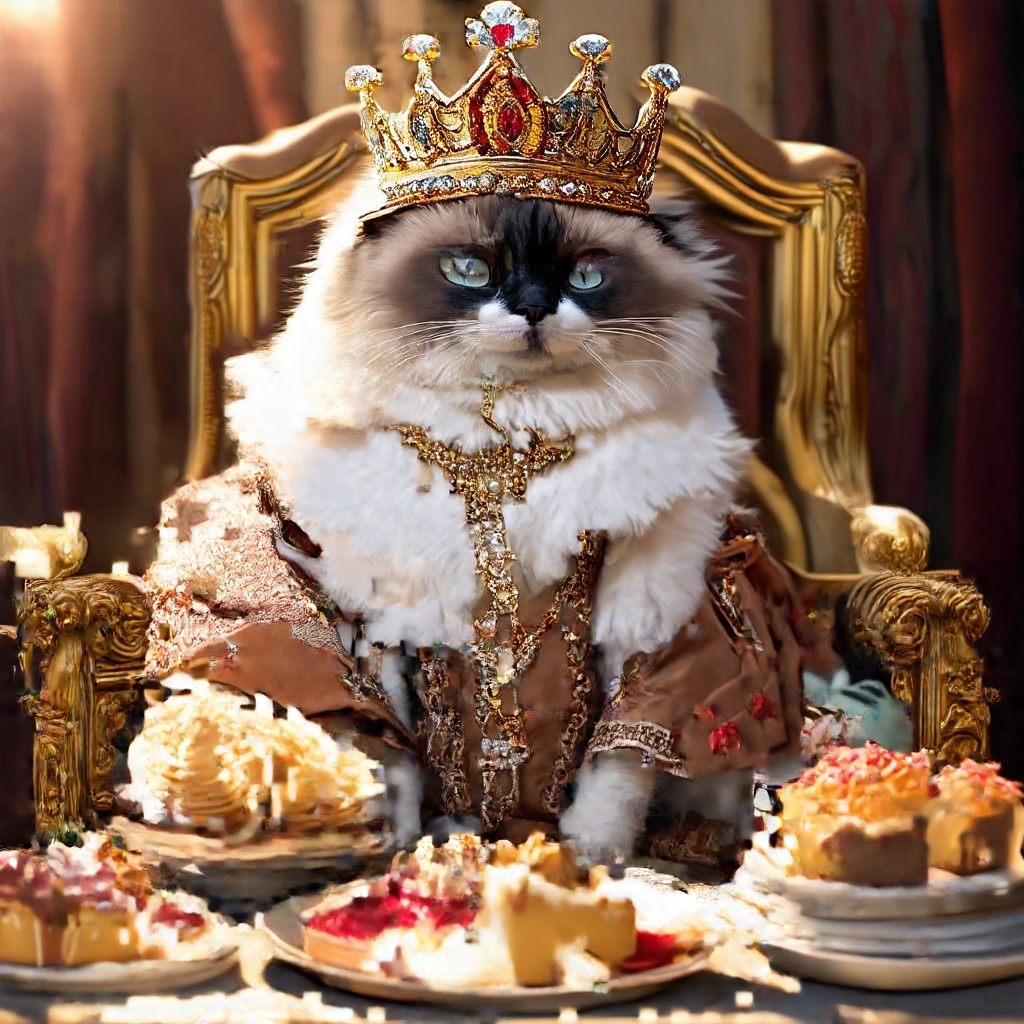}
        \caption*{NeuralSort}
    \end{subfigure}\hspace{0.01\textwidth}
    \begin{subfigure}[b]{0.13\textwidth}
        \includegraphics[width=\linewidth]{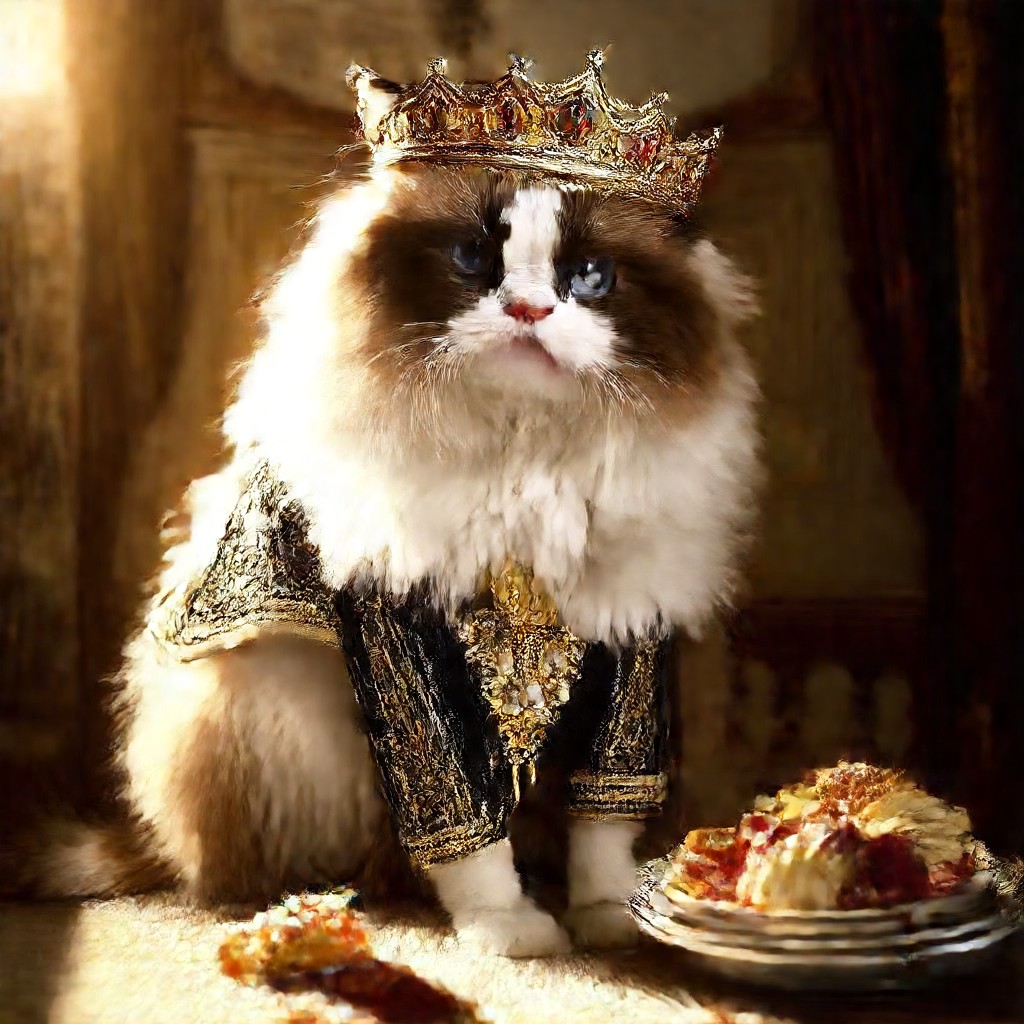}
        \caption*{SparseDiT}
    \end{subfigure}\hspace{0.01\textwidth}
    \begin{subfigure}[b]{0.13\textwidth}
        \includegraphics[width=\linewidth]{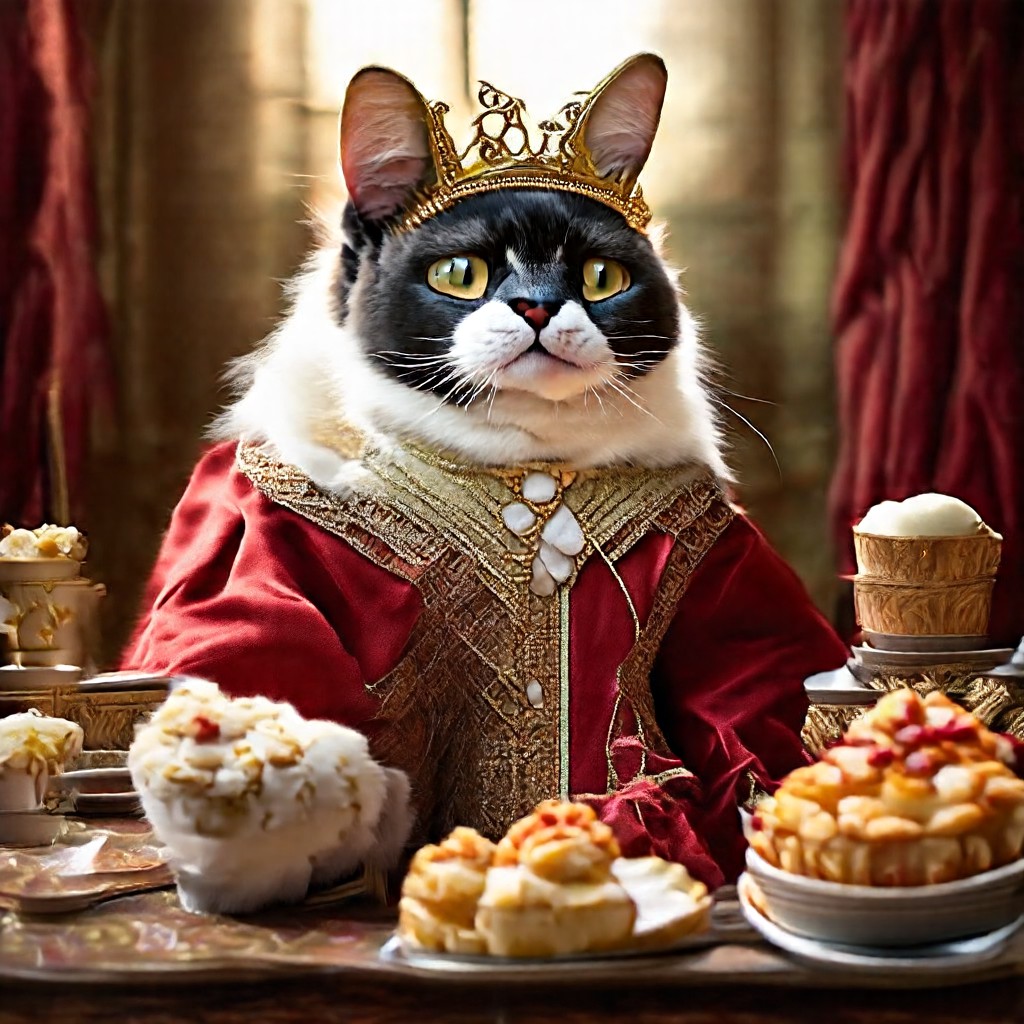}
        \caption*{DyDiT}
    \end{subfigure}

    \vspace{0.2em}

    \begin{subfigure}[b]{0.13\textwidth}
        \includegraphics[width=\linewidth]{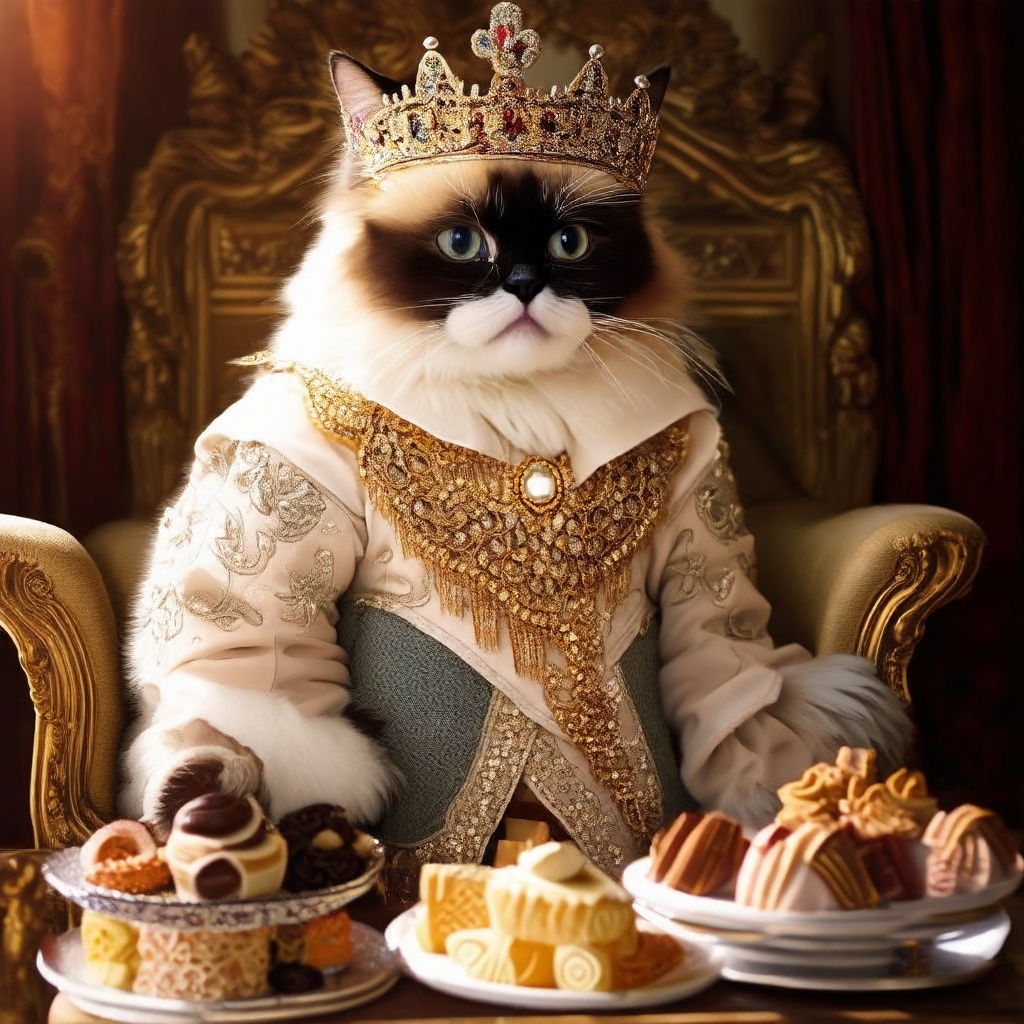}
        \caption*{\textbf{Shiva-90}}
    \end{subfigure}\hspace{0.01\textwidth}
    \begin{subfigure}[b]{0.13\textwidth}
        \includegraphics[width=\linewidth]{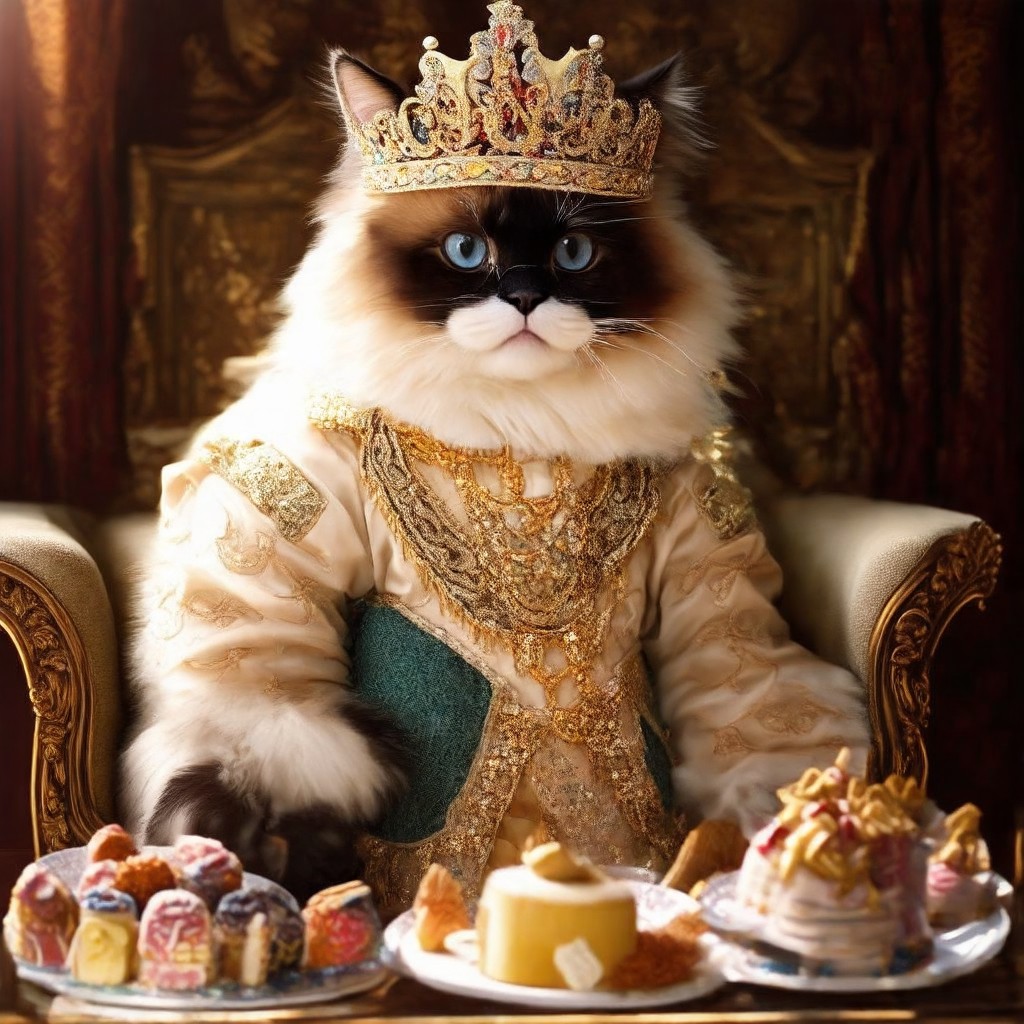}
        \caption*{\textbf{Shiva-80}}
    \end{subfigure}\hspace{0.01\textwidth}
    \begin{subfigure}[b]{0.13\textwidth}
        \includegraphics[width=\linewidth]{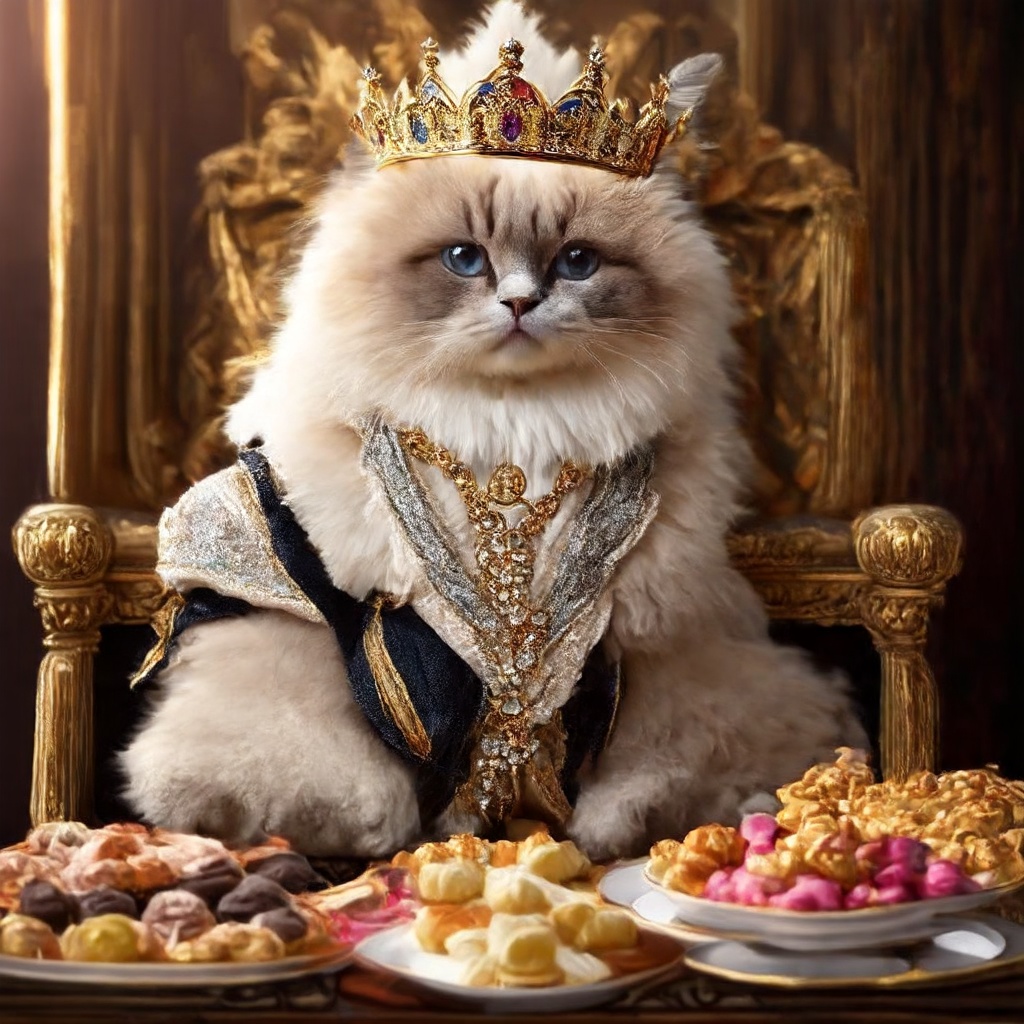}
        \caption*{\textbf{Shiva-70}}
    \end{subfigure}\hspace{0.01\textwidth}
    \begin{subfigure}[b]{0.13\textwidth}
        \includegraphics[width=\linewidth]{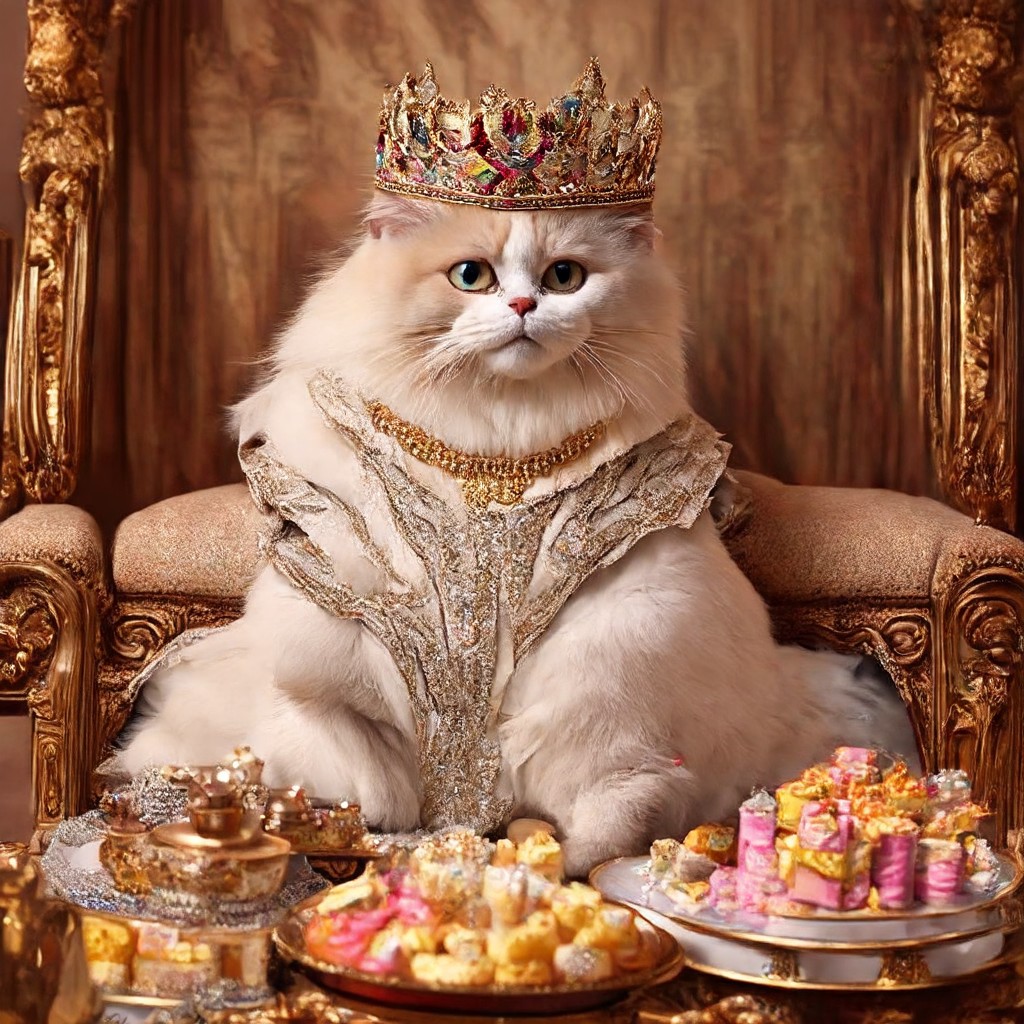}
        \caption*{\textbf{Shiva-60}}
    \end{subfigure}\hspace{0.01\textwidth}
    \begin{subfigure}[b]{0.13\textwidth}
        \includegraphics[width=\linewidth]{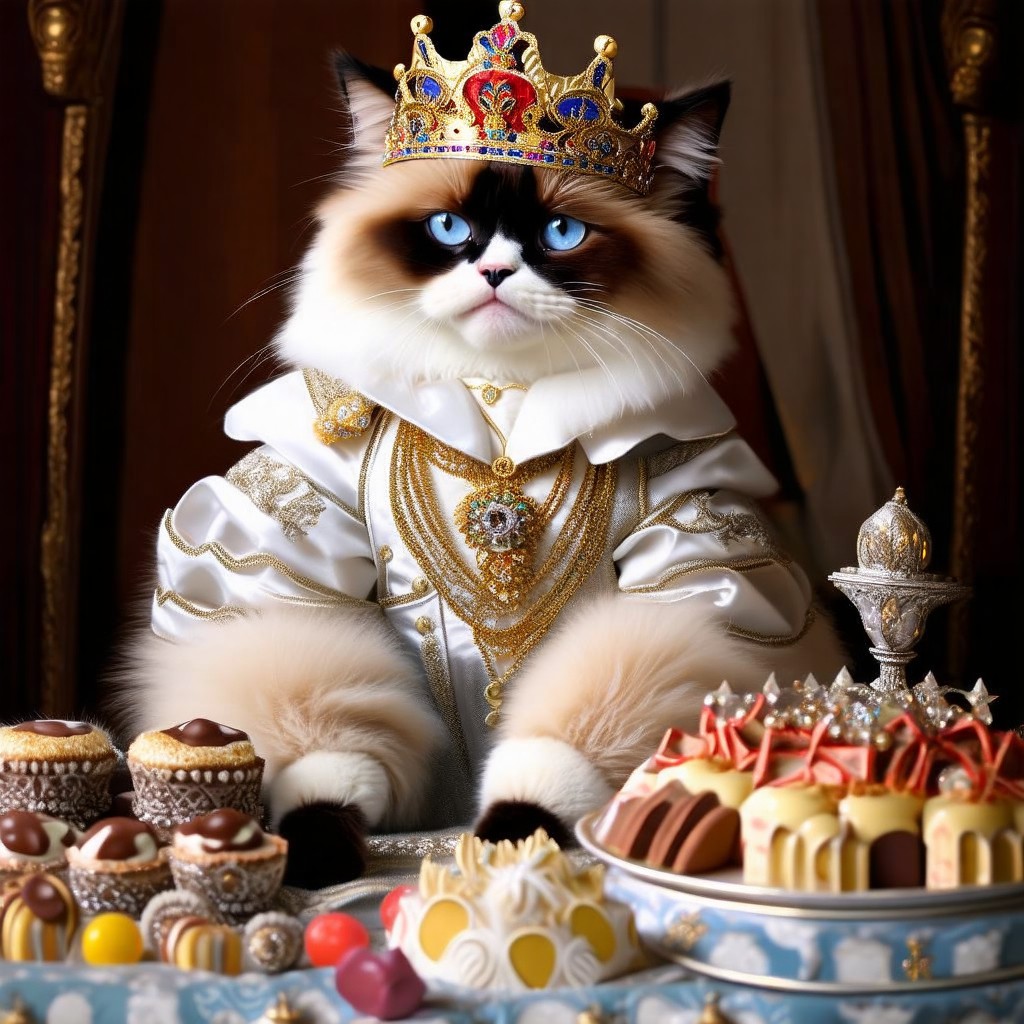}
        \caption*{Finetuned}
    \end{subfigure}

    \caption{Prompt: \textit{a ragdoll cat dressed in magnificent court attire, worn a crown, sparkling necklace, seated on a throne, surrounded by plates of sweets, realistic direct sun lighting, ultra detailed, super realistic, digital photography.}}
    \label{fig:appendix_fig_2}
\end{figure}

\begin{figure}[H]
    \centering
    \begin{subfigure}[b]{0.13\textwidth}
        \includegraphics[width=\linewidth]{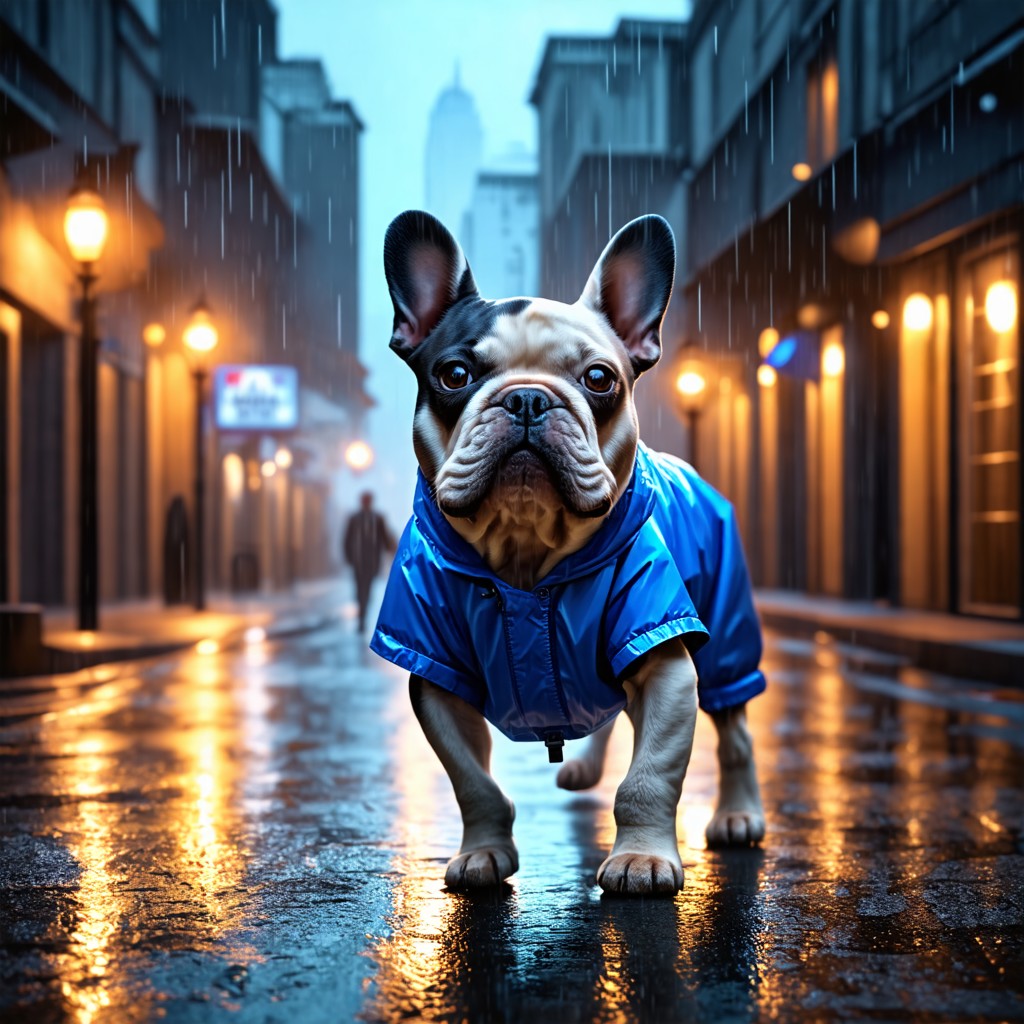}
        \caption*{Vanilla}
    \end{subfigure}\hspace{0.01\textwidth}
    \begin{subfigure}[b]{0.13\textwidth}
        \includegraphics[width=\linewidth]{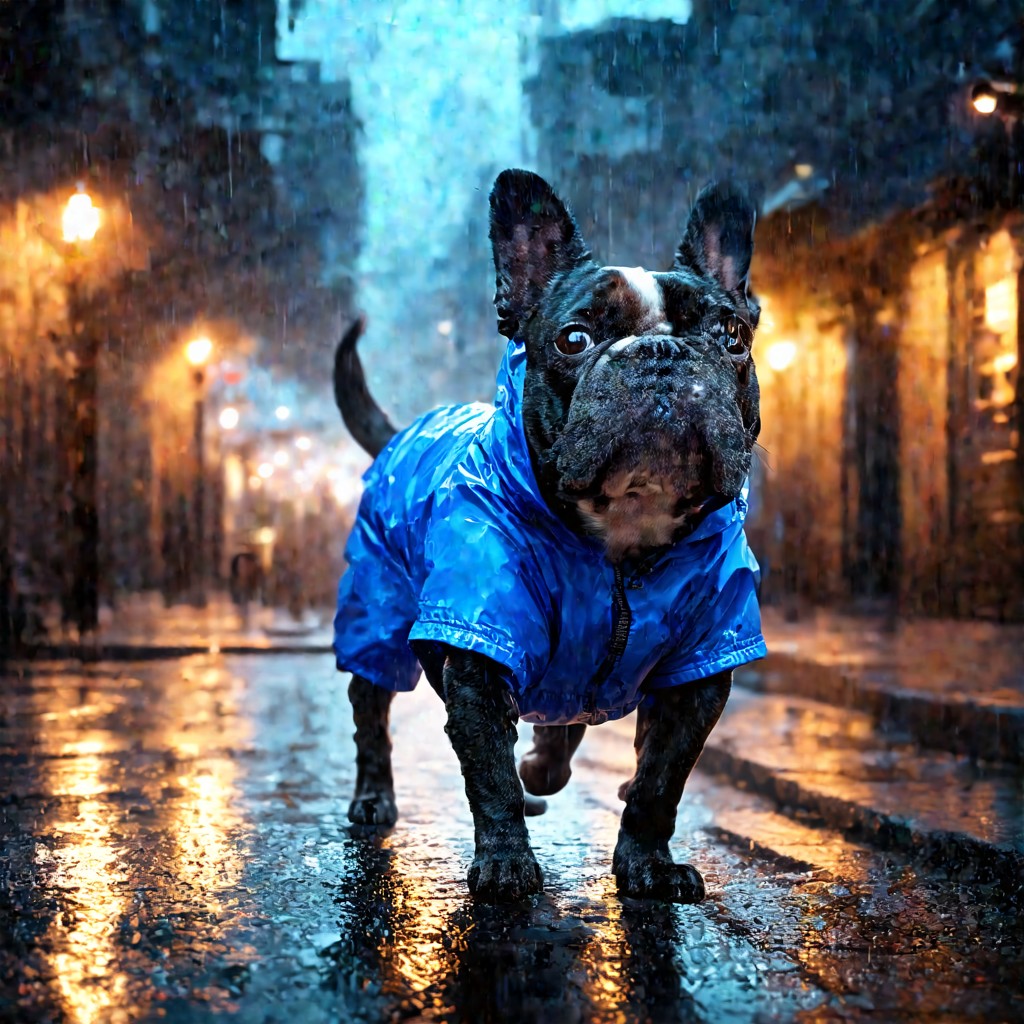}
        \caption*{ToMeSD}
    \end{subfigure}\hspace{0.01\textwidth}
    \begin{subfigure}[b]{0.13\textwidth}
        \includegraphics[width=\linewidth]{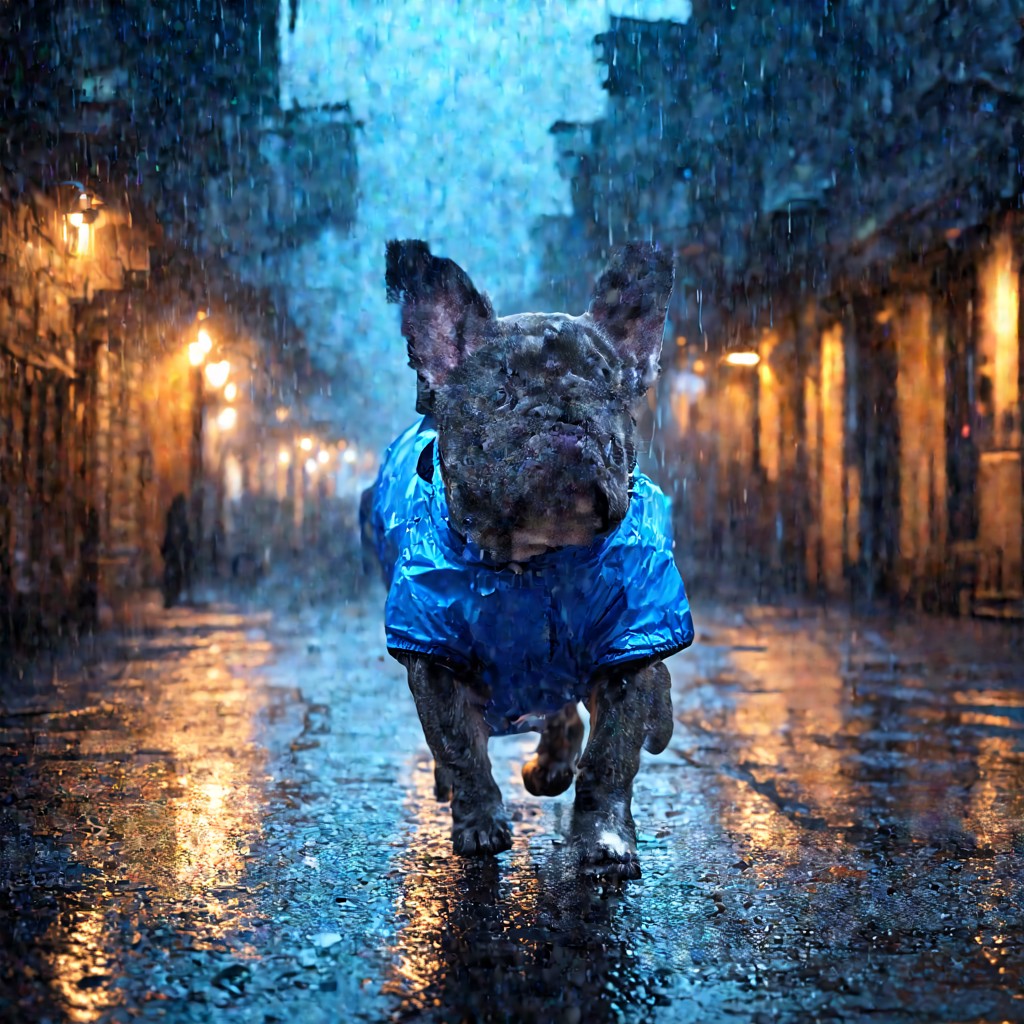}
        \caption*{ToFu}
    \end{subfigure}\hspace{0.01\textwidth}
    \begin{subfigure}[b]{0.13\textwidth}
        \includegraphics[width=\linewidth]{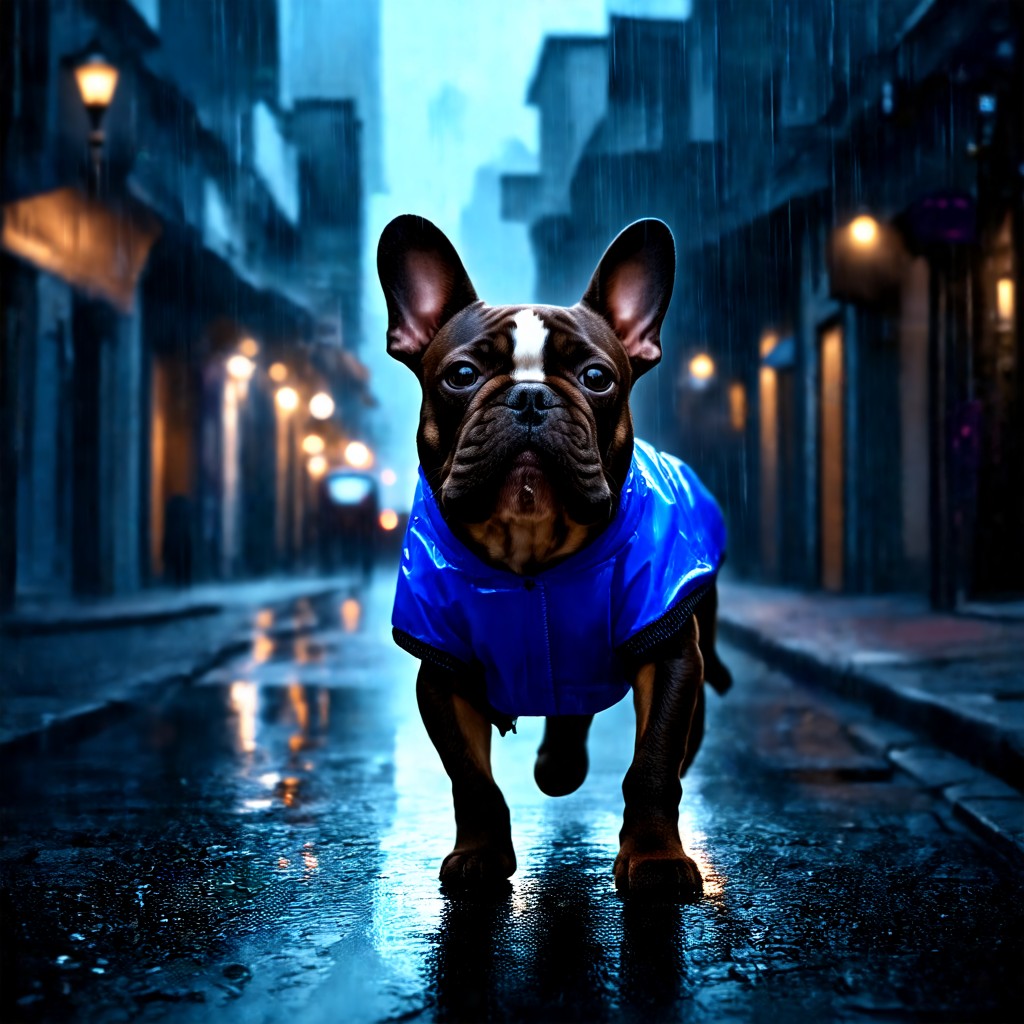}
        \caption*{SDTM}
    \end{subfigure}\hspace{0.01\textwidth}
    \begin{subfigure}[b]{0.13\textwidth}
        \includegraphics[width=\linewidth]{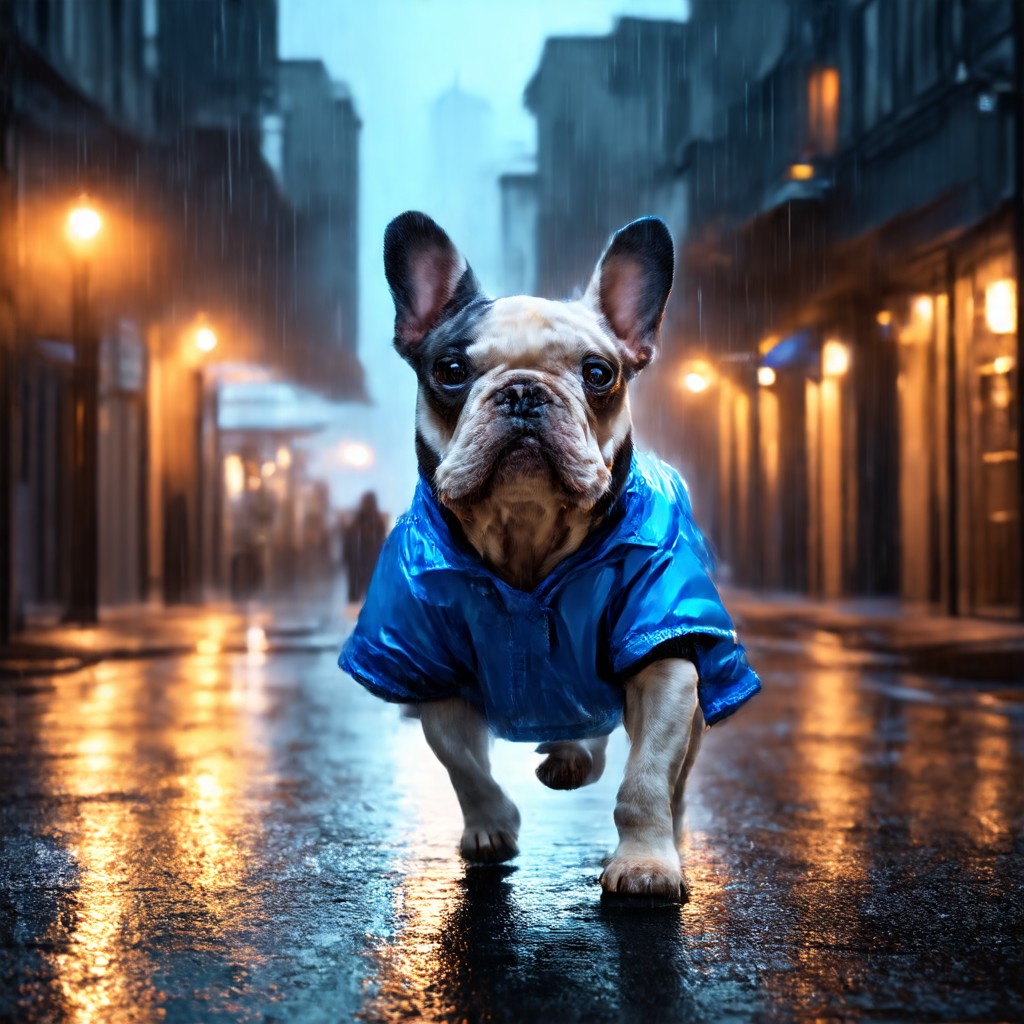}
        \caption*{IBTM}
    \end{subfigure}

    \vspace{0.2em}

    \begin{subfigure}[b]{0.13\textwidth}
        \includegraphics[width=\linewidth]{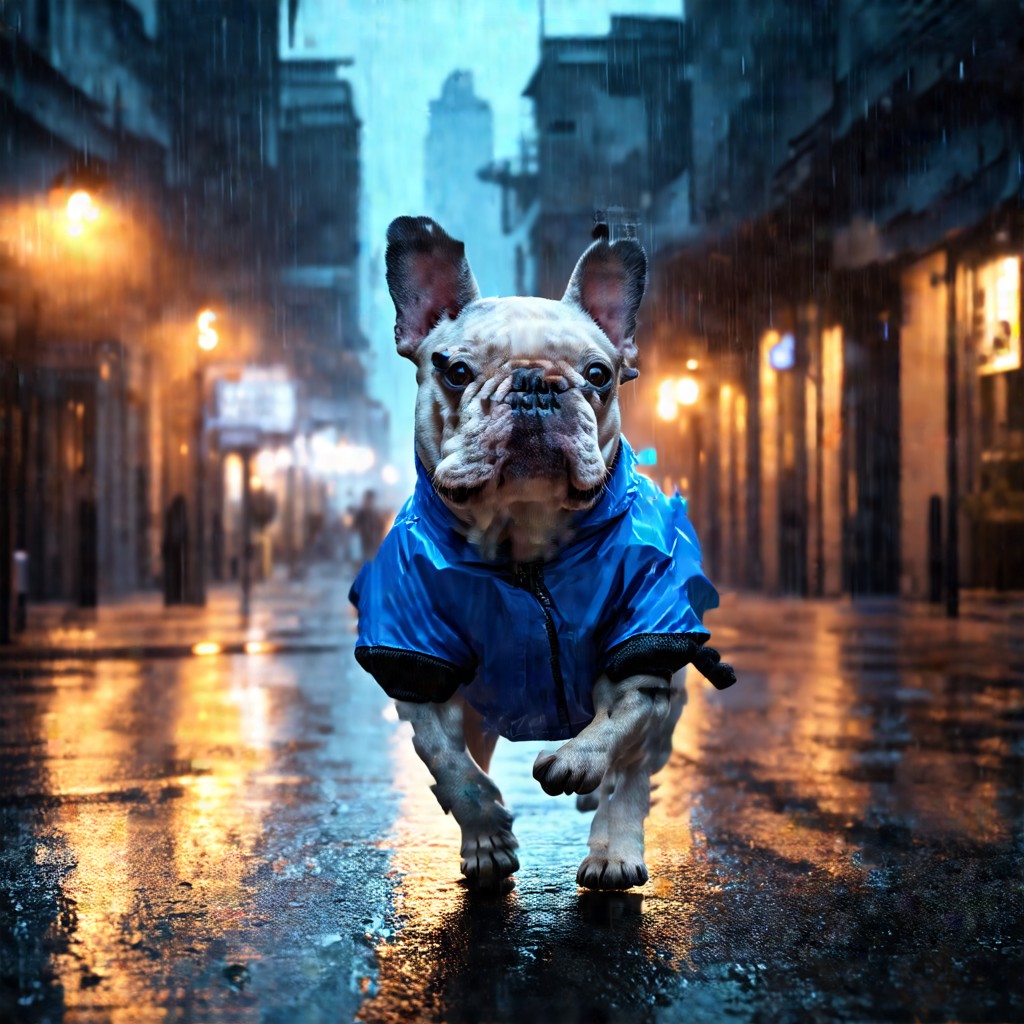}
        \caption*{ToMA}
    \end{subfigure}\hspace{0.01\textwidth}
    \begin{subfigure}[b]{0.13\textwidth}
        \includegraphics[width=\linewidth]{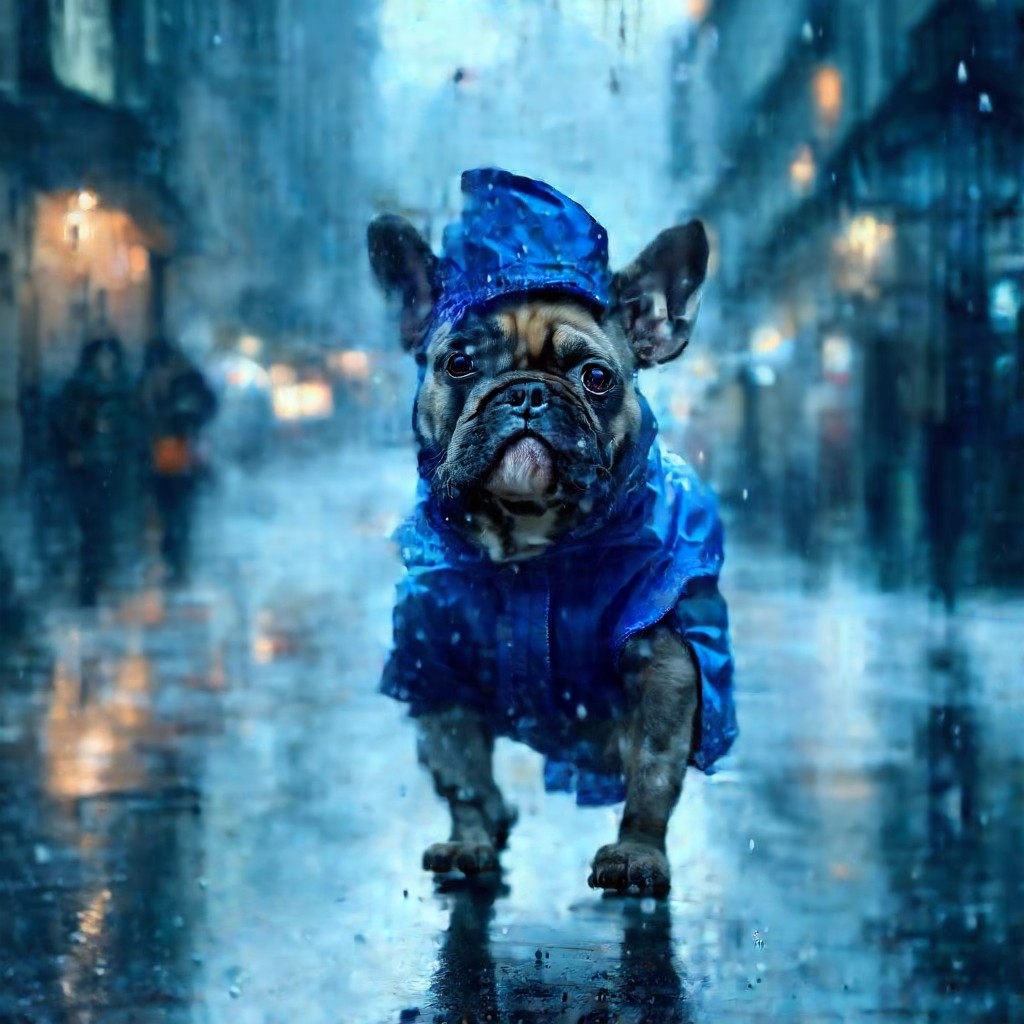}
        \caption*{DiffCR}
    \end{subfigure}\hspace{0.01\textwidth}
    \begin{subfigure}[b]{0.13\textwidth}
        \includegraphics[width=\linewidth]{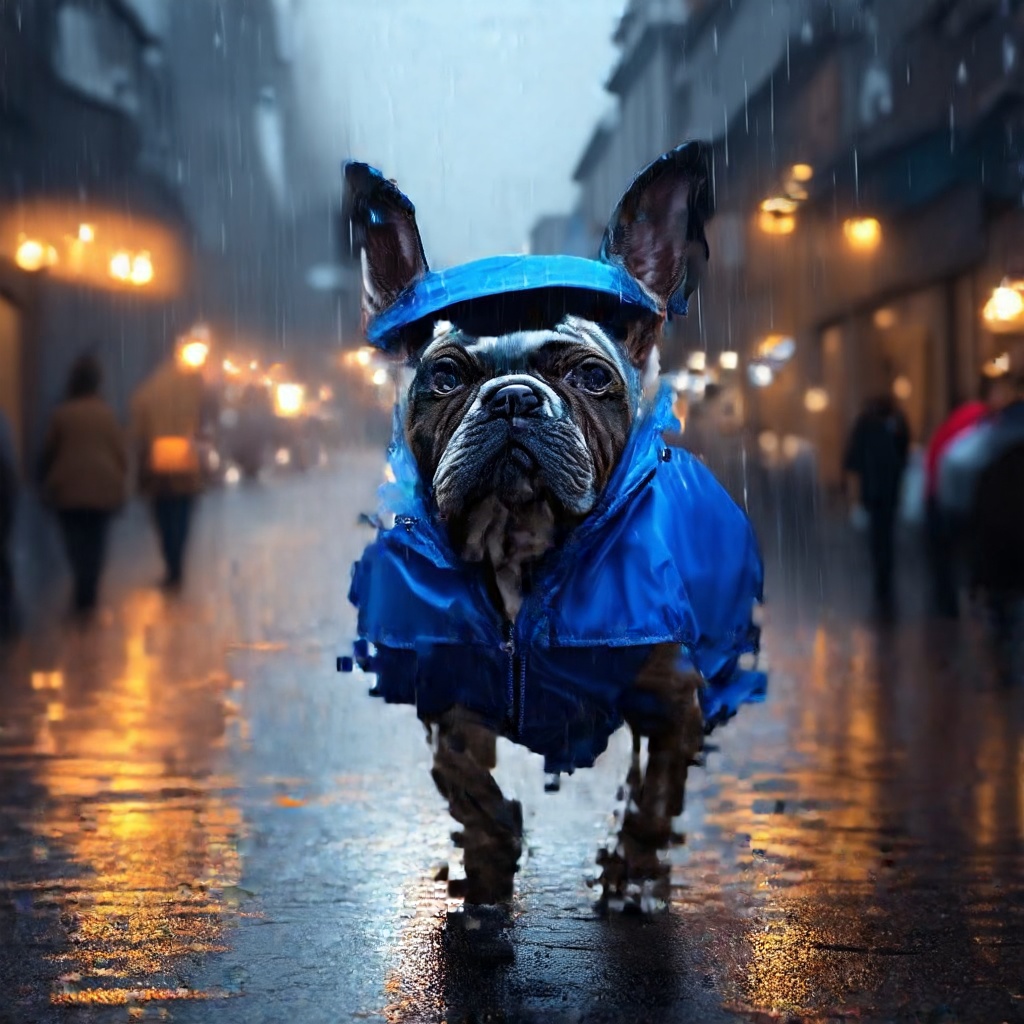}
        \caption*{NeuralSort}
    \end{subfigure}\hspace{0.01\textwidth}
    \begin{subfigure}[b]{0.13\textwidth}
        \includegraphics[width=\linewidth]{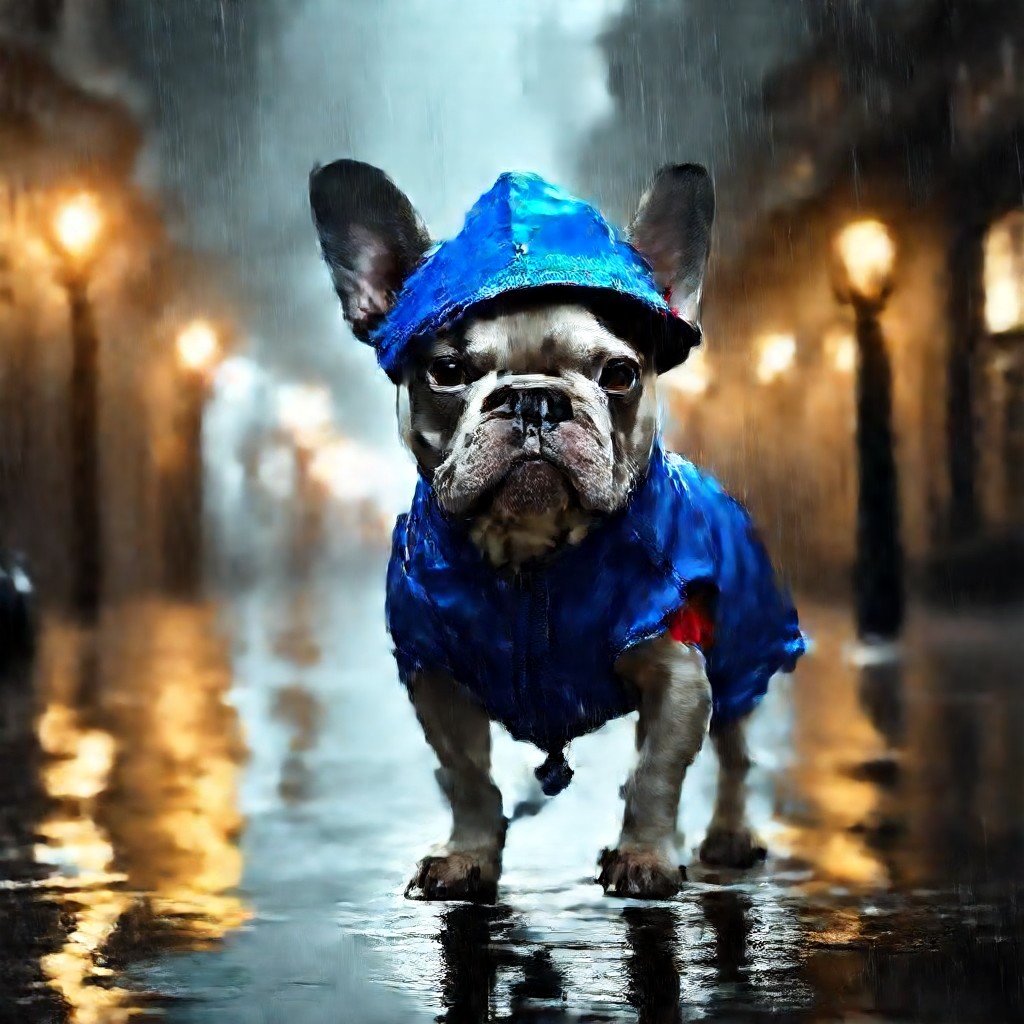}
        \caption*{SparseDiT}
    \end{subfigure}\hspace{0.01\textwidth}
    \begin{subfigure}[b]{0.13\textwidth}
        \includegraphics[width=\linewidth]{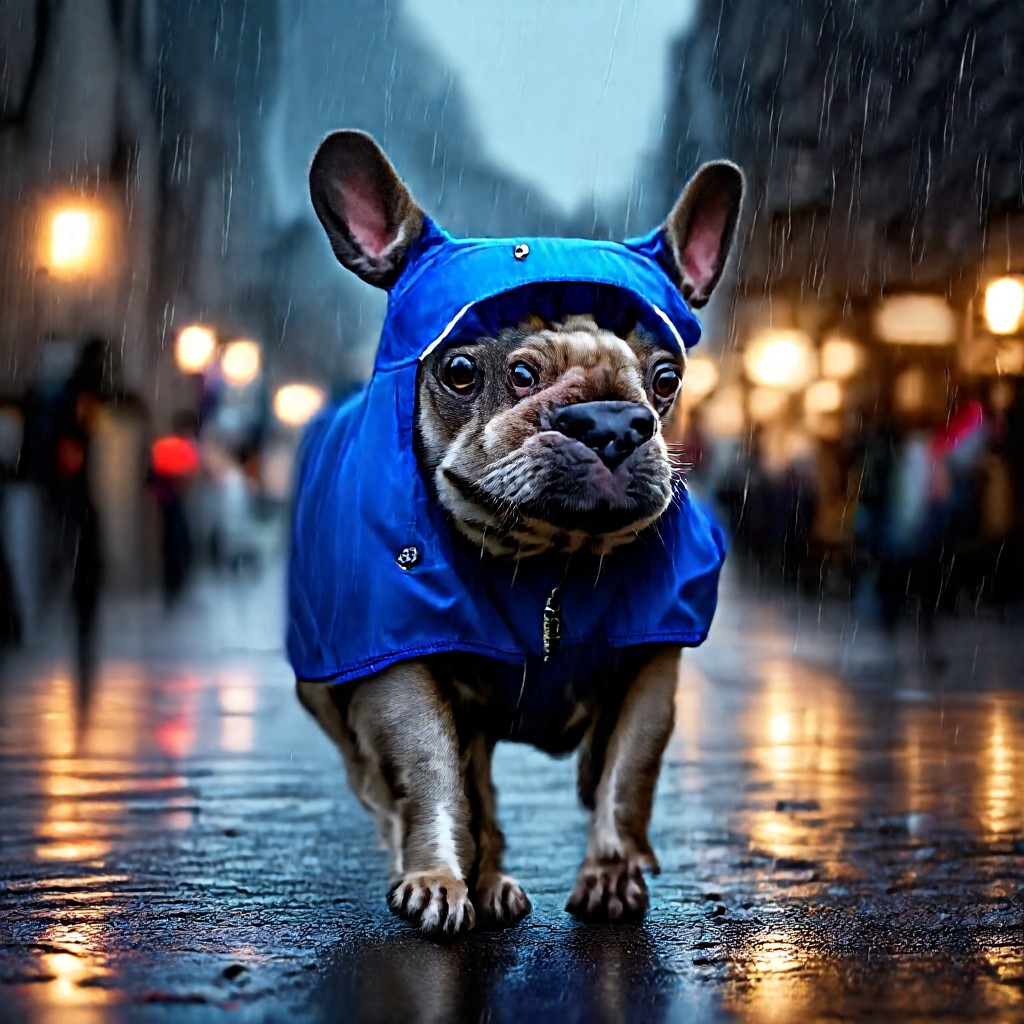}
        \caption*{DyDiT}
    \end{subfigure}

    \vspace{0.2em}

    \begin{subfigure}[b]{0.13\textwidth}
        \includegraphics[width=\linewidth]{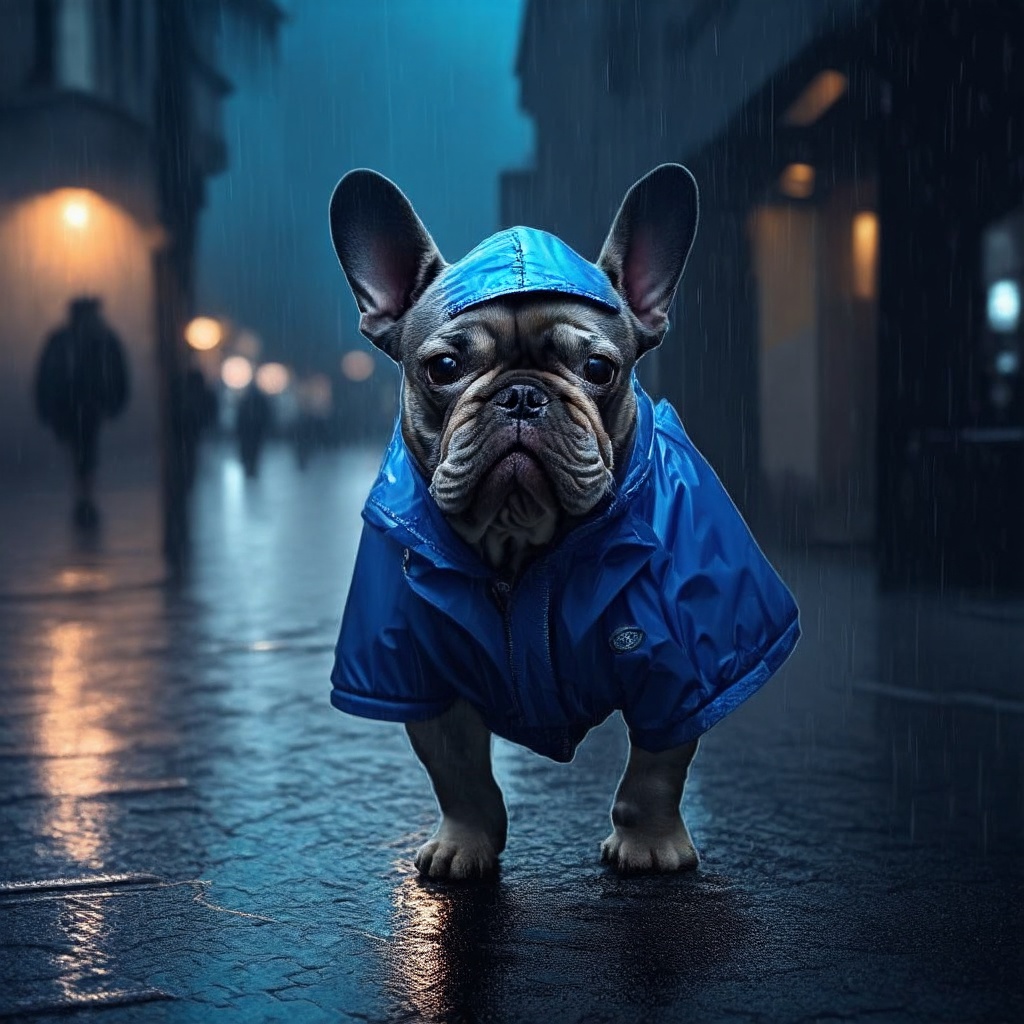}
        \caption*{\textbf{Shiva-90}}
    \end{subfigure}\hspace{0.01\textwidth}
    \begin{subfigure}[b]{0.13\textwidth}
        \includegraphics[width=\linewidth]{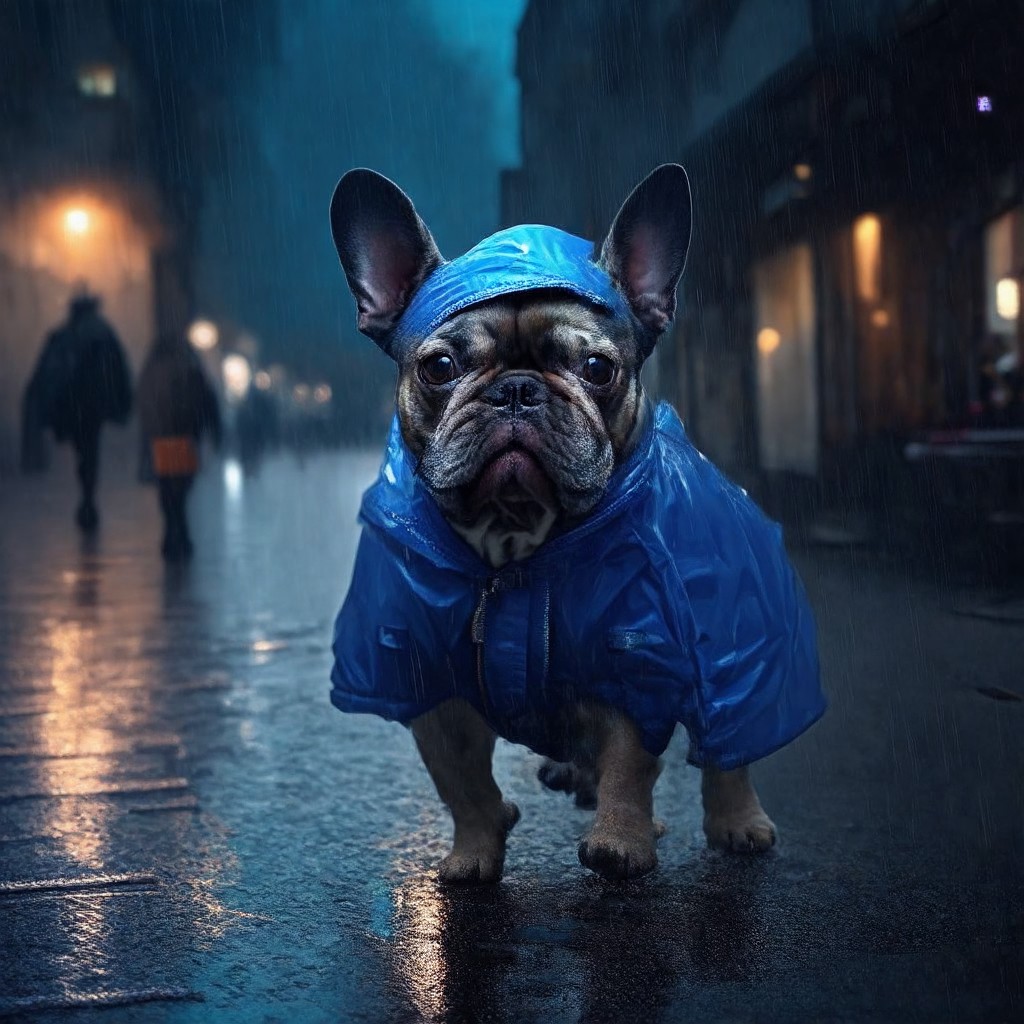}
        \caption*{\textbf{Shiva-80}}
    \end{subfigure}\hspace{0.01\textwidth}
    \begin{subfigure}[b]{0.13\textwidth}
        \includegraphics[width=\linewidth]{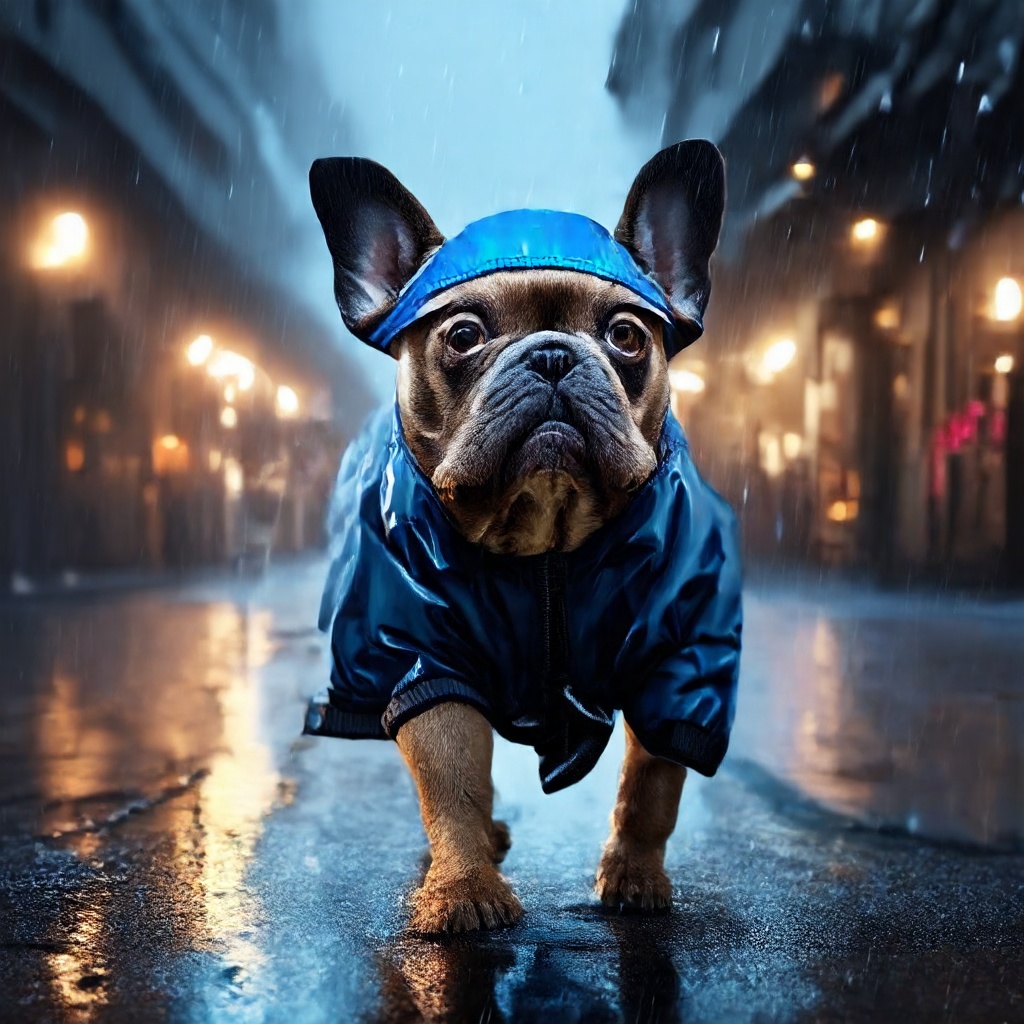}
        \caption*{\textbf{Shiva-70}}
    \end{subfigure}\hspace{0.01\textwidth}
    \begin{subfigure}[b]{0.13\textwidth}
        \includegraphics[width=\linewidth]{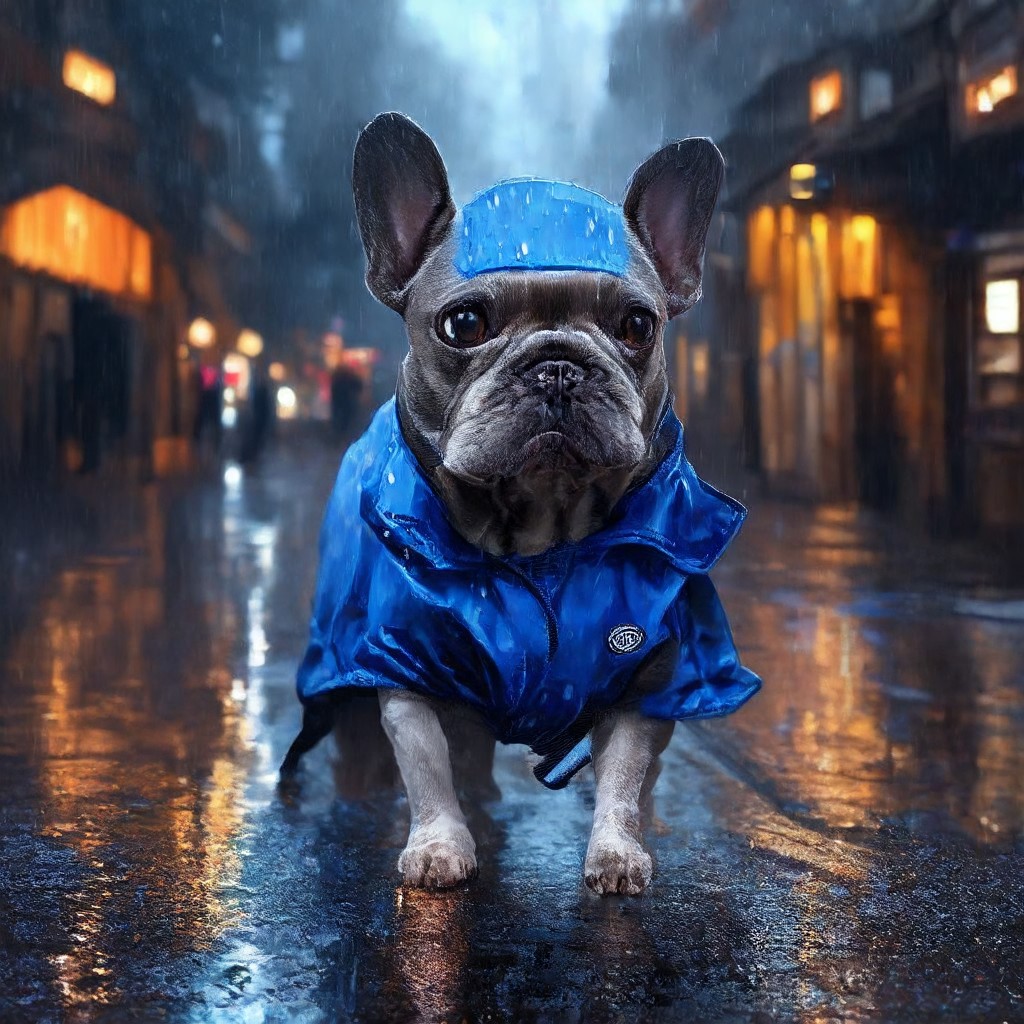}
        \caption*{\textbf{Shiva-60}}
    \end{subfigure}\hspace{0.01\textwidth}
    \begin{subfigure}[b]{0.13\textwidth}
        \includegraphics[width=\linewidth]{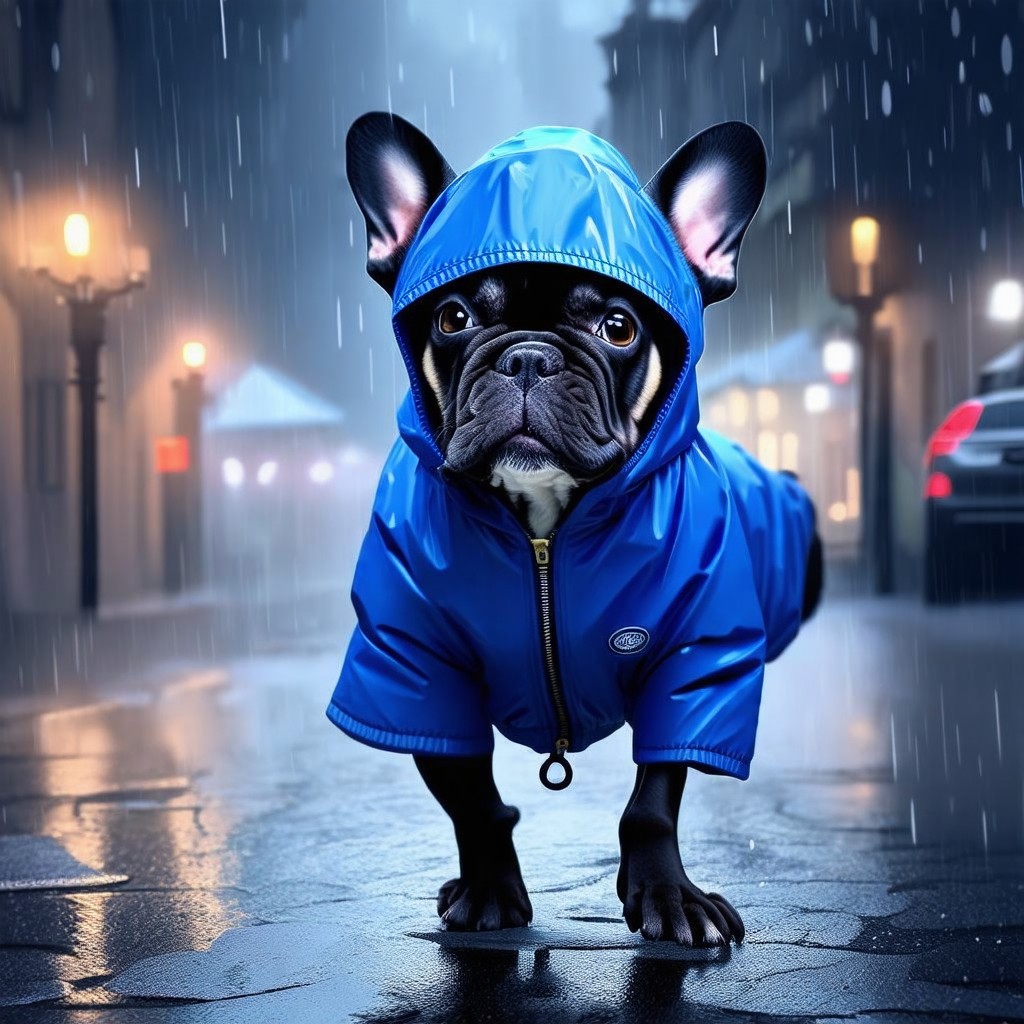}
        \caption*{Finetuned}
    \end{subfigure}

    \caption{Prompt: \textit{Design a superrealistic digital painting of a French Bulldog in a rainy urban scene, wearing a rain jacket. Use dim and moody lighting with street lights and reflections on the pavement to create the perfect atmosphere. Make the rain jacket bright blue and the background dark blue and grey to contrast. Emphasize the urban environment and the rain with a midshot of the dog walking towards the camera.}}
    \label{fig:appendix_fig_3}
\end{figure}

Figures~\ref{fig:appendix_fig_1}, \ref{fig:appendix_fig_2}, and~\ref{fig:appendix_fig_3} compare SD3-Medium outputs across methods. The Shiva outputs retain the bracelet links and red starfish pendant in Figure~\ref{fig:appendix_fig_1}, the cat's crown and garment details in Figure~\ref{fig:appendix_fig_2}, and the blue raincoat and wet-street reflections in Figure~\ref{fig:appendix_fig_3}.

\FloatBarrier
\subsection{Flux.1-dev and PixArt-\texorpdfstring{$\Sigma$}{Sigma}}
\label{app:add_models}

\begin{figure}[H]
    \centering
    \begin{subfigure}[b]{0.155\textwidth}
        \includegraphics[width=\linewidth]{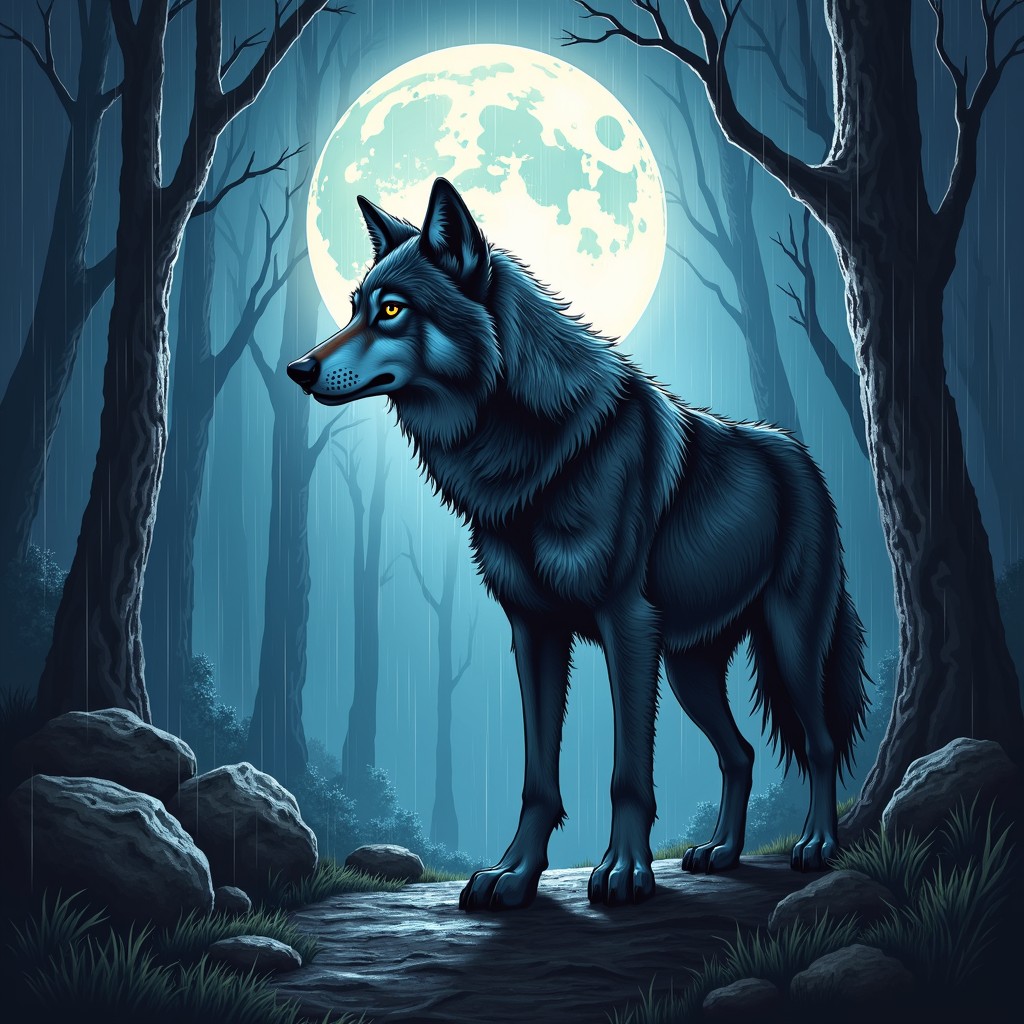}
        \caption*{Vanilla-Flux}
    \end{subfigure}
    \hfill
    \begin{subfigure}[b]{0.155\textwidth}
        \includegraphics[width=\linewidth]{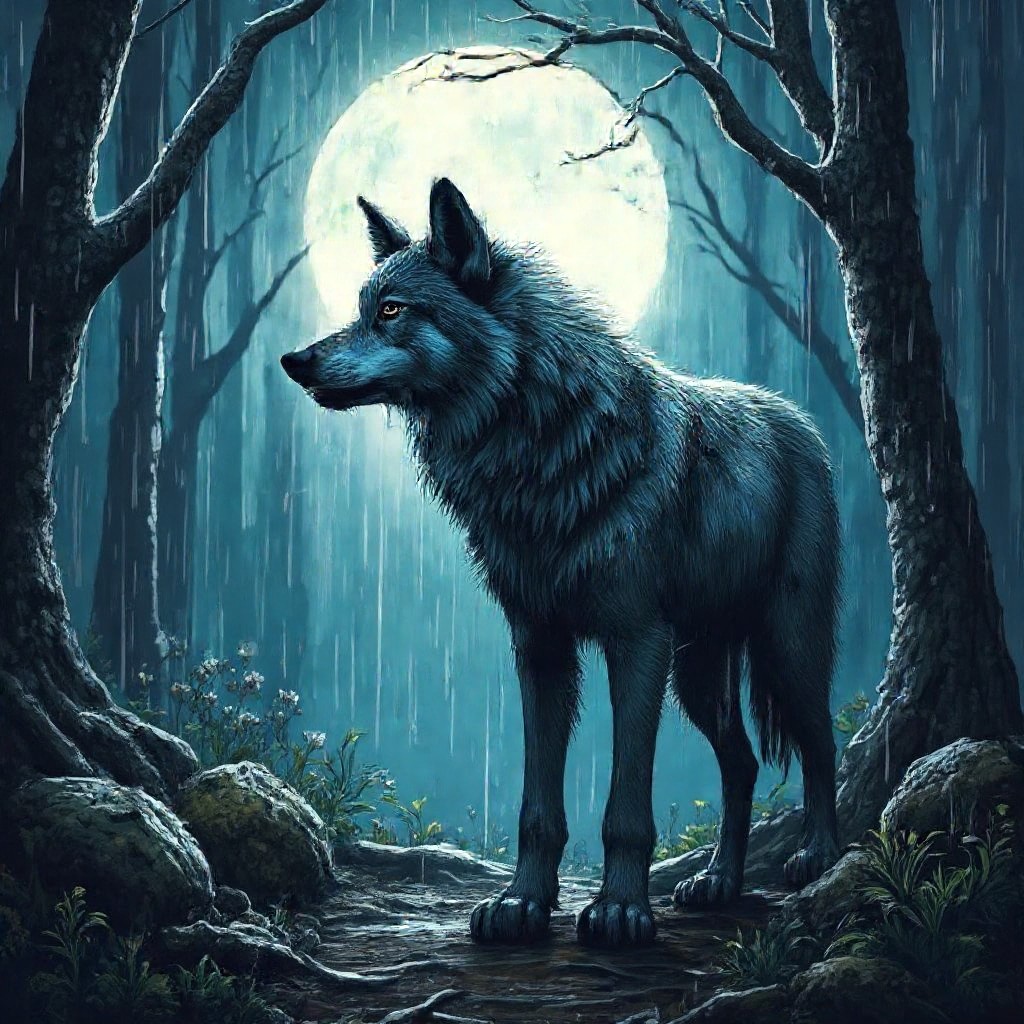}
        \caption*{\textbf{Shiva-Flux}}
    \end{subfigure}
    \hfill
    \begin{subfigure}[b]{0.155\textwidth}
        \includegraphics[width=\linewidth]{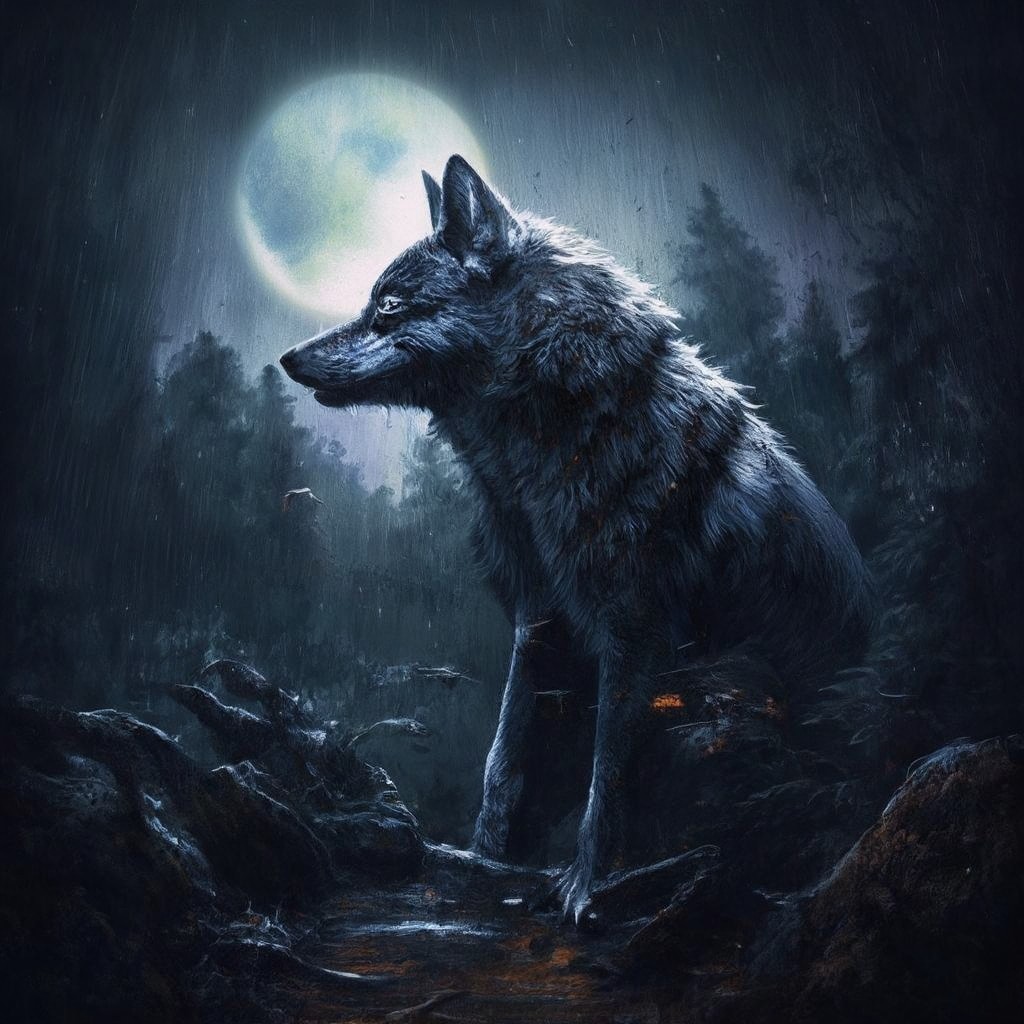}
        \caption*{Finetuned-Flux}
    \end{subfigure}
    \hfill
    \begin{subfigure}[b]{0.155\textwidth}
        \includegraphics[width=\linewidth]{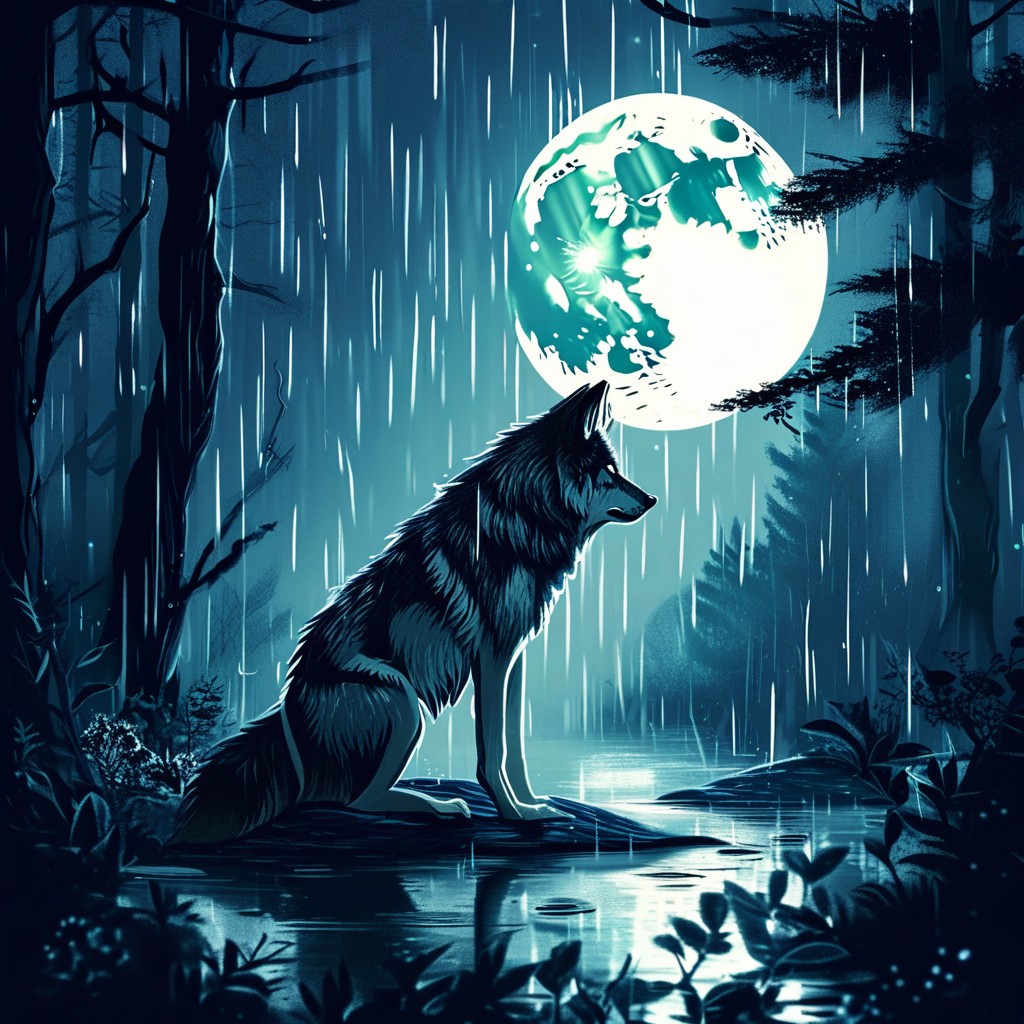}
        \caption*{Vanilla-Pixart}
    \end{subfigure}
    \hfill
    \begin{subfigure}[b]{0.155\textwidth}
        \includegraphics[width=\linewidth]{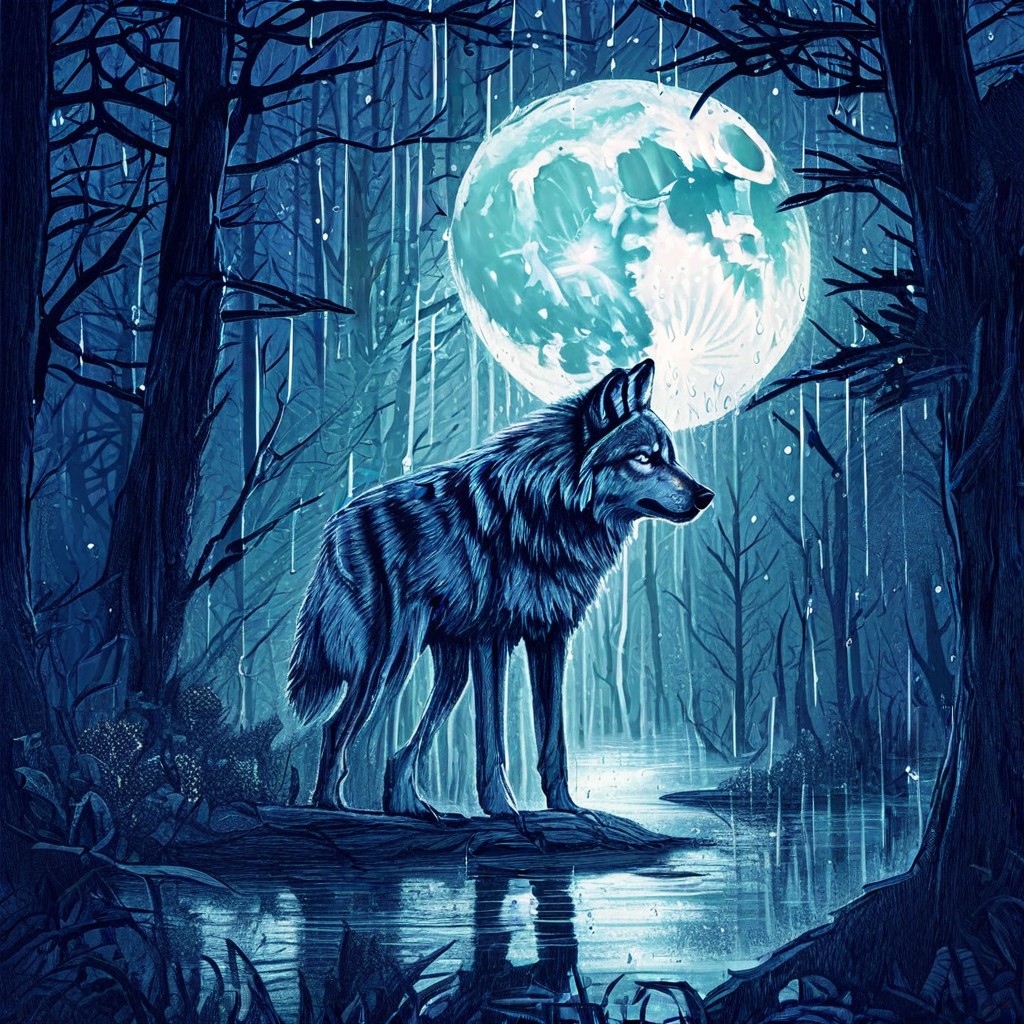}
        \caption*{\textbf{Shiva-Pixart}}
    \end{subfigure}
    \hfill
    \begin{subfigure}[b]{0.155\textwidth}
        \includegraphics[width=\linewidth]{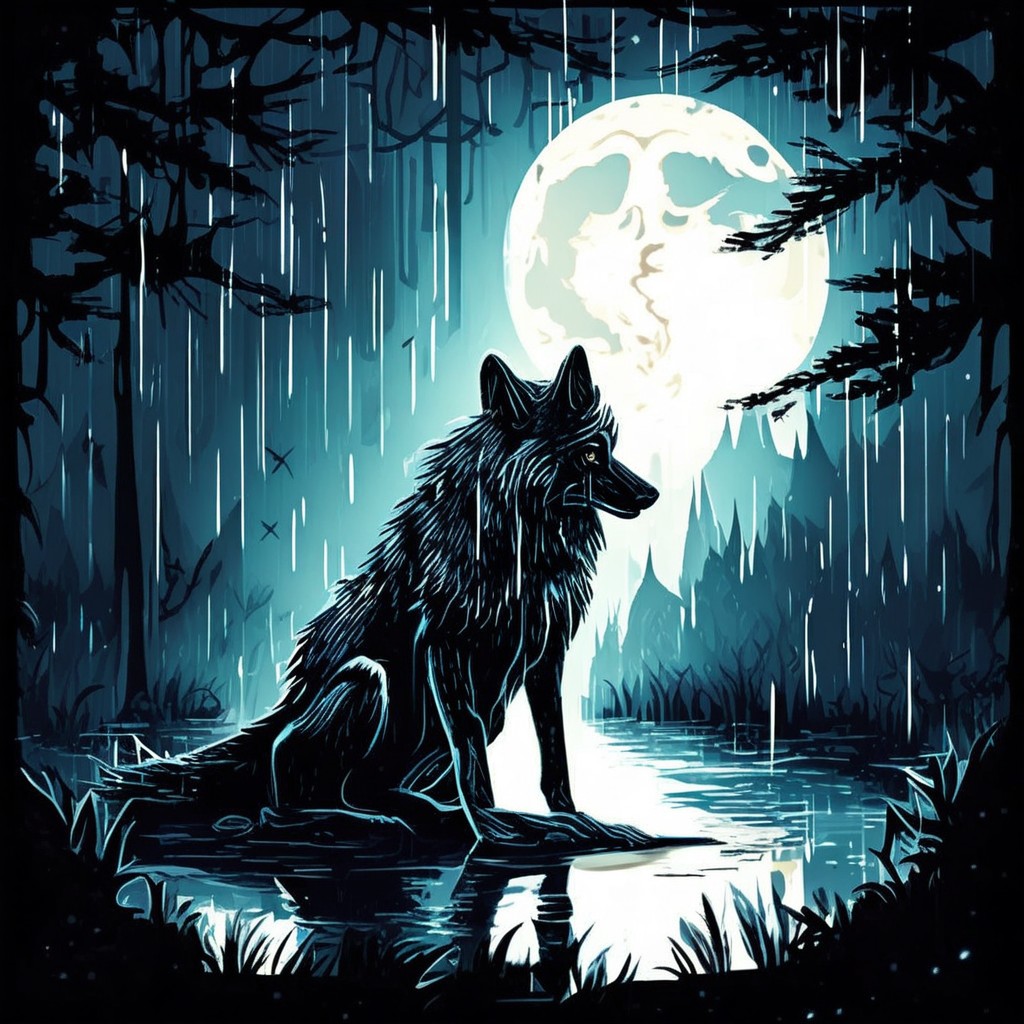}
        \caption*{Finetuned-Pixart}
    \end{subfigure}
    
    \caption{Prompt: \textit{Wolf in fantasy forest, moonlight with rain, inked style.}}
    \label{fig:appendix_fig_4}
\end{figure}

\begin{figure}[H]
    \centering
    \begin{subfigure}[b]{0.155\textwidth}
        \includegraphics[width=\linewidth]{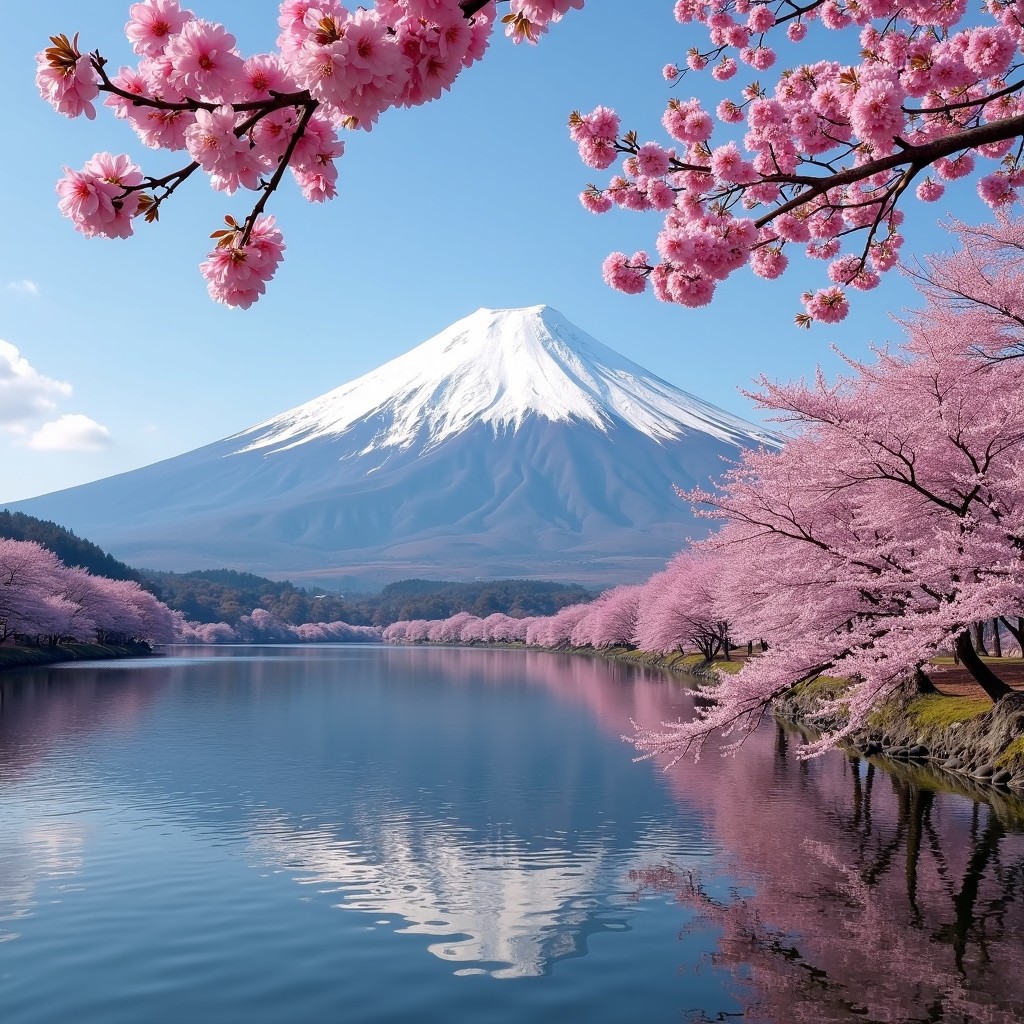}
        \caption*{Vanilla-Flux}
    \end{subfigure}
    \hfill
    \begin{subfigure}[b]{0.155\textwidth}
        \includegraphics[width=\linewidth]{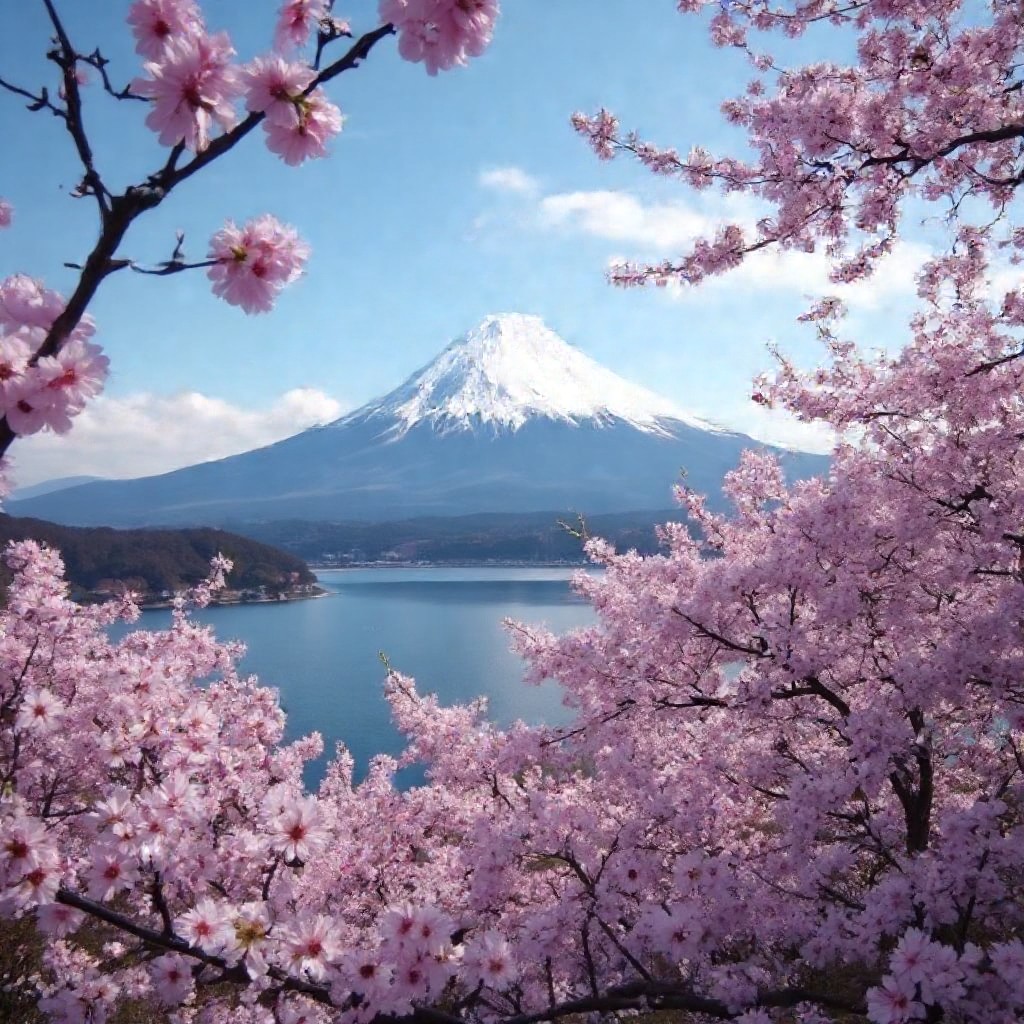}
        \caption*{\textbf{Shiva-Flux}}
    \end{subfigure}
    \hfill
    \begin{subfigure}[b]{0.155\textwidth}
        \includegraphics[width=\linewidth]{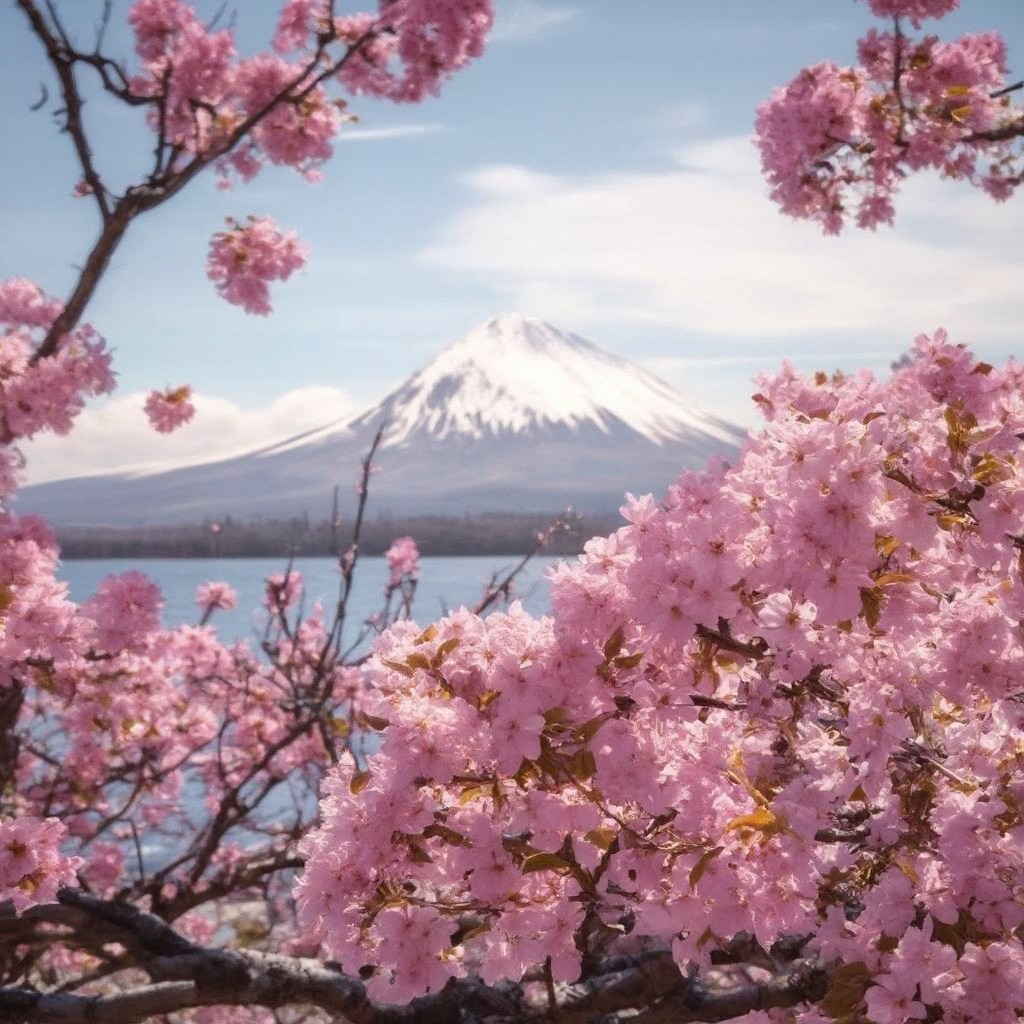}
        \caption*{Finetuned-Flux}
    \end{subfigure}
    \hfill
    \begin{subfigure}[b]{0.155\textwidth}
        \includegraphics[width=\linewidth]{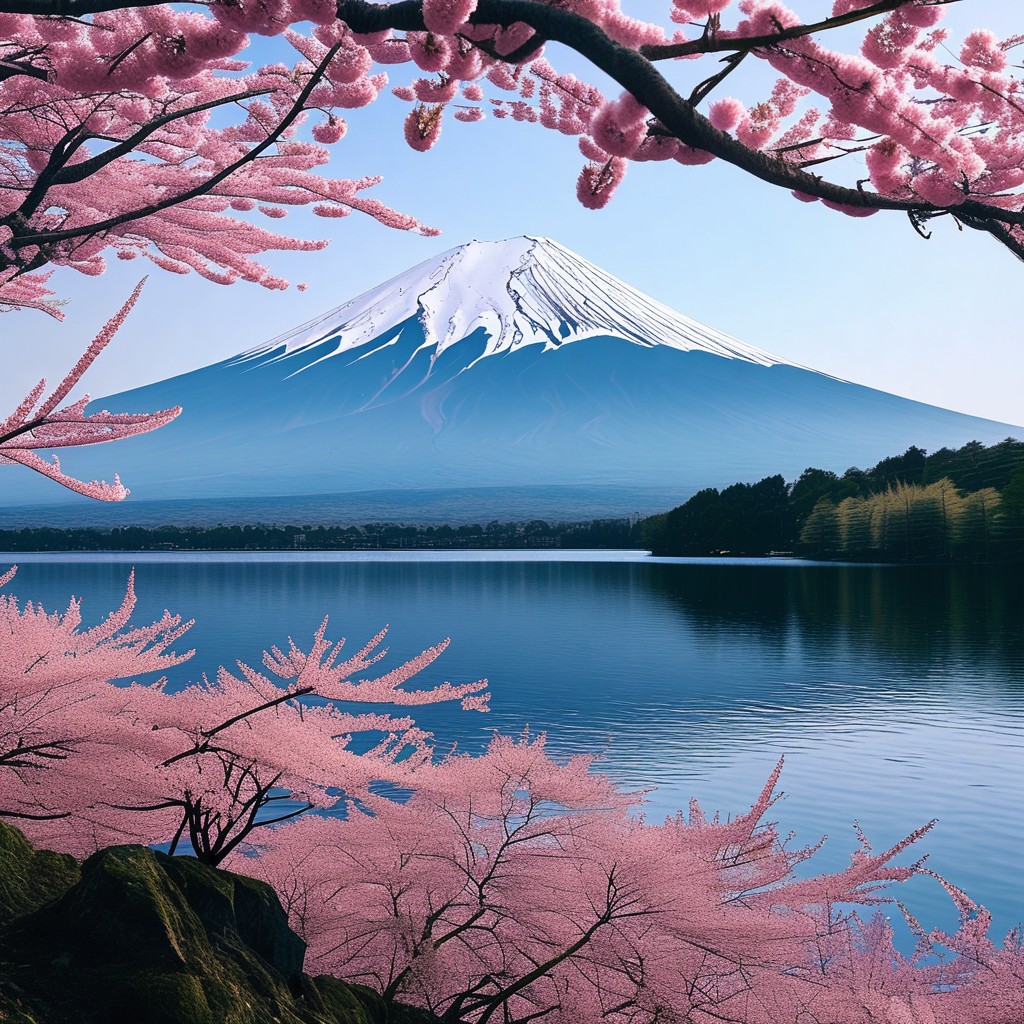}
        \caption*{Vanilla-Pixart}
    \end{subfigure}
    \hfill
    \begin{subfigure}[b]{0.155\textwidth}
        \includegraphics[width=\linewidth]{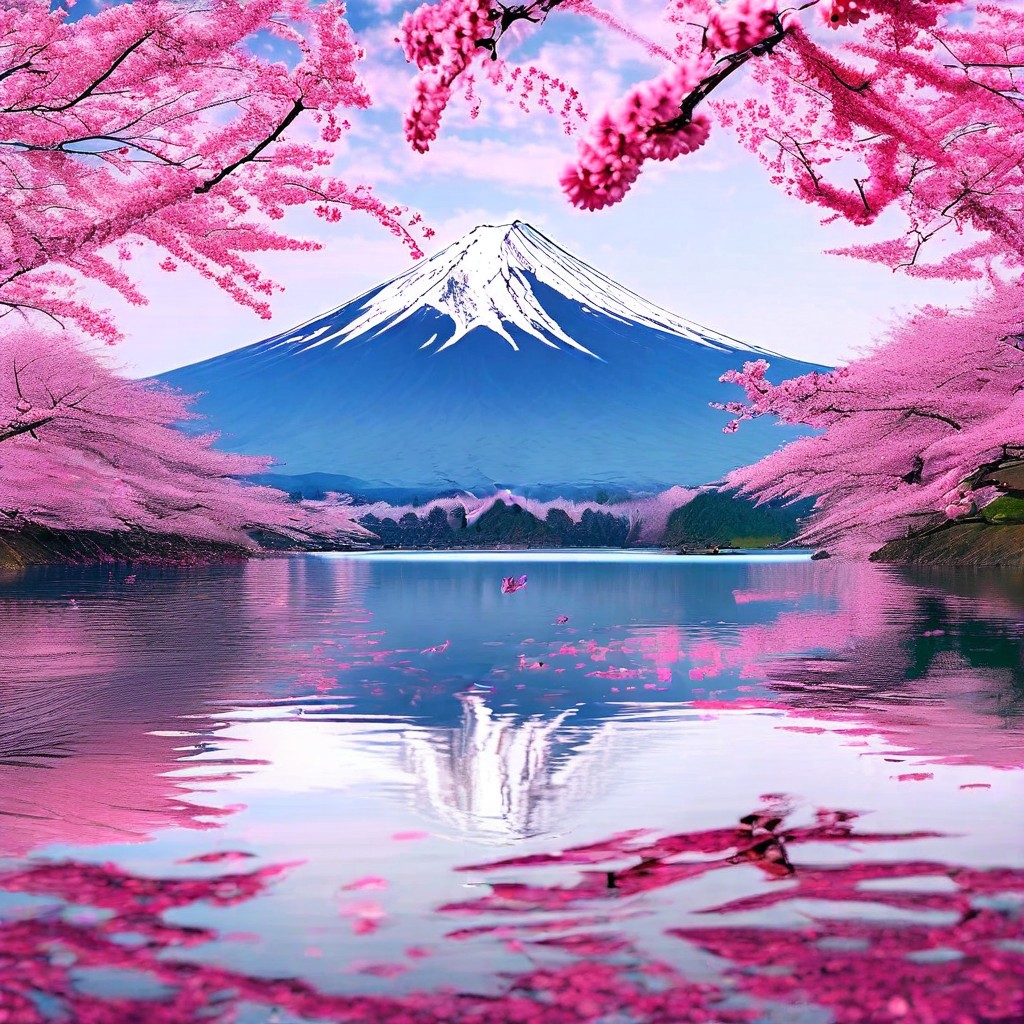}
        \caption*{\textbf{Shiva-Pixart}}
    \end{subfigure}
    \hfill
    \begin{subfigure}[b]{0.155\textwidth}
        \includegraphics[width=\linewidth]{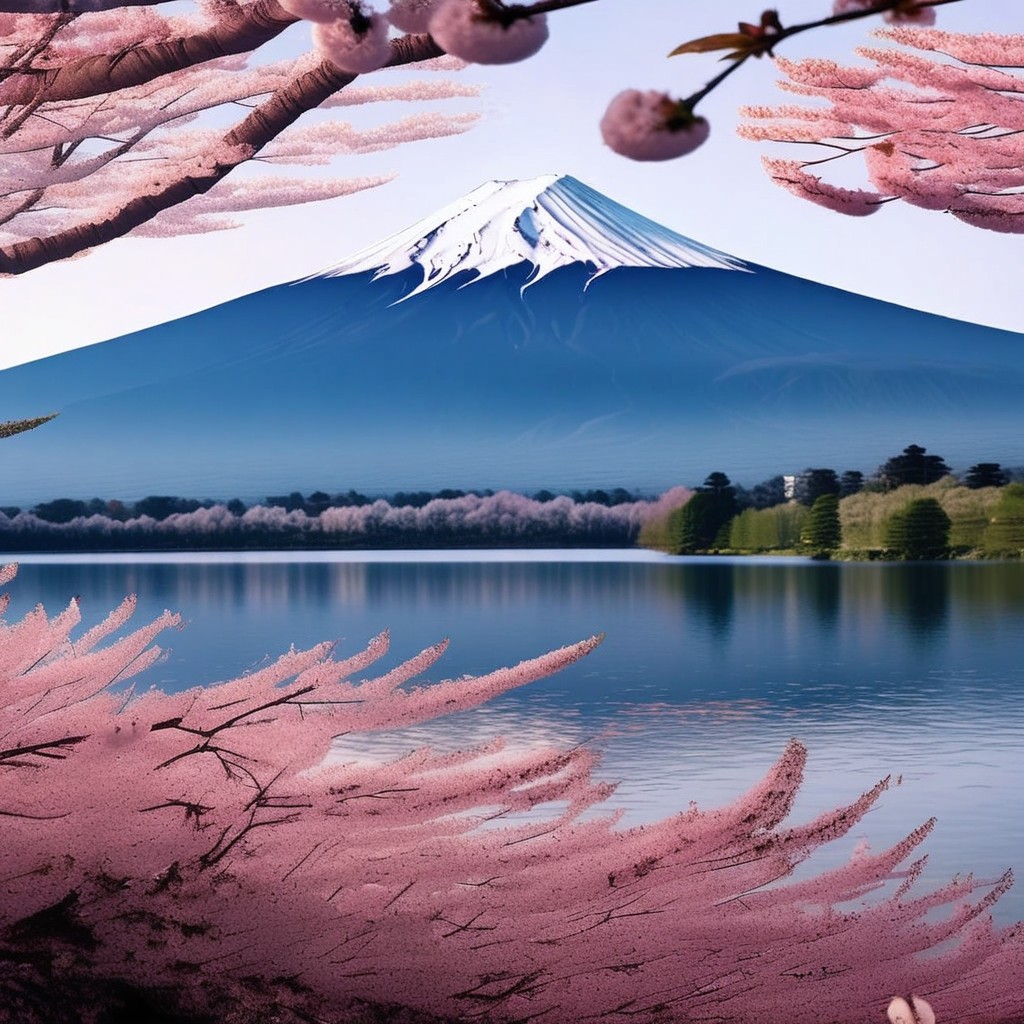}
        \caption*{Finetuned-Pixart}
    \end{subfigure}
    
    \caption{Prompt: \textit{Hyperdetailed photography, Mount Fuji behind cherry blossoms overlooking Saiko lake.}}
    \label{fig:appendix_fig_5}
\end{figure}

\begin{figure}[H]
    \centering
    \begin{subfigure}[b]{0.155\textwidth}
        \includegraphics[width=\linewidth]{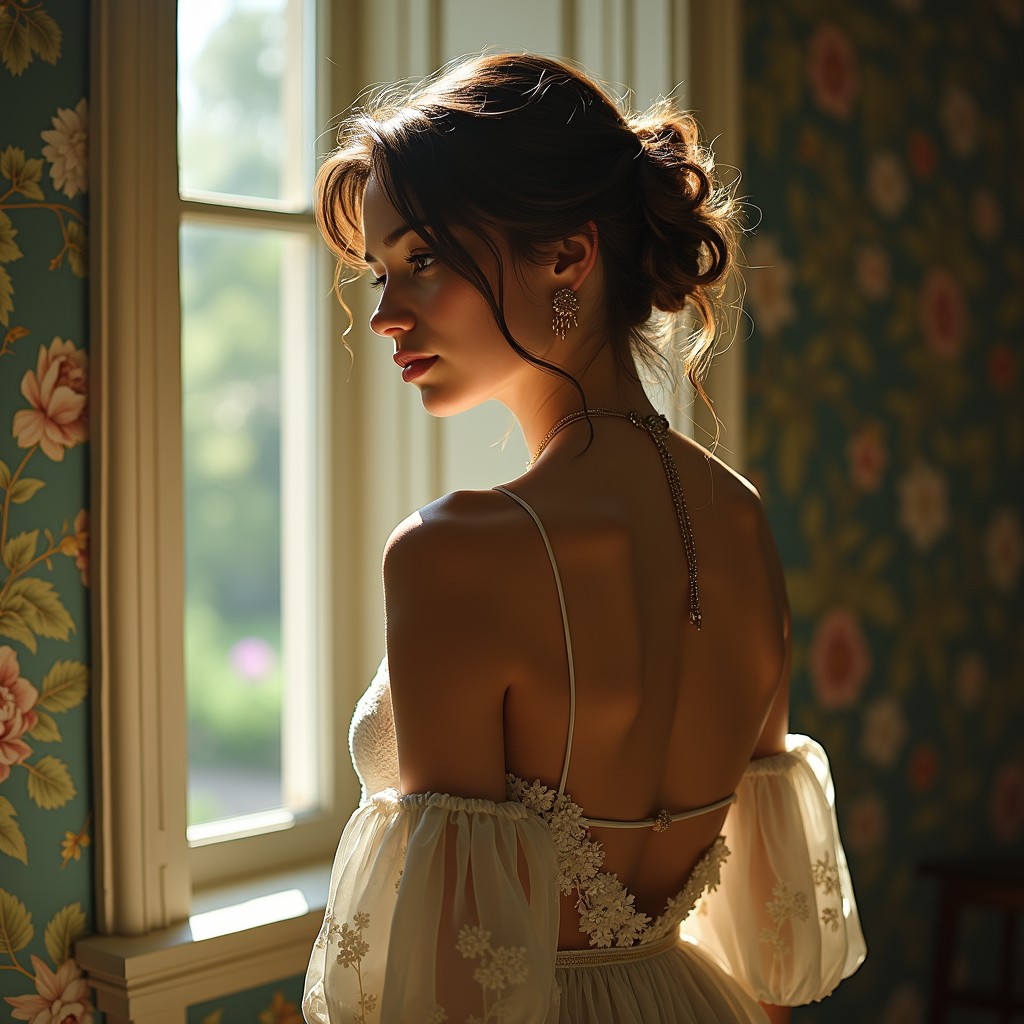}
        \caption*{Vanilla-Flux}
    \end{subfigure}
    \hfill
    \begin{subfigure}[b]{0.155\textwidth}
        \includegraphics[width=\linewidth]{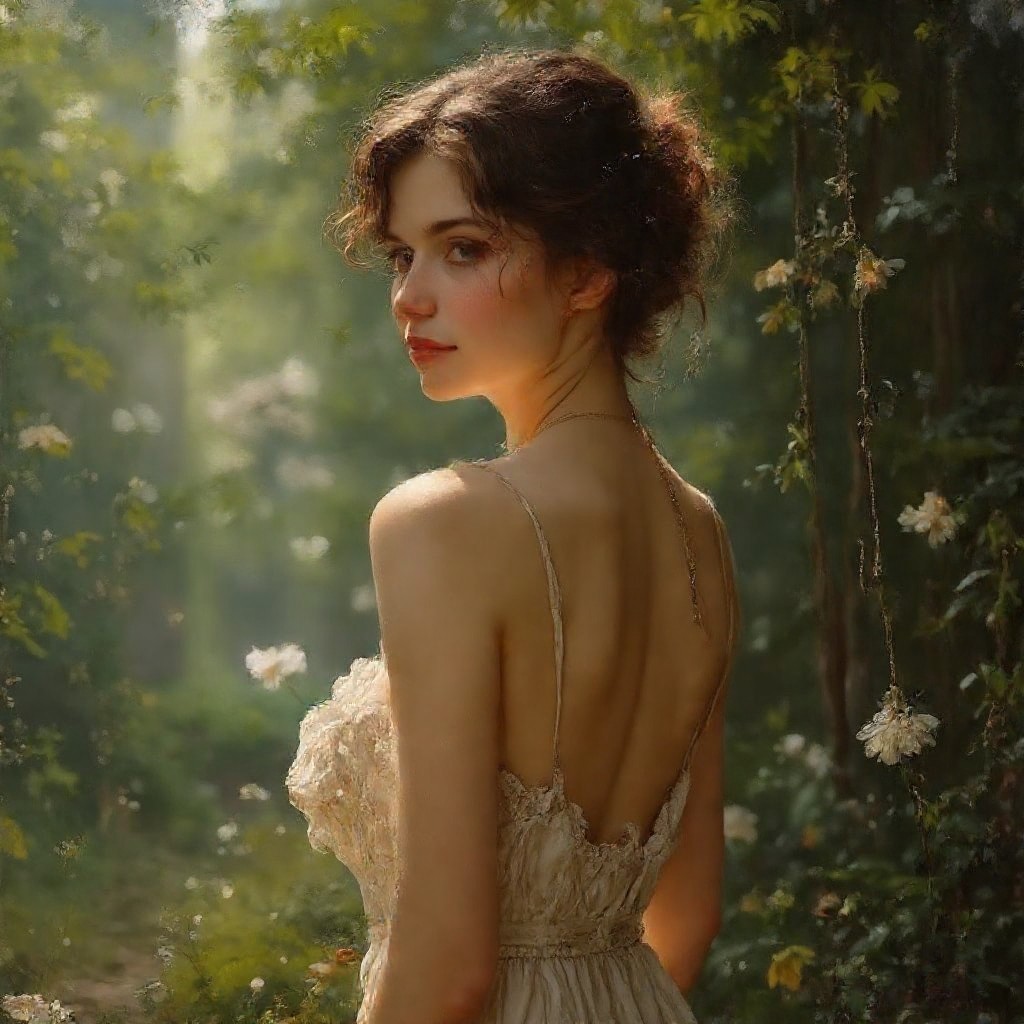}
        \caption*{\textbf{Shiva-Flux}}
    \end{subfigure}
    \hfill
    \begin{subfigure}[b]{0.155\textwidth}
        \includegraphics[width=\linewidth]{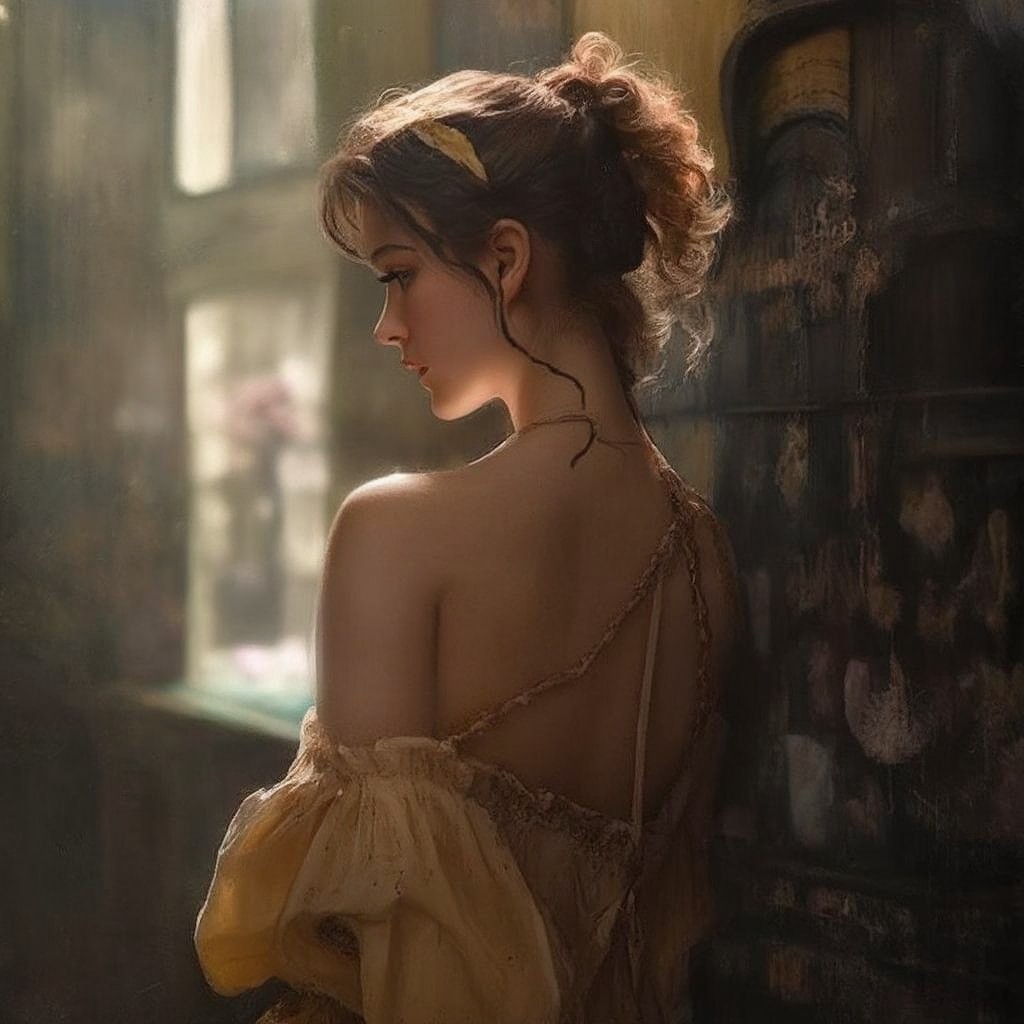}
        \caption*{Finetuned-Flux}
    \end{subfigure}
    \hfill
    \begin{subfigure}[b]{0.155\textwidth}
        \includegraphics[width=\linewidth]{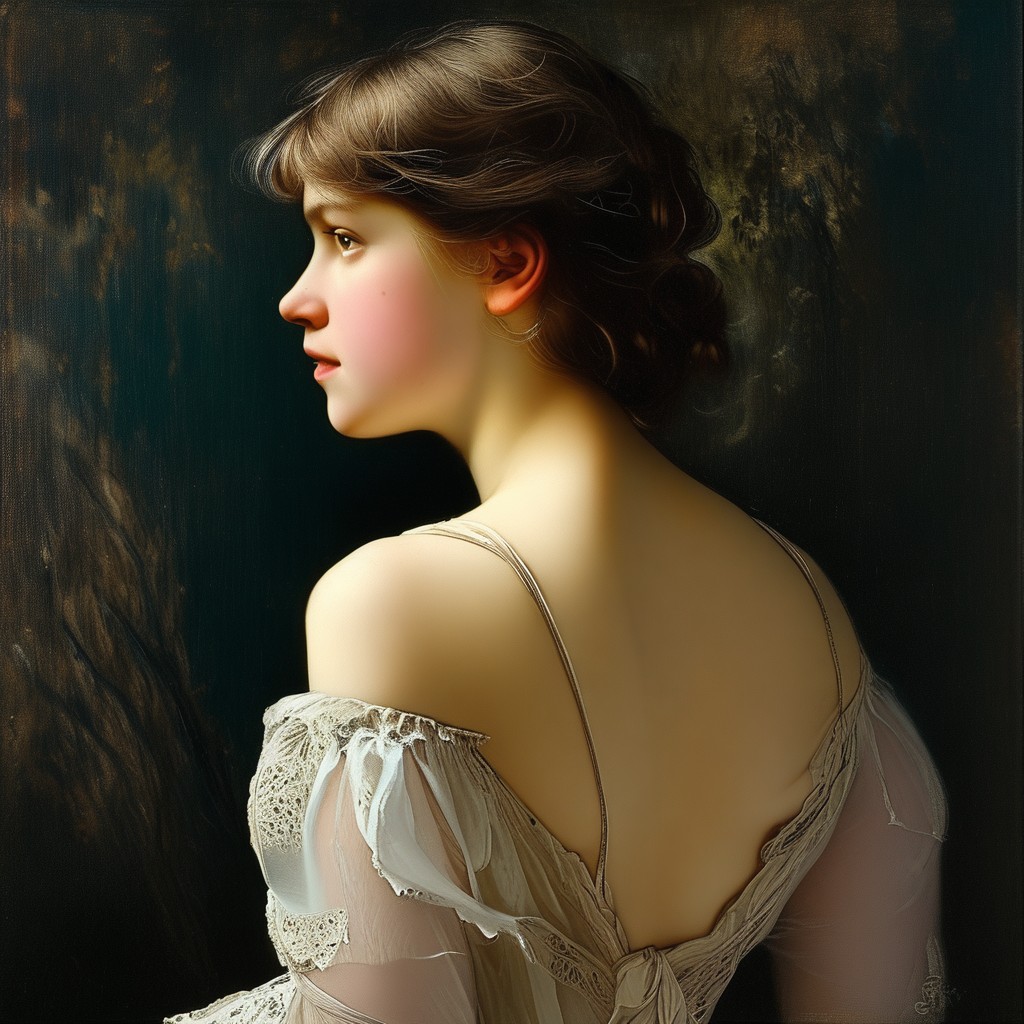}
        \caption*{Vanilla-Pixart}
    \end{subfigure}
    \hfill
    \begin{subfigure}[b]{0.155\textwidth}
        \includegraphics[width=\linewidth]{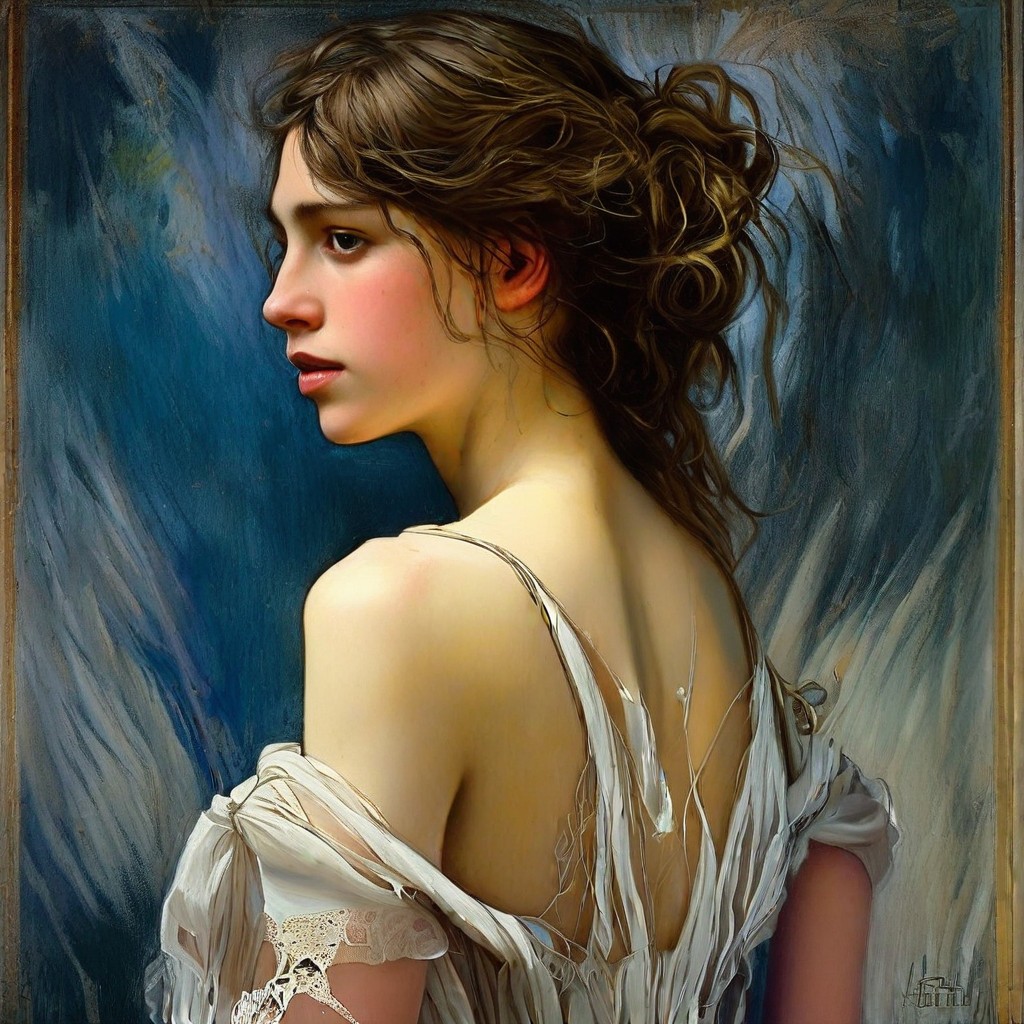}
        \caption*{\textbf{Shiva-Pixart}}
    \end{subfigure}
    \hfill
    \begin{subfigure}[b]{0.155\textwidth}
        \includegraphics[width=\linewidth]{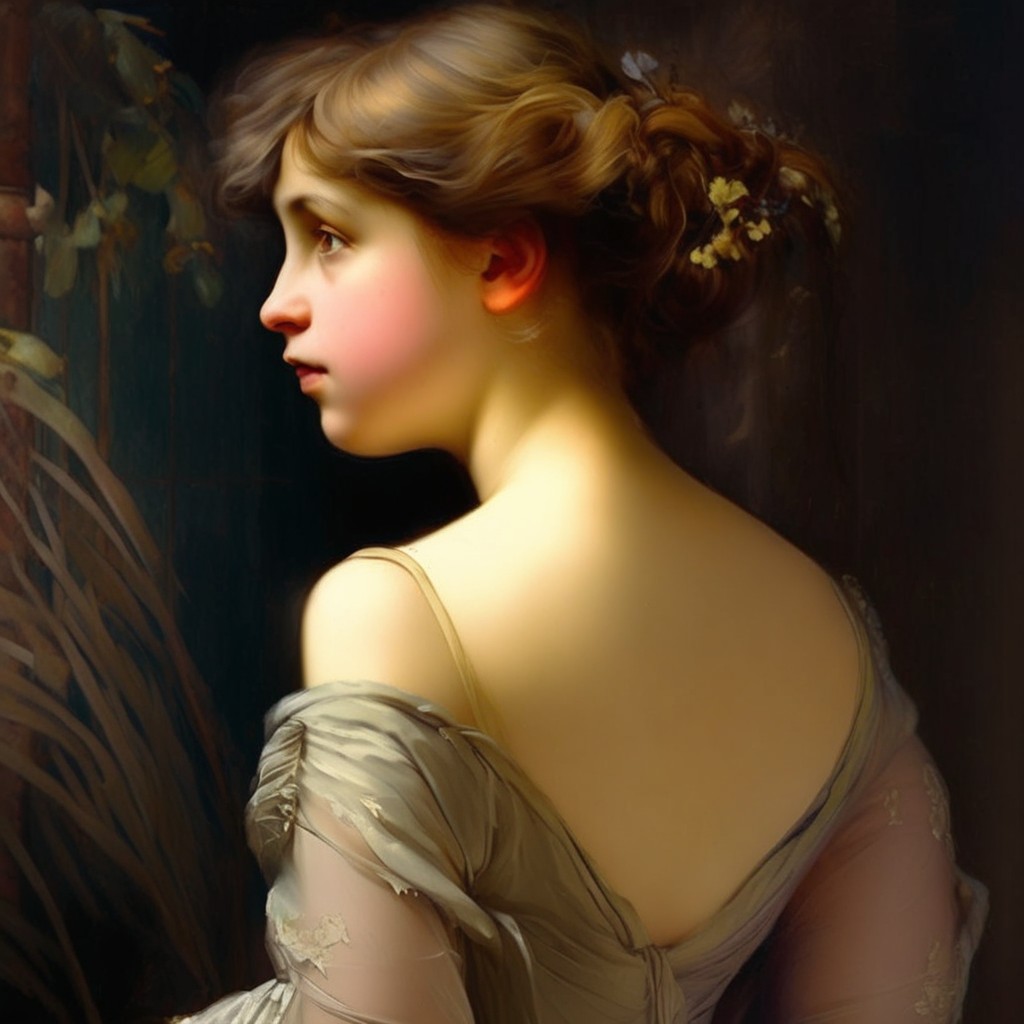}
        \caption*{Finetuned-Pixart}
    \end{subfigure}
    
    \caption{Prompt: \textit{Young woman in an openback summer dress, looking longingly, visible shoulders, by Henri Gervex, Édouard Manet, JeanHonoré Fragonard, Alfons Mucha, high fantasy, cinematic lighting, romantic atmosphere, ultrarealistic, hyperrealistic, intricate detail.}}
    \label{fig:appendix_fig_6}
\end{figure}

\begin{figure}[H]
    \centering
    \begin{subfigure}[b]{0.155\textwidth}
        \includegraphics[width=\linewidth]{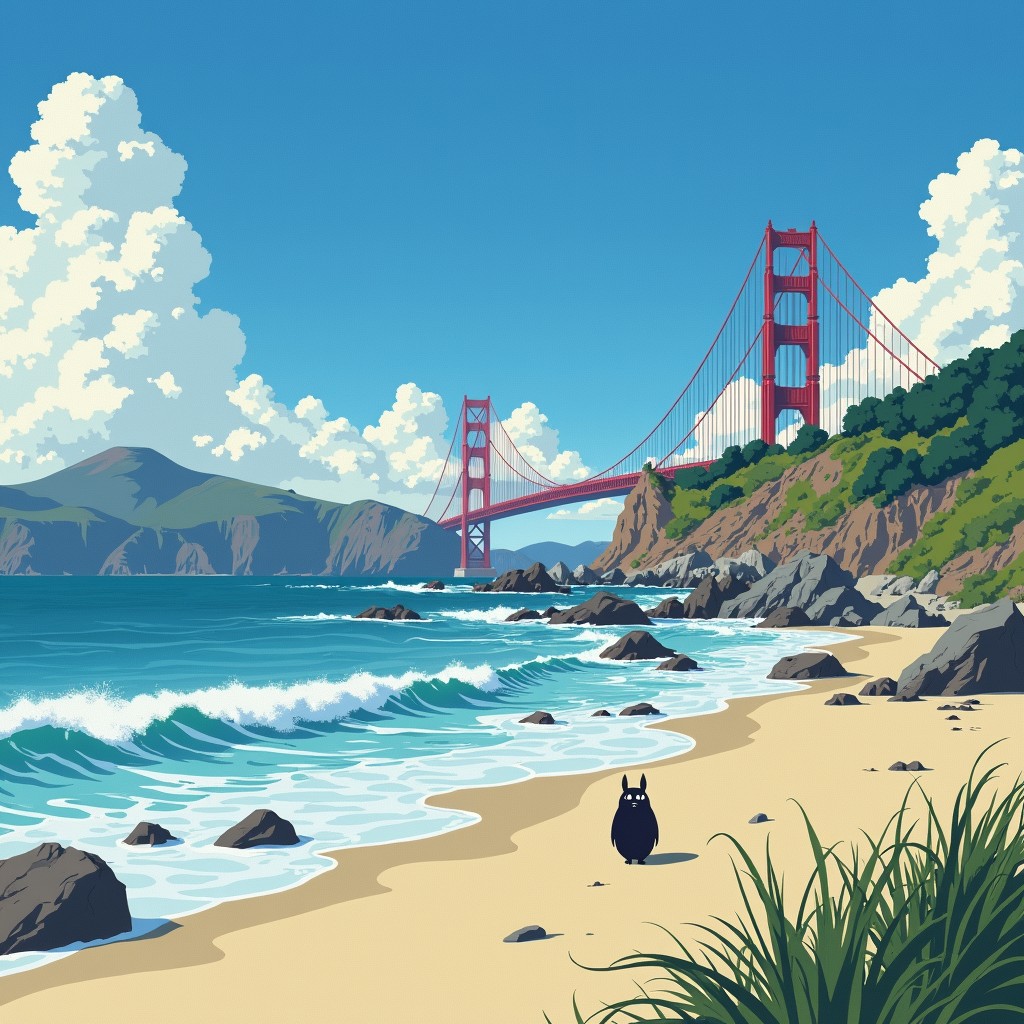}
        \caption*{Vanilla-Flux}
    \end{subfigure}
    \hfill
    \begin{subfigure}[b]{0.155\textwidth}
        \includegraphics[width=\linewidth]{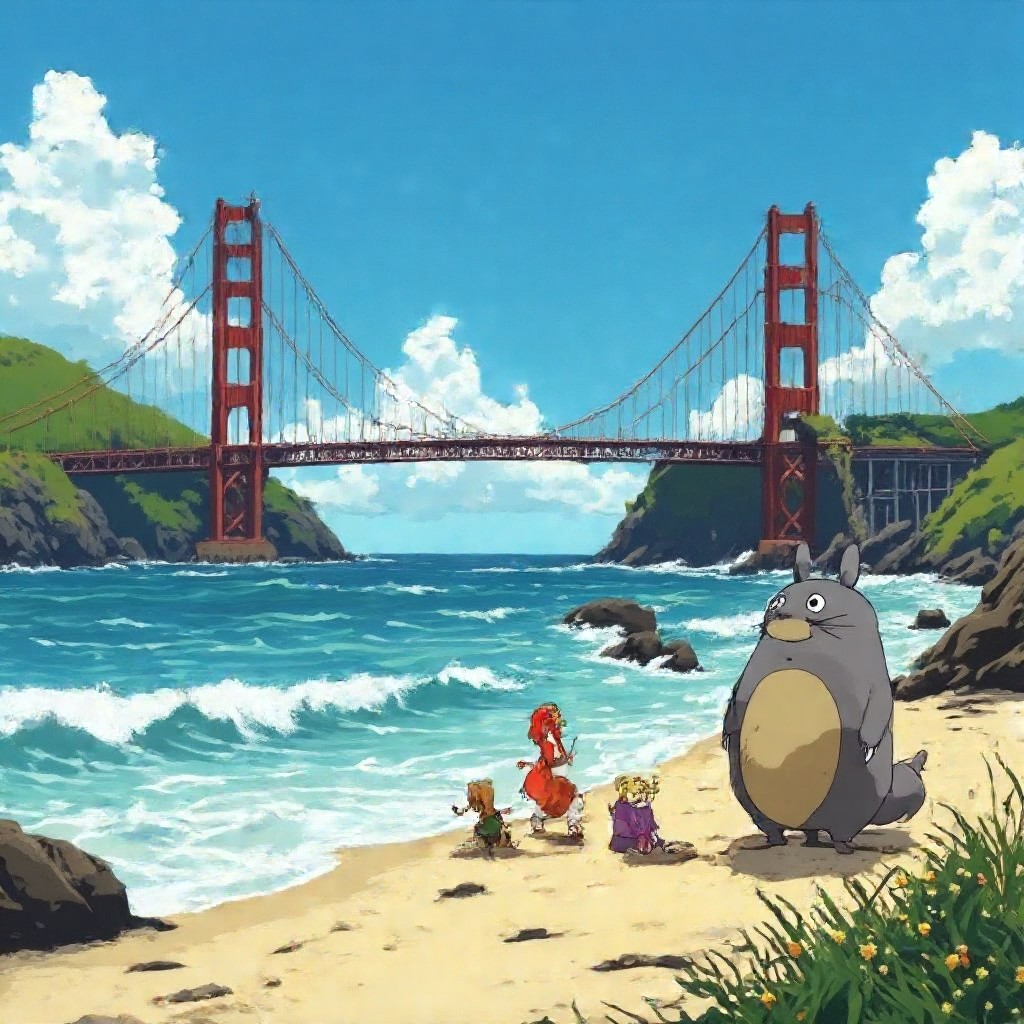}
        \caption*{\textbf{Shiva-Flux}}
    \end{subfigure}
    \hfill
    \begin{subfigure}[b]{0.155\textwidth}
        \includegraphics[width=\linewidth]{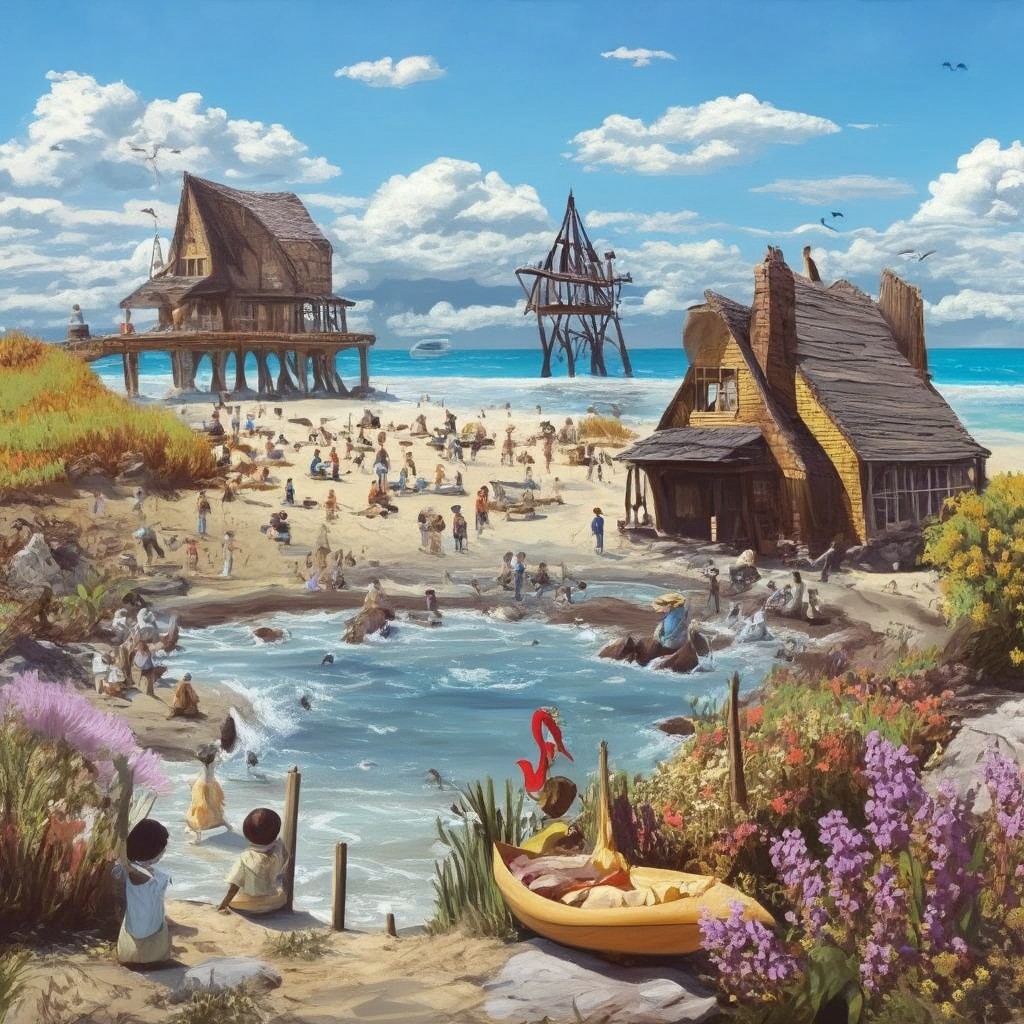}
        \caption*{Finetuned-Flux}
    \end{subfigure}
    \hfill
    \begin{subfigure}[b]{0.155\textwidth}
        \includegraphics[width=\linewidth]{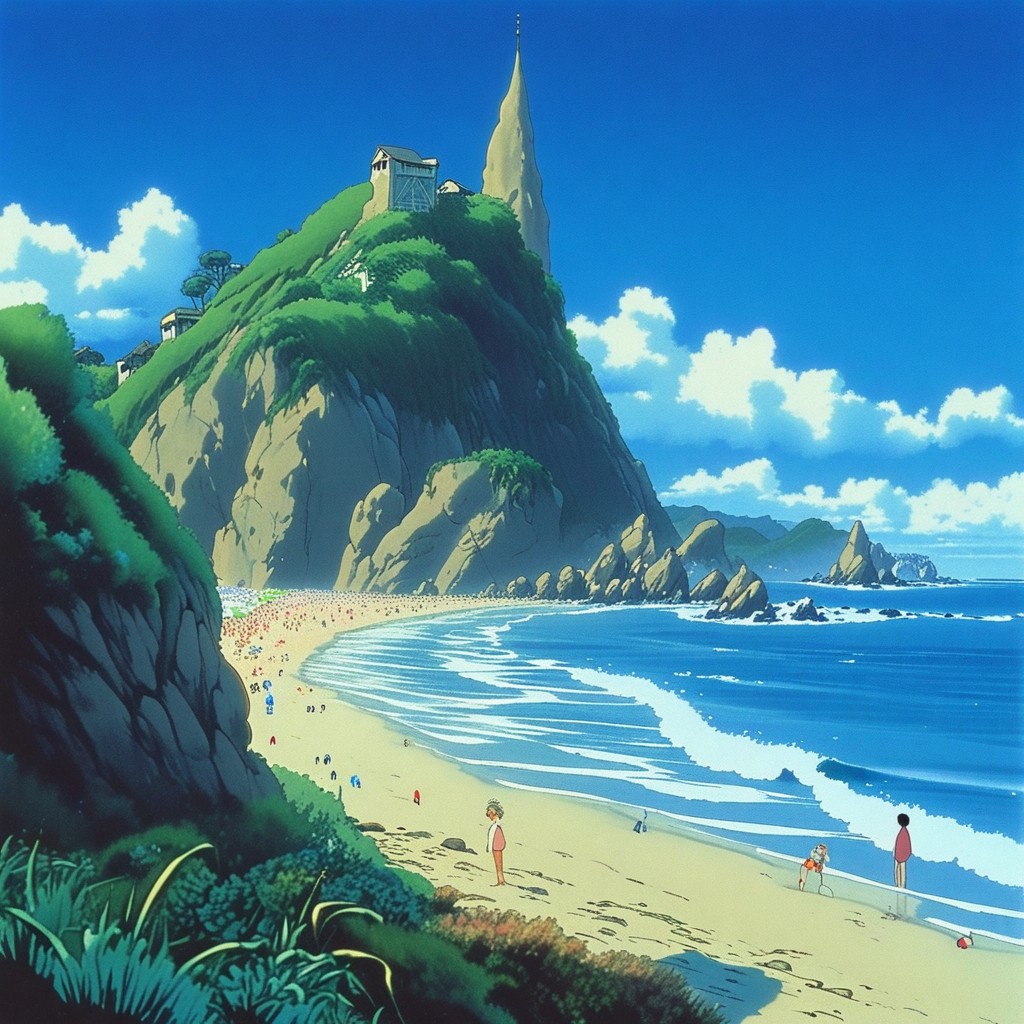}
        \caption*{Vanilla-Pixart}
    \end{subfigure}
    \hfill
    \begin{subfigure}[b]{0.155\textwidth}
        \includegraphics[width=\linewidth]{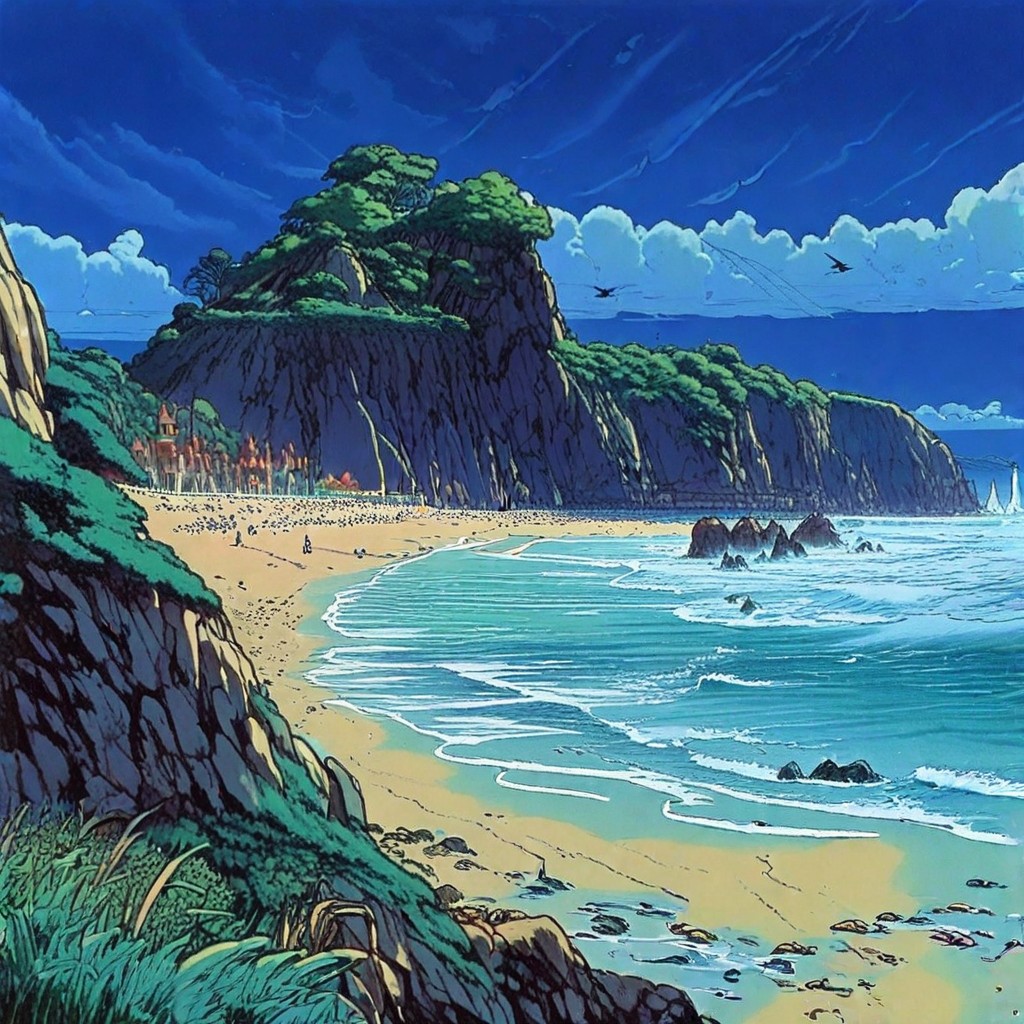}
        \caption*{\textbf{Shiva-Pixart}}
    \end{subfigure}
    \hfill
    \begin{subfigure}[b]{0.155\textwidth}
        \includegraphics[width=\linewidth]{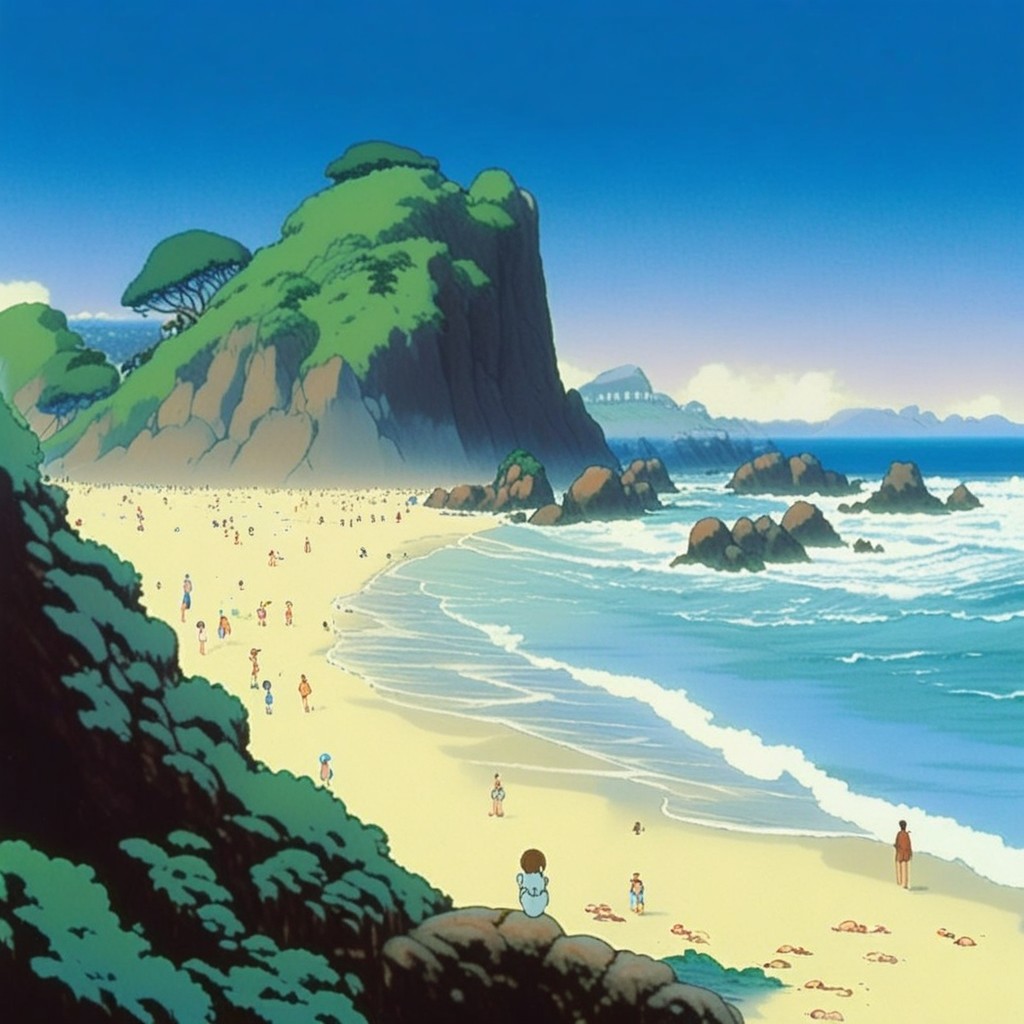}
        \caption*{Finetuned-Pixart}
    \end{subfigure}
    
    \caption{Prompt: \textit{1989 Studio Ghibli anime movie, mythical ocean beach san francisco.}}
    \label{fig:appendix_fig_7}
\end{figure}

\begin{figure}[H]
    \centering
    \begin{subfigure}[b]{0.155\textwidth}
        \includegraphics[width=\linewidth]{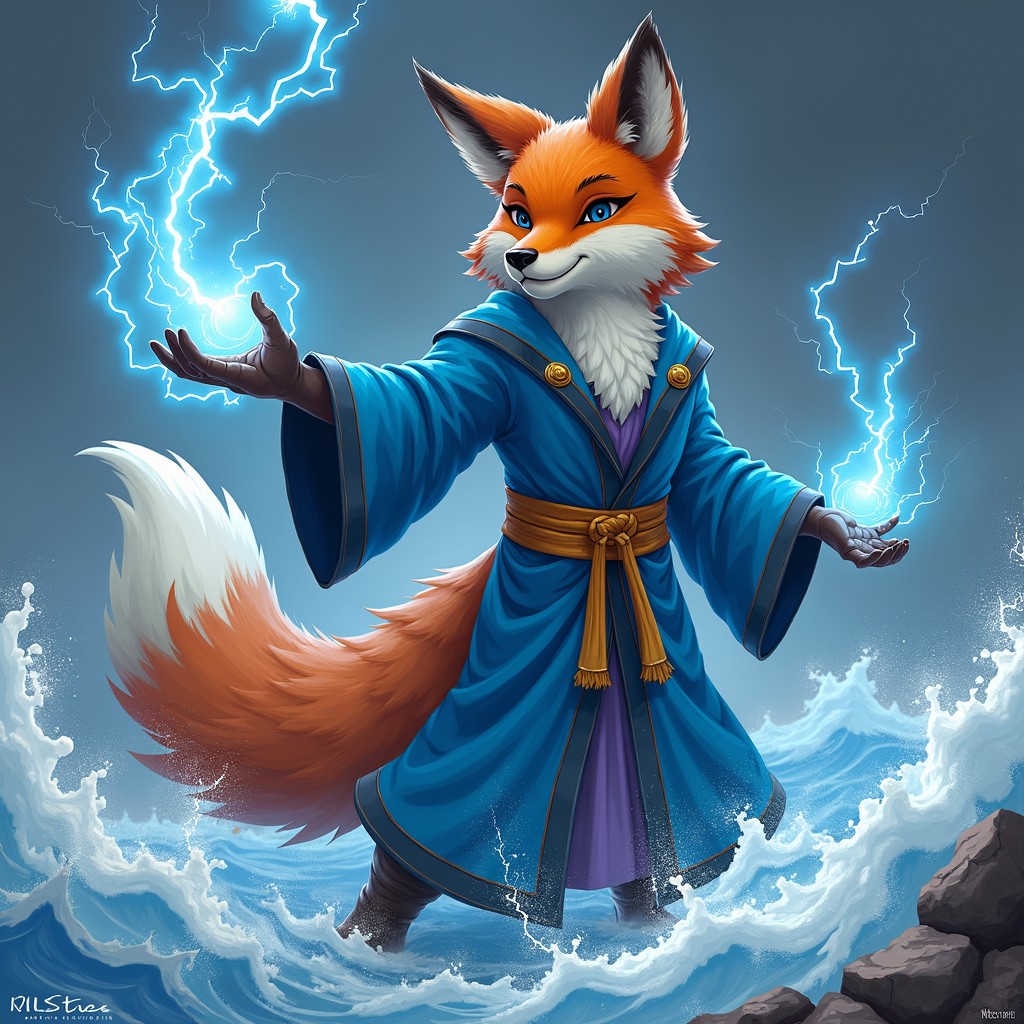}
        \caption*{Vanilla-Flux}
    \end{subfigure}
    \hfill
    \begin{subfigure}[b]{0.155\textwidth}
        \includegraphics[width=\linewidth]{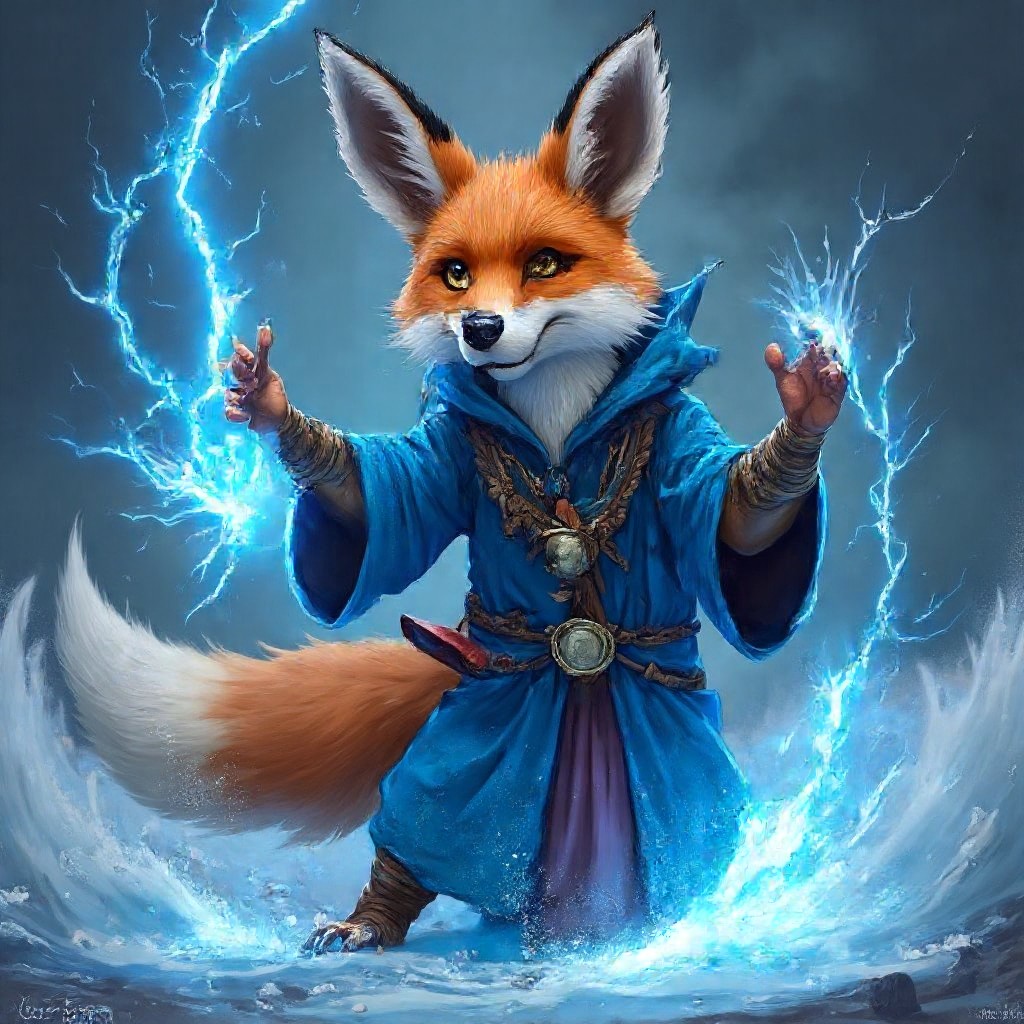}
        \caption*{\textbf{Shiva-Flux}}
    \end{subfigure}
    \hfill
    \begin{subfigure}[b]{0.155\textwidth}
        \includegraphics[width=\linewidth]{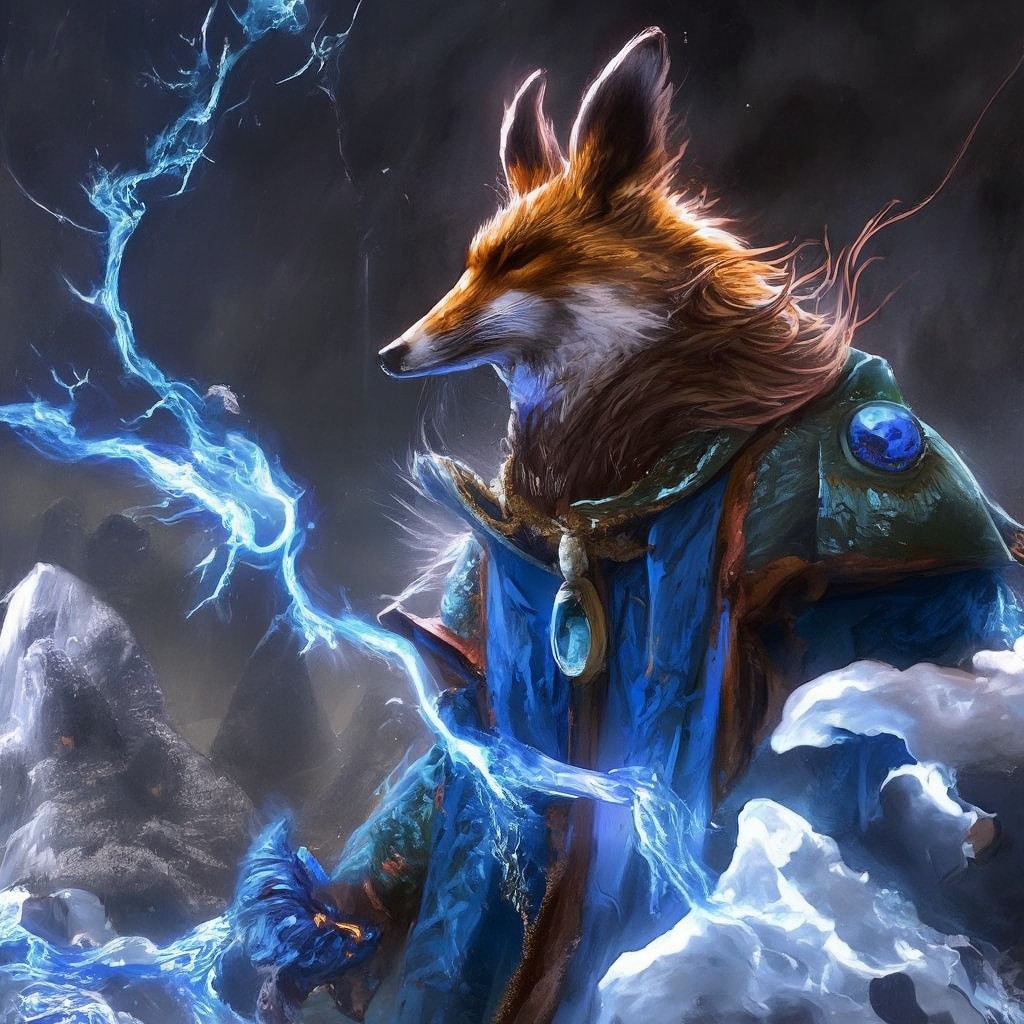}
        \caption*{Finetuned-Flux}
    \end{subfigure}
    \hfill
    \begin{subfigure}[b]{0.155\textwidth}
        \includegraphics[width=\linewidth]{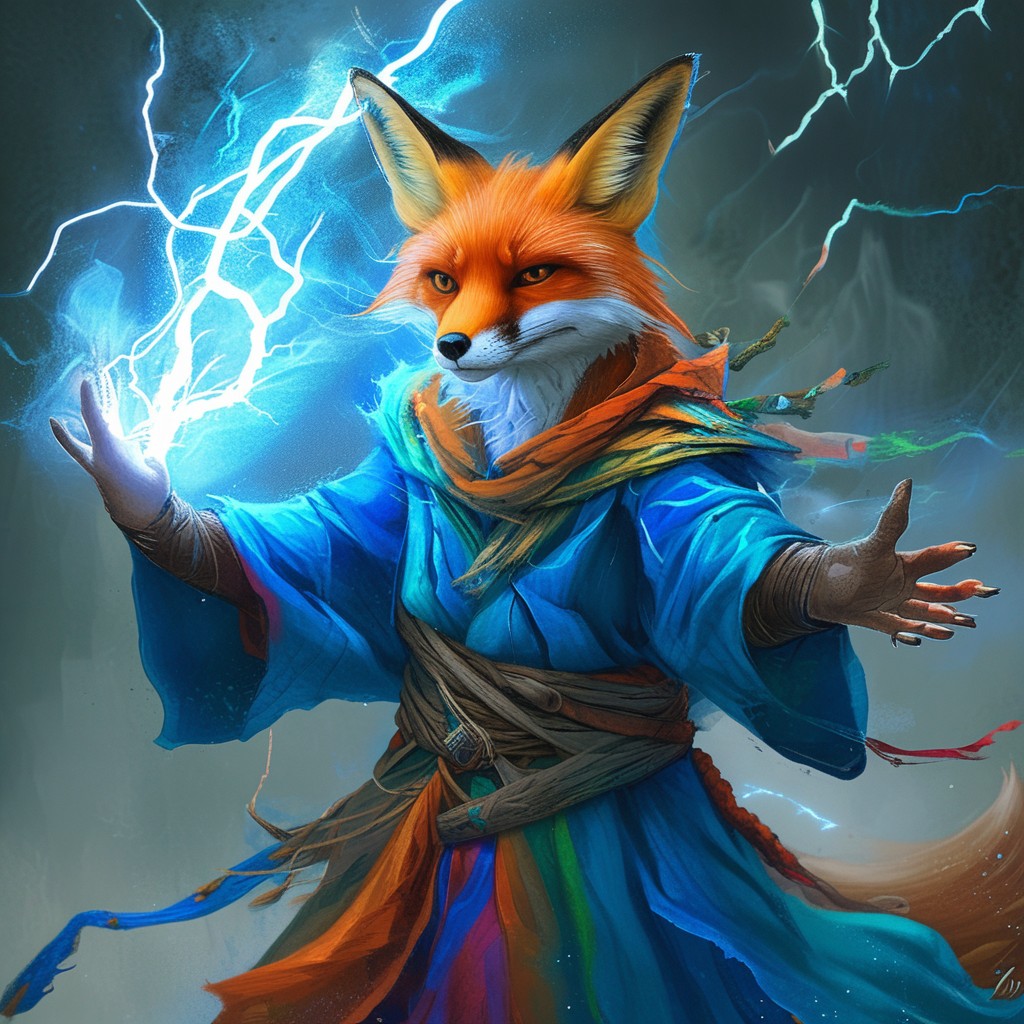}
        \caption*{Vanilla-Pixart}
    \end{subfigure}
    \hfill
    \begin{subfigure}[b]{0.155\textwidth}
        \includegraphics[width=\linewidth]{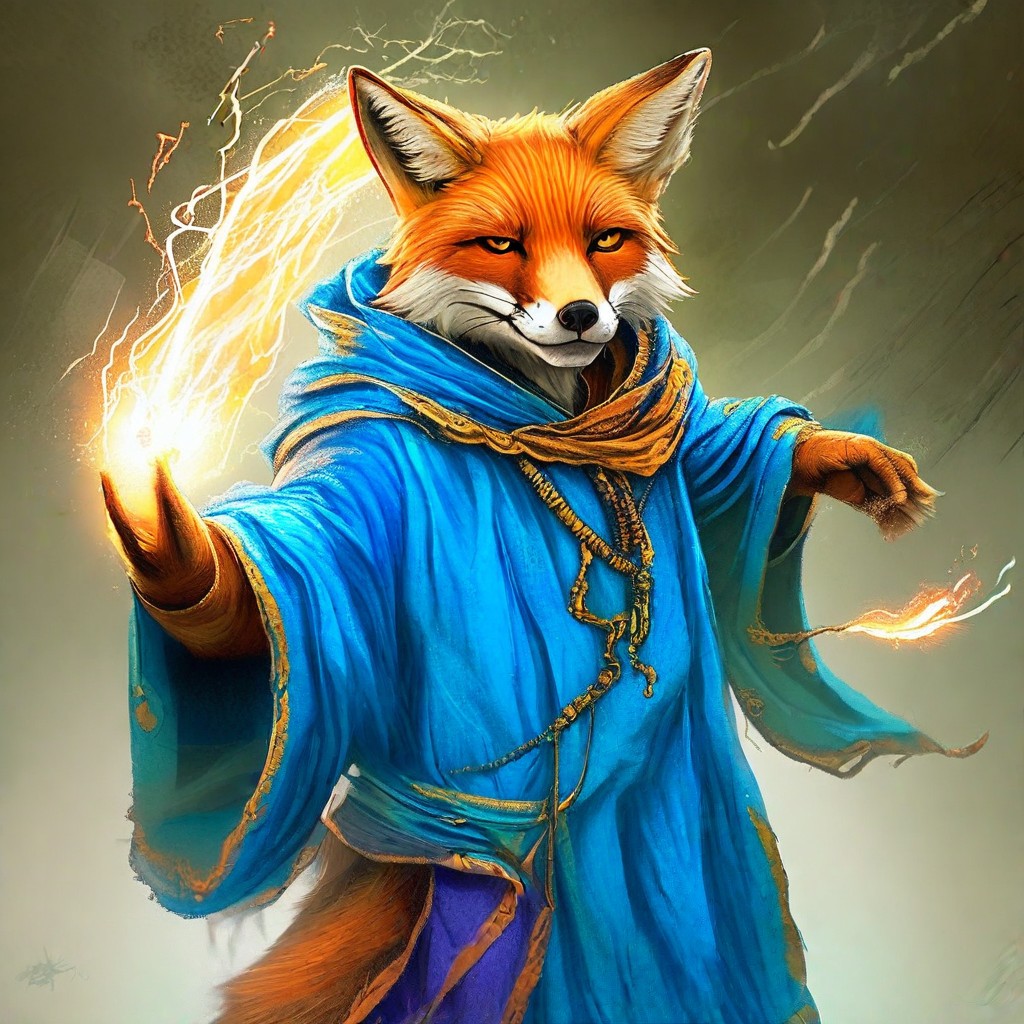}
        \caption*{\textbf{Shiva-Pixart}}
    \end{subfigure}
    \hfill
    \begin{subfigure}[b]{0.155\textwidth}
        \includegraphics[width=\linewidth]{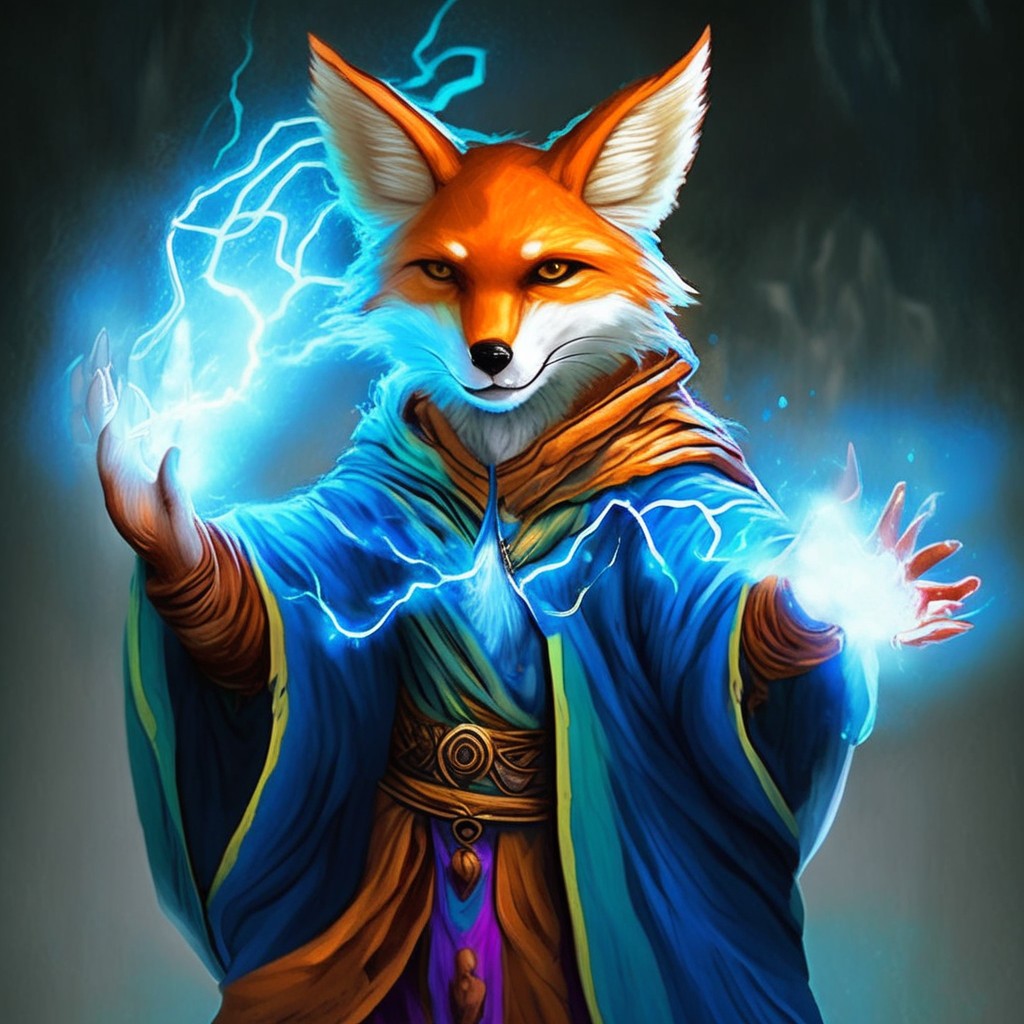}
        \caption*{Finetuned-Pixart}
    \end{subfigure}
    
    \caption{Prompt: \textit{A Fox druid wearing blue colorful robes casting thunder Wave.}}
    \label{fig:appendix_fig_8}
\end{figure}

\begin{figure}[H]
    \centering
    \begin{subfigure}[b]{0.155\textwidth}
        \includegraphics[width=\linewidth]{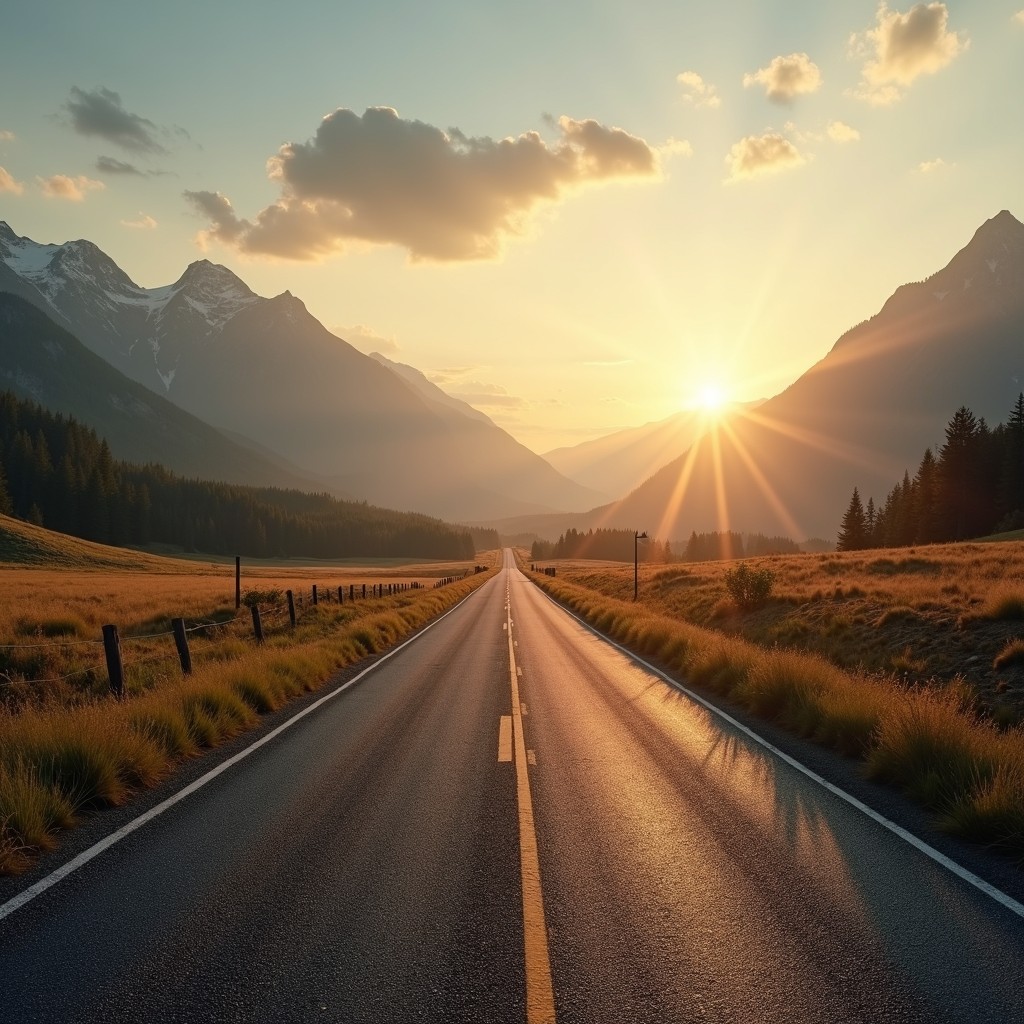}
        \caption*{Vanilla-Flux}
    \end{subfigure}
    \hfill
    \begin{subfigure}[b]{0.155\textwidth}
        \includegraphics[width=\linewidth]{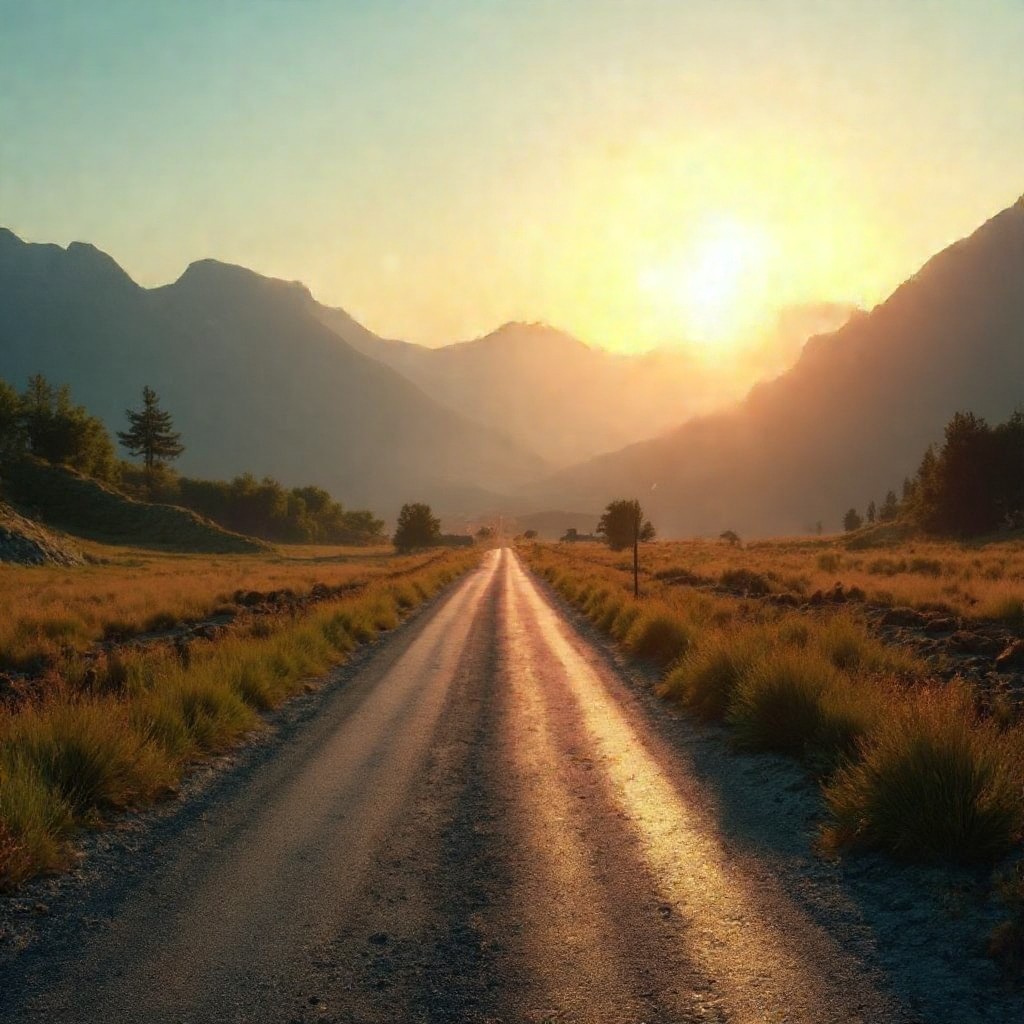}
        \caption*{\textbf{Shiva-Flux}}
    \end{subfigure}
    \hfill
    \begin{subfigure}[b]{0.155\textwidth}
        \includegraphics[width=\linewidth]{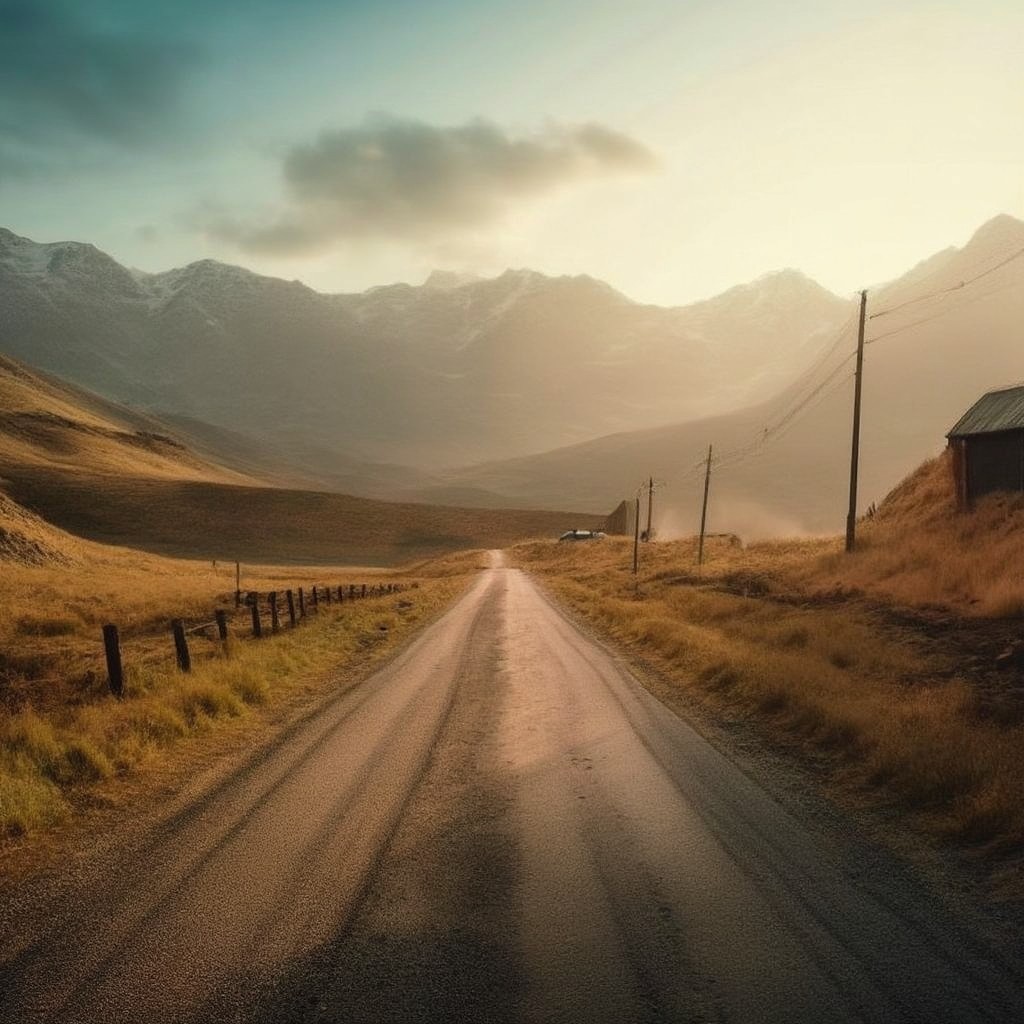}
        \caption*{Finetuned-Flux}
    \end{subfigure}
    \hfill
    \begin{subfigure}[b]{0.155\textwidth}
        \includegraphics[width=\linewidth]{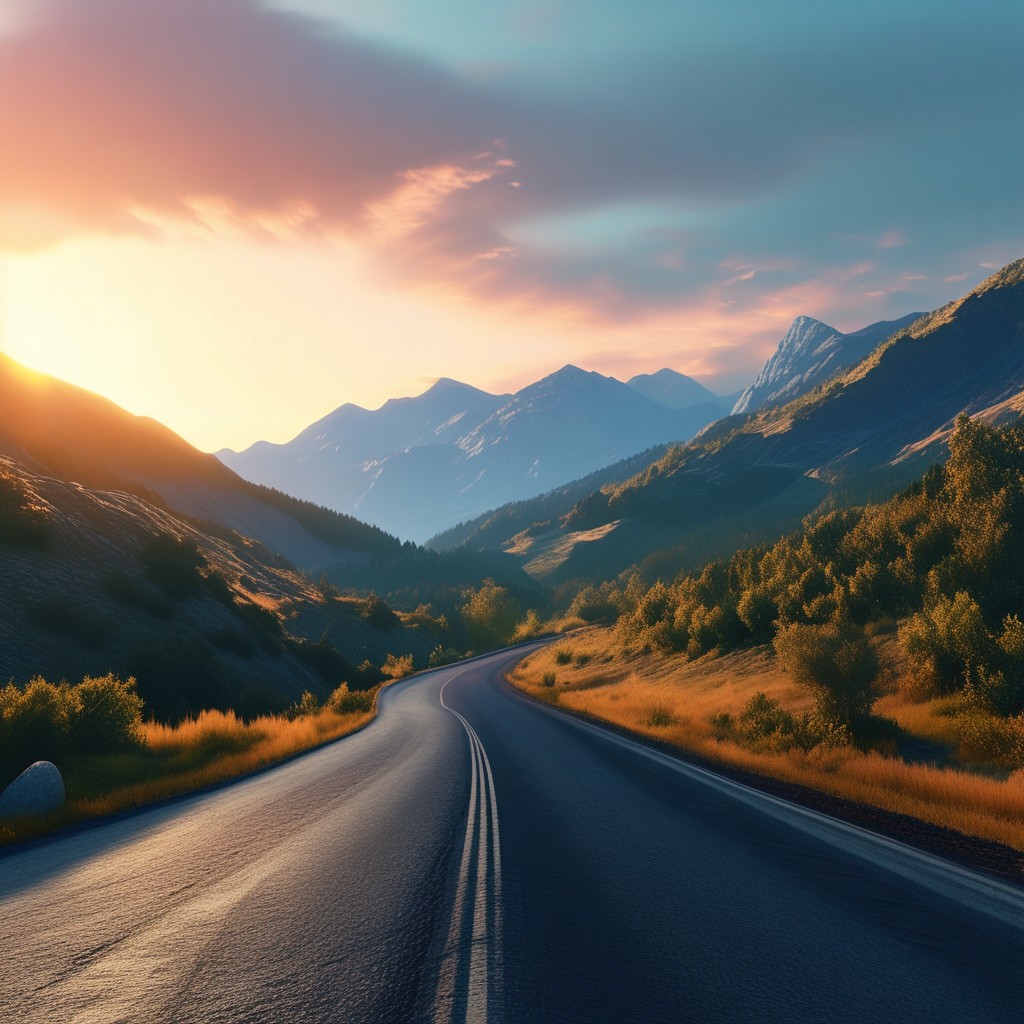}
        \caption*{Vanilla-Pixart}
    \end{subfigure}
    \hfill
    \begin{subfigure}[b]{0.155\textwidth}
        \includegraphics[width=\linewidth]{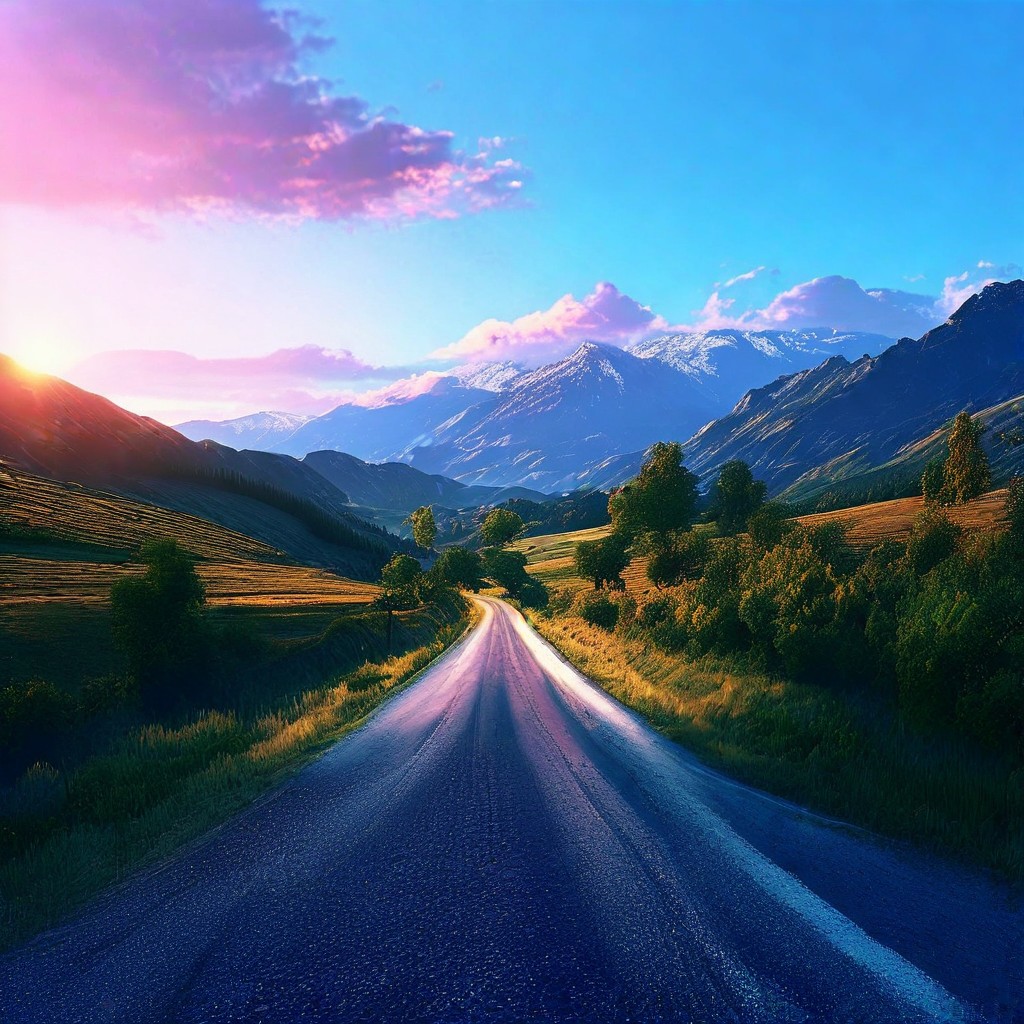}
        \caption*{\textbf{Shiva-Pixart}}
    \end{subfigure}
    \hfill
    \begin{subfigure}[b]{0.155\textwidth}
        \includegraphics[width=\linewidth]{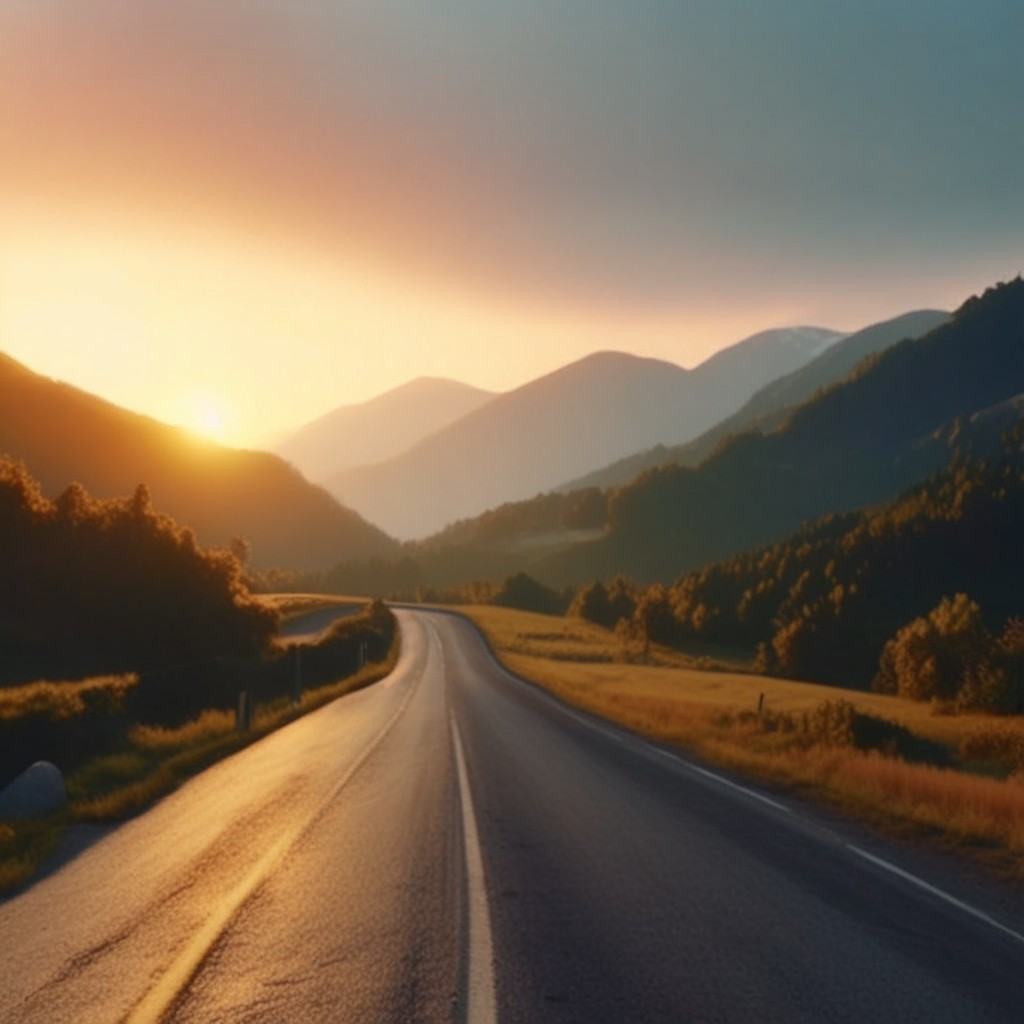}
        \caption*{Finetuned-Pixart}
    \end{subfigure}
    
    \caption{Prompt: \textit{Country road with mountains in the background at sunrise. realistic footage, ultra details, 4k.}}
    \label{fig:appendix_fig_9}
\end{figure}

\begin{figure}[H]
    \centering
    \begin{subfigure}[b]{0.155\textwidth}
        \includegraphics[width=\linewidth]{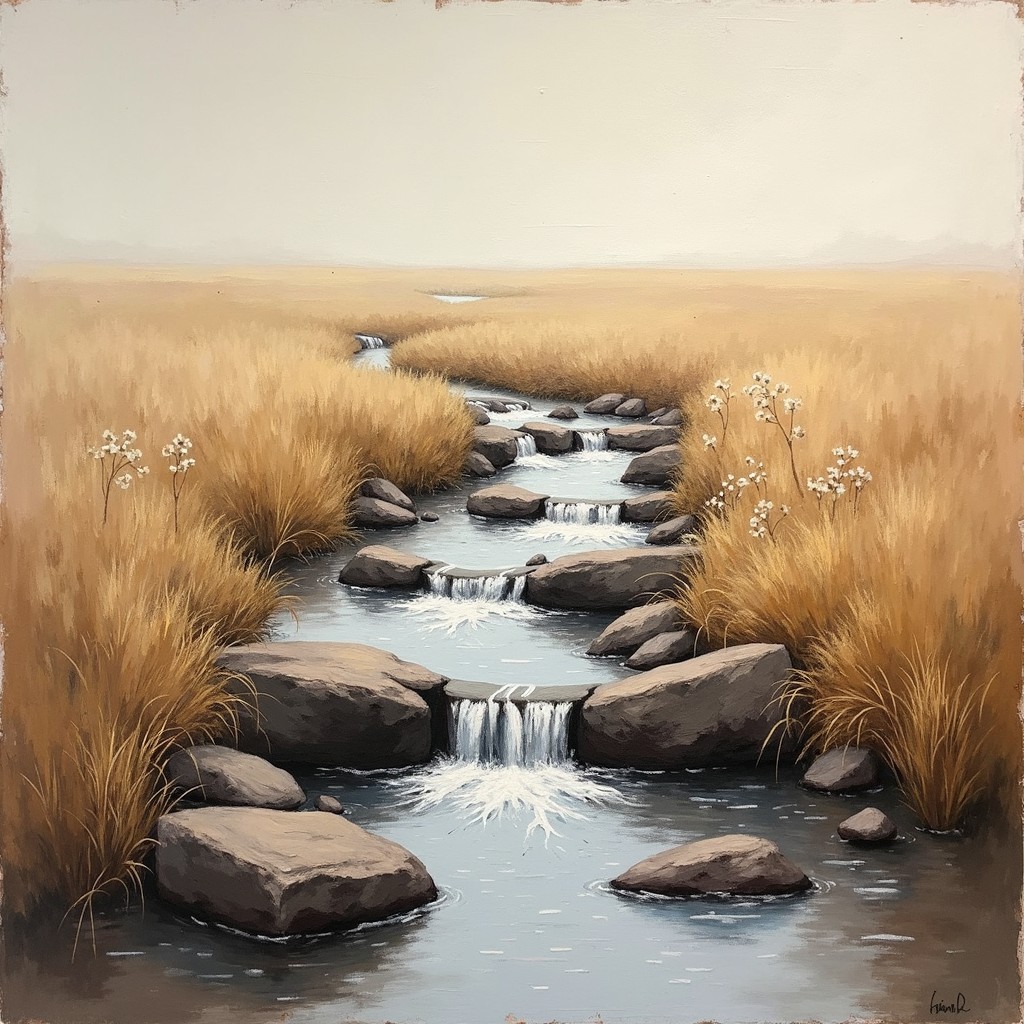}
        \caption*{Vanilla-Flux}
    \end{subfigure}
    \hfill
    \begin{subfigure}[b]{0.155\textwidth}
        \includegraphics[width=\linewidth]{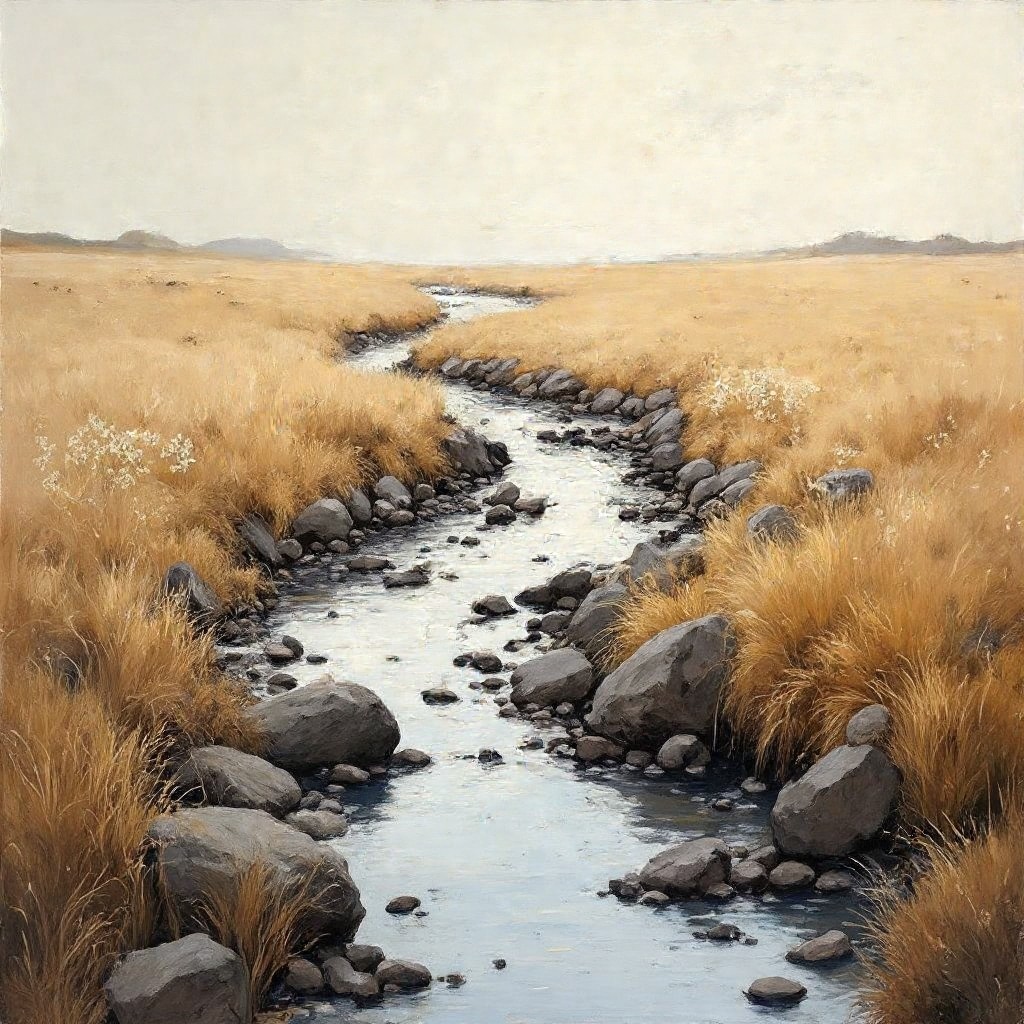}
        \caption*{\textbf{Shiva-Flux}}
    \end{subfigure}
    \hfill
    \begin{subfigure}[b]{0.155\textwidth}
        \includegraphics[width=\linewidth]{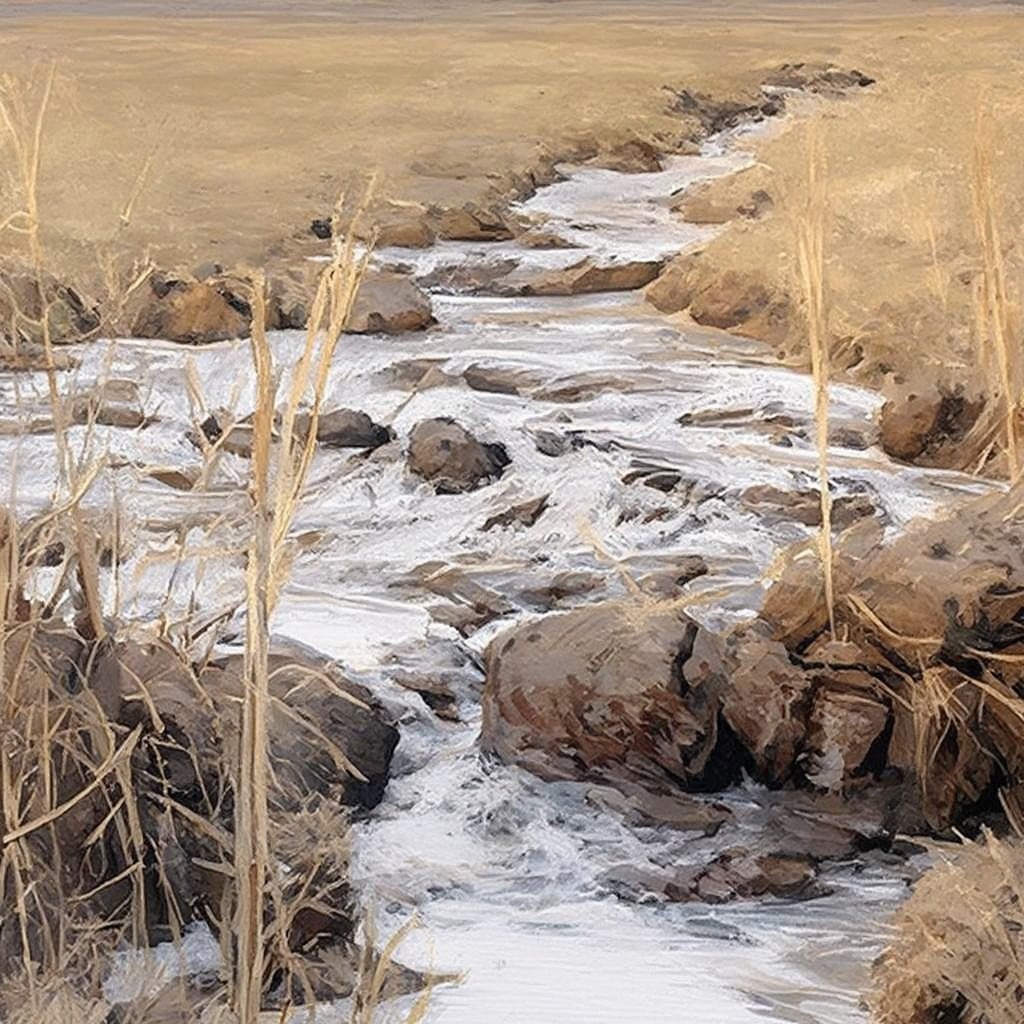}
        \caption*{Finetuned-Flux}
    \end{subfigure}
    \hfill
    \begin{subfigure}[b]{0.155\textwidth}
        \includegraphics[width=\linewidth]{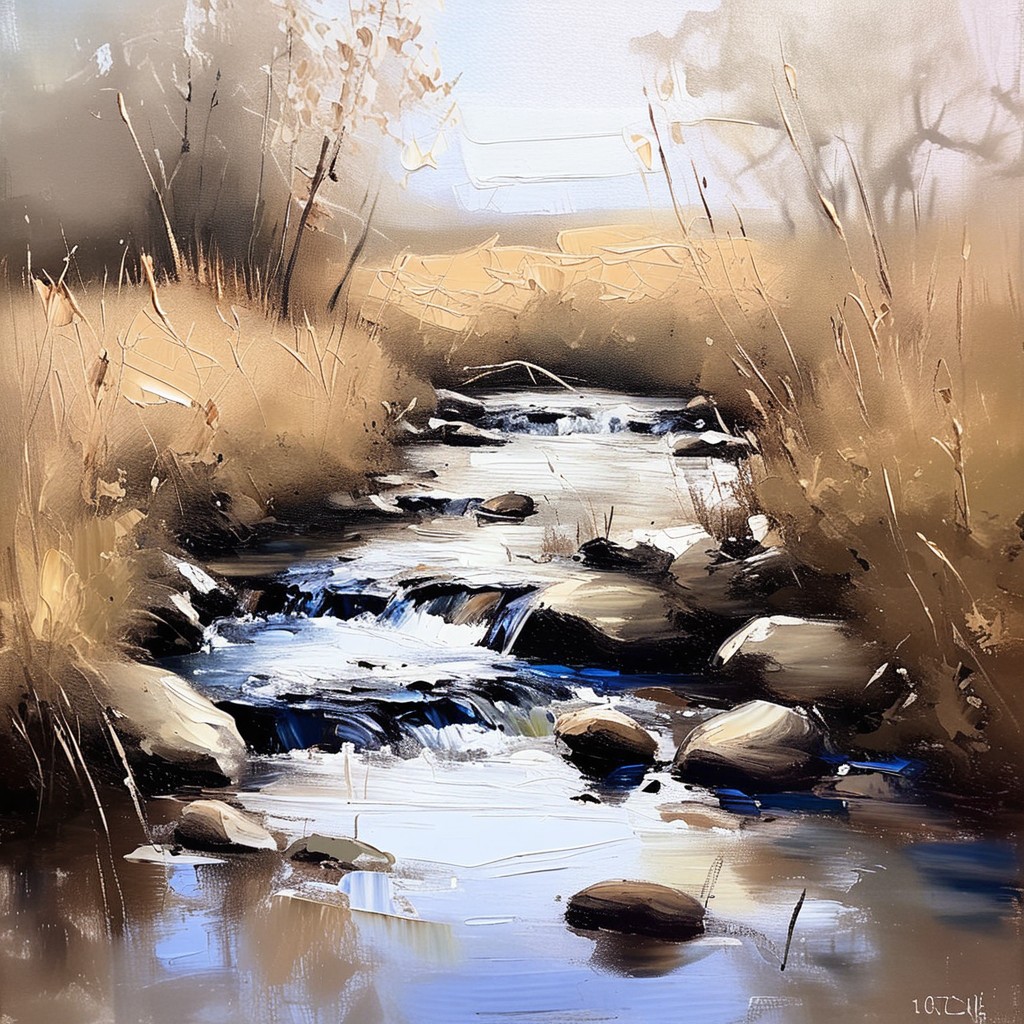}
        \caption*{Vanilla-Pixart}
    \end{subfigure}
    \hfill
    \begin{subfigure}[b]{0.155\textwidth}
        \includegraphics[width=\linewidth]{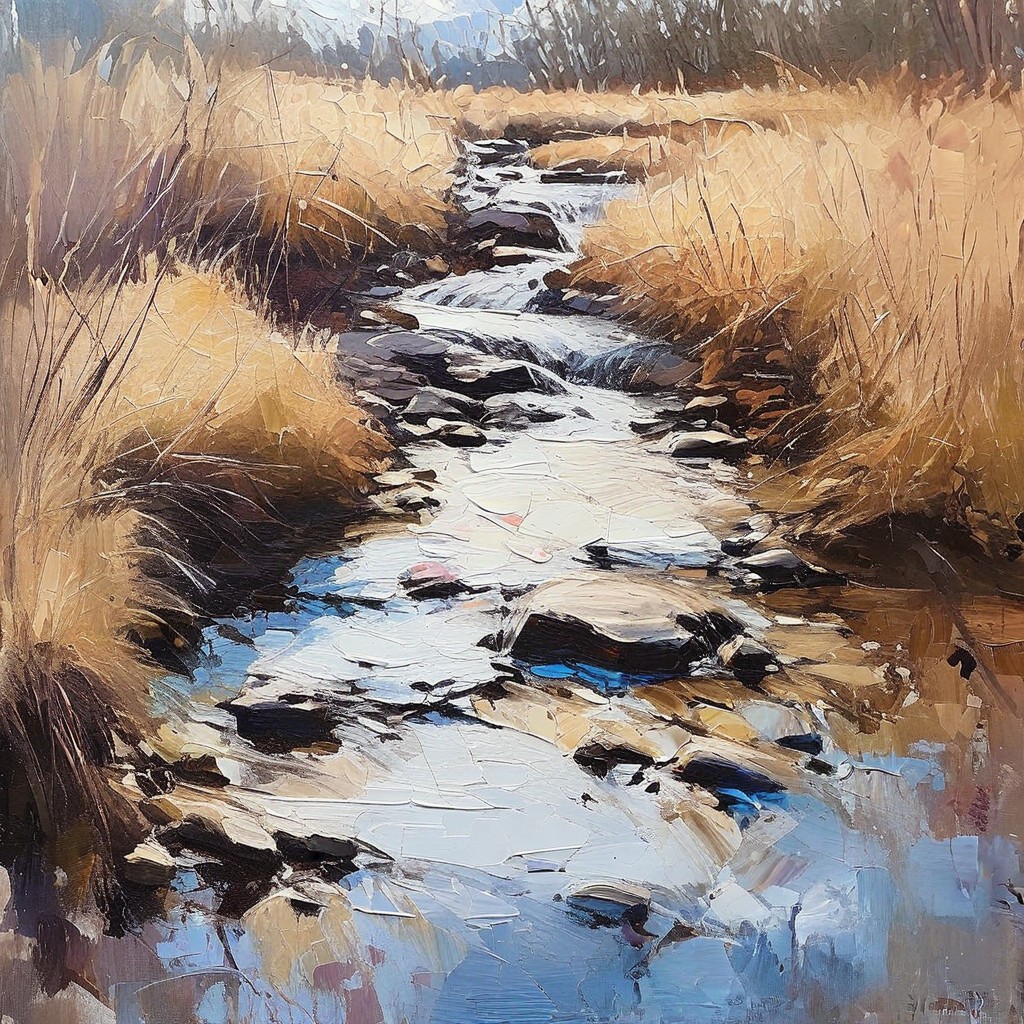}
        \caption*{\textbf{Shiva-Pixart}}
    \end{subfigure}
    \hfill
    \begin{subfigure}[b]{0.155\textwidth}
        \includegraphics[width=\linewidth]{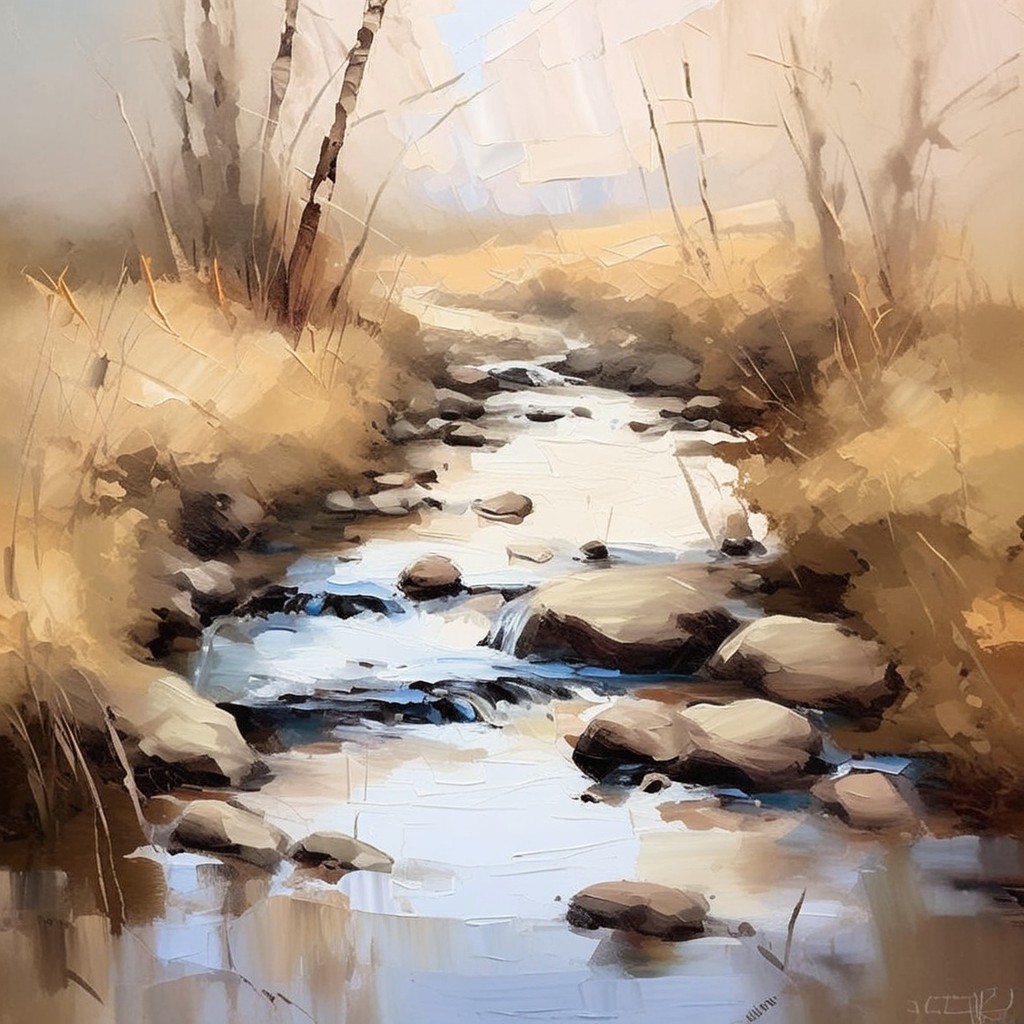}
        \caption*{Finetuned-Pixart}
    \end{subfigure}
    
    \caption{Prompt: \textit{Plein air painting, palette knife, loose brushwork, slightly abstract, a thin creek spilling over rocks, drying grass field, soft lighting, soft colors, beige, white, brown, serene, vintage.}}
    \label{fig:appendix_fig_10}
\end{figure}

Additional qualitative comparisons are presented in Figures~\ref{fig:appendix_fig_4}, \ref{fig:appendix_fig_5}, \ref{fig:appendix_fig_6}, \ref{fig:appendix_fig_7}, \ref{fig:appendix_fig_8}, \ref{fig:appendix_fig_9}, and \ref{fig:appendix_fig_10}.

\end{document}